%% file: main.tex
\documentclass[10pt,a4paper,singlecolumn,google]{google}
\uselogo{}

\usepackage{letltxmacro}        
\usepackage{xspace}             
\usepackage{breakcites}         
\usepackage{epsfig,subfigure,float,wrapfig}                   
\usepackage{amsmath,amsfonts,amsthm,bm,nicefrac,colonequals}  
\usepackage{multirow,colortbl,booktabs,array,tabularx}        
\usepackage{nameref}            
\usepackage{appendix,titletoc}  
\usepackage{minted}
\usepackage{upquote}            
\usepackage[capitalize,noabbrev]{cleveref}
\usepackage[nodisplayskipstretch]{setspace}
\usepackage[normalem]{ulem}     
\usepackage{algorithm,algpseudocode}

\input{math_commands}
\input{macros}

\title{Muir:%
~Linear Attention with Sensitivity-Allocated Gating and Matched-Filter Readout
}
\reportnumber{}  

\author[1,2]{Tushaar~Gangavarapu}
\author[2]{Gil~Tabak}
\author[2]{Clemens~JS~Schaefer}

\correspondingauthor{cjsschaefer@google.com}

\affil[1]{University of Texas at Austin}
\affil[2]{\thepa{}{}}

\begin{abstract}
    \input{sections/0_abstract}
\end{abstract}

\begin{document}
\maketitle

\input{sections/1_intro}
\input{sections/2_prelim}
\input{sections/3_theory_v2}
\input{sections/4_method}
\input{sections/5_results}
\input{sections/6_analysis}
\input{sections/7_concl}
\input{sections/acks}

\bibliographystyle{plainnat}
\bibliography{paperpile}


\onecolumn
\newpage
\section*{Appendix}
\startcontents[appendix]
\printcontents[appendix]{l}{1}{\setcounter{tocdepth}{3}}

\begin{appendices}
    \input{apps/a_related}
    \input{apps/b_notation}
    \input{apps/c_optim}

    \input{apps/d_theory_v2}
    \input{apps/e_toy}
    \input{apps/f_kernel}
    \input{apps/g_train}
    \input{apps/h_eval}
    \input{apps/i_analysis}
    \input{apps/j_code}
\end{appendices}

\end{document}

%% file: math_commands.tex
\def\Figref#1{Figure~\ref{#1}}

\def\eqref#1{(\ref{#1})}
\def\Eqref#1{Equation~(\ref{#1})}

\def\1{\bm{1}}

\def\eps{{\epsilon}}

\def\vzero{{\bm{0}}}
\def\vone{{\bm{1}}}

\def\valpha{{\bm{\alpha}}}

\def\vgamma{{\bm{\gamma}}}
\def\vxi{{\bm{\xi}}}

\def\vpi{{\bm{\pi}}}

\def\va{{\bm{a}}}
\def\vb{{\bm{b}}}
\def\vc{{\bm{c}}}

\def\ve{{\bm{e}}}

\def\vg{{\bm{g}}}
\def\vh{{\bm{h}}}

\def\vk{{\bm{k}}}

\def\vo{{\bm{o}}}

\def\vq{{\bm{q}}}
\def\vr{{\bm{r}}}
\def\vs{{\bm{s}}}

\def\vu{{\bm{u}}}
\def\vv{{\bm{v}}}
\def\vw{{\bm{w}}}
\def\vx{{\bm{x}}}
\def\vy{{\bm{y}}}
\def\vz{{\bm{z}}}

\def\evalpha{{\alpha}}
\def\evbeta{{\beta}}

\def\evlambda{{\lambda}}

\def\evmu{{\mu}}

\def\evsigma{{\sigma}}
\def\evtheta{{\theta}}
\def\evgamma{{\gamma}}
\def\evrho{{\rho}}

\def\evpi{{\pi}}
\def\eva{{a}}
\def\evb{{b}}
\def\evc{{c}}

\def\evg{{g}}

\def\evk{{k}}

\def\evm{{m}}

\def\evp{{p}}
\def\evq{{q}}

\def\evs{{s}}

\def\evu{{u}}
\def\evv{{v}}
\def\evw{{w}}

\def\mZero{{\bm{0}}}
\def\mA{{\bm{A}}}

\def\mE{{\bm{E}}}

\def\mG{{\bm{G}}}

\def\mK{{\bm{K}}}
\def\mL{{\bm{L}}}
\def\mM{{\bm{M}}}

\def\mO{{\bm{O}}}
\def\mP{{\bm{P}}}
\def\mQ{{\bm{Q}}}

\def\mS{{\bm{S}}}
\def\mT{{\bm{T}}}
\def\mU{{\bm{U}}}
\def\mV{{\bm{V}}}
\def\mW{{\bm{W}}}
\def\mX{{\bm{X}}}
\def\mY{{\bm{Y}}}
\def\mZ{{\bm{Z}}}

\def\mPhi{{\bm{\Phi}}}
\def\mLambda{{\bm{\Lambda}}}

\def\mAlpha{{\bm{\alpha}}}

\DeclareMathAlphabet{\mathsfit}{\encodingdefault}{\sfdefault}{m}{sl}
\SetMathAlphabet{\mathsfit}{bold}{\encodingdefault}{\sfdefault}{bx}{n}

\def\gH{{\mathcal{H}}}
\def\gI{{\mathcal{I}}}
\def\gJ{{\mathcal{J}}}

\def\gL{{\mathcal{L}}}

\def\gN{{\mathcal{N}}}
\def\gO{{\mathcal{O}}}

\def\gR{{\mathcal{R}}}

\def\gT{{\mathcal{T}}}

\def\gY{{\mathcal{Y}}}

\def\emLambda{{\Lambda}}

\def\emM{{M}}

\newcommand{\E}{\mathbb{E}}

\newcommand{\R}[1]{\mathbb{R}^{#1}}   
\newcommand{\Rvec}[1]{\mathbb{R}^{1\times{#1}}}   
\newcommand{\bS}[1]{\mathbb{S}^{#1}}  
\newcommand{\I}{\mathbf{I}}        

\newcommand{\sigmoid}{\sigma}
\newcommand{\softplus}{\mathrm{softplus}}

\DeclareMathOperator*{\argmin}{arg\,min}

\DeclareMathOperator{\sign}{sign}
\DeclareMathOperator{\Tr}{Tr}

\DeclareMathOperator{\zipf}{Zipf}
\DeclareMathOperator{\Diag}{Diag}    
\DeclareMathOperator{\stril}{stril}  
\DeclareMathOperator{\tril}{tril}    
\DeclareMathOperator{\triu}{triu}
\newcommand{\rms}{\mathrm{RMSNorm}}
\newcommand{\silu}{\mathrm{SiLU}}
\newcommand{\gelu}{\mathrm{GELU}}
\newcommand{\ffn}{\mathrm{FFN}}
\newcommand{\conv}{\mathrm{Conv}}

\newcommand{\inner}[2]{\langle#1,#2\rangle}
\newcommand{\inv}{^{-1}}
\newcommand{\tr}{^{\top}}
\newcommand{\norm}[1]{\|#1\|}
\newcommand{\abs}[1]{|#1|}

\newcommand{\D}{\mathrm{D}}  

\newcommand{\defas}{\colonequals}

\newcommand{\sdiv}{\mkern-0.5mu\smash{\raisebox{0.55pt}{\scalebox{0.78}{$\scriptstyle/$}}}\mkern-0.5mu}
\newcommand{\suppar}[1]{\smash{\raisebox{0.48pt}{\scalebox{0.80}{$\scriptstyle($}}}#1\smash{\raisebox{0.48pt}{\scalebox{0.80}{$\scriptstyle)$}}}}

%% file: macros.tex
\makeatletter
\g@addto@macro\normalsize{%
  \setlength{\abovedisplayskip}{3pt plus 1pt minus 1pt}%
  \setlength{\belowdisplayskip}{3pt plus 1pt minus 1pt}%
  \setlength{\abovedisplayshortskip}{0pt plus 1.5pt minus 0.5pt}%
  \setlength{\belowdisplayshortskip}{2.5pt plus 1pt minus 1pt}%
}
\normalsize
\makeatother
\AtBeginDocument{\normalsize}
\allowdisplaybreaks[2] 

\titlespacing*{\section}{0pt}{1.4ex plus 0.4ex minus 0.2ex}{0.6ex plus 0.2ex}
\titlespacing*{\subsection}{0pt}{1.1ex plus 0.3ex minus 0.2ex}{0.4ex plus 0.1ex}
\titlespacing*{\subsubsection}{0pt}{0.8ex plus 0.2ex minus 0.1ex}{0.3ex plus 0.1ex}
\titlespacing*{\paragraph}{0pt}{0.4ex plus 0.2ex minus 0.1ex}{0.6em}

\usepackage{etoolbox}
\makeatletter
\patchcmd{\ttl@straight@ii}{\clubpenalty\@M}{\clubpenalty\z@}{}{} 
\patchcmd{\@afterheading}{\clubpenalty \@M}{\clubpenalty\z@}{}{}  
\makeatother

\allowdisplaybreaks[4]       

\AtBeginEnvironment{appendices}{
    \crefalias{section}{appendix}
    \crefalias{subsection}{appendix}
}

\newcommand{\tbf}[1]{{\fontseries{b}\selectfont #1}}
\newcommand{\accn}{acc\kern0.04em\rule[-0.15ex]{0.28em}{0.4pt}\kern0.03em n}
\newcommand{\multirowrotate}[3]{\multirow{#1}{*}{\rotatebox[origin=c]{#2}{#3}}}

\newcommand{\tf}{Transformer\xspace}
\newcommand{\gdntwoxp}{Gated DeltaNet-2\xspace}  
\newcommand{\gdntwo}{GDN-2\xspace}
\newcommand{\oursxp}{Muir\xspace}
\newcommand{\ours}{Muir\xspace}

\algrenewcommand\algorithmicrequire{\textbf{Input:}}
\algrenewcommand\algorithmicensure{\textbf{Output:}}

\newcommand{\pycode}[1]{{\fontfamily{lmtt}\selectfont #1}}
\setminted[python]{
    fontfamily=lmtt,
    fontsize=\small,
    baselinestretch=0.98,
    linenos=true,
    numbersep=9pt,
    xleftmargin=16pt,
    autogobble=false,
    stripnl=false,
}

\newcommand{\listingcaption}[2]{%
    \needspace{6\baselineskip}%
    \begingroup\captionsetup{hypcap=false,aboveskip=0pt,belowskip=0pt}%
    \captionof{listing}{#1}\label{#2}\endgroup\nopagebreak\vspace{-4pt}%
}
\crefname{listing}{Listing}{Listings}
\Crefname{listing}{Listing}{Listings}

\newcommand{\cut}[1]{}
\newcommand{\xhdr}[1]{\paragraph{#1.}}
\makeatletter
\newcommand{\xxhdr}[1]{%
  \leavevmode\uline{\textit{#1}}\@addpunct{.}%
  \nobreak\hspace{1em}
  \ignorespaces%
}
\makeatother

\renewcommand{\Figref}[2]{Figure~\hyperref[#1]{\ref*{#1}#2}}

\makeatletter
\let\oldr@@t\r@@t
\def\r@@t#1#2{%
\setbox0=\hbox{$\oldr@@t#1{#2\,}$}\dimen0=\ht0
\advance\dimen0-0.2\ht0
\setbox2=\hbox{\vrule height\ht0 depth -\dimen0}%
{\box0\lower0.4pt\box2}}
\LetLtxMacro{\oldsqrt}{\sqrt}
\renewcommand*{\sqrt}[2][\ ]{\oldsqrt[#1]{#2}}
\makeatother

\theoremstyle{plain}
\newtheorem{theorem}{Theorem}[section]
\newtheorem{proposition}[theorem]{Proposition}
\newtheorem{lemma}[theorem]{Lemma}
\newtheorem{corollary}[theorem]{Corollary}
\theoremstyle{definition}

\theoremstyle{remark}
\newtheorem{remark}[theorem]{Remark}

\AddToHook{env/proposition/begin}{\crefalias{theorem}{proposition}}
\AddToHook{env/lemma/begin}{\crefalias{theorem}{lemma}}
\AddToHook{env/corollary/begin}{\crefalias{theorem}{corollary}}
\AddToHook{env/definition/begin}{\crefalias{theorem}{definition}}
\AddToHook{env/objective/begin}{\crefalias{theorem}{objective}}
\AddToHook{env/assumption/begin}{\crefalias{theorem}{assumption}}
\AddToHook{env/remark/begin}{\crefalias{theorem}{remark}}
\crefname{objective}{Objective}{Objectives}
\Crefname{objective}{Objective}{Objectives}
\crefname{assumption}{Assumption}{Assumptions}
\Crefname{assumption}{Assumption}{Assumptions}

%% file: sections/0_abstract.tex
Linear attention compresses context into a fixed-size key-value state for linear-time training and constant-memory inference.
Gated delta-rule models decay this state and erase its prediction for an incoming key before writing a new value; \gdntwoxp (\gdntwo) generalizes this with channel-wise decay and decoupled, full-rank erase and write gates.
However, we show that state updates are far less sensitive to erasure than to writes: key activation and unit normalization concentrate key energy onto a sparse subset of channels, whereas value writes scale all channels, so the erase gate can be parameterized at lower rank.
Furthermore, slow-decaying state rows accumulate much larger norms than fast rows, dominating unweighted readout and forcing stronger write-time erasure to suppress interference.
Guided by this analysis, we introduce \emph{\oursxp}, which fuses all gates into a sensitivity-allocated low-rank projection, re-weights state rows with a matched-filter readout gate, and reinvests the saved parameters in a wider feed-forward network.
Because standard chunkwise algorithms become numerically unstable at the large chunk sizes needed to saturate wide matrix units, we design a parallel update for stable, efficient training.
Across $166\text{M}\text{--}1.34\text{B}$ parameters, \ours lowers perplexity by $1.8\text{--}2.0\%$ over \gdntwo and $2.5\text{--}3.6\%$ over \tf, raises commonsense accuracy by up to $1.9$ points, and improves extractive recall over \gdntwo.
At $1.34\text{B},$ \ours outperforms both baselines on multi-hop and multi-needle recall, extrapolates $2{,}560\times$ from its $4\text{K}$ training length to $10\text{M}$-token streams without perplexity drift, and maintains near-$100\%$ passkey retrieval to $1\text{M}$ tokens with static readout.
\ours achieves these gains while matching \gdntwo throughput, reaching its $100\text{B}$-token validation loss $1.2\times$ faster, and training $2.7\times$ faster than \tf at $32\text{K}.$\looseness=-1


%% file: sections/1_intro.tex
\section{Introduction}
\label{sec:intro}

As large language models scale to long-document comprehension, multi-hop reasoning, and in-context retrieval, the \tf architecture \citep{Vaswani2017-el} faces a fundamental scaling bottleneck: softmax self-attention requires training compute that grows quadratically with sequence length and an inference cache that grows linearly.
Linear attention removes this bottleneck by compressing variable-length context into a fixed-size key-value state, enabling linear-time training and constant-memory inference \citep{Katharopoulos2020-km,Choromanski2020-tg,Gu2021-lj,Orvieto2023-hc,Sun2023-af,Gu2023-wx,Yang2023-pq,Dao2024-cx,De2024-hf}.
However, because a finite-dimensional state admits only finitely many orthogonal key directions, packing more keys into the state forces cross-key interference \citep{Johnson1984-va,Smolensky1990-kt,Arora2023-tw,Jelassi2024-ns,Arora2024-lm}.
Delta-rule recurrences \citep{Widrow1960-rx,Schlag2021-pg,Yang2024-ja,Grazzi2024-rh,Siems2025-fj} mitigate this interference by erasing the current prediction for an incoming key before writing a new value.
In particular, Gated DeltaNet \citep{Yang2025-dr} pairs scalar exponential decay with delta-rule erasure, Kimi Linear \citep{Kimi-Team2025-rg} introduces channel-wise decay so state rows evolve at independent rates, and \gdntwoxp (\gdntwo; \citealp{Hatamizadeh2026-cw}) generalizes both by decoupling erasure and writes into channel-wise gates so each state row and value channel updates independently.

\begin{figure}[!t]
    \centering
    \includegraphics[width=\textwidth]{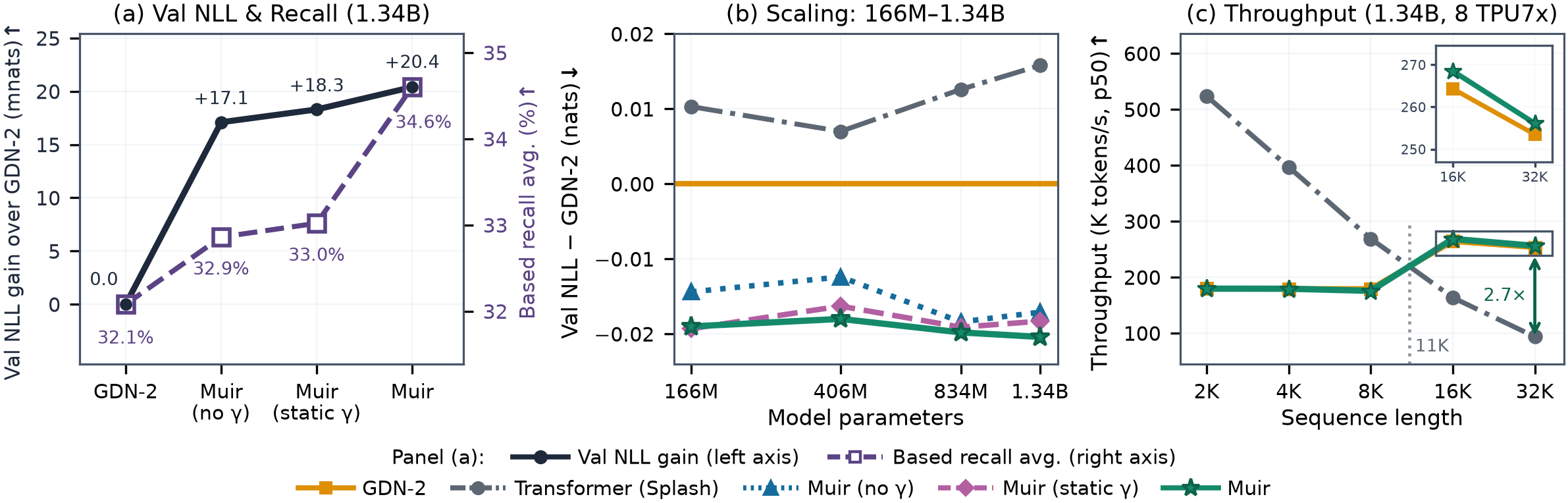}
    \caption{%
        \textbf{\ours outperforms \gdntwo and \tf on language modeling and long-context throughput while improving recall over \gdntwo.}
        (a) Sensitivity-allocated low-rank gates with FFN expansion drive the bulk of the validation NLL gain over \gdntwo, while dynamic matched-filter readout drives the gain on extractive recall.
        (b) Across $166\text{M}\text{--}1.34\text{B}$ parameters, \ours lowers validation NLL by $18.0\text{--}20.4\text{ millinats}$ over \gdntwo and $25.0\text{--}36.2\text{ millinats}$ over \tf.
        (c) On $8$ TPU7x chips ($0.5\text{M}$ tokens/step) at $1.34\text{B}$ with our chunkwise kernel, \ours matches \gdntwo throughput despite an additional readout gate and a $30.4\%$ wider FFN, and trains $1.64\times$ and $2.74\times$ faster than \tf at $16\text{K}$ and $32\text{K},$ respectively.\looseness=-1
    }
    \label{fig:teaser}
\end{figure}

Parameterizing decay, erasure, and writes with channel-wise gates, however, introduces three bottlenecks across parameter allocation, state readout, and parallel training.
%
First, when decoupling erasure and writes per channel, \gdntwo treats both gates symmetrically, allocating full-rank $d \times d$ projections to each while compressing the decay and output gates through low-rank bottlenecks; these two full-rank matrices alone consume $2d^2$ parameters (one-sixth of the $12d^2$ layer budget), so the gates together shrink the feed-forward network (FFN) width by over $25\%$ to remain iso-parameter with a \tf.
Second, because state rows decay at different rates, slow-decaying rows accumulate value writes over hundreds to thousands of tokens while fast-decaying rows retain only recent inputs, so state row norms span more than an order of magnitude.
Under unweighted query readout, high-norm slow rows dominate the output and force stronger write-time erasure to suppress their interference, which prematurely overwrites stored associations.
Third, because channels decay at independent rates, chunkwise parallel training \citep{Sun2023-af,Yang2023-pq,Yang2024-ja,Yang2025-dr} must divide keys by cumulative decay prefixes to compute within-chunk interactions via matrix multiplication and solve a lower-triangular system per chunk, which breaks down on wide-matrix-unit accelerators.
%
While GPU kernels \citep{Yang2024-vn,Kimi-Team2025-rg} avoid full-chunk division and solve diagonal triangular sub-blocks sequentially in SRAM at small chunks matched to $16 \times 16$ TensorCores, accelerators that pair wide systolic matrix units with a much slower vector unit (such as AWS Trainium and Google TPUs, reaching $256 \times 256$ matrix units with vector units over $60\times$ slower on modern TPUs; \citealp{Austin2025-yc,Jiang2026-cm}) require large chunks for high matrix utilization.
At these chunk sizes, cumulative decay prefixes underflow on fast-decaying channels and overflow the division; alternatively, materializing pairwise channel decays or solving the triangular system sequentially falls back to the slow vector unit.

To address these challenges, we analyze channel-wise gated delta recurrences and show that state updates are far less sensitive to erasure than to writes: rectifying key activations (such as $\silu$; \citealp{Ramachandran2018-gn}) and unit normalization concentrate key energy onto a sparse subset of channels, whereas value writes scale all channels, so the erase gate requires a much lower-rank projection than the write gate.
We further derive a closed-form scaling law for state row norms and show that weighting queries by a matched-filter readout gain balances row contributions, which lowers the write-time erasure needed to suppress interference and extends the memory horizon.
%
Guided by this analysis, we introduce \emph{\oursxp}, a gated delta-rule model that fuses all recurrence and output gates into a sensitivity-allocated low-rank projection, modulates queries with a matched-filter readout gate (combining a static channel profile with dynamic per-token modulation) to re-weight state rows, and reinvests the saved gate parameters under a fixed budget to widen the FFN and expand its capacity for factual associations \citep{Geva2021-vm,Meng2022-qp}.
%
Finally, to train stably at the large chunk sizes needed to saturate wide matrix units, we design a boundary-pivoted chunkwise kernel that splits relative decays without division or full-chunk 3D decay tensors and solves within-chunk triangular systems via block matrix multiplications.

\Cref{fig:teaser} summarizes our main findings.
%
Across model scales from $166\text{M}$ to $1.34\text{B}$ parameters, \ours outperforms both \gdntwo and \tf on language modeling and commonsense reasoning, and improves over \gdntwo on real-world extractive recall \citep{Arora2024-lm,Arora2024-at}.
%
At $1.34\text{B},$ \ours improves synthetic multi-hop and multi-needle retrieval \citep{Hsieh2024-ji} over both baselines, and extrapolates from $4\text{K}$ to $10\text{M}$-token streams in a $25\text{ MB}$ \texttt{float32} state without perplexity drift, whereas Dynamic YaRN \tf \citep{Peng2024-tz} diverges past $8\text{K}$ and needs a $2.1\text{ TB}$ \texttt{bfloat16} KV cache at $10\text{M}.$
On passkey retrieval, our static matched-filter readout achieves near-$100\%$ accuracy to $1\text{M}$ tokens, whereas \gdntwo collapses past $4\text{K}\text{--}8\text{K}.$
Using our chunkwise kernel for both recurrent models and Splash Attention (TPU implementation of FlashAttention; \citealp{Dao2022-vw,Dao2023-ln}) for \tf, \ours matches \gdntwo training throughput across sequence lengths despite a wider FFN and an additional gate, reaches the $100\text{B}$-token validation loss of \gdntwo $1.2\times$ faster, and trains $2.7\times$ faster than \tf at $32\text{K}.$
%
Mechanistic analysis confirms that state updates are more sensitive to writes than to erasure; state row norms closely follow our scaling law; and the learned readout balances state rows, lowers write-time erasure, and retains associations over longer contexts.

%% file: sections/2_prelim.tex
\section{Preliminaries}
\label{sec:prelim}

\xhdr{Notation}

Throughout the paper, scalars and scalar entries of vectors or matrices are denoted by lightface letters ($m, T, \evv_j, \emM_{i,j}$), vectors by boldface lowercase letters ($\vv\in\Rvec{m}$), and matrices by boldface uppercase letters ($\mM\in\R{m\times n}$).
All vectors are row vectors, so the Euclidean inner product of $\vu,\vv\in\Rvec{m}$ is $\inner{\vu}{\vv}\defas\vu\vv\tr$ and the rank-one outer product of $\vu\in\Rvec{m}$ and $\vv\in\Rvec{n}$ is $\vu\tr\vv\in\R{m\times n}.$
Additionally, for vectors $\vx,\vy\in\Rvec{m},$ $\vx\odot\vy,$ $\vx\oslash\vy,$ and $\vx^{\odot p}$ denote elementwise multiplication, division, and $p$-th power, respectively, and $\Diag(\vx)\in\R{m\times m}$ is the diagonal matrix with diagonal $\vx.$
%
For a matrix $\mM\in\R{m\times n},$ $\mM_{i,:}$ and $\mM_{:,j}$ denote its $i$-th row and $j$-th column, and $\tril(\mM),$ $\stril(\mM),$ and $\triu(\mM)$ denote its lower-triangular, strictly lower-triangular, and upper-triangular parts.
We denote the Kronecker delta by $\delta_{ij}$ ($\delta_{ij} = 1$ if $i = j$ and $0$ otherwise).
For a differentiable map $f,$ $\D_{\vx}f(\vx)[\delta\vx]\defas\lim_{\eps\to 0}\eps\inv(f(\vx+\eps\,\delta\vx)-f(\vx))$ denotes the directional derivative at $\vx$ along perturbation $\delta\vx.$
Finally, a hat ($\hat{\cdot}$) denotes gate-modulated quantities (e.g., $\hat\vk\defas\vb\odot\vk$), a bar ($\bar{\cdot}$) denotes expectations or empirical averages, and an asterisk (${}^\ast$) denotes stationary fixed points or optimal solutions.
(See \Cref{app:notation} for a summary of notation.)


\subsection{Channel-Wise Gated Delta Rule}
\label{sec:prelim:gdn2}

Given an input sequence $\vx_1, \dotsc, \vx_T \in \Rvec{d},$ linear-time sequence models avoid the $\gO(T^2)$ training cost and $\gO(T)$ key-value cache of softmax attention by compressing the token history into a fixed-size associative memory $\mS_t \in \R{d_k \times d_v}$ per head across $N_h$ heads of dimension $d_k = d_v = d/N_h$ (typically $d_k = d_v = 128$).
Following standard practice in modern linear attention models, where local feature mixing before state updates is critical for in-context recall \citep{Gu2023-wx,Arora2023-tw,Yang2024-ja,Allen-Zhu2025-zv}, each token $\vx_t$ is projected by $\mW_q, \mW_k, \mW_v \in \R{d \times d},$ passed through a depthwise causal convolution and $\silu$ activation, and partitioned into $N_h$ heads.
%
Within each head, queries and keys are $\ell_2$-normalized to the unit sphere ($\vq_t, \vk_t \in \bS{d_k-1},$ so $\norm{\vq_t}_2 = \norm{\vk_t}_2 = 1$; \citealp{Schlag2021-pg,Yang2024-ja}) for bounded addressing and erasure stability, whereas values $\vv_t \in \Rvec{d_v}$ enter additively and remain unnormalized to preserve feature magnitude.
Standard linear attention accumulates outer products additively ($\mS_t = \mS_{t-1} + \vk_t\tr\vv_t$; \citealp{Katharopoulos2020-km}) or with exponential decay ($\mS_t = \evalpha_t\mS_{t-1} + \vk_t\tr\vv_t$ for scalar $\evalpha_t$ \citep{Sun2023-af,Dao2024-cx}; or $\Diag(\valpha_t)\mS_{t-1} + \vk_t\tr\vv_t$ for channel-wise $\valpha_t$ \citep{Yang2023-pq}).
By contrast, the delta rule \citep{Widrow1960-rx,Schlag2021-pg,Yang2024-ja} updates $\mS_t$ via one step of online gradient descent on the instantaneous key-value reconstruction loss $\frac{1}{2}\norm{\vk_t\mS - \vv_t}_2^2,$ querying the state along $\vk_t$ and erasing its current prediction before writing the new value (see \Cref{sec:related} for additional related work).


\xhdr{\gdntwoxp}

\gdntwo \citep{Hatamizadeh2026-cw} extends scalar-decay Gated DeltaNet \citep{Yang2025-dr} by promoting the per-head decay factor and erase step size to channel-wise vectors alongside a channel-wise write gate (\Cref{app:optim} derives this update via a left preconditioner $\Diag(\vb_t)\inv$ on the gated reconstruction loss).
%
Specifically, at timestep $t,$ each head computes four channel-wise gates from $\vx_t$: a decay (retention) gate $\valpha_t \in (0, 1)^{1\times d_k},$ an erase gate $\vb_t \in (0, 2)^{1\times d_k},$ a write gate $\vw_t \in (0, 1)^{1\times d_v},$ and a post-recurrence output gate $\vg_t \in \Rvec{d_v}.$
Given pre-activation logits $\vz^f_t, \vz^b_t \in \Rvec{d_k}$ and $\vz^w_t \in \Rvec{d_v},$ the recurrence gates are parameterized as $\log\valpha_t = -\exp(a_{\log})\,\softplus(\vz^f_t),$ $\vb_t = 2\sigmoid(\vz^b_t),$ and $\vw_t = \sigmoid(\vz^w_t),$ where $\vz^f_t$ includes a bias $\vc_f \in \Rvec{d_k}$ and $a_{\log} \in \mathbb{R}$ is a per-head log-rate.

With the decayed prior state $\hat\mS_t \defas \Diag(\valpha_t)\mS_{t-1}$ (whose $j$-th row $\hat\mS_{t,j,:} = \evalpha_{t,j}\mS_{t-1,j,:}$ scales key channel $j \in \{1, \dotsc, d_k\}$), the erase-gated key $\hat\vk_t \defas \vb_t \odot \vk_t,$ and the write-gated value $\hat\vv_t \defas \vw_t \odot \vv_t,$ the \gdntwo state update and readout are given by
\begin{equation}
\label{eq:gdn2}
    \mS_t = \big(\I - \vk_t\tr(\vb_t\odot\vk_t)\big)\Diag(\valpha_t)\mS_{t-1} + \vk_t\tr(\vw_t\odot\vv_t)
    = \hat\mS_t - \vk_t\tr(\hat\vk_t\hat\mS_t) + \vk_t\tr\hat\vv_t,\qquad
    \vo_t = \vq_t\mS_t.
\end{equation}
Here, $\ell_2$-normalization of $\vk_t$ ensures that $\inner{\vk_t}{\vb_t \odot \vk_t} \in (0, 2)$ for any $\vb_t \in (0, 2)^{1\times d_k},$ so the erasure factor $\I - \vk_t\tr(\vb_t \odot \vk_t)$ \emph{contracts} the state along $\vk_t$ with eigenvalue $1 - \inner{\vk_t}{\vb_t \odot \vk_t} \in (-1, 1)$ \citep{Grazzi2024-rh,Hatamizadeh2026-cw}.
The head outputs $\vo_t$ are normalized via per-head $\rms,$ modulated elementwise by $\silu(\vg_t)$ \citep{Sun2023-af,Yang2023-pq,Qiu2025-vl}, concatenated across the $N_h$ heads, and projected back to the residual stream by $\mW_o \in \R{d \times d}.$


\xhdr{Gate Parameterization and Parameter Budget}

A language model stacks $L$ residual layers, each pairing the sequence-mixing block above with a feed-forward network $\ffn(\vy) = \gelu(\vy\mW_{\mathrm{in}})\mW_{\mathrm{out}}$ \citep{Hendrycks2016-oq} of intermediate width $d_{\mathrm{ffn}}$ ($\mW_{\mathrm{in}} \in \R{d \times d_{\mathrm{ffn}}},$ $\mW_{\mathrm{out}} \in \R{d_{\mathrm{ffn}} \times d}$).
A standard \tf layer with projections $\mW_q, \mW_k, \mW_v, \mW_o \in \R{d \times d}$ ($4d^2$ parameters) and FFN width $d_{\mathrm{ffn}} = 4d$ ($2d\,d_{\mathrm{ffn}} = 8d^2$ parameters) contains $12d^2$ non-embedding parameters.
%
Across all $N_h$ heads, \gdntwo parameterizes the decay logit $\vz^f_t$ and output gate $\vg_t$ with rank-$r_0$ bottleneck projections ($\mW_{f\downarrow}, \mW_{g\downarrow} \in \R{d \times r_0}$ and $\mW_{f\uparrow}, \mW_{g\uparrow} \in \R{r_0 \times d}$ with $r_0 = 128,$ totaling $4d r_0 = 512d$ parameters), whereas the erase logit $\vz^b_t = \vx_t\mW_b$ and write logit $\vz^w_t = \vx_t\mW_w$ use unbiased full-rank (dense) matrices $\mW_b, \mW_w \in \R{d \times d}$ totaling $2d^2$ parameters (one-sixth of the $12d^2$ layer budget).
Following standard practice of adjusting $d_{\mathrm{ffn}}$ to maintain a $12d^2$ per-layer parameter budget for iso-parameter comparisons \citep{Shazeer2020-lf,Sun2023-af,Qiu2025-vl,Hatamizadeh2026-cw}, \gdntwo sets $d_{\mathrm{ffn}} = 3d - 2r_0 = 3d - 256$ (shrinking $d_{\mathrm{ffn}}$ from $8{,}192$ to $5{,}888$ at $d = 2{,}048$).


\subsection{Chunkwise Parallel Form}
\label{sec:prelim:chunk}

Evaluating \eqref{eq:gdn2} sequentially over $T$ timesteps performs $\gO(d_k d_v)$ vector operations per token per head, which has low arithmetic intensity and is memory-bandwidth-bound on modern hardware.
Defining the \emph{effective write} $\tilde\vv_t \defas \hat\vv_t - \hat\vk_t\hat\mS_t = \hat\vv_t - \hat\vk_t\Diag(\valpha_t)\mS_{t-1} \in \Rvec{d_v}$ (gated memory write minus gated memory read) rewrites \eqref{eq:gdn2} in the Gated Linear Attention (GLA; \citealp{Yang2023-pq}) form:
\[
    \mS_t = \big(\I - \vk_t\tr\hat\vk_t\big)\Diag(\valpha_t)\mS_{t-1} + \vk_t\tr\hat\vv_t
    = \Diag(\valpha_t)\mS_{t-1} + \vk_t\tr\big(\hat\vv_t - \hat\vk_t\Diag(\valpha_t)\mS_{t-1}\big)
    = \Diag(\valpha_t)\mS_{t-1} + \vk_t\tr\tilde\vv_t.
\]
To expose matrix-matrix multiplications during training, chunkwise-parallel algorithms \citep{Sun2023-af,Yang2023-pq,Yang2024-ja,Yang2025-dr} partition the sequence into $N_c = T/C$ chunks of length $C.$
Consider a single chunk indexed by $t \in \{1, \dotsc, C\}$ starting from chunk-initial state $\mS_0 \in \R{d_k \times d_v}.$
Let $\vpi_t \defas \prod_{r=1}^t \valpha_r$ and $\vpi_{t,s} \defas \prod_{r=s+1}^t \valpha_r \in (0, 1]^{1\times d_k}$ (with $\vpi_{t,t} = \vone$) denote the elementwise cumulative decay prefixes and relative decays; unrolling $\mS_{t-1}$ from $\mS_0$ in $\tilde\vv_t = \hat\vv_t - \hat\vk_t\Diag(\valpha_t)\mS_{t-1}$ and using $\hat\vk_t\Diag(\vpi_{t,s})\vk_s\tr = (\hat\vk_t \odot \vpi_{t,s})\vk_s\tr = \inner{\hat\vk_t}{\vpi_{t,s} \odot \vk_s}$ gives
\[
    \tilde\vv_t = \hat\vv_t - (\vpi_t \odot \hat\vk_t)\mS_0 - \sum_{s=1}^{t-1} \inner{\hat\vk_t}{\vpi_{t,s} \odot \vk_s}\tilde\vv_s, \qquad t \in \{1, \dotsc, C\}.
\]
Stacking $\tilde\vv_t,$ $\hat\vv_t,$ $\vpi_t \odot \hat\vk_t,$ and $\vk_t \oslash \vpi_t$ across $t \in \{1, \dotsc, C\}$ as the rows of $\tilde\mV, \hat\mV \in \R{C \times d_v}$ and $\hat\mK_{\odot\pi}, \mK_{\oslash\pi} \in \R{C \times d_k}$ casts this coupled recurrence in matrix form as
\[
    \tilde\mV = \hat\mV - \hat\mK_{\odot\pi}\mS_0 - \stril\!\big(\hat\mK_{\odot\pi}\mK_{\oslash\pi}\tr\big)\tilde\mV
    \quad\Longleftrightarrow\quad
    \mT\tilde\mV = \hat\mV - \hat\mK_{\odot\pi}\mS_0,
    \qquad
    \mT \defas \I + \mL_{\mathrm{strict}} \in \R{C \times C},
\]
where $\mL_{\mathrm{strict}}$ is strictly lower-triangular with entries $(\mL_{\mathrm{strict}})_{t,s} = \inner{\hat\vk_t}{\vpi_{t,s} \odot \vk_s}$ ($t > s$).
Solving $\mT\begin{bmatrix}\mU & \mY\end{bmatrix} = \begin{bmatrix}\hat\mV & \hat\mK_{\odot\pi}\end{bmatrix}$ yields $\tilde\mV = \mT\inv\hat\mV - (\mT\inv\hat\mK_{\odot\pi})\mS_0 = \mU - \mY\mS_0.$
Once $\tilde\mV$ is known, unrolling the GLA recurrence $\mS_t = \Diag(\valpha_t)\mS_{t-1} + \vk_t\tr\tilde\vv_t$ across the chunk \citep{Yang2023-pq} gives the end-of-chunk state $\mS_C \in \R{d_k \times d_v}$ (passed as $\mS_0$ to the next chunk) and chunk outputs $\mO \in \R{C \times d_v}$ via matrix multiplications:
\[
    \mS_C = \Diag(\vpi_C)\mS_0 + \tilde\mK\tr\tilde\mV,
    \qquad
    \mO = \mQ_{\odot\pi}\mS_0 + \mM\tilde\mV,
\]
where $\tilde\mK, \mQ_{\odot\pi} \in \R{C \times d_k}$ have rows $\vpi_{C,s} \odot \vk_s$ and $\vpi_t \odot \vq_t,$ and $\mM \in \R{C \times C}$ is lower-triangular with entries $\emM_{t,s} = \inner{\vq_t}{\vpi_{t,s} \odot \vk_s}$ for $t \ge s.$

In Gated DeltaNet \citep{Yang2025-dr}, where $\valpha_t = \evalpha_t\vone$ and $\vb_t = \evbeta_t\vone,$ scalar multiplication commutes with rank-one erasure: $\evalpha_t(\I - \evbeta_t\vk_t\tr\vk_t) = (\I - \evbeta_t\vk_t\tr\vk_t)\evalpha_t,$ so the scalar decay ratio $\evpi_t/\evpi_s$ factors out of the inner product $\inner{\hat\vk_t}{\vk_s}.$
Under channel-wise decay $\valpha_t$ in \eqref{eq:gdn2}, however, the diagonal decay operator $\Diag(\valpha_t)$ and the rank-one erasure operator $\I - \vk_t\tr\hat\vk_t$ do not commute whenever $\valpha_t$ varies across channels:
\[
    \big(\I - \vk_t\tr\hat\vk_t\big)\Diag(\valpha_t) \neq \Diag(\valpha_t)\big(\I - \vk_t\tr\hat\vk_t\big).
\]
This non-commutativity introduces two obstacles to executing the chunkwise solve on wide-matrix-unit accelerators.
First, forming $\mT$ and $\mM$ faces a tradeoff between efficiency and numerical stability: factoring $\mL_{\mathrm{strict}} = \stril(\hat\mK_{\odot\pi}\mK_{\oslash\pi}\tr)$ and $\mM = \tril(\mQ_{\odot\pi}\mK_{\oslash\pi}\tr)$ as matrix multiplications divides keys by $\vpi_s,$ which underflows \texttt{bfloat16} and \texttt{float32} (minimum positive normal $\approx 1.18 \times 10^{-38}$) over long chunks (e.g., $\evpi_{s,j} \le 2^{-256} \approx 10^{-77}$ at $C = 256$ whenever $\evalpha_{r,j} \le 0.5$) and overflows $\mK_{\oslash\pi}.$
Conversely, evaluating $\vpi_{t,s} = \exp(\mLambda_t - \mLambda_s)$ from the cumulative log-decay matrix $\mLambda \in \R{C \times d_k}$ (with rows $\mLambda_t \defas \sum_{r=1}^t \log\valpha_r \le \vzero$) is numerically stable for $t \ge s,$ but materializing the 3D tensor $\vpi_{t,s} \odot \vk_s \in \R{C \times C \times d_k}$ ($32\text{ MB}$ per head in \texttt{float32} at $C = 256$) and reducing over $d_k$ executes on the vector unit.
Second, once $\mT = \I + \mL_{\mathrm{strict}}$ is formed, solving $\mT\begin{bmatrix}\mU & \mY\end{bmatrix} = \begin{bmatrix}\hat\mV & \hat\mK_{\odot\pi}\end{bmatrix}$ via forward substitution is sequential across the $C$ steps of the chunk and also executes on the vector unit.\looseness=-1


\xhdr{Wide-Matrix-Unit Architecture and Tiling}

We study wide-matrix-unit execution on Google TPU7x Ironwood accelerators \citep{Jiang2026-cm}, where each chip comprises two TensorCores totaling $192\text{ GB}$ of High-Bandwidth Memory (HBM) at $7{,}380\text{ GB/s}$ and $128\text{ MB}$ of on-chip Vector Memory (VMEM).
Within each TensorCore, two Matrix Multiply Units (MXUs) operate on $256 \times 256$ systolic arrays (\texttt{bfloat16} inputs accumulated in \texttt{float32}, $2{,}307\text{ TFLOP/s}$ per chip), whereas the Vector Processing Unit (VPU, a $128\text{-lane} \times 8\text{-sublane}$ SIMD processor) has roughly $64\times$ lower \texttt{bfloat16} throughput ($\approx\!36\text{ TFLOP/s}$ in \texttt{bfloat16} and $\approx\!18\text{ TFLOP/s}$ in \texttt{float32} per chip; \citealp{Austin2025-yc}).
However, when either dimension of a matrix multiplication is smaller than $256,$ the MXU pads unused lanes with zeros, reducing MXU utilization proportionally.
Across HBM, VMEM, and Vector Registers (VREG), XLA tiles the last two dimensions of every tensor into 2D tiles of shape \texttt{T(8,\,128)} for $32$-bit words (\texttt{T(16,\,128)} in packed \texttt{bfloat16}), matching the $128$ VPU lanes; elementwise operations that preserve the trailing dimensions fuse in VREGs, whereas permutations that alter the tiled dimensions incur an HBM round-trip.

%% file: sections/3_theory_v2.tex
\section{Channel-Wise Gated Delta Memory Dynamics}
\label{sec:theory}

In this section, we analyze how the channel-wise gates $(\valpha_t, \vb_t, \vw_t)$ and unweighted readout $\vo_t = \vq_t\mS_t$ in the \gdntwo recurrence govern state updates and associative retrieval.
In \S\ref{sec:theory:theory}, we study this recurrence from single-step updates to long-horizon stationary dynamics: we first quantify the single-step sensitivity of $\mS_t$ to perturbations in $\vb_t$ and $\vw_t,$ next characterize how channel-wise decay and erasure determine the expected squared norms of the state rows over long sequences, and finally analyze how weighting state rows at readout affects retrieval error and interacts with write-time erasure.
We then test our theoretical predictions in controlled synthetic memory experiments in \S\ref{sec:theory:exp}.
(All proofs are in \Cref{app:theory}.)


\subsection{Gate Sensitivity and Readout Dynamics}
\label{sec:theory:theory}

In \gdntwo, both the erase and write gates are parameterized by dense projections, yet they enter the state update asymmetrically: $\vb_t$ modulates the $\silu$-activated, unit-norm key $\vk_t$ against the decayed prior state $\hat\mS_t,$ whereas $\vw_t$ scales the unnormalized value $\vv_t.$
We first formalize the resulting asymmetry in state sensitivity.


\begin{proposition}[State sensitivity to erase and write gates]
\label{prop1}
    Consider the \gdntwo state update \eqref{eq:gdn2} at step $t$ as a function $\mS_t(\vb_t,\vw_t)$ of the erase gate $\vb_t$ and write gate $\vw_t,$ with decayed prior state $\hat\mS_t=\Diag(\valpha_t)\mS_{t-1}.$
    \begin{enumerate}[label=(\roman*),leftmargin=2em,ref=\getrefnumber{prop1}(\roman*)]
        \item For any perturbations $\delta\vb$ and $\delta\vw,$ the directional derivatives of $\mS_t$ with respect to $\vb_t$ and $\vw_t$ are
        \begin{equation}
        \label{prop1:eq1}
            \D_{\vb}\mS_t[\delta\vb] = -\vk_t\tr\big((\delta\vb\odot\vk_t)\hat\mS_t\big),\qquad
            \D_{\vw}\mS_t[\delta\vw] = \vk_t\tr(\delta\vw\odot\vv_t).
        \end{equation}
        \item Conditioned on $(\vk_t,\vv_t,\hat\mS_t),$ the expected squared Frobenius sensitivities under isotropic perturbations $\delta\vb\sim\gN(\vzero,\I_{d_k})$ and $\delta\vw\sim\gN(\vzero,\I_{d_v})$ are
        \begin{equation}
        \label{prop1:eq2}
            \E_{\delta\vb}\!\left[\norm{\D_{\vb}\mS_t[\delta\vb]}_F^2\right] = \sum_{j=1}^{d_k} \evk_{t,j}^2\norm{\hat\mS_{t,j,:}}_2^2,\qquad
            \E_{\delta\vw}\!\left[\norm{\D_{\vw}\mS_t[\delta\vw]}_F^2\right] = \norm{\vv_t}_2^2.
        \end{equation}
        \item\label{prop1:item3} Decompose the gated key $\hat\vk_t=\vb_t\odot\vk_t$ into its parallel erase strength $\evmu_t\defas\inner{\vk_t}{\hat\vk_t}$ along $\vk_t$ and its orthogonal residual $\vs_t\defas\hat\vk_t-\evmu_t\vk_t$ ($\inner{\vs_t}{\vk_t}=0$), whose squared Euclidean norm equals the $\vk_t^{\odot 2}$-weighted within-head coordinate variance of $\vb_t$ given by
        \[
            \norm{\vs_t}_2^2 = \norm{\hat\vk_t}_2^2 - \evmu_t^2 = \sum_{j=1}^{d_k} \evk_{t,j}^2\big(\evb_{t,j}-\evmu_t\big)^2.
        \]
        Moreover, the gradient $\nabla_{\vb}\evmu_t=\vk_t^{\odot 2}=(\evk_{t,1}^2,\dotsc,\evk_{t,d_k}^2)$ has squared Euclidean norm
        \[
            \norm{\nabla_{\vb}\evmu_t}_2^2 = \norm{\vk_t}_4^4 = \frac{1}{\eta_{\mathrm{eff}}(\vk_t)}\in\left[\frac{1}{d_k},1\right],\quad\text{where}\quad
            \eta_{\mathrm{eff}}(\vk_t)\defas\frac{\big(\sum_{j=1}^{d_k}\evk_{t,j}^2\big)^2}{\sum_{j=1}^{d_k}\evk_{t,j}^4} = \frac{1}{\norm{\vk_t}_4^4}\in[1,d_k]
        \]
        is the per-token key participation ratio.
    \end{enumerate}
\end{proposition}



\begin{remark}[Origin of the write-erase sensitivity asymmetry]
\label{prop1:rem1}
    \Cref{prop1:eq1,prop1:eq2} show that the write gate $\vw_t$ and the erase gate $\vb_t$ influence the state update through different mechanisms.
    A write perturbation $\delta\vw$ scales the incoming value $\vv_t$ directly in $\vk_t\tr(\delta\vw\odot\vv_t)$; because $\norm{\vk_t}_2=1,$ $\vw_t$ modifies $\mS_t$ unconditionally at every step, even when $\hat\mS_t$ is empty.
    By contrast, an erase perturbation $\delta\vb$ is weighted across channels by both the squared key $\evk_{t,j}^2$ and the decayed squared row norm (row energy) $\norm{\hat\mS_{t,j,:}}_2^2,$ so a channel of $\vb_t$ affects $\mS_t$ only when two conditions hold simultaneously: (i) the corresponding memory row already stores non-negligible energy, and (ii) the current key $\vk_t$ places non-negligible weight on that channel.
    Because keys are constructed by applying a $\silu$ nonlinearity prior to $\ell_2$-normalization, negative pre-activations are suppressed and $\vk_t$ concentrates its unit norm onto $\eta_{\mathrm{eff}}(\vk_t)\ll d_k$ active channels; on the remaining $d_k-\eta_{\mathrm{eff}}(\vk_t)$ inactive channels, as well as on any empty or recently decayed state rows, perturbations to $\vb_t$ leave $\mS_t$ nearly unchanged.
\end{remark}


\begin{remark}[Regularization of oblique erase variance by low-rank projections]
\label{prop1:rem2}
    Proposition~\ref{prop1:item3} yields an exact decomposition of the erase transition $\I-\vk_t\tr\hat\vk_t = (\I-\evmu_t\vk_t\tr\vk_t)-\vk_t\tr\vs_t$ for every $\vk_t\in\bS{d_k-1}$ and $\vb_t\in(0,2)^{1\times d_k}.$
    All contraction along $\vk_t$ is governed by the scalar $\evmu_t=\inner{\vb_t}{\vk_t^{\odot 2}},$ whose gradient $\nabla_{\vb}\evmu_t=\vk_t^{\odot 2}$ is supported on the $\eta_{\mathrm{eff}}(\vk_t)\ll d_k$ active key channels.
    Consequently, any component of $\vb_t$ in the orthogonal complement $(\vk_t^{\odot 2})^\perp\subset\Rvec{d_k}$ contributes identically zero to $\evmu_t$ and enters the state update exclusively through the oblique residual $\vs_t.$
    Although the eigenvalues of $\I-\vk_t\tr\hat\vk_t$ remain in $(-1,1],$ any non-zero residual $\vs_t\neq\vzero$ contributes the non-normal off-diagonal term $-\vk_t\tr\vs_t,$ which strictly inflates the transition spectral norm above $1$ ($\norm{\I-\vk_t\tr\hat\vk_t}_2\ge\sqrt{1+\norm{\vs_t}_2^2}>1$; \Cref{lem1}).
    These properties motivate parameterizing the erase logit via a biased low-rank bottleneck $\vx_t\mW_{b\downarrow}\mW_{b\uparrow}+\vc_b$ ($r_b\ll d_k$) instead of an unconstrained dense projection $\vx_t\mW_b.$
    %
    Specifically, the channel bias $\vc_b$ sets distinct baseline erase rates across the $d_k$ rows (avoiding reduction to a uniform scalar gate $\evb_t\vone$), while the $r_b$ bottleneck directions modulate $\evmu_t$ across the $\eta_{\mathrm{eff}}(\vk_t)\ll d_k$ active key channels without injecting high-rank variations in $(\vk_t^{\odot 2})^\perp$ that would inflate $\norm{\vs_t}_2^2.$
\end{remark}


Having shown that single-step state updates are far less sensitive to $\vb_t$ than to $\vw_t,$ we now analyze how channel-wise decay and erasure shape the expected energies $\E[\norm{\mS_{t,j,:}}_2^2]$ of the state rows $\mS_{t,j,:}$ over long sequences.
For the rest of this section, let \eqref{eq:gdn2} be driven by a stationary i.i.d.\ stream $(\vk_t,\hat\vv_t,\valpha_t,\vb_t)$ whose components are mutually independent at each timestep $t$ (and hence independent of $\mS_{t-1}$).
Additionally, let $\E[\hat\vv_t]=\vzero$ and $\E[\norm{\hat\vv_t}_2^2]=\evsigma_v^2,$ keys $\vk_t$ have uncorrelated channels $\E[\evk_{t,i}\evk_{t,j}]=\delta_{ij}\evc_j$ with second moments $\evc_j>0$ ($\sum_{j=1}^{d_k}\evc_j=1$), and the gates have mean squared decay factor $\evlambda_j\defas\E[\evalpha_{t,j}^2]\in(0,1)$ and mean channel erase rate $\bar\evb_j\defas\E[\evb_{t,j}]\in(0,2)$ satisfying $\bar\evb_j\evc_j<1.$

\begin{proposition}[Stationary row-energy law]
\label{prop2}
    Let $\evm_{t,j}\defas\E[\norm{\mS_{t,j,:}}_2^2]$ denote the expected energy of state row $j\in\{1,\dotsc,d_k\}$ at timestep $t.$
    Under the stationary stream above, the uncorrelated-row condition $\E[\inner{\mS_{t-1,i,:}}{\mS_{t-1,j,:}}]=\delta_{ij}\evm_{t-1,j},$ and the second-order closures $\E[\evk_{t,i}^2\evk_{t,j}^2]\approx\evc_i\evc_j$ and $\E[\evb_{t,i}^2]\approx\bar\evb_i^2,$ the row energy $\evm_{t,j}$ of \eqref{eq:gdn2} evolves according to the linear recurrence
    \begin{equation}
    \label{prop2:eq1}
        \evm_{t,j} = \evlambda_j(1-2\bar\evb_j\evc_j)\,\evm_{t-1,j} + \evc_j\sum_{i=1}^{d_k}\bar\evb_i^2\evc_i\evlambda_i\,\evm_{t-1,i} + \evc_j\evsigma_v^2.
    \end{equation}
    %
    Because the cross-row sum $\sum_{i=1}^{d_k}\bar\evb_i^2\evc_i\evlambda_i\evm_{t-1,i}$ is a scalar shared across all rows $j$ (rescaling $\evsigma_v^2$ by a constant factor without altering relative row energies), dropping this cross-row coupling term yields the unique stationary row energy $\evm_j^\ast\defas\lim_{t\to\infty}\evm_{t,j}$ and stationary row root-mean-square (RMS) norm $\evrho_j\defas\sqrt{\smash{\evm_j^\ast}\vphantom{\evm}}$ in closed form:
    \begin{equation}
    \label{prop2:eq2}
        \evm_j^\ast = \evrho_j^2 = \frac{\evc_j\evsigma_v^2}{1-\evlambda_j+2\evlambda_j\bar\evb_j\evc_j} = \evc_j\evsigma_v^2\,\tau_{\mathrm{eff},j}(\bar\evb_j),\qquad
        \tau_{\mathrm{eff},j}(\bar\evb_j)\defas\frac{1}{1-\evlambda_j+2\evlambda_j\bar\evb_j\evc_j}.
    \end{equation}
\end{proposition}



\begin{remark}[Two-regime scaling of stationary row energy]
\label{prop2:rem1}
    \Eqref{prop2:eq2} shows that stationary row energy $\evm_j^\ast$ is the product of the write injection rate $\evc_j\evsigma_v^2$ and the effective memory horizon $\tau_{\mathrm{eff},j}(\bar\evb_j)=(1-\evlambda_j+2\evlambda_j\bar\evb_j\evc_j)\inv,$ whose denominator combines two competing dissipation mechanisms: exponential decay $1-\evlambda_j$ and key-triggered erasure $2\evlambda_j\bar\evb_j\evc_j.$
    This yields two distinct scaling regimes across channels.
    In the \emph{erasure-dominated limit} ($1-\evlambda_j\ll 2\evlambda_j\bar\evb_j\evc_j,$ i.e., $\evlambda_j\to 1^-$), the key-energy factor $\evc_j$ cancels between numerator and denominator to give $\evm_j^\ast\to\evsigma_v^2/(2\bar\evb_j),$ so key-triggered erasure self-equalizes row energy only at the cost of heavy erasure $\bar\evb_j$ that over-erases slow channels and caps their memory horizon at $\tau_{\mathrm{eff},j}\approx(2\bar\evb_j\evc_j)\inv.$
    Conversely, in the \emph{decay-dominated regime} ($2\evlambda_j\bar\evb_j\evc_j\ll 1-\evlambda_j$), $\evm_j^\ast\approx\evc_j\evsigma_v^2/(1-\evlambda_j)$ compounds post-$\silu$ key anisotropy $\evc_j$ with multi-timescale decay $(1-\evlambda_j)\inv,$ which causes stationary row energies $\evm_j^\ast$ to span multiple orders of magnitude across rows $j\in\{1,\dotsc,d_k\}$ within a single head.
\end{remark}


Under the standard (unweighted) readout $\vo_t=\vq_t\mS_t,$ rows with large stationary RMS $\evrho_j$ dominate the inner product with $\vq_t$ and mask low-energy, fast-decaying channels.
To characterize how a per-channel query gain $\vgamma_t\in\Rvec{d_k}$ remedies this imbalance, consider an associative retrieval setting in which a key-value pair $(\vk_{t-\ell},\hat\vv_{t-\ell})$ was written at timestep $t-\ell$ and the model queries $\mS_t$ at timestep $t$ along $\vq_t=\vk_{t-\ell}$ to recover $\hat\vv_{t-\ell}$; we refer to the elapsed timestep count $\ell\ge 1$ as the \emph{retrieval lag}.

\begin{proposition}[Lag-dependent matched-filter readout]
\label{prop3}
    Suppose at timestep $t,$ the state row $j\in\{1,\dotsc,d_k\}$ decomposes at retrieval lag $\ell\ge 1$ as $\mS_{t,j,:}=\eva_{\ell,j}\evk_{t-\ell,j}\hat\vv_{t-\ell}+\vxi_{t,j},$ where $\eva_{\ell,j}\defas\prod_{s=t-\ell+1}^t\evalpha_{s,j}(1-\evb_{s,j}\evk_{s,j}^2)$ is the cumulative diagonal retention over the $\ell$ intervening timesteps (with off-diagonal cross-row key products centered in $\vxi_{t,j}$ under $\E[\evk_{s,i}\evk_{s,j}]=\delta_{ij}\evc_j$) and $\vxi_{t,j}$ is zero-mean multi-write interference from timesteps $s\neq t-\ell,$ uncorrelated with $\hat\vv_{t-\ell}$ and satisfying $\E[\inner{\vxi_{t,i}}{\vxi_{t,j}}\mid\vk_{t-\ell}]=\delta_{ij}\evm_j^\ast.$
    Assume $\eva_{\ell,j}$ is independent of $(\vk_{t-\ell},\hat\vv_{t-\ell})$ with mean $\bar\eva_{\ell,j}\defas\E[\eva_{\ell,j}]\in(0,1)$ and small gate variance $\E[\eva_{\ell,i}\eva_{\ell,j}]\approx\bar\eva_{\ell,i}\bar\eva_{\ell,j}$ (so $\bar\eva_{\ell,j}\approx\evlambda_j^{\ell\sdiv 2}$ when $\bar\evb_j\evc_j\ll 1$).
    For a query $\vq_t=\vk_{t-\ell}$ and gain-modulated readout $\vo_t(\vgamma)\defas(\vgamma\odot\vq_t)\mS_t,$ the following hold:
    \begin{enumerate}[label=(\roman*),leftmargin=2em,ref=\getrefnumber{prop3}(\roman*)]
        \item Under the second-order closure $\E[\evk_{t-\ell,i}^2\evk_{t-\ell,j}^2]\approx\evc_i\evc_j,$ the expected retrieval mean-squared error (MSE) $\gL_\ell(\vgamma)\defas\E[\norm{\vo_t(\vgamma)-\hat\vv_{t-\ell}}_2^2]$ is
        \begin{equation}
        \label{prop3:eq1}
            \gL_\ell(\vgamma) = \evsigma_v^2\!\left(1-\sum_{j=1}^{d_k}\evgamma_j\evc_j\bar\eva_{\ell,j}\right)^{\!2} + \sum_{j=1}^{d_k}\evgamma_j^2\evc_j\evm_j^\ast.
        \end{equation}
        \item\label{prop3:item2} The unit-signal interference variance $\sum_{j=1}^{d_k}\evgamma_j^2\evc_j\evm_j^\ast$ (subject to $\sum_{j=1}^{d_k}\evgamma_j\evc_j\bar\eva_{\ell,j}=1$) and the unconstrained MSE $\gL_\ell(\vgamma)$ are both uniquely minimized along the matched-filter direction $\evgamma_j\propto\bar\eva_{\ell,j}/\evm_j^\ast,$ with unconstrained minimizer $\evgamma_j^\ast(\ell)$ ($\kappa_\ell\defas\evsigma_v^2/\big(1+\evsigma_v^2\sum_{i=1}^{d_k}\evc_i\bar\eva_{\ell,i}^2/\evm_i^\ast\big)$):
        \begin{equation}
        \label{prop3:eq2}
            \evgamma_j^\ast(\ell) = \kappa_\ell\,\frac{\bar\eva_{\ell,j}}{\evm_j^\ast} \propto \frac{\bar\eva_{\ell,j}}{\evrho_j^2} \approx \frac{\evlambda_j^{\ell\sdiv 2}}{\evrho_j^2},\qquad j\in\{1,\dotsc,d_k\}.
        \end{equation}
    \end{enumerate}
\end{proposition}



\begin{remark}[Incompatibility of static gain across retrieval lags]
\label{prop3:rem1}
    For any two distinct retrieval lags $1\le\ell_1<\ell_2,$ the elementwise ratio of optimal gains in \eqref{prop3:eq2} is $\evgamma_j^\ast(\ell_2)/\evgamma_j^\ast(\ell_1) = (\kappa_{\ell_2}/\kappa_{\ell_1})\,\bar\eva_{\ell_2,j}/\bar\eva_{\ell_1,j} \propto \evlambda_j^{\suppar{\ell_2-\ell_1}\sdiv 2}.$
    Whenever decay factors $\evlambda_j$ vary across channels ($\evlambda_j\neq\evlambda_{j'}$), the vectors $\vgamma^\ast(\ell_1)$ and $\vgamma^\ast(\ell_2)$ are non-collinear and reverse their correlation with the decay factor.
    Specifically, short-lag queries ($\ell_1=\gO(1),$ where $\evlambda_j^{\ell_1\sdiv 2}\approx 1$) follow $\evgamma_j^\ast(\ell_1)\propto 1/\evm_j^\ast$ and emphasize fast-decaying, low-interference rows (correlating negatively with $\evlambda_j$), whereas long-lag queries ($\ell_2\gg 1,$ where $\evlambda_j^{\ell_2\sdiv 2}\to 0$ on fast rows) must upweight slow-decaying rows ($\evlambda_j\to 1^-$) where the target signal survives (correlating positively with $\evlambda_j$).
    Consequently, no static vector $\vgamma$ can achieve the matched-filter optimum across multiple retrieval horizons simultaneously, which motivates a token-dependent dynamic readout gain $\vgamma_t.$
\end{remark}


When a head is restricted to a static gain vector $\vgamma$ shared across timesteps, marginalizing \eqref{prop3:eq1} over a distribution of retrieval lags characterizes the optimal time-invariant compromise.

\begin{corollary}[Horizon-marginalized static gain and fractional power law]
\label{prop3:cor1}
    Suppose a static gain $\vgamma$ minimizes the interference variance in \eqref{prop3:eq1} subject to the horizon-averaged signal constraint $\E_\ell\big[\sum_{j=1}^{d_k}\evgamma_j\evc_j\bar\eva_{\ell,j}\big]=1$ under a geometric lag distribution of mean horizon $H > 1,$ with $\bar\eva_{\ell,j}=\evlambda_j^{\ell\sdiv 2}.$
    Replacing $\bar\eva_{\ell,j}$ in \eqref{prop3:eq2} by the horizon-averaged signal survival $g_H(\evlambda_j)\defas\E_\ell[\evlambda_j^{\ell\sdiv 2}]$ yields the static matched filter
    \[
        \evgamma_j^{\mathrm{stat}} \propto \frac{g_H(\evlambda_j)}{\evm_j^\ast} = \frac{g_H(\evlambda_j)}{\evrho_j^2},\qquad
        g_H(\evlambda_j) = \frac{\sqrt{\evlambda_j}}{1+(H-1)(1-\sqrt{\evlambda_j})},\qquad j\in\{1,\dotsc,d_k\}.
    \]
    
    In the decay-dominated regime ($2\evlambda_j\bar\evb_j\evc_j\ll 1-\evlambda_j\ll 1$ at fixed $\evc_j$), where $\evm_j^\ast=\evrho_j^2\approx\evc_j\evsigma_v^2\tau_{\mathrm{eff},j}$ with $\tau_{\mathrm{eff},j}=(1-\evlambda_j)\inv,$ the local log-slope of $\evgamma_j^{\mathrm{stat}}$ with respect to the row RMS $\evrho_j\propto\tau_{\mathrm{eff},j}^{1\sdiv 2}$ satisfies
    \[
        -\frac{\mathrm{d}\log\evgamma_j^{\mathrm{stat}}}{\mathrm{d}\log\evrho_j} \approx \frac{4\tau_{\mathrm{eff},j}}{2\tau_{\mathrm{eff},j}+H-1} \defas p(\tau_{\mathrm{eff},j})\in(0,2),
    \]
    which satisfies $\evgamma_j^{\mathrm{stat}}\propto\evrho_j^{-p}$ with an intermediate exponent $p\in(0,1)$ (strictly between unit gain $p=0$ and full inverse-RMS whitening $p=1$) for all channels with $\tau_{\mathrm{eff},j}<(H-1)/2.$
\end{corollary}



\begin{remark}[Suboptimality of full inverse-RMS whitening]
\label{prop3:cor1:rem1}
    \Cref{prop3:cor1} shows that naive row normalization $\evgamma_j\propto\evrho_j\inv$ ($p=1$) or inverse-variance weighting $\evgamma_j\propto\evrho_j^{-2}$ ($p=2,$ which corresponds to the memoryless $H\to 1$ limit where $g_1(\evlambda_j)=\evlambda_j^{1\sdiv 2}\approx 1$) is suboptimal for multi-step associative retrieval: rows with large $\evrho_j$ are the slow-decaying channels ($\evlambda_j\to 1^-,$ large $\tau_{\mathrm{eff},j}$) that preserve long-range signal $g_H(\evlambda_j).$
    Dividing by $\evrho_j$ or $\evrho_j^2$ without accounting for the signal survival numerator $g_H(\evlambda_j)$ over-amplifies low-energy, fast-decaying rows whose target signal has vanished while suppressing high-energy, slow-decaying channels, whereas the horizon-marginalized matched filter $\evgamma_j^{\mathrm{stat}}\propto g_H(\evlambda_j)/\evrho_j^2$ damps high-energy rows with a compressed exponent $p\in(0,1)$ that balances interference suppression against long-horizon signal preservation.
\end{remark}


Finally, we connect query-time readout back to write-time erasure.
Consider a head with channel-uniform mean erase rate $\bar\evb_j=\bar\evb\in(0,2)$ in \eqref{prop2:eq2} (where $\evb_{t,j}=\bar\evb$ implies $\evmu_t=\sum_{j=1}^{d_k}\evb_{t,j}\evk_{t,j}^2=\bar\evb$).
Choosing $\bar\evb$ faces a fundamental tension: increasing $\bar\evb$ suppresses multi-write interference by contracting every row's effective memory horizon $\tau_{\mathrm{eff},j}(\bar\evb)$ in \eqref{prop2:eq2}, but simultaneously attenuates stored target associations by a factor of $1-\bar\evb\evc_j$ on each intervening write step (\Cref{prop3}).
Because matched-filter readout already damps interference from high-energy rows at query time, it relaxes the erasure required at write time.

\begin{proposition}[Erase-readout substitution]
\label{prop4}
    Consider a head with channel-uniform mean erase rate $\bar\evb_j=\bar\evb\in(0,2)$ and horizon-averaged signal survival $g_H(\evlambda_j).$
    For any readout gain $\vgamma\in\Rvec{d_k}$ with expected horizon-averaged signal gain $S(\vgamma)\defas\sum_{j=1}^{d_k}\evgamma_j\evc_jg_H(\evlambda_j)>0,$ define the normalized interference-to-signal ratio:
    \[
        \gI(\bar\evb,\vgamma) \defas \frac{\sum_{j=1}^{d_k}\evgamma_j^2\evc_j\evm_j^\ast(\bar\evb)}{\evsigma_v^2\,S(\vgamma)^2}
        = \sum_{j=1}^{d_k}\frac{\evgamma_j^2\evc_j^2}{S(\vgamma)^2\big(1-\evlambda_j+2\evlambda_j\bar\evb\evc_j\big)}
        = \sum_{j=1}^{d_k}\frac{\evgamma_j^2\evc_j^2}{S(\vgamma)^2}\,\tau_{\mathrm{eff},j}(\bar\evb).
    \]
    Let $\bar\tau(\bar\evb)\defas\gI(\bar\evb,\vone)$ denote the baseline interference ratio under unit readout ($\vgamma=\vone$), and let $B\colon(0,2)\to\mathbb{R}$ be a twice-differentiable erasure cost (capturing target-signal attenuation and, for $\bar\evb>1,$ over-relaxation error) satisfying $B'(\bar\evb)>0$ and $B''(\bar\evb)\ge 0.$
    For any readout interference multiplier $\Phi>0$ relative to $\bar\tau(\bar\evb),$ define the equilibrium erase rate:
    \[
        \bar\evb^\ast(\Phi) \defas \argmin_{\bar\evb\in(0,2)} \;\gJ(\bar\evb;\Phi),\qquad
        \gJ(\bar\evb;\Phi) \defas \Phi\,\bar\tau(\bar\evb) + B(\bar\evb).
    \]
    Then, at any $\bar\evb\in(0,2)$ for which the channel noise-to-signal ratios $\evm_j^\ast(\bar\evb)/g_H(\evlambda_j)$ vary across $j,$ the static matched filter $\vgamma^\ast$ satisfying $S(\vgamma^\ast)=S(\vone)$ achieves a strict interference reduction $\Phi^\ast\defas\gI(\bar\evb,\vgamma^\ast)/\bar\tau(\bar\evb)<1,$ and any interior minimizer $\bar\evb^\ast(\Phi)$ satisfies
    \[
        \bar\evb^\ast(\Phi^\ast) < \bar\evb^\ast(1),\qquad
        \tau_{\mathrm{eff},j}\!\big(\bar\evb^\ast(\Phi^\ast)\big) > \tau_{\mathrm{eff},j}\!\big(\bar\evb^\ast(1)\big)\quad\forall\, j\in\{1,\dotsc,d_k\}.
    \]
\end{proposition}



\begin{remark}[Mechanism of memory-horizon extension]
\label{prop4:rem1}
    Under unit readout ($\vgamma=\vone$), the only mechanism available to suppress multi-write interference $\gI(\bar\evb,\vone)$ in high-energy rows is to elevate the erase rate $\bar\evb,$ which reduces $\tau_{\mathrm{eff},j}(\bar\evb)$ and prematurely flushes older associations across all channels.
    Introducing matched-filter readout gain suppresses interference at query time ($\Phi^\ast<1$), which shifts the equilibrium erase rate downward ($\bar\evb^\ast(\Phi^\ast)<\bar\evb^\ast(1)$) and extends the effective memory horizon $\tau_{\mathrm{eff},j}$ of every state channel.
\end{remark}


\subsection{Synthetic Memory Experiments}
\label{sec:theory:exp}

To test our theoretical predictions under controlled data-generating processes that satisfy our theoretical assumptions, we design two synthetic associative-memory tasks:
a \emph{single-head stationary retrieval task} that isolates \Cref{prop2,prop3,prop4} and \Cref{prop3:cor1}, and an \emph{end-to-end learned multi-query associative recall task} that tests \Cref{prop1,prop3,prop4} under gradient descent.
(Additional details in \Cref{app:toy}.)


\subsubsection{Synthetic Task Setup}

\xhdr{Single-Head Stationary Retrieval}

In this task, we simulate a single gated-delta head ($d_k = d_v = 64$) driven by a stationary stream of unit-norm keys $\vk_t \in \bS{d_k-1}$ with anisotropic power-law channel second moments $\evc_j = \E[\evk_{t,j}^2] > 0$ ($\sum_{j=1}^{d_k} \evc_j = 1$) and isotropic values $\vv_t \sim \gN(\vzero, d_v\inv\I_{d_v})$ ($\evsigma_v^2 = 1$).
%
To isolate stationary row-energy accumulation and query readout from learned gate projections, we fix the write gate to $\vw_t = \vone$ (so $\hat\vv_t = \vv_t$), the erase gate to a uniform scalar $\vb_t = \evmu\vone$ with $\evmu = \bar\evb \in (0, 2)$ (so $\evmu_t \equiv \evmu$ and $\vs_t \equiv \vzero$), and the decay vector $\valpha \in (0, 1)^{1\times d_k}$ to time-invariant factors with log-spaced timescales $\tau_j$ ($\evalpha_j = e^{-1\sdiv\tau_j},$ $\evlambda_j = \evalpha_j^2$).
At each timestep $t,$ the state $\mS_t$ updates via \eqref{eq:gdn2}, and we sample a retrieval lag $\ell_t \sim p(\ell)$ and query $\mS_t$ along the historical key $\vq_t = \vk_{t-\ell_t}$ with readout gain $\vgamma$ to recover $\vv_{t-\ell_t}$ under normalized MSE $\gL(\vgamma) \defas \evsigma_v^{-2}\E[\norm{\vo_t(\vgamma) - \vv_{t-\ell_t}}_2^2]$:\looseness=-1
\[
    \mS_t = \big(\I - \evmu\vk_t\tr\vk_t\big)\Diag(\valpha)\mS_{t-1} + \vk_t\tr\vv_t,
    \qquad
    \vo_t(\vgamma) = \big(\vgamma \odot \vk_{t-\ell_t}\big)\mS_t.
\]

We consider two lag distributions $p(\ell).$
First, under a \emph{geometric lag horizon} of mean $H,$ we sweep the erase strength $\evmu \in (0, 2)$ across the stability interval of the rank-one erasure contraction $\I - \evmu\vk_t\tr\vk_t$ and compare five static readout profiles (each scaled by a least-squares scalar prefactor $\kappa^\ast$ fitted on a disjoint split, corresponding to $\kappa_\ell$ in \eqref{prop3:eq2}):
unit gain $\evgamma_j = 1$ ($p = 0$; \gdntwo), half-power scaling $\evgamma_j \propto 1/\sqrt{\smash{\evrho_j}\vphantom{\evrho}}$ ($p = 1/2$), full inverse-RMS whitening $\evgamma_j \propto 1/\evrho_j$ ($p = 1$), the \Cref{prop3:cor1} horizon filter $\evgamma_j^{\mathrm{stat}} \propto g_H(\evlambda_j)/\evm_j^\ast,$ and the unconstrained $d_k$-parameter least-squares static oracle $\vgamma^{\mathrm{LS}}.$
Second, under a \emph{bimodal two-lag mixture} $p(\ell) = \tfrac{1}{2}\delta_{\ell,2} + \tfrac{1}{2}\delta_{\ell,80}$ at the unit-gain optimal erase strength $\evmu = \evmu_{\mathrm{unit}}^\ast,$ we compare the shared $d_k$-parameter static oracle $\vgamma^{\mathrm{LS}}$ against two lag-dependent readouts: the \Cref{prop3} analytic filter $\evgamma_j^\ast(\ell) \propto \evlambda_j^{\ell\sdiv 2}/\evm_j^\ast$ (one scalar prefactor per lag) and the lag-conditioned least-squares oracle $\vgamma^{\mathrm{LS}}(\ell).$
%


\xhdr{Learned Multi-Query Associative Recall}

We train a single-layer, two-head model ($N_h = 2,$ $d_k = d_v = 32,$ $d = 128$) end-to-end on \emph{streaming} multi-query associative recall~\citep{Arora2023-tw}.
Each sequence spans $T = 256$ write/read events over $K = 256$ keys and $V = 64$ values, activating $K_{\mathrm{live}} = 96$ keys ($1.5\times$ the total state rank $N_h d_k = 64$) with $\zipf(1.0)$ frequencies.
Each timestep $t \in \{1, \dotsc, T\}$ is a write (prob.\ $0.7$) binding an active key to a fresh value in $\{1, \dotsc, V\}$ (overwriting prior bindings), or a read (prob.\ $0.3$) querying a written key for its latest value.
To ensure the write gate ($\vw_t \odot \vv_t$) must dynamically suppress inactive value channels, we partition values across four orthogonal subspaces of $\Rvec{d}$ and inject isotropic Gaussian noise into the inactive subspaces.
From RMS-normalized $\vx_t,$ each head $h \in \{1, 2\}$ computes unit-norm $\silu$ queries and keys ($\vq_t^{\suppar{h}}$ and $\vk_t^{\suppar{h}}$), values $\vv_t^{\suppar{h}},$ and rank-$8$ decay gates $\valpha_t^{\suppar{h}},$ projecting concatenated head outputs to $V$-way read logits.

We vary two architectural axes:
%
(i) to test \Cref{prop1}, we parameterize the erase gate $\vb_t$ and write gate $\vw_t$ across heads either with a dense projection $\vx_t\mW_m$ ($m \in \{b, w\},$ full rank $N_h d_k = 64$) or a rank-$r_m$ bottleneck $\vx_t\mW_{m\downarrow}\mW_{m\uparrow} + \vc_m,$ sweeping $r_m \in \{1, 2, 4, 8, 16, 32\}$ for one gate while keeping the other dense; and
(ii) to test \Cref{prop3,prop4}, we compare unit ($\vgamma_t = \vone$), static ($\vgamma_t = 2\sigmoid(\vc_q)$), and rank-$8$ dynamic ($\vgamma_t = 2\sigmoid(\vx_t\mW_{q\downarrow}\mW_{q\uparrow} + \vc_q)$) readout gains (partitioned across heads into $\vb_t^{\suppar{h}}, \vw_t^{\suppar{h}}, \vgamma_t^{\suppar{h}}$):
\begin{gather*}
    \vb_t = 2\sigmoid(\vx_t\mW_{b\downarrow}\mW_{b\uparrow} + \vc_b),\qquad
    \vw_t = \sigmoid(\vx_t\mW_{w\downarrow}\mW_{w\uparrow} + \vc_w),\qquad
    \vgamma_t = 2\sigmoid(\vx_t\mW_{q\downarrow}\mW_{q\uparrow} + \vc_q);\\
    \vo_t^{\suppar{h}} = \big(\vgamma_t^{\suppar{h}} \odot \vq_t^{\suppar{h}}\big)\mS_t^{\suppar{h}}.
\end{gather*}
%


\begin{figure}[!t]
    \centering
    \includegraphics[width=\textwidth]{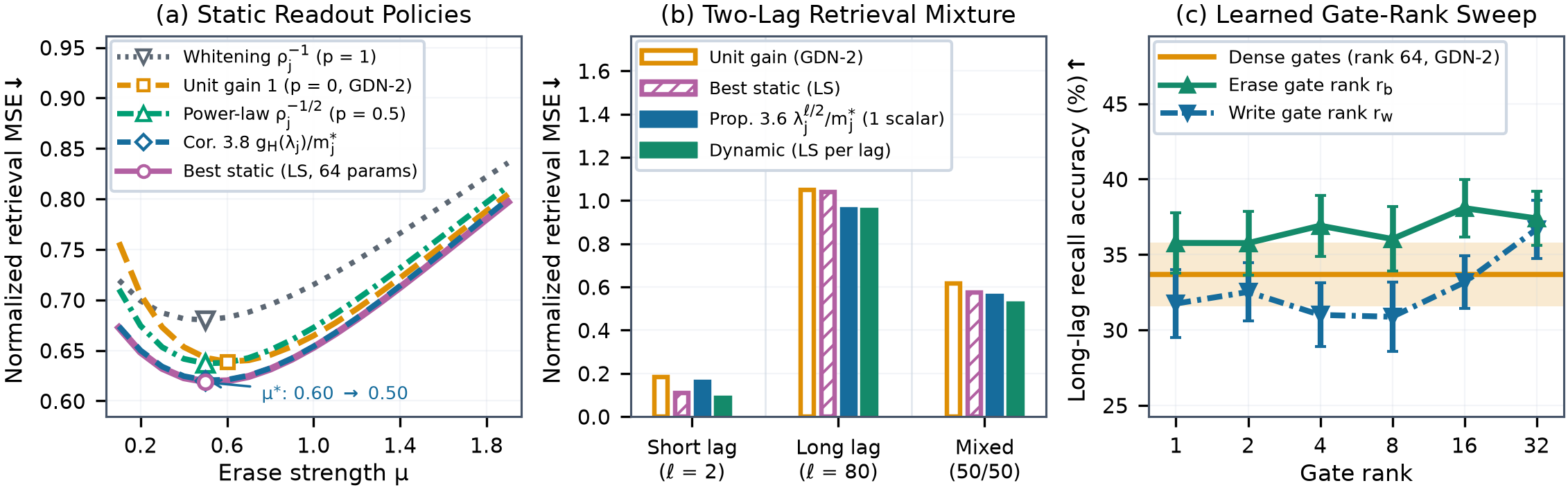}
    \caption{%
        \textbf{Matched-filter readout and low-rank erase gates improve synthetic associative retrieval and lower optimal erase strength.}
        (a) Full whitening worsens retrieval error relative to unit gain (\gdntwo), whereas half-power scaling and the \Cref{prop3:cor1} filter reduce error and shift $\evmu^\ast$ from $0.60$ to $0.50,$ with the \Cref{prop3:cor1} filter matching the static oracle.
        (b) On a $50/50$ mixture of lags $\ell \in \{2, 80\}$ at $\evmu = 0.60,$ the lag-matched oracle outperforms the static oracle across all horizons, while the \Cref{prop3} filter matches the lag-matched oracle at $\ell = 80$ and beats the static oracle on long and mixed horizons.
        %
        (c) Low-rank erase gates $r_b$ outperform dense gates at every bottleneck rank on long-lag recall, whereas low-rank write gates $r_w$ degrade below dense at $r_w \le 8$ and recover once $r_w \ge 16.$%
    }
    \label{fig:toys}
\end{figure}

\subsubsection{Synthetic Task Results}
\label{sec:theory:exp:results}

\Cref{fig:toys} summarizes the results across both tasks over $20$ seeds (full tables in \Cref{app:toy}).
In the single-head stationary setting at $\evmu = \evmu_{\mathrm{unit}}^\ast = 0.60,$ the closed-form law \eqref{prop2:eq2} (\Cref{prop2}) matches the empirical row energies $\evm_j$ across a $52.4\times$ dynamic range with Spearman and log-Pearson correlations of $1.0.$
Sweeping $\evmu \in (0, 2)$ under the geometric lag horizon of mean $H = 27$ (\Figref{fig:toys}{a}), full inverse-RMS whitening $1/\evrho_j$ ($p = 1$) worsens retrieval MSE relative to unit gain because dividing by $\evrho_j$ amplifies fast-decaying rows whose target signal has vanished (\Cref{prop3:cor1:rem1}).
By contrast, half-power scaling $1/\sqrt{\smash{\evrho_j}\vphantom{\evrho}}$ improves over unit gain and the \Cref{prop3:cor1} horizon filter $\evgamma_j^{\mathrm{stat}} \propto g_H(\evlambda_j)/\evm_j^\ast$ matches the $64$-parameter static oracle $\vgamma^{\mathrm{LS}},$ with both filters shifting the optimal erase strength from $\evmu^\ast = 0.60$ down to $0.50$ (\Cref{prop4}).
On the bimodal $\ell_t \in \{2, 80\}$ mixture (\Figref{fig:toys}{b}), the lag-conditioned oracle $\vgamma^{\mathrm{LS}}(\ell)$ correlates negatively with the channel decay factor $\evalpha_j$ at $\ell = 2$ (Spearman $r_s = -0.98$) and reverses sign at $\ell = 80$ ($r_s = +0.83$), in accordance with \Cref{prop3:rem1}.
Consequently, lag-matched readout $\vgamma^{\mathrm{LS}}(\ell)$ ($64$ parameters per lag) outperforms the static oracle $\vgamma^{\mathrm{LS}}$ ($64$ parameters) across all horizons.
Meanwhile, the \Cref{prop3} filter, fitting just one scalar per lag, matches $\vgamma^{\mathrm{LS}}(\ell)$ at $\ell=80$ and outperforms the static oracle on long and mixed horizons.

In the learned associative recall task (\Figref{fig:toys}{c}), sweeping the bottleneck rank for $\vb_t$ vs.\ $\vw_t$ confirms \Cref{prop1}: compressing $\vb_t$ (keeping $\vw_t$ dense) improves long-lag recall over dense \gdntwo at every rank while cutting erase-logit token variance on writes ($3.0\%\text{--}15.6\%$ vs.\ $27.9\%$; \Cref{prop1:rem2}).
%
Conversely, compressing $\vw_t$ (with $\vb_t$ dense) to $r_w \le 8$ degrades long-lag recall below the dense baseline, whereas $r_w = 16$ approaches dense recall and $r_w = 32$ surpasses it.
%
Finally, when both gates are compressed to $(r_b, r_w) = (4, 4),$ replacing unit readout ($\vgamma_t = \vone$) with static $\vgamma$ and dynamic $\vgamma_t$ monotonically raises long-lag recall ($+4.18$ points over unit readout, $+2.00$ over dense gates) while lowering final-timestep state energy $\E[\norm{\mS_T}_F^2]$ by $45.7\%$ compared to unit readout ($64.3\%$ vs.\ dense gates), matching \Cref{prop3,prop4}.

%% file: sections/4_method.tex
\section{\oursxp}
\label{sec:method}

\begin{figure}[!t]
    \centering
    \includegraphics[width=\textwidth]{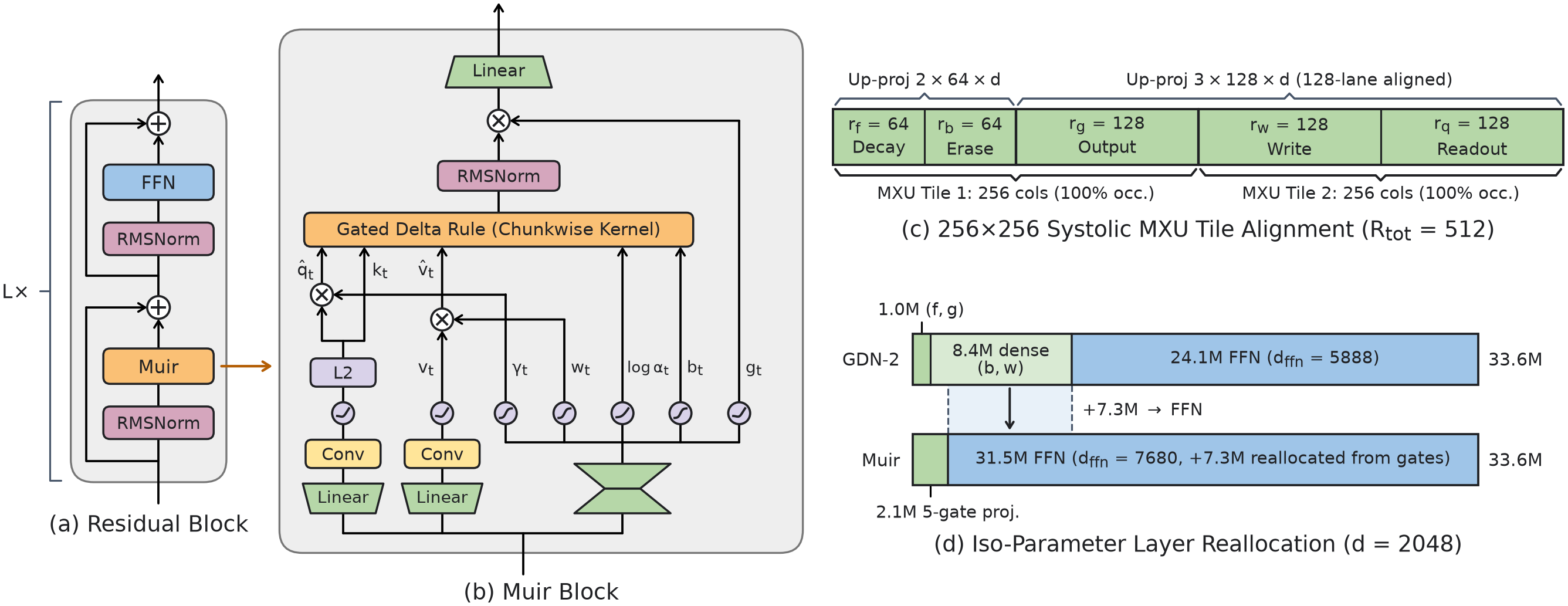}
    \caption{%
        \textbf{\ours co-designs asymmetric low-rank gate parameterization and matched-filter query readout with MXU tile alignment and iso-parameter FFN reallocation.}
        (a) Standard residual block with a \ours token mixer.
        (b) Within the \ours block, $\vx_t$ feeds linear-$\conv$-$\silu$ pathways for $(\vq_t, \vk_t, \vv_t)$ alongside a single fused low-rank gate projection that produces all five gates $(\log\valpha_t, \vb_t, \vg_t, \vw_t, \vgamma_t).$
        (c) Asymmetric gate ranks $(r_f, r_b, r_g, r_w, r_q) = (64, 64, 128, 128, 128)$ sum to $R_{\mathrm{tot}} = 512 = 2 \times 256,$ which fills two $256$-column MXU tiles in the fused down-projection and aligns $(f, b)$ and $(g, w, q)$ on $128$-lane VMEM tile boundaries for batched up-projections.
        (d) At $d = 2{,}048,$ replacing the dense $(b, w)$ gate matrices in \gdntwo cuts gate-projection parameters by $4.5\times$; this either reduces per-layer parameters and FLOPs by $14.6\%$ at fixed FFN width, or frees $+7.3\text{M}$ parameters per layer to widen the FFN under an iso-parameter budget.%
    }
    \label{fig:arch}
\end{figure}

In this section, we translate our theoretical findings in \S\ref{sec:theory} into the \ours architecture (\Cref{fig:arch}) and its TPU-efficient chunkwise training algorithm (\Cref{fig:kernel}).\footnote{%
    We name \emph{\ours} (pronounced \emph{mewr}) after Muir Woods National Monument and its ancient coastal redwoods, whose rings preserve centuries of history, echoing long-horizon state memory.%
}
Following \Cref{prop2,prop3,prop3:rem1,prop4}, each \ours head augments the four channel-wise gates of \gdntwo ($\valpha_t, \vb_t, \vw_t, \vg_t$) with a fifth channel-wise gate, a dynamic query readout gain $\vgamma_t \in \Rvec{d_k}$ that scales the $\ell_2$-normalized query prior to state readout.
With the gain-modulated query $\hat\vq_t \defas \vgamma_t \odot \vq_t$ alongside $\hat\mS_t = \Diag(\valpha_t)\mS_{t-1},$ $\hat\vk_t = \vb_t \odot \vk_t,$ and $\hat\vv_t = \vw_t \odot \vv_t,$ the \ours state update and readout are
\begin{equation}
\label{eq:ours}
    \mS_t = \big(\I - \vk_t\tr(\vb_t \odot \vk_t)\big)\Diag(\valpha_t)\mS_{t-1} + \vk_t\tr(\vw_t \odot \vv_t)
    = \hat\mS_t - \vk_t\tr\big(\hat\vk_t\hat\mS_t\big) + \vk_t\tr\hat\vv_t,\qquad
    \vo_t = \hat\vq_t\mS_t = (\vgamma_t \odot \vq_t)\mS_t,
\end{equation}
followed by per-head $\rms(\vo_t) \odot \silu(\vg_t)$ and the output projection $\mW_o.$


\subsection{Low-Rank Gating and Matched-Filter Readout}
\label{sec:method:arch}

While \gdntwo already parameterizes the decay and output gates $(f, g)$ in low rank, it uses dense $d \times d$ matrices for the erase and write gates $(b, w)$ to provide full channel-wise flexibility.
However, \Cref{prop1,prop1:rem2} establish that a low-rank bottleneck on $\vb_t$ suffices to preserve the channel-wise erase control of \gdntwo along the active key subspace while projecting out the oblique erase variance $\norm{\vs_t}_2^2$ that inflates the state-transition spectral norm above $1,$ whereas the write gate $\vw_t$ modulates all $d_v$ value channels directly and requires higher projection rank than $\vb_t.$
We therefore replace the separate low-rank $(f, g)$ and dense $(b, w)$ projections in \gdntwo with a single \emph{fused} low-rank gate projection covering all five gates: decay $f,$ erase gate $b,$ output gate $g,$ write gate $w,$ and query readout gain $q.$

Given the token representation $\vx_t \in \Rvec{d},$ we compute a single fused down-projection $\mW_{\downarrow} \in \R{d \times R_{\mathrm{tot}}}$ into a bottleneck of total rank $R_{\mathrm{tot}} \defas r_f + r_b + r_g + r_w + r_q,$ followed by five per-gate up-projections $\mW_{m\uparrow} \in \R{r_m \times d}$ with learned channel biases $\vc_m \in \Rvec{d}$:
\begin{equation}
\label{eq:ours_proj}
    \vh_t = \vx_t \mW_{\downarrow} \in \Rvec{R_{\mathrm{tot}}},\quad 
    \begin{bmatrix}
        \vh^f_t & \vh^b_t & \vh^g_t & \vh^w_t & \vh^q_t
    \end{bmatrix} = \vh_t,\quad
    \vz^m_t = \vh^m_t \mW_{m\uparrow} + \vc_m \;\;\text{for } m \in \{f, b, g, w, q\}.
\end{equation}
After partitioning each $\vz^m_t$ across the $N_h$ heads, the five gate vectors in \eqref{eq:ours} are parameterized as
\[
    \log\valpha_t = -\exp(a_{\log})\,\softplus(\vz^f_t),\quad
    \vb_t = 2\sigmoid(\vz^b_t),\quad
    \vw_t = \sigmoid(\vz^w_t),\quad
    \vg_t = \vz^g_t,\quad
    \vgamma_t = \vgamma^{\mathrm{stat}} \odot 2\sigmoid(\vz^q_t).
\]
Unlike the \emph{unbiased} dense $(b, w)$ projections in \gdntwo, including a bias $\vc_m$ on every up-projection gives each channel an independent operating point despite the low-rank bottleneck ($\vc_f, \vc_g$ are inherited from \gdntwo, while $\vc_b, \vc_w, \vc_q$ are newly added in \ours).

For the readout gain $\vgamma_t,$ we initialize $\vgamma^{\mathrm{stat}} = \vone$ and $\mW_{q\uparrow} = \mZero$ (with $\vc_q = \vzero$) so that $2\sigmoid(\vz^q_t) = \vone$ and $\vgamma_t = \vone$ at initialization.
Scaling the sigmoid into $(0, 2)$ centers the dynamic multiplier at $1$ so each query can either attenuate ($<\!1$) or amplify ($>\!1$) state rows relative to $\evgamma_j^{\mathrm{stat}}$ in $\vo_t = \sum_{j=1}^{d_k}\evgamma_{t,j}\evq_{t,j}\mS_{t,j,:},$ whereas an unscaled sigmoid in $(0, 1)$ could only damp $\evgamma_j^{\mathrm{stat}}.$
This factored form lets $\vgamma^{\mathrm{stat}}$ learn the horizon-marginalized power-law profile (\Cref{prop3:cor1}) and $2\sigmoid(\vz^q_t)$ learn the lag-dependent matched filter (\Cref{prop3}), while scale-invariant $\rms(\vo_t)$ absorbs the global factor $\kappa_\ell$ in \eqref{prop3:eq2}.
%
Unlike learned per-channel query-key scaling \citep{Hua2022-di,Dehghani2023-iu,Lahoti2026-fc}, $\vgamma_t$ scales only $\vq_t$ at readout so $\vk_t$ remains unit-norm ($\norm{\vk_t}_2 = 1$) for contractive erasure.


\xhdr{Rank Allocation and Hardware Alignment}

We set the bottleneck ranks $(r_f, r_b, r_g, r_w, r_q)$ in \eqref{eq:ours_proj} to satisfy two constraints:
(i) motivated by the sensitivity asymmetry in \Cref{prop1}, the gates that modulate full-dimensional vectors directly (output gate $g,$ write gate $w,$ and query readout gain $q$) receive twice the rank of the gates governing state decay and key-contracted erasure (decay $f$ and erase gate $b$), and
(ii) the total rank $R_{\mathrm{tot}}$ is an exact multiple of the $256$-column TPU7x MXU tile width.
Setting $(r_f, r_b, r_g, r_w, r_q) = (64, 64, 128, 128, 128)$ gives $R_{\mathrm{tot}} = 512 = 2 \times 256.$\footnote{%
    In ablations that omit the token-dependent factor $2\sigmoid(\vz^q_t)$ ($r_q = 0,$ so $\vgamma_t = \vgamma^{\mathrm{stat}}$ or $\vone$), those $128$ rank units are assigned to the write gate ($r_w = 256$) to hold $R_{\mathrm{tot}} = 512$ and the layer parameter count invariant.
}

We also order the columns of $\mW_{\downarrow} \in \R{d \times 512}$ to match the $256$-column MXU tile width and the $128$-lane TPU VMEM tile boundary (\Figref{fig:arch}{c}).
Placing the two rank-$64$ gates $(f, b)$ contiguously at columns $[0\!:\!128]$ and the three rank-$128$ gates $(g, w, q)$ contiguously at columns $[128\!:\!512]$ executes the fused down-projection $\vh_t = \vx_t\mW_{\downarrow}$ across two full $256 \times 256$ MXU tiles at $100\%$ column occupancy without zero-padding.
At the same time, placing the split between the rank-$64$ and rank-$128$ groups at column $128$ aligns with an exact $128$-lane VMEM tile boundary, so the five up-projections $\vh^m_t\mW_{m\uparrow}$ execute as two batched matrix multiplications (stacking the two $64 \times d$ matrices for $(f, b)$ and the three $128 \times d$ matrices for $(g, w, q)$) instead of five separate narrow matrix multiplications.


\xhdr{Iso-Parameter FFN Reallocation}

The fused low-rank gate projection \eqref{eq:ours_proj} uses only $\sum_m 2d r_m = 2d R_{\mathrm{tot}}$ parameters instead of $2d^2 + 512d$ in \gdntwo.
Under the $12d^2$ per-layer non-embedding parameter budget of the \tf, setting $d_{\mathrm{ffn}}(R_{\mathrm{tot}}) = 4d - R_{\mathrm{tot}}$ gives $4d^2 + 2d R_{\mathrm{tot}} + 2d(4d - R_{\mathrm{tot}}) = 12d^2$ for every $d$ and $R_{\mathrm{tot}},$ expanding the FFN width relative to \gdntwo ($d_{\mathrm{ffn}} = 3d - 256$) by $\Delta d_{\mathrm{ffn}} = d - 256$ channels at $R_{\mathrm{tot}} = 512.$
At $d = 2{,}048$ and $R_{\mathrm{tot}} = 512,$ replacing the gate projections in \gdntwo ($9.44\text{M}$ parameters) with \eqref{eq:ours_proj} ($2.10\text{M}$ parameters) reduces gate-projection parameters and FLOPs by $4.5\times$ ($-77.8\%$) and token-mixer projection parameters and FLOPs by $28.0\%$ ($26.21\text{M} \to 18.87\text{M}$ parameters/layer).
At fixed $d_{\mathrm{ffn}} = 5{,}888,$ this reduces full-layer non-embedding parameters and linear FLOPs by $14.6\%$ ($50.33\text{M} \to 42.99\text{M}$ parameters/layer); under our primary iso-parameter budget, reinvesting the $7.34\text{M}$ saved parameters per layer widens $d_{\mathrm{ffn}}$ from $5{,}888$ to $7{,}680 = 30 \times 256$ ($+1{,}792$ channels, or $+30.4\%$; \Figref{fig:arch}{d}).


\subsection{Boundary-Pivoted Chunkwise Kernel}
\label{sec:method:kernel}

Within each chunk of length $C$ indexed by $t \in \{1, \dotsc, C\},$ starting from chunk-initial state $\mS_0 \in \R{d_k \times d_v},$ replacing $\vq_t$ with the gain-modulated query $\hat\vq_t = \vgamma_t \odot \vq_t$ reduces the \ours recurrence \eqref{eq:ours} to the chunkwise linear system of \gdntwo in \S\ref{sec:prelim:chunk}.
Let $\mLambda_t \defas \sum_{r=1}^t \log\valpha_r \le \vzero$ denote the cumulative log-decay at timestep $t,$\footnote{%
    In practice, instead of running a sequential prefix scan on the VPU, we multiply the chunk log-decay matrix $\log\mAlpha \in \R{C\times d_k}$ (with rows $\log\valpha_t$) by a lower-triangular all-ones matrix on the MXU: $\mLambda = \tril(\vone_{C\times C})\log\mAlpha.$%
}
%
with chunk-prefix and pairwise relative decays $\vpi_t = \exp(\mLambda_t),\, \vpi_{t,s} = \exp(\mLambda_t - \mLambda_s) \in (0, 1]^{1\times d_k}$ ($t \ge s$).
Evaluating a chunk requires three stages:
(i) constructing the unit lower-triangular system matrix $\mT = \I + \mL_{\mathrm{strict}}$ with strictly lower-triangular entries $(\mL_{\mathrm{strict}})_{t,s} = \inner{\hat\vk_t}{\vpi_{t,s}\odot\vk_s}$ ($t > s$) and the lower-triangular causal readout matrix $\mM$ with entries $\emM_{t,s} = \inner{\hat\vq_t}{\vpi_{t,s}\odot\vk_s}$ ($t \ge s$);
(ii) solving the triangular system $\mT\mZ = \begin{bmatrix}\hat\mV & \hat\mK_{\odot\pi}\end{bmatrix}$ for $\mZ \defas \begin{bmatrix}\mU & \mY\end{bmatrix} \in \R{C\times(d_v+d_k)},$ where $\hat\mV$ and $\hat\mK_{\odot\pi}$ have rows $\hat\vv_t$ and $\vpi_t \odot \hat\vk_t,$ respectively; and
(iii) substituting $\tilde\mV = \mU - \mY\mS_0$ into the chunkwise readout and state update to advance the chunk outputs $\mO$ and end-of-chunk state $\mS_C$ as affine maps of $\mS_0$:
\begin{equation}
\label{eq:chunk_state_out}
    \mO = \big(\hat\mQ_{\odot\pi} - \mM \mY\big)\mS_0 + \mM \mU,
    \qquad
    \mS_C = \big(\Diag(\vpi_C) - \tilde\mK\tr \mY\big)\mS_0 + \tilde\mK\tr \mU,
\end{equation}
where $\hat\mQ_{\odot\pi}, \tilde\mK \in \R{C\times d_k}$ have rows $(\hat\mQ_{\odot\pi})_{t,:} \defas \vpi_t \odot \hat\vq_t$ and $\tilde\mK_{s,:} = \vpi_{C,s} \odot \vk_s.$


\xhdr{Divide-and-Conquer Block Decomposition}
As noted in \S\ref{sec:prelim:chunk}, executing stages (i) and (ii) directly over the full chunk is bottlenecked by the $C \times C \times d_k$ 3D decay tensor (or the underflowing division $\vpi_t \oslash \vpi_s$) when forming $\mT$ and $\mM,$ and by sequential $C$-step forward substitution when solving with $\mT.$
%
Instead of forward substitution, we construct $\mT\inv$ and $\mM$ via a divide-and-conquer recursion \citep{Heller1974-mg,Bini1994-bt,Sobczyk2026-lr} over a binary tree of depth $r$ and apply $\mT\inv$ as a matrix multiplication $\begin{bmatrix}\mU & \mY\end{bmatrix} = \mT\inv\begin{bmatrix}\hat\mV & \hat\mK_{\odot\pi}\end{bmatrix}.$
%
Whenever any contiguous token interval is split into left ($\gL$) and right ($\gR$) halves of equal length, the corresponding sub-blocks of $\mT,$ its inverse $\mT\inv,$ and $\mM$ partition into $2\times 2$ block form:
\begin{equation}
\label{eq:block_inv}
    \mT_{\mathrm{blk}} = \left[\begin{array}{c|c}
        \mT_\gL & \mZero \\[2pt]
        \hline\\[-10pt]
        \mT_{\mathrm{cross}} & \mT_\gR
    \end{array}\right],
    \quad
    \mT_{\mathrm{blk}}\inv = \left[\begin{array}{c|c}
        \mT_\gL\inv & \mZero \\[2pt]
        \hline\\[-10pt]
        -\mT_\gR\inv\mT_{\mathrm{cross}}\mT_\gL\inv & \mT_\gR\inv
    \end{array}\right];
    \qquad
    \mM_{\mathrm{blk}} = \left[\begin{array}{c|c}
        \mM_\gL & \mZero \\[2pt]
        \hline\\[-10pt]
        \mM_{\mathrm{cross}} & \mM_\gR
    \end{array}\right].
\end{equation}
Here, the diagonal sub-blocks $(\mT_\gL, \mT_\gR)$ and their inverses $(\mT_\gL\inv, \mT_\gR\inv)$ remain unit lower-triangular and $(\mM_\gL, \mM_\gR)$ remain lower-triangular, whereas the bottom-left cross-blocks $\mT_{\mathrm{cross}}$ and $\mM_{\mathrm{cross}}$ (whose row indices $t \in \gR$ and column indices $s \in \gL$ satisfy $t > s$ for every entry) are dense.
Because the diagonal sub-blocks preserve the exact same lower-triangular structure on half the interval, bisecting the chunk $r$ times reduces stages (i) and (ii) to a base case on $2^r$ leaf intervals of length $n_0 \defas C/2^r,$ followed by $r$ bottom-up merge levels ($m = 0, \dotsc, r-1$) across child intervals of length $n_m \defas n_0 2^m.$


\xxhdr{Base Case: Diagonal Leaf Blocks ($n_0 = C/2^r$)}
Within each of the $2^r$ leaf intervals ($\lceil t/n_0\rceil = \lceil s/n_0\rceil$), we evaluate the pairwise decay $\vpi_{t,s} = \exp(\mLambda_t - \mLambda_s)$ directly from the non-positive log-differences $\mLambda_t - \mLambda_s \le \vzero$ ($t \ge s$) to form the diagonal leaf blocks $\mT_{\mathrm{leaf}} = \I + \mL_{\mathrm{leaf}}$ and $\mM_{\mathrm{leaf}}.$
Because each leaf has size $n_0 \times n_0,$ forming the leaf decay tensors across all $2^r$ leaves materializes only $2^r(n_0^2 d_k) = C n_0 d_k$ elements per chunk, a $2^r\times$ reduction compared to the full $C^2 d_k$ 3D decay tensor.
Moreover, because the strictly lower-triangular matrix $\mL_{\mathrm{leaf}} \in \R{n_0 \times n_0}$ is nilpotent of index $n_0$ ($\mL_{\mathrm{leaf}}^{n_0} = \mZero$), its Neumann series terminates at degree $n_0 - 1$ and factors into $\log_2 n_0$ multiplicative stages ($\log_2 n_0 - 1$ squaring steps; \citealp{Orcutt1974-js,Heller1978-ni}):
\begin{equation}
\label{eq:neumann}
    \mT_{\mathrm{leaf}}\inv
    = \sum_{k=0}^{n_0-1}(-\mL_{\mathrm{leaf}})^k
    = \prod_{j=0}^{\log_2 n_0-1}\big(\I + (-\mL_{\mathrm{leaf}})^{2^j}\big),
\end{equation}
where $n_0 \ll C$ bounds nilpotent matrix powers in \texttt{bfloat16} \citep{Sobczyk2026-lr}.


\xxhdr{Inductive Step: Bottom-Up Merge via Boundary-Pivoted Splitting ($n_m = n_0 2^m$)}
At each merge level ($m = 0, \dotsc, r-1$), a parent interval of length $2n_m$ merges adjacent left ($\gL$) and right ($\gR$) child intervals of length $n_m,$ for which the diagonal blocks $(\mT_\gL\inv, \mM_\gL)$ and $(\mT_\gR\inv, \mM_\gR)$ have already been computed.
By \eqref{eq:block_inv}, completing the parent blocks requires forming the dense $n_m \times n_m$ cross-blocks $\mT_{\mathrm{cross}}$ and $\mM_{\mathrm{cross}}$ and multiplying $-\mT_\gR\inv\mT_{\mathrm{cross}}\mT_\gL\inv$ for the bottom-left block of $\mT_{\mathrm{blk}}\inv.$
To compute $\mT_{\mathrm{cross}}$ and $\mM_{\mathrm{cross}}$ via matrix multiplications without dividing by $\vpi_s = \exp(\mLambda_s),$ we exploit the fact that every cross-interval pair $(t, s) \in \gR \times \gL$ straddles the interval boundary between $\gL$ and $\gR.$
%
Extending sub-chunk boundary decay splitting \citep{Yang2023-pq,Yang2024-vn,Kimi-Team2025-rg} to every merge level of the tree, we choose the \emph{boundary pivot} $p \defas \mathrm{last}(\gL)$ with cumulative log-decay $\mLambda_p \in \Rvec{d_k}$ and split the log-decay difference across $p$:
\[
    \vpi_{t,s}
    = \exp(\mLambda_t - \mLambda_s)
    = \exp\big((\mLambda_t - \mLambda_p) + (\mLambda_p - \mLambda_s)\big)
    = \underbrace{\exp(\mLambda_t - \mLambda_p)}_{\vpi_{t,p}} \odot \underbrace{\exp(\mLambda_p - \mLambda_s)}_{\vpi_{p,s}},
    \qquad t \in \gR,\; s \in \gL.
\]
Because cumulative log-decay is monotonically non-increasing ($\mLambda_1 \ge \dotsb \ge \mLambda_C$) and $s \le p < t$ for all $s \in \gL$ and $t \in \gR,$ both $\mLambda_t - \mLambda_p \le \vzero$ and $\mLambda_p - \mLambda_s \le \vzero$ hold identically, keeping both exponential factors in $(0, 1]$ without division or overflow.
Each cross-block entry therefore decouples across $p$ as $(\mL_{\mathrm{strict}})_{t,s} = (\hat\vk_t \odot \vpi_{t,p})(\vk_s \odot \vpi_{p,s})\tr$ and $\emM_{t,s} = (\hat\vq_t \odot \vpi_{t,p})(\vk_s \odot \vpi_{p,s})\tr.$
Stacking $\vpi_{t,p}$ ($t \in \gR$) and $\vpi_{p,s}$ ($s \in \gL$) as the rows of $\vpi_\gR, \vpi_\gL \in (0, 1]^{n_m \times d_k}$ computes both $\mT_{\mathrm{cross}}$ and $\mM_{\mathrm{cross}}$ in a single row-packed matrix multiplication sharing $(\mK_\gL \odot \vpi_\gL)\tr,$ followed by two matrix multiplications for the bottom-left inverse block $\mX_{21} \defas (\mT_{\mathrm{blk}}\inv)_{\gR,\gL}$ in \eqref{eq:block_inv}:
\[
    \begin{bmatrix}
        \mT_{\mathrm{cross}} \\
        \mM_{\mathrm{cross}}
    \end{bmatrix}
    =
    \begin{bmatrix}
        \hat\mK_\gR \odot \vpi_\gR \\
        \hat\mQ_\gR \odot \vpi_\gR
    \end{bmatrix}
    \big(\mK_\gL \odot \vpi_\gL\big)\tr,
    \qquad
    \mX_{21} = \big(\mT_{\mathrm{blk}}\inv\big)_{\gR,\gL} = -\mT_\gR\inv\mT_{\mathrm{cross}}\mT_\gL\inv.
\]
%
After $r$ merge levels, $\mT\inv$ and $\mM$ are assembled across the full chunk without materializing $\mT$ via prefix division, yielding $\begin{bmatrix}\mU & \mY\end{bmatrix} = \mT\inv\begin{bmatrix}\hat\mV & \hat\mK_{\odot\pi}\end{bmatrix}$ and the affine state update \eqref{eq:chunk_state_out}.


\xhdr{Custom Backward Pass}

In the backward pass, differentiating the chunk readout and state update \eqref{eq:chunk_state_out} in stage (iii) requires only matrix multiplications to compute the gradients $\mathrm{d}\mZ = \begin{bmatrix}\mathrm{d}\mU & \mathrm{d}\mY\end{bmatrix}$ and $\mathrm{d}\mM$ (where $\mathrm{d}\mX \defas \partial\gJ/\partial\mX$ for scalar loss $\gJ$).
However, naively differentiating through stages (i) and (ii) via automatic differentiation would retain all $r$ tree levels of intermediate inverse sub-blocks and scaled key/query tensors in HBM and backpropagate sequentially through the $r$ block-inversion levels \eqref{eq:block_inv} and $\log_2 n_0-1$ leaf-squaring steps.
Instead, we define a custom backward pass for stages (i) and (ii) that saves only the assembled inverse $\mT\inv$ and solved right-hand side $\mZ$ along with the chunk inputs.

First, to backpropagate through stage (ii), differentiating $\mT\mZ = \begin{bmatrix}\hat\mV & \hat\mK_{\odot\pi}\end{bmatrix}$ with $\mT = \I + \mL_{\mathrm{strict}}$ yields the gradients $\mathrm{d}\!\begin{bmatrix}\hat\mV & \hat\mK_{\odot\pi}\end{bmatrix}$ and $\mathrm{d}\mL_{\mathrm{strict}} \in \R{C\times C}$ directly from the saved $\mT\inv$ and $\mZ$ in two full-chunk MXU matrix multiplications without backpropagating through the forward inversion tree:
\[
    \mathrm{d}\!\begin{bmatrix}\hat\mV & \hat\mK_{\odot\pi}\end{bmatrix} = \mT^{-\top}\mathrm{d}\mZ, \qquad
    \mathrm{d}\mL_{\mathrm{strict}} = -\stril\!\left(\mathrm{d}\!\begin{bmatrix}\hat\mV & \hat\mK_{\odot\pi}\end{bmatrix}\mZ\tr\right),
\]
%
where $\mathrm{d}\!\begin{bmatrix}\hat\mV & \hat\mK_{\odot\pi}\end{bmatrix}\mZ\tr$ uses three-pass \texttt{bfloat16} MXU multiplication with \texttt{float32} accumulation so that the log-decay gradients driven by $\mathrm{d}\mL_{\mathrm{strict}}$ retain high accuracy without falling back to the VPU.

Second, to backpropagate $(\mathrm{d}\mL_{\mathrm{strict}}, \mathrm{d}\mM)$ through stage (i) to $(\mathrm{d}\hat\mQ, \mathrm{d}\mK, \mathrm{d}\hat\mK, \mathrm{d}\mLambda)$ without storing intermediate tree activations in HBM, we traverse the same depth-$r$ boundary-pivoted binary tree and differentiate the $2^r$ diagonal leaves of size $n_0$ in parallel across leaves.
At each merge level $m \in \{0, \dotsc, r-1\}$ with child size $n_m = n_0 2^m,$ we recompute the boundary-pivot scalings $\vpi_\gR = \exp(\mLambda_\gR - \mLambda_p)$ and $\vpi_\gL = \exp(\mLambda_p - \mLambda_\gL)$ on the fly in VMEM from $\mLambda$ and compute the scaled query and key gradients from the off-diagonal blocks $\mathrm{d}\mT_{\mathrm{cross}} \defas (\mathrm{d}\mL_{\mathrm{strict}})_{\gR,\gL}$ and $\mathrm{d}\mM_{\mathrm{cross}} \defas (\mathrm{d}\mM)_{\gR,\gL} \in \R{n_m \times n_m}$ in two row-packed MXU matrix multiplications:
\[
    \begin{bmatrix}
        \mathrm{d}(\hat\mK_\gR \odot \vpi_\gR) \\
        \mathrm{d}(\hat\mQ_\gR \odot \vpi_\gR)
    \end{bmatrix}
    =
    \begin{bmatrix}
        \mathrm{d}\mT_{\mathrm{cross}} \\
        \mathrm{d}\mM_{\mathrm{cross}}
    \end{bmatrix}
    \big(\mK_\gL \odot \vpi_\gL\big),
    \qquad
    \mathrm{d}(\mK_\gL \odot \vpi_\gL)
    =
    \begin{bmatrix}
        \mathrm{d}\mT_{\mathrm{cross}} \\
        \mathrm{d}\mM_{\mathrm{cross}}
    \end{bmatrix}\tr
    \begin{bmatrix}
        \hat\mK_\gR \odot \vpi_\gR \\
        \hat\mQ_\gR \odot \vpi_\gR
    \end{bmatrix},
\]
before applying the elementwise chain rule through $\vpi_\gR$ and $\vpi_\gL$ into $(\mathrm{d}\hat\mQ, \mathrm{d}\mK, \mathrm{d}\hat\mK, \mathrm{d}\mLambda)$ and converting $\mathrm{d}\mLambda$ to $\mathrm{d}\log\mAlpha = \tril(\vone_{C\times C})\tr\mathrm{d}\mLambda = \triu(\vone_{C\times C})\mathrm{d}\mLambda$ on the MXU (full stage-(iii) backward scan equations and elementwise exponential derivatives are in \Cref{app:kernel:bwd}).


\begin{figure}[!t]
    \centering
    \includegraphics[width=\textwidth]{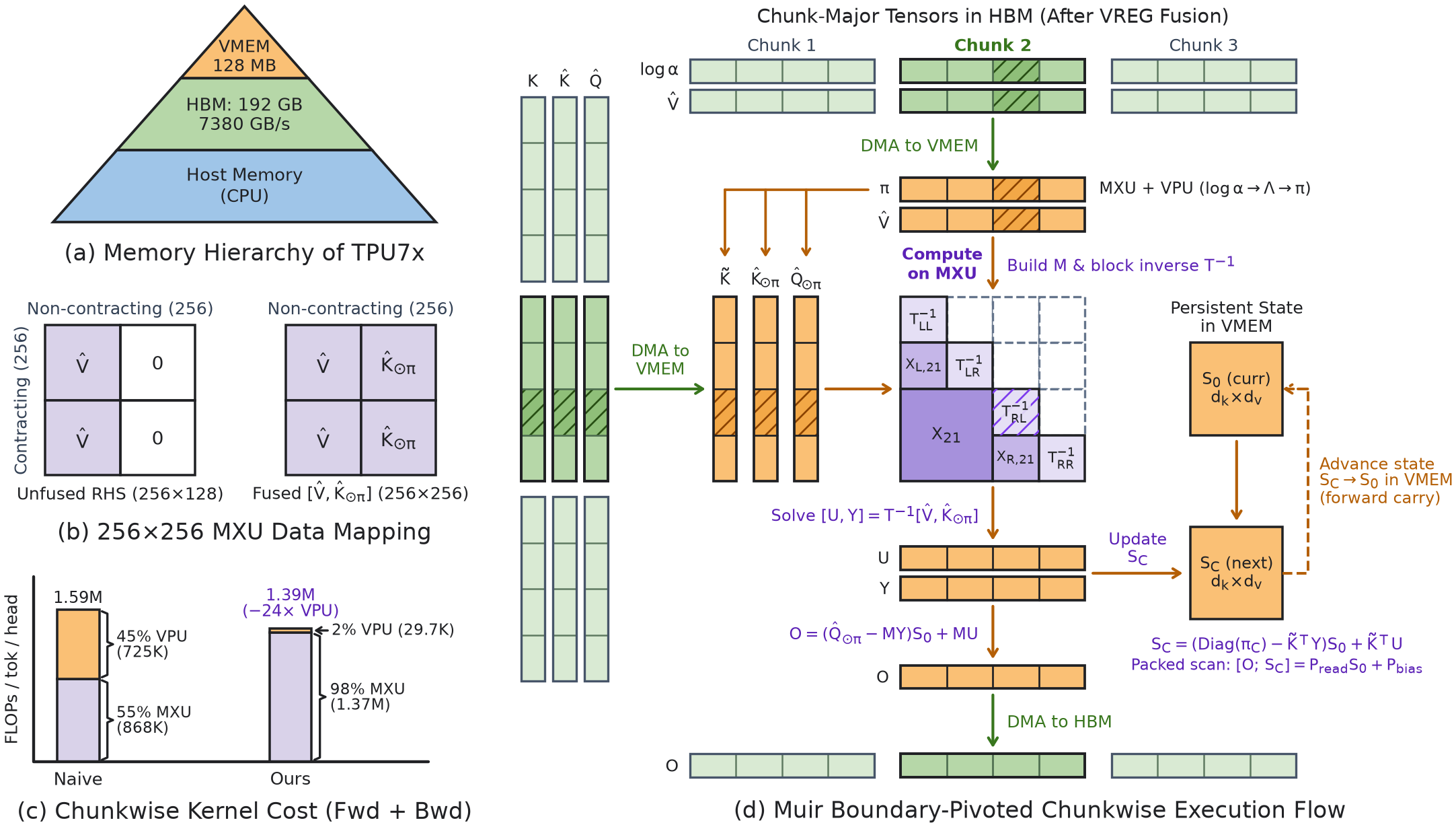}
    \caption{%
        \textbf{Boundary-pivoted chunkwise kernel eliminates 3D decay tensors, division overflow, and VPU bottlenecks.}
        (a) Memory hierarchy of a TPU7x chip.
        (b) On the $256 \times 256$ MXU, an unfused $256 \times 128$ right-hand side $\hat\mV$ achieves only $50\%$ utilization due to zero-padded lanes, whereas the fused solve $\begin{bmatrix}\mU & \mY\end{bmatrix} = \mT\inv\begin{bmatrix}\hat\mV & \hat\mK_{\odot\pi}\end{bmatrix}$ achieves $100\%$ utilization.
        %
        (c) At chunk size $C = 256,$ leaf size $8,$ and $d_k = d_v = 128,$ replacing naive $C \times C \times d_k$ 3D VPU decay and sequential VPU triangular solves with boundary-pivoted MXU matrix multiplications, recursive block inversion, and a custom backward pass cuts forward-and-backward VPU compute by $24\times$ and total kernel FLOPs by $12\%,$ routing $98\%$ of chunkwise arithmetic onto the MXU.
        %
        (d) Boundary-pivoted chunkwise execution combines pre-permutation VREG fusion for $\hat\vq_t$ and $\hat\vv_t,$ MXU cumulative log-decay, numerically stable VPU decay splitting, recursive block inversion of $\mT$ on the MXU, and a packed MXU scan carrying $\mS_0 \to \mS_C$ in VMEM (saving $\mS_0$ only for backward).%
    }
    \label{fig:kernel}
\end{figure}

\xhdr{TPU Implementation and Batched Execution}

\Figref{fig:kernel}{d} traces how this algorithm maps onto the TPU7x memory hierarchy (\Figref{fig:kernel}{a}).
The projections and depthwise convolutions produce activations in token-major order $(B, T, N_h, d_k),$ whereas chunkwise execution operates on chunks of length $C$ per head and requires permuting axes into chunk-major order $(N_c, B, N_h, C, d_k)$ with $T = N_c C.$
%
%
Because permuting the trailing tiled dimensions from $(N_h, d_k)$ to $(C, d_k)$ incurs an HBM round-trip per tensor, and \eqref{eq:ours} depends on $(\vq_t, \vgamma_t)$ and $(\vv_t, \vw_t)$ only through $\hat\vq_t = \vgamma_t \odot \vq_t$ and $\hat\vv_t = \vw_t \odot \vv_t,$ we fuse $\hat\vq_t$ and $\hat\vv_t$ in VREGs before permutation (\Figref{fig:arch}{b}) alongside $\hat\vk_t = \vb_t \odot \vk_t$ (retaining $\vk_t$), reducing the token-to-chunk permutation from seven tensors to five: $(\log\mAlpha, \hat\mV, \mK, \hat\mK, \hat\mQ).$

We set $C = 256$ to match the $256 \times 256$ MXU tile size and $n_0 = 8$ ($r = 5$ levels, $32$ leaves per chunk) to match the $8$-row $32$-bit \texttt{T(8,\,128)} VMEM tile for \texttt{float32} leaf buffers during \texttt{bfloat16} MXU squaring (before casting $\mT_{\mathrm{leaf}}\inv$ to \texttt{bfloat16}); \Figref{fig:kernel}{d} illustrates a two-level tree with four leaves $\mT_{\gL\gL}\inv, \mT_{\gL\gR}\inv, \mT_{\gR\gL}\inv, \mT_{\gR\gR}\inv$ and off-diagonal inverse blocks $\mX_{\gL,21}, \mX_{\gR,21},$ and $\mX_{21}.$
To evaluate all nodes at a given tree depth simultaneously, reshaping the chunk axis into $(32, 8)$ at the leaf level folds all $32$ leaves into the leading batch dimensions $(N_c, B, N_h, 32),$ so leaf evaluation and the two Neumann squaring steps run in parallel across all leaves, chunks, and heads.
Likewise, at each merge level $m \in \{0, \dotsc, 4\},$ reshaping the chunk axis into $(N_m, 2n_m)$ with $N_m = 2^{4-m}$ parent nodes folds all $N_m$ merges across all chunks and heads into batched MXU contractions, so the full tree completes in $r = 5$ sequential stages in both the forward and backward passes.

Once $\mT\inv$ and $\mM$ are assembled, concatenating $\begin{bmatrix}\hat\mV & \hat\mK_{\odot\pi}\end{bmatrix} \in \R{256 \times 256}$ along columns (for $d_k = d_v = 128$) computes $\begin{bmatrix}\mU & \mY\end{bmatrix} = \mT\inv\begin{bmatrix}\hat\mV & \hat\mK_{\odot\pi}\end{bmatrix}$ (and $\mT^{-\top}\mathrm{d}\mZ$ in the backward pass) in a single $256 \times 256$ MXU matrix multiplication per chunk at $100\%$ tile utilization, whereas unfused $256 \times 128$ solves drop to $50\%$ utilization (\Figref{fig:kernel}{b}).
Similarly, stacking $\mM$ and $\tilde\mK\tr$ along rows into a $(C+d_k)\times C$ matrix multiplies $\begin{bmatrix}\mU & \mY\end{bmatrix}$ in one packed MXU contraction across all chunks in parallel to form $(\mP_{\mathrm{read}}, \mP_{\mathrm{bias}})$ of \eqref{eq:chunk_state_out}:
\[
    \begin{bmatrix} \mO \\ \mS_C \end{bmatrix}
    = \underbrace{\left(\begin{bmatrix} \hat\mQ_{\odot\pi} \\ \Diag(\vpi_C) \end{bmatrix} - \begin{bmatrix} \mM \\ \tilde\mK\tr \end{bmatrix}\mY\right)}_{\mP_{\mathrm{read}}}\mS_0
    + \underbrace{\begin{bmatrix} \mM \\ \tilde\mK\tr \end{bmatrix}\mU}_{\mP_{\mathrm{bias}}}.
\]
%
Because the per-head state $\mS_0 \in \R{128\times 128}$ (\texttt{float32}, $64\text{ KB}$ per head) fits in on-chip VMEM ($128\text{ MB}$ per chip), this packed scan carries $\mS_0 \to \mS_C$ in VMEM during the forward sweep and writes chunk-boundary $\mS_0$ to HBM only for the backward pass.
Over a full training step (\Figref{fig:kernel}{c} and \Cref{app:kernel:flops}), this kernel executes $1.37\text{M}$ MXU FLOPs ($98\%$) and only $29.7\text{K}$ VPU FLOPs ($2\%$) per token per head, which reduces VPU compute by $24\times$ and total chunkwise FLOPs by $12\%$ relative to naive chunkwise execution.

%% file: sections/5_results.tex
\section{Experiments}
\label{sec:exp}

We detail our experimental setup in \S\ref{sec:exp:setup}, present main results on language modeling, recall, length extrapolation, training throughput, and numerical stability in \S\ref{sec:exp:results}, and ablate \ours in \S\ref{sec:exp:abl}.


\subsection{Experimental Setup}
\label{sec:exp:setup}

\xhdr{Models and Baselines}

We evaluate all models across four iso-parameter scales: $166\text{M}$ ($L = 8,$ $d = 1{,}024$), $406\text{M}$ ($L = 11,$ $d = 1{,}536$), $834\text{M}$ ($L = 14,$ $d = 2{,}048$), and $1.34\text{B}$ ($L = 24,$ $d = 2{,}048$), where all architectures at a given scale share depth $L,$ residual width $d,$ head dimension $d_k = d_v = 128$ ($N_h = d/128$ heads), a $\gelu$ FFN, and untied input and output embeddings (full per-scale configurations are provided in \Cref{app:train:pretrain}).
Under the $12d^2$ per-layer non-embedding parameter budget, we compare five model variants across all four scales:
(i) \tf, with rotary position embeddings ($\mathrm{RoPE},$ $\theta = 10{,}000$; \citealp{Su2021-oc}), FFN width $d_{\mathrm{ffn}} = 4d,$ and TPU Pallas Splash Attention \citep{Dao2022-vw};\footnote{\url{https://github.com/jax-ml/jax/tree/main/jax/experimental/pallas/ops/tpu/splash_attention}.}
(ii) \gdntwo, the channel-wise gated delta recurrence with rank-$128$ $(f, g)$ gates, dense $d \times d$ $(b, w)$ gates, and $d_{\mathrm{ffn}} = 3d - 256$;\footnote{%
    Under the same $1.34\text{B}$ FineWeb-Edu setup, \citet{Hatamizadeh2026-cw} showed that \gdntwo consistently outperforms scalar-decay Gated DeltaNet \citep{Yang2025-dr}, Kimi Linear \citep{Kimi-Team2025-rg}, and Mamba-3 \citep{Lahoti2026-fc} across language modeling and retrieval, so we use \gdntwo as our recurrent baseline.%
}
(iii) \ours (no $\vgamma$), which replaces the gate projections in \gdntwo with the fused low-rank projection of ranks $(r_f, r_b, r_g, r_w, r_q) = (64, 64, 128, 256, 0)$ and unit readout ($\vgamma_t = \vone$);
(iv) \ours (static $\vgamma$), which adds the learned static readout gain $\vgamma_t = \vgamma^{\mathrm{stat}}$ at the same gate ranks; and
(v) \ours (ours), which splits the $256$-column write budget into $(r_w, r_q) = (128, 128)$ to parameterize the dynamic readout gain $\vgamma_t.$
Across all recurrent models (ii)--(v), the chunkwise recurrence is executed with chunk length $C = 256,$ and all three \ours variants (iii)--(v) share total gate rank $R_{\mathrm{tot}} = 512$ and FFN width $d_{\mathrm{ffn}} = 4d - 512.$


\xhdr{Pretraining Recipe}

Following \citet{Hatamizadeh2026-cw}, we pretrain all models from scratch on $100\text{B}$ tokens sampled from FineWeb-Edu \citep{Penedo2024-cr}, tokenized with the $32{,}000$-token Llama-2 SentencePiece tokenizer \citep{Touvron2023-qi} and presented in identical token order across all runs.
We optimize all models using AdamW \citep{Loshchilov2017-co} ($\beta_1 = 0.9,$ $\beta_2 = 0.999$) with weight decay $0.1$ on weight matrices ($0.0$ on additive biases, $a_{\log},$ and $\vgamma^{\mathrm{stat}}$), global gradient norm clipping at $1.0,$ and a $1\text{B}$-token ($1\%$) linear warmup to a peak learning rate of $4\times 10^{-4}$ followed by cosine decay to $4\times 10^{-5}$ ($0.1\times$ peak).
Training runs on $8$ TPU7x chips in \texttt{bfloat16} mixed precision (with \texttt{float32} master weights, recurrent state $\mS_t,$ and cumulative log-decay $\mLambda_t$) at sequence length $T = 4{,}096$ and a global batch size of $524{,}288\text{ tokens/step}.$


\xhdr{Evaluation}

We evaluate language modeling and commonsense reasoning using the \texttt{lm-evaluation-harness} \citep{Gao2024-jc}.
We report held-out FineWeb-Edu validation \emph{token} negative log-likelihood (NLL, over $100\text{M}$ tokens) and \emph{word-level} perplexity on WikiText \citep{Merity2016-re} and LAMBADA \citep{Paperno2016-vx}.
We also report zero-shot accuracy on LAMBADA, PIQA \citep{Bisk2020-yj}, HellaSwag \citep{Zellers2019-ht}, WinoGrande \citep{Sakaguchi2021-vw}, ARC-Easy and ARC-Challenge \citep{Clark2018-be}, OpenBookQA \citep{Mihaylov2018-pm}, and Social IQA \citep{Sap2019-ai}, as well as 5-shot WinoGrande and closed-book TriviaQA \citep{Joshi2017-fl}.
Following \citet{Yang2025-dr,Lahoti2026-fc}, we report length-normalized accuracy for HellaSwag and ARC-Challenge, and unnormalized accuracy for all other tasks.\footnote{%
    We omit zero-shot BoolQ \citep{Clark2019-zw} from the main text because it hovers near its $62.2\%$ majority-class baseline with high single-checkpoint variance across sub-$1.5\text{B}$ models \citep{Madaan2024-bx,Miller2024-aw,Heineman2025-ji}; BoolQ accuracy and the 9-task average are reported in \Cref{app:eval:lm}.%
}

To evaluate associative recall, we test synthetic retrieval and state tracking on six tasks from RULER \citep{Hsieh2024-ji} ($500$ examples per task and context length, spanning uniform needle depths $0\text{--}100\%$): single-needle retrieval (S-NIAH-1/2/3, testing $7$-digit numbers and $36$-character UUIDs in synthetic noise and Paul Graham essays), as well as $5$-hop variable tracking (VT), at context lengths $1\text{K}\text{--}8\text{K}$ ($2\times$ the $4\text{K}$ training window), alongside multi-value (MV-NIAH, $4$ values bound to one key) and multi-query (MQ-NIAH, $4$ distinct query keys) retrieval across the in-distribution lengths $1\text{K}\text{--}4\text{K}$ (additional details in \Cref{app:eval:recall}).
We also evaluate real-world extractive recall at $2\text{K}$ context on the six Based/JRT benchmarks \citep{Arora2024-lm,Arora2024-at}: SWDE \citep{Lockard2019-hy}, SQuAD \citep{Rajpurkar2018-za}, FDA \citep{Arora2023-nn}, TriviaQA \citep{Joshi2017-fl}, Natural Questions \citep{Kwiatkowski2019-gp}, and DROP \citep{Dua2019-lu}.

For length extrapolation, we test the $1.34\text{B}$ models both zero-shot from $4\text{K}$ pretraining and after a $300$-step $64\text{K}$ cooldown ($0.315\text{B}$ held-out FineWeb-Edu tokens, or $0.3\%$ of pretraining, using a $70/30$ token mix of $\ge\!16\text{K}$ and standard documents; \citealp{Fu2024-bl}).
%
Our \tf baseline uses training-free Dynamic YaRN \citep{Peng2024-tz} beyond $4\text{K}$ in the zero-shot regime and raises the RoPE base to $\theta = 5\times 10^5$ during cooldown \citep{Xiong2024-rg} with Dynamic YaRN beyond $64\text{K}$ (\Cref{app:train:longctx}).
%
Language-modeling extrapolation covers $64$ contiguous $128\text{K}$-token documents ($S = 128\text{K},$ $T \in \{4\text{K}, \dotsc, 128\text{K}\}$) and three continuous $10\text{M}$-token streams ($S = 10\text{M},$ $T \in \{4\text{K}, \dotsc, 10\text{M}\}$) from FineWeb-Edu validation, where each sequence is partitioned into $S / T$ non-overlapping reset windows so every $T$ scores identical target tokens; to separate long-range state drift from fewer boundary resets, we pair $128\text{K}$ perplexity with $\Delta_{\mathrm{tail}} \defas \mathrm{NLL}_{[127\text{K},128\text{K})}(T = 128\text{K}) - \mathrm{NLL}_{[127\text{K},128\text{K})}(T = 4\text{K}).$
%
Long-context passkey retrieval is evaluated on RULER S-NIAH-1 from $1\text{K}$ to $1\text{M}$ tokens ($160$ examples across $10$ uniform depth bins at each of $11$ context lengths).

Finally, we measure end-to-end training throughput ($\text{K tokens/s}$) on $8$ TPU7x chips at a fixed batch size of $524{,}288\text{ tokens/step}$ across sequence lengths $T \in \{2\text{K}, 4\text{K}, 8\text{K}, 16\text{K}, 32\text{K}\}$ for \ours against \gdntwo and \tf.
We also verify the numerical accuracy and conditioning of the boundary-pivoted $C = 256$ chunkwise kernel against an exact \texttt{float64} sequential reference.


\subsection{Results}
\label{sec:exp:results}

\subsubsection{Language Modeling and Commonsense Reasoning}

\begin{figure}[!t]
    \centering
    \includegraphics[width=\textwidth]{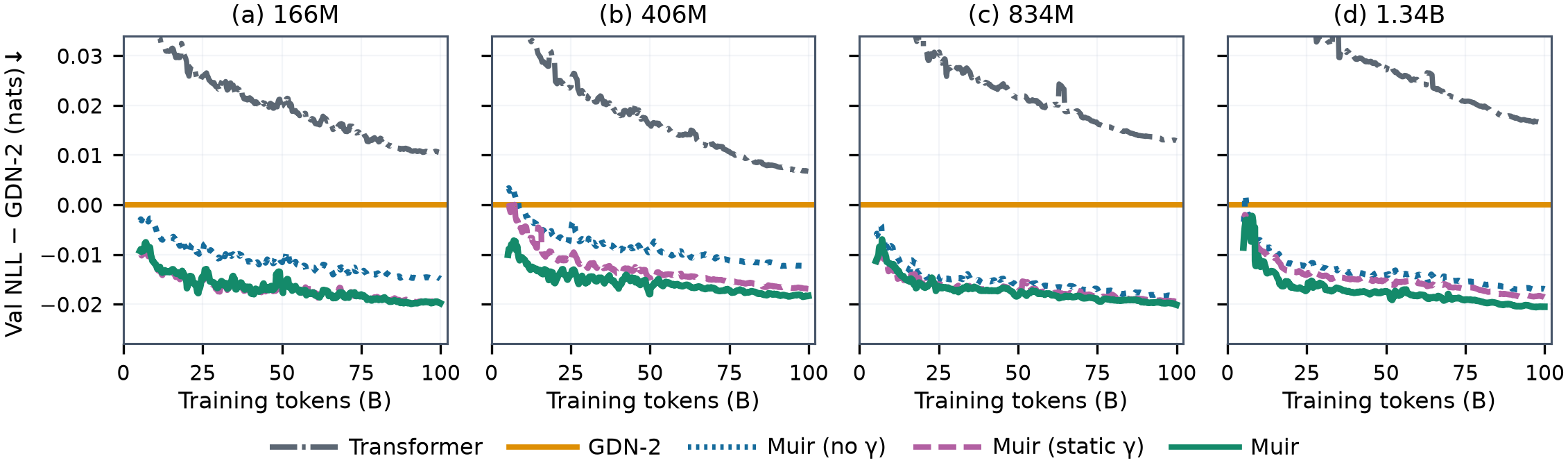}
    \caption{%
        \textbf{\ours consistently lowers validation NLL relative to \gdntwo and \tf throughout $\boldsymbol{100\text{B}}$-token pretraining.}
        Held-out FineWeb-Edu validation token NLL difference relative to \gdntwo across $166\text{M}\text{--}1.34\text{B}$ parameters as a function of training tokens.
        All three \ours variants improve over \gdntwo within the first $10\text{B}$ tokens and reach a $12\text{--}20\text{ mnat}$ reduction by $100\text{B}$ tokens, while static and dynamic matched-filter readout provide additional reductions over \ours (no $\vgamma$).%
    }
    \label{fig:nll_curves}
\end{figure}

Across all four model scales (\Cref{fig:nll_curves,tab:main_lm}), all three \ours variants outperform \gdntwo and \tf on validation NLL, WikiText perplexity, and the 8-task zero-shot average.


\xhdr{Pretraining and Scaling Dynamics}

In \Cref{fig:nll_curves}, all three \ours variants improve over \gdntwo within the first $10\text{B}$ tokens and reach a $12\text{--}20\text{ millinat (mnat)}$ reduction by $100\text{B}$ tokens because reallocating the freed gate parameters to widen the FFN (from $d_{\mathrm{ffn}} = 3d - 256$ to $4d - 512$) expands token-level representation capacity from the start of optimization.
%
As training proceeds and the recurrent heads specialize their decay gates $\valpha_t$ across timescales, the resulting stationary row-energy imbalance (\Cref{prop2}) creates headroom for the matched-filter readout (\Cref{prop3,prop3:cor1}), and static and dynamic $\vgamma_t$ maintain consistent additional gains over \ours (no $\vgamma$) throughout training.
Across all four scales, \ours reaches the $100\text{B}$-token validation NLL of \gdntwo by roughly $85\text{B}$ tokens ($1.2\times$ faster), and at $1.34\text{B}$ (\Cref{tab:main_lm}), full \ours lowers final validation NLL by $20.4\text{ mnats}$ relative to \gdntwo and by $36.2\text{ mnats}$ relative to \tf, with matched-filter readout accounting for $3.3\text{ mnats}$ of the improvement over \ours (no $\vgamma$).

\input{tables/tab1_main_lm}


\xhdr{Short-Context Parametric Knowledge and Context-Dependent Readout}

Standard zero-shot multiple-choice benchmarks (PIQA, HellaSwag, WinoGrande, ARC, OBQA, and SIQA) use short prompts ($<\!100$ tokens) that place little load on long-range state retention and mainly probe world knowledge in FFN weights \citep{Geva2021-vm,Meng2022-qp}.
%
%
Whereas \gdntwo improves temporal mixing over \tf ($d_{\mathrm{ffn}} = 4d$) at the cost of a $28\text{--}31\%$ narrower FFN, \ours retains the gated-delta mixer while restoring near-\tf FFN width; with this shared FFN expansion, all three \ours variants achieve comparable gains over \gdntwo and \tf across scales.
%
At $1.34\text{B},$ \ours outperforms \gdntwo on all eight zero-shot tasks and raises the 8-task average by $+1.38$ points over \gdntwo ($+1.79$ over \tf), while the three \ours variants remain within single-checkpoint sampling noise ($0.4\text{--}0.7$ points) at smaller scales.

By contrast, tasks that require resolving predictions from extended context isolate the readout filter.
On LAMBADA word-level perplexity (predicting the final word of a passage from the preceding context) and on 5-shot WinoGrande and closed-book TriviaQA (extracting task formatting and entity-answer patterns across five in-context demonstrations while suppressing interference from earlier examples), FFN expansion alone (\ours no $\vgamma$) provides almost no gain over \gdntwo at $1.34\text{B}$ ($10.26$ vs.\ $10.38$ on LAMBADA perplexity; $16.52\%$ vs.\ $16.48\%$ on 5-shot TriviaQA).
Introducing the dynamic readout gain $\vgamma_t$ drives the improvement on these context-sensitive evaluations, lowering $1.34\text{B}$ LAMBADA perplexity to $10.00$ (achieving the lowest perplexity at $166\text{M},$ $834\text{M},$ and $1.34\text{B}$)\footnote{%
    Because LAMBADA scores only the final word of $5{,}153$ passages, its perplexity has higher variance at smaller scales ($\pm 0.67$ SE at $406\text{M},$ where \ours no $\vgamma$ attains $18.11$ vs.\ $20.00$ for \ours despite the lower validation NLL of \ours).%
}
and raising 5-shot TriviaQA accuracy to $19.32\%.$
Consistent with \Cref{prop3,prop3:rem1}, processing a multi-shot prompt requires querying both local syntactic context at short lags and demonstration patterns at lags of several hundred tokens, a bimodal retrieval mixture that dynamic $\vgamma_t$ resolves where a static gain vector cannot.


\begin{figure}[!t]
    \centering
    \includegraphics[width=\textwidth]{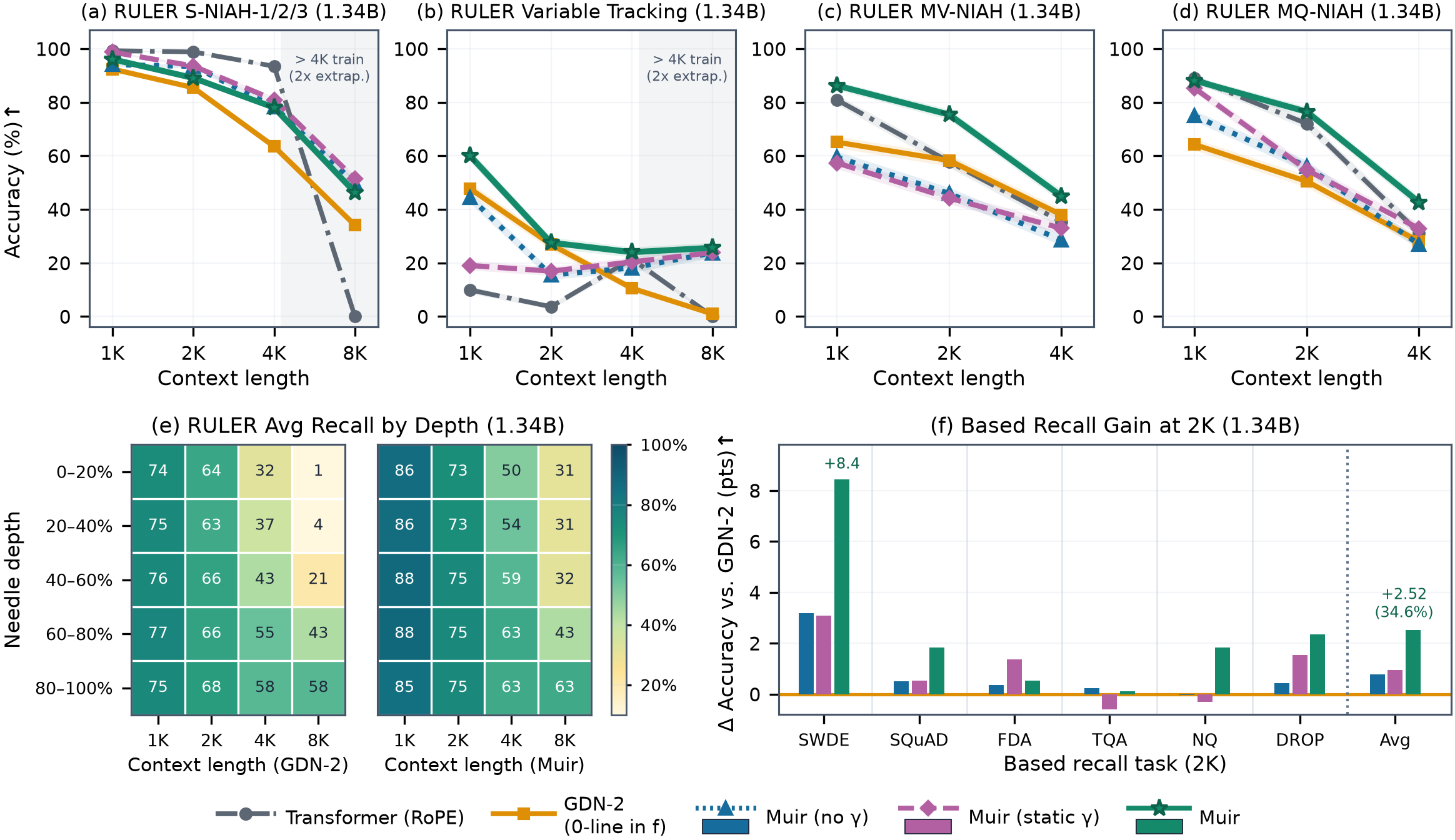}
    \caption{%
        \textbf{\ours improves synthetic single-needle, multi-hop, multi-needle, and real-world retrieval over \gdntwo.}
        (a) Averaged across the three single-needle tasks, all \ours variants mitigate the $4\text{K}$ degradation of \gdntwo and extrapolate to $8\text{K},$ where unextended RoPE \tf degrades beyond its $4\text{K}$ training window.
        (b) On $5$-hop variable-binding chains, static $\vgamma$ fails to match the varying hop lags, whereas dynamic $\vgamma_t$ leads all models at every context length.
        (c, d) When retrieving four values bound to a single key or four distinct keys queried jointly, dynamic $\vgamma_t$ matches or outperforms both \gdntwo and \tf across $1\text{K}\text{--}4\text{K}.$
        (e) Averaged across the RULER tasks (a--d) by needle depth and context length, \ours achieves the largest gains over \gdntwo on early needles ($0\text{--}40\%$ depth) at $4\text{K}\text{--}8\text{K}.$
        (f) On real-world document extraction at $2\text{K},$ \ours improves over \gdntwo on all six tasks, with dynamic $\vgamma_t$ driving $69\%$ of the gain.%
    }
    \label{fig:recall}
\end{figure}

\subsubsection{Associative Recall}
\label{sec:exp:results:recall}

\Cref{fig:recall} evaluates synthetic retrieval across the six RULER tasks and real-world document extraction across the six Based/JRT tasks at $1.34\text{B}$ parameters (complete per-task and per-length tables are in \Cref{app:eval:recall}).


\xhdr{Single-Needle, Multi-Hop, and Multi-Needle Retrieval}

Across the three single-needle tasks (S-NIAH-1/2/3; \Figref{fig:recall}{a}), unweighted readout in \gdntwo degrades as context length increases.\footnote{%
    While our \gdntwo baseline matches \citet{Hatamizadeh2026-cw} on language modeling and zero-shot tasks, S-NIAH scores differ because we use \texttt{lm-evaluation-harness} instead of \texttt{NVIDIA/RULER} (\Cref{app:eval:recall}).%
}
As established in \Cref{prop4,prop4:rem1}, unit readout ($\vgamma_t = \vone$) forces the recurrence to rely on heavier write-time erasure $\bar\evb$ to keep multi-write interference bounded in high-energy rows, which shortens the effective memory horizon $\tau_{\mathrm{eff},j}(\bar\evb)$ and flushes distant needles over thousands of intervening tokens.
%
Calibrating channel readout (\Cref{prop3,prop3:cor1}) lowers the equilibrium erase rate and raises the $1\text{K}\text{--}8\text{K}$ single-needle mean over both baselines, with \ours holding $100\%$ on S-NIAH-1 through $8\text{K}.$
Because a single needle is queried at one depth at the end of the prompt, a fixed channel filter (\ours static $\vgamma$) suffices to suppress fast-row noise on single-key lookups, and pairing static $\vgamma$ with the wider write gate ($r_w = 256$ vs.\ $128$ in full \ours at fixed $R_{\mathrm{tot}} = 512$) reaches a slightly higher S-NIAH-1/2/3 average, particularly on S-NIAH-3 where writing a $36$-character UUID across roughly $34$ tokens benefits from higher write-gate rank.

By contrast, resolving a $5$-hop pointer chain on VT (\Figref{fig:recall}{b}) or sequentially retrieving four values on MV-NIAH (\Figref{fig:recall}{c}) and MQ-NIAH (\Figref{fig:recall}{d}) requires querying associations stored at multiple depths and shifting the readout filter across state channels as each answer token is generated.
Because a single static gain vector cannot match multiple retrieval lags simultaneously (\Cref{prop3:rem1}), static $\vgamma$ falls below unweighted \ours on VT and below \gdntwo on MV-NIAH.
Reallocating half of the write-gate rank to parameterize token-dependent $\vgamma_t$ trades single-UUID write capacity for lag-adaptive readout, enabling \ours to outperform all recurrent and attention baselines and achieve the highest overall RULER suite average.


\xhdr{Depth Horizon and Real-World Extractive Recall}

In the depth-by-length breakdown (\Figref{fig:recall}{e}), both \gdntwo and \ours maintain uniform recall across needle depths at $1\text{K}\text{--}2\text{K}$ (with \ours leading by $7\text{--}12$ points), whereas decoupling query-time noise suppression from write-time erasure (\Cref{prop4}) prevents the premature overwriting that causes \gdntwo accuracy to drop sharply on early-context needles ($0\text{--}40\%$ depth) at $4\text{K}\text{--}8\text{K}.$
Similarly, on Based/JRT (\Figref{fig:recall}{f}), extracting semi-structured attributes (SWDE, FDA) and passage spans (SQuAD, TriviaQA, NQ, DROP) from natural documents requires resolving query-target lags that vary widely across examples.
Consequently, FFN expansion (\ours no $\vgamma$) and static scaling (\ours static $\vgamma$) yield only modest gains over \gdntwo, whereas dynamic $\vgamma_t$ adapts the channel filter per query token to drive the bulk of the six-task improvement and surpass \tf on DROP.


\begin{figure}[!t]
    \centering
    \includegraphics[width=\textwidth]{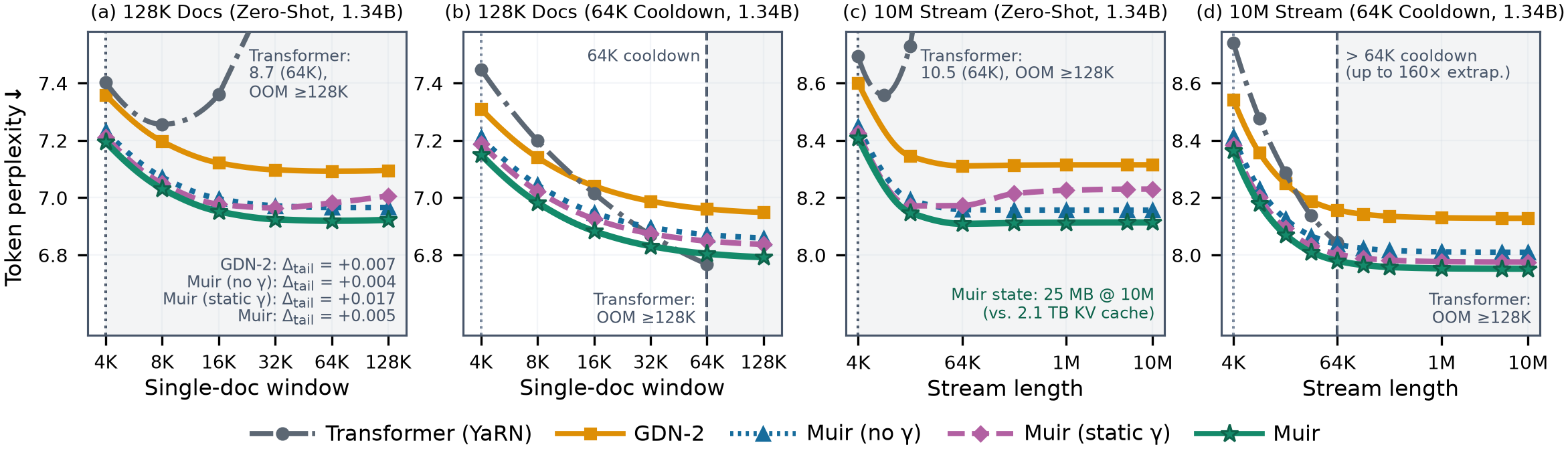}
    \caption{%
        \textbf{\ours extrapolates to $\boldsymbol{128\text{K}}$-token documents and $\boldsymbol{10\text{M}}$-token streams with the lowest perplexity at a constant $\boldsymbol{25\text{ MB}}$ state.}
        %
        (a, c) Zero-shot from $4\text{K}$ pretraining, Dynamic YaRN \tf diverges past $8\text{K}$ and unchunked evaluation exceeds HBM at $\ge\!128\text{K},$ while \ours (static $\vgamma$) drifts upward past $32\text{K}$; dynamic $\vgamma_t$ avoids this static-gain drift and outperforms \gdntwo across $128\text{K}$ documents and $10\text{M}$-token streams ($2{,}560\times$ training length).
        %
        (b, d) After a $64\text{K}$ cooldown ($0.3\%$ of pretraining), \tf improves through $64\text{K}$ before unchunked evaluation exceeds HBM at $\ge\!128\text{K},$ static-$\vgamma$ drift vanishes, and all \ours variants improve monotonically below \gdntwo through $10\text{M}$ tokens ($160\times$ beyond cooldown).%
    }
    \label{fig:longctx}
\end{figure}

\subsubsection{Length Extrapolation}
\label{sec:exp:results:longctx}

\begin{figure}[!t]
    \centering
    \includegraphics[width=\textwidth]{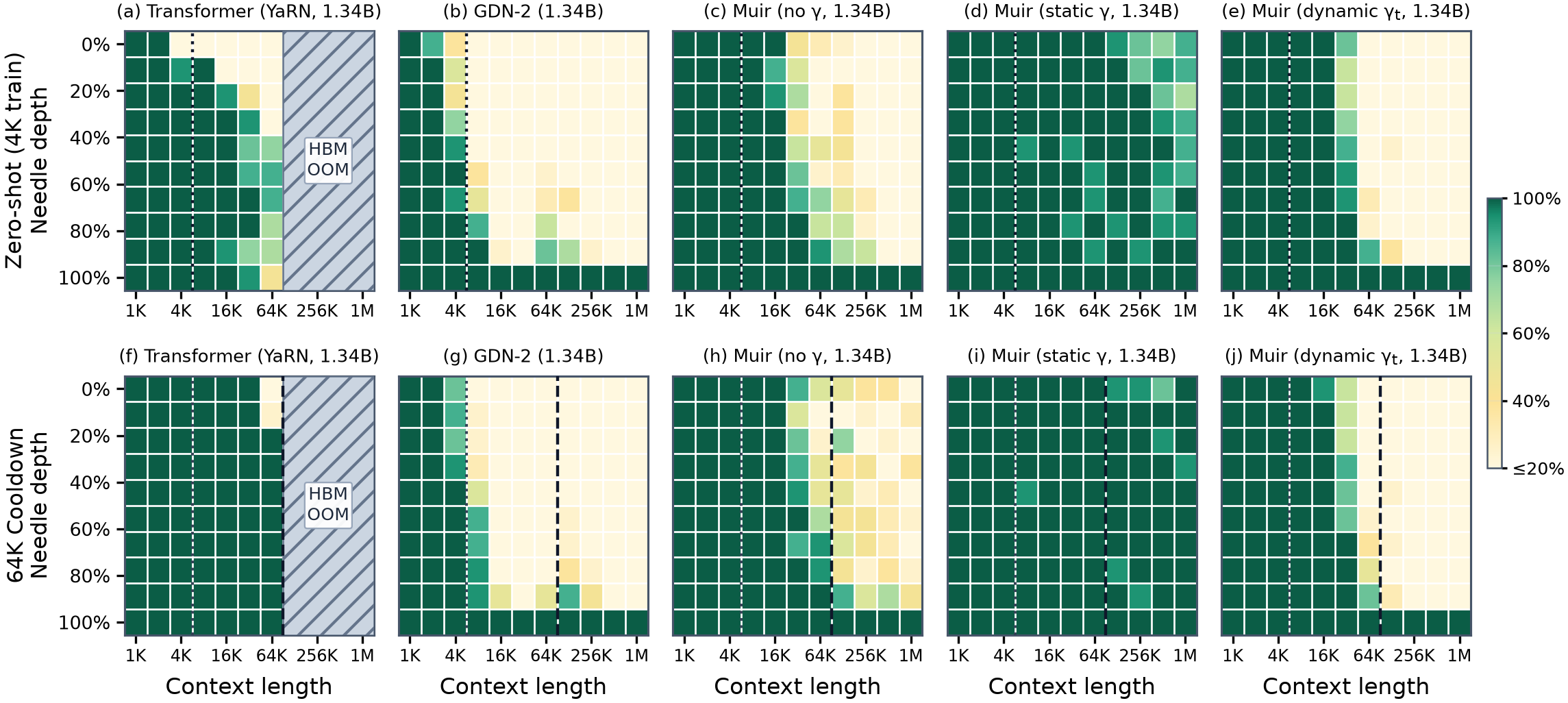}
    \caption{%
        \textbf{Static matched-filter readout extends passkey retrieval to $\boldsymbol{1\text{M}}$ tokens both zero-shot and after a $\boldsymbol{64\text{K}}$ cooldown.}
        S-NIAH-1 accuracy across $10$ needle-depth bins and context lengths $1\text{K}\text{--}1\text{M}$ ($256\times$ pretraining length) at $1.34\text{B}$ scale under (top) zero-shot evaluation from $4\text{K}$ pretraining and (bottom) pure-text continual cooldown at $64\text{K}.$
        %
        Cooldown extends Dynamic YaRN \tf from $8\text{K}$ to $32\text{K}$ before early-depth accuracy drops at $64\text{K}$ and unchunked prefill exceeds HBM at $\ge\!128\text{K},$ whereas recurrent horizons change little under pure-text cooldown: \gdntwo collapses past $4\text{K}\text{--}8\text{K},$ \ours (no $\vgamma$) fades past $32\text{K},$ \ours (dynamic $\vgamma_t,$ splitting $(r_w, r_q) = (128, 128)$ for multi-lag recall) holds through $32\text{K}$ before tapering at $\ge\!64\text{K},$ and \ours (static $\vgamma,$ keeping $r_w = 256$ for single-lag lookup) maintains near-perfect recall to $1\text{M}$ tokens in both regimes.\looseness=-1%
    }
    \label{fig:longctx_niah}
\end{figure}

\Cref{fig:longctx,fig:longctx_niah} evaluate length extrapolation up to $10\text{M}$-token language-modeling streams and $1\text{M}$-token passkey retrieval at $1.34\text{B}$ parameters (full tables in \Cref{app:eval:longctx}).
%
In zero-shot language modeling (\Figref{fig:longctx}{a,c}), Dynamic YaRN slows \tf divergence past $4\text{K},$ but perplexity rises past $8\text{K}$ and unchunked evaluation exceeds HBM at $\ge\!128\text{K}$; even if chunked, a $10\text{M}$-token KV cache needs $2.1\text{ TB}$ in \texttt{bfloat16}, or $0.32\text{ TB}$ at $2.5$ bits \citep{Zandieh2026-ms}, compared to a $25\text{ MB}$ \texttt{float32} state across all $24$ layers.
%
Whereas EOS-separated packed streams reset discourse context and stay flat ($\Delta_{\mathrm{tail}} \approx 0$), continuous topical accumulation in $128\text{K}$ documents and $10\text{M}$ streams causes \ours (static $\vgamma$) to drift upward past $32\text{K}$ ($\Delta_{\mathrm{tail}} = +0.017\text{ nats}$) because a fixed gain calibrated at $4\text{K}$ over-weights slow-decaying rows once their energy builds over longer horizons (\Cref{prop2,prop3:rem1}).
%
%
%
Adapting the channel filter per query via dynamic $\vgamma_t$ cuts tail drift by $3.6\times$ (vs.\ static $\vgamma$) and achieves the lowest perplexity at every horizon up to $10\text{M}$ tokens.
After the $64\text{K}$ cooldown (\Figref{fig:longctx}{b,d}), \tf improves through $64\text{K}$ before unchunked evaluation exceeds HBM at $\ge\!128\text{K},$ whereas recalibrated slow-channel gains eliminate static-$\vgamma$ drift ($\Delta_{\mathrm{tail}} = -0.012\text{ nats}$) and full \ours outperforms \tf and \gdntwo across the entire $10\text{M}$ stream.

On RULER S-NIAH-1 (\Cref{fig:longctx_niah}), unseen RoPE positions limit zero-shot \tf past $8\text{K},$ and raising the RoPE base during cooldown restores $100\%$ recall through $32\text{K}$ before early-depth accuracy drops at $64\text{K}$ and unchunked prefill exceeds HBM at $\ge\!128\text{K}.$
By contrast, recurrent models have no positional encodings to recalibrate, so pure-text cooldown without synthetic passkeys only refines slow-channel statistics and leaves the retrieval horizons learned at $4\text{K}$ largely unchanged.
%
Under unweighted readout ($\vgamma_t = \vone$), heavy write-time erasure overwrites distant needles (\Cref{prop4:rem1}): \gdntwo collapses past $4\text{K}\text{--}8\text{K},$ and even with low-rank erase regularization (\Cref{prop1:rem2}), \ours (no $\vgamma$) fades on early-depth needles past $32\text{K}.$
Adding matched-filter readout relaxes this erasure; because S-NIAH-1 queries a single needle at the end of the prompt, a fixed filter suffices while keeping the wider $r_w = 256$ write gate, so \ours (static $\vgamma$) reaches $90.0\%$ zero-shot and $99.4\%$ post-cooldown recall at $1\text{M}$ tokens ($256\times$ pretraining length).
Conversely, splitting $(r_w, r_q) = (128, 128)$ under a fixed rank budget in full \ours trades half of that write rank for the per-token routing that resolves multi-lag recall and prevents $10\text{M}$-stream drift (\Cref{fig:recall,fig:longctx}), retaining near-perfect average passkey recall across $1\text{K}\text{--}32\text{K}$ before tapering at $\ge\!64\text{K}.$
%


\subsubsection{Training Throughput and Numerical Stability}

\xhdr{Training Throughput}

\Figref{fig:teaser}{c} compares training throughput ($\mathrm{p50}$ of $100$-step intervals) on $8$ TPU7x chips at a fixed global batch size of $B\times T = 524{,}288\text{ tokens/step}$ across sequence lengths $T$ from $2\text{K}$ to $32\text{K}$ at $1.34\text{B}$ scale, where both \gdntwo and \ours use our boundary-pivoted chunkwise kernel (complete throughput, compiled XLA FLOP, and HBM traffic breakdowns are in \Cref{app:kernel:perf}).
Replacing the two dense $d\times d$ gate projections in \gdntwo with the fused low-rank projection allows \ours to widen the FFN by $+30.4\%$ (from $d_{\mathrm{ffn}} = 5{,}888$ to $7{,}680$) and add the dynamic readout gain $\vgamma_t$ while keeping compiled per-step FLOPs ($-1.8\%$) and HBM traffic ($-0.7\%\text{ to }-\!1.4\%$) nearly unchanged relative to \gdntwo.
Consequently, \ours matches \gdntwo at the $4\text{K}$ pretraining length ($179.4$ vs.\ $179.3\text{ K tokens/s}$) and runs $+1.0\%\text{--}1.6\%$ faster at $16\text{K}\text{--}32\text{K}.$

At $T \le 8\text{K},$ \tf is faster because $\gO(T^2 d)$ causal attention is small and \tf uses a hand-tuned TPU Pallas Splash-Attention kernel fused with block GEMMs, whereas our pure-JAX chunkwise recurrence incurs XLA kernel-boundary HBM reads and writes between chunk stages.
As $T$ increases from $4\text{K}$ to $32\text{K},$ quadratic attention nearly doubles \tf per-step FLOPs ($348 \to 652\text{ TFLOP/step/device}$) and reduces throughput by $4.2\!\times\!.$
By contrast, because recurrent per-step FLOPs are invariant to $T,$ both \gdntwo and \ours maintain flat throughput across $2\text{K}\text{--}8\text{K}.$
%
At $T \ge 16\text{K}$ ($N_c \ge 64$ chunks per sequence at $B \le 32$), deeper per-sequence chunk loops allow XLA to fuse boundary temporaries (cutting HBM traffic by $330\text{--}356\text{ GB/step/device}$ at $32\text{K}$ vs.\ $4\text{K}$) and overlap asynchronous HBM transfers with MXU execution across chunks, raising memory-bandwidth utilization and increasing throughput by $48\text{--}53\%$ at $16\text{K}$ vs.\ $8\text{K}.$
Consequently, \ours crosses over \tf at $T \approx 11\text{K}$ and trains $1.64\times$ and $2.74\times$ faster at $16\text{K}$ and $32\text{K}.$


\begin{figure}[!t]
    \centering
    \includegraphics[width=\textwidth]{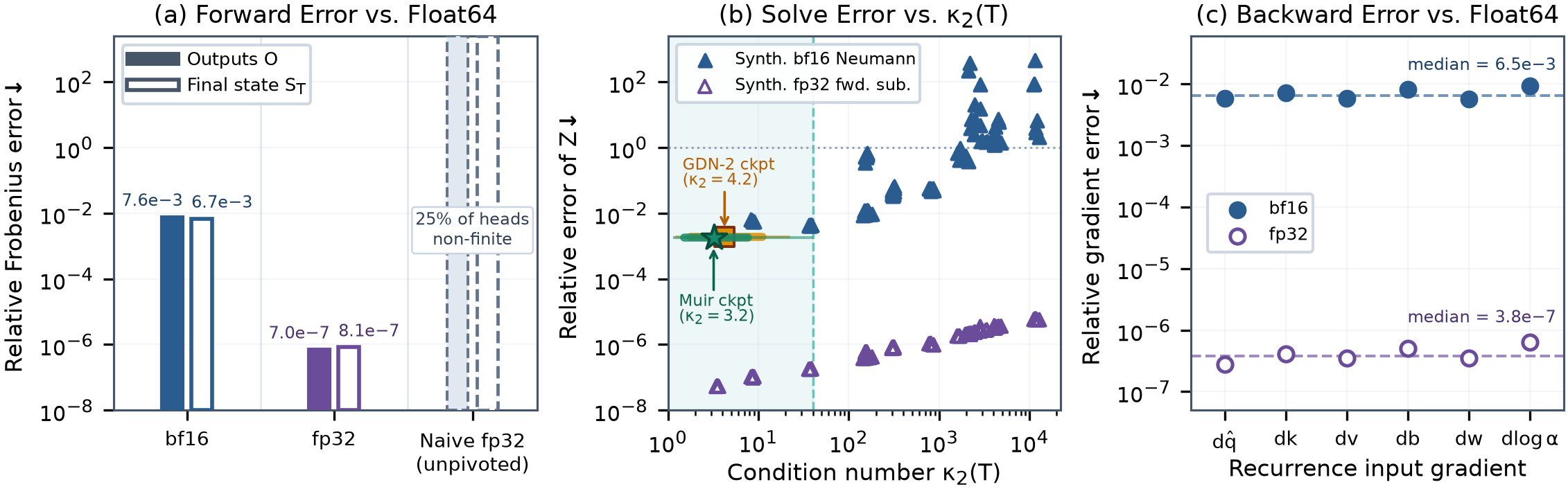}
    \caption{%
        \textbf{Boundary-pivoted chunkwise execution eliminates decay overflow and maintains well-conditioned forward and backward passes.}
        (a) Against an exact \texttt{float64} sequential reference, naive unpivoted \texttt{float32} decay factorization overflows and produces \texttt{NaN} on $25\%$ of heads, whereas the boundary-pivoted \texttt{bfloat16} and \texttt{float32} kernels match the reference to machine precision on both chunk outputs and final state.
        %
        (b) Across synthetic stress regimes spanning $\kappa_2(\mT) \in [1, 10^4],$ the \texttt{bfloat16} block-Neumann inverse stays near its precision floor throughout the well-conditioned regime ($\kappa_2(\mT) < 40,$ shaded), which contains the sampled $100\text{B}$-token checkpoints of \gdntwo and \ours (markers show medians).
        %
        (c) Single-layer backward gradients ($T\!=\!128, C\!=\!64, d_k\!=\!16$) match the \texttt{float64} reference across all six inputs without log-decay drift.%
    }
    \label{fig:kernel_num}
\end{figure}

\xhdr{Numerical Stability}

\Cref{fig:kernel_num} verifies the numerical accuracy and conditioning of the boundary-pivoted chunkwise kernel (full tables in \Cref{app:kernel:perf}).
%
In \Figref{fig:kernel_num}{a}, forming $\mL_{\mathrm{strict}}$ and $\mM$ via the unpivoted split $\vpi_{t,s} = \vpi_t \oslash \vpi_s$ divides keys by prefix decay $\vpi_s = \exp(\mLambda_s)$ in $\mK_{\oslash\pi}$; on fast-decaying channels ($25\%$ of heads) where $\emLambda_{s,j} < -\ln(\mathrm{FLT\_MAX}) \approx -88.7,$ $\mK_{\oslash\pi}$ overflows \texttt{float32} to $\pm\infty,$ producing \texttt{NaN} in $\hat\mK_{\odot\pi}\mK_{\oslash\pi}\tr$ even though every relative decay $\vpi_{t,s} = \exp(\mLambda_t - \mLambda_s)$ lies in $(0, 1].$
%
Splitting $\vpi_{t,s} = \vpi_{t,p} \odot \vpi_{p,s}$ across each sub-block boundary $p$ keeps both exponential factors in $(0, 1]$ without division, and maintaining $\mLambda_t$ and the recurrent state carry $\mS_t$ in \texttt{float32} keeps forward errors at the precision floor of the GEMM inputs.

In \Figref{fig:kernel_num}{b}, the \texttt{bfloat16} block-Neumann inverse multiplies 7-bit-mantissa MXU operands across leaf squaring and sub-block merges, so rounding error grows with $\kappa_2(\mT)$ and exceeds $10^0$ once $\kappa_2(\mT) \gtrsim 10^3,$ whereas sequential \texttt{float32} forward substitution remains accurate across $\kappa_2(\mT) \in [1, 10^4].$
In trained models, however, exponential decay $\vpi_{t,s}$ across lags $t - s$ and small cross-step inner products between $\ell_2$-normalized keys damp the strictly lower-triangular entries $(\mL_{\mathrm{strict}})_{t,s} = \inner{\hat\vk_t}{\vpi_{t,s}\odot\vk_s}$ and keep $\kappa_2(\mT)$ well within the stable regime ($\max < 40$ for both \gdntwo and \ours across $1{,}536$ chunk systems from the $166\text{M}$ checkpoints).
Moreover, \ours attains lower $\mathrm{p90}$ $\kappa_2(\mT)$ than \gdntwo ($7.7$ vs.\ $11.1$) because matched-filter readout lowers erasure (\Cref{prop4}), shrinking $\hat\vk_t = \vb_t \odot \vk_t$ and further contracting $\norm{\mL_{\mathrm{strict}}}_2.$

Finally, reusing $\mT\inv$ from the forward solve and recomputing the boundary-pivoted decay factors across the merge tree keeps $\mathrm{d}\log\valpha$ at the same precision floor as the query, key, value, and gate gradients (\Figref{fig:kernel_num}{c}), so pretraining runs without loss spikes or \texttt{NaN}s.


\subsection{Ablations}
\label{sec:exp:abl}

\Cref{tab:abl} ablates the gate-rank allocation across all four model scales by comparing our sensitivity-allocated \ours ranks ($R_{\mathrm{tot}} = 512$) against wider output-gate allocations ($r_g = 256$) and uniform gate ranks ($128$ across active gates).
We also decouple low-rank gate compression from FFN expansion by varying gate ranks and FFN width independently at $406\text{M}$ (full control tables are in \Cref{app:eval:abl}).

\input{tables/tab2_abl}


\xhdr{Output-Gate Rank Allocation}

Doubling the output-gate rank from $r_g = 128$ to $256$ requires $R_{\mathrm{tot}} = 640$ for no or static $\vgamma$ and $R_{\mathrm{tot}} = 768$ for dynamic $\vgamma_t,$ which shrinks $d_{\mathrm{ffn}}$ by $128$ and $256$ channels, respectively.
While $(64, 64, 256, 256, 128)$ achieves slightly lower validation NLL at $166\text{M}$ and $834\text{M},$ \ours surpasses it at $406\text{M}$ and outperforms all three $r_g = 256$ variants at $1.34\text{B}$ across validation NLL, zero-shot commonsense reasoning, Based recall, and 5-shot TriviaQA.
At larger scale, preserving $d_{\mathrm{ffn}} = 4d - 512$ and two-tile $R_{\mathrm{tot}} = 512$ alignment outweighs a wider output gate.


\xhdr{Sensitivity-Allocated vs.\ Uniform Gate Ranks}

When readout gain is omitted or static ($r_q = 0$), four uniform rank-$128$ gates $(128, 128, 128, 128, 0)$ sum to $R_{\mathrm{tot}} = 512$ and perform comparably to the sensitivity-allocated $(64, 64, 128, 256, 0)$ variants across scales.
However, adding a fifth rank-$128$ gate for dynamic $\vgamma_t$ under uniform allocation $(128, 128, 128, 128, 128)$ raises the total rank to $R_{\mathrm{tot}} = 640,$ which breaks $256$-column MXU tile alignment and narrows $d_{\mathrm{ffn}}$ by $128$ channels.
Because state updates are much less sensitive to the decay and erase gates than to the write gate (\Cref{prop1,prop1:rem1,prop1:rem2}), compressing $(r_f, r_b)$ to $(64, 64)$ in \ours fits the rank-$128$ dynamic readout gain within $R_{\mathrm{tot}} = 512$ without reducing $d_{\mathrm{ffn}},$ and outperforms the uniform dynamic-$\vgamma_t$ variant on validation NLL at every scale and on $1.34\text{B}$ zero-shot commonsense reasoning, Based recall, and 5-shot TriviaQA.


\xhdr{Decoupling Gate Compression from FFN Expansion}

To separate the effect of low-rank gating from FFN expansion under the $12d^2$ budget, we vary gate ranks and FFN width independently at $406\text{M}$ relative to \gdntwo ($d_{\mathrm{ffn}} = 4{,}352$).
%
Compressing the dense $(b, w)$ gates into rank-$128$ projections alongside rank-$128$ $\vgamma_t$ ($R_{\mathrm{tot}} = 640$) while holding the FFN fixed at $d_{\mathrm{ffn}} = 4{,}352$ cuts gate-projection FLOPs by $2.8\times$ and reduces non-embedding parameters and FLOPs by $12.5\%$ ($-38.9\text{M}$ parameters across all layers) at a validation NLL cost of only $+7.4\text{ mnats}.$
Conversely, adding the same $+38.9\text{M}$ parameters ($+12.5\%$ non-embedding capacity, $d_{\mathrm{ffn}} = 5{,}504$) to widen the \gdntwo FFN with dense gates improves validation NLL by $-20.5\text{ mnats}$ ($2.8\times$ larger NLL reduction per parameter).
Reinvesting the freed parameters into $d_{\mathrm{ffn}}$ and low-rank $\vgamma_t$ in iso-parameter \ours achieves $-18.0\text{ mnats},$ recovering $88\%$ of the unconstrained wide-FFN gain at zero parameter or FLOP overhead.
%
By contrast, parameterizing $\vgamma_t$ with a dense $d \times d$ projection in \gdntwo forces $d_{\mathrm{ffn}}$ down to $3{,}584$ under the iso-parameter budget and degrades validation NLL by $+8.3\text{ mnats},$ which confirms that under an iso-parameter budget, dynamic readout requires a low-rank rather than dense parameterization.

%% file: tables/tab1_main_lm.tex
\begin{table}[!t]
    \centering
    \caption{%
        \textbf{\ours improves language modeling and downstream reasoning over \tf and \gdntwo across model scale.}
        We report FineWeb-Edu validation token NLL (FW-Edu), WikiText (Wiki.) and LAMBADA (LMB.) word-level perplexity, zero-shot accuracy across eight commonsense reasoning benchmarks and their mean (Avg.), and 5-shot WinoGrande (Wino.) and TriviaQA (TQA) accuracy after $100\text{B}$ training tokens.
        Best results per scale are in \textbf{bold} and second-best are \underline{underlined}.%
    }
    \label{tab:main_lm}
    \resizebox{\textwidth}{!}{%
        \setlength{\tabcolsep}{3pt}%
        \renewcommand{\arraystretch}{1.08}%
        \begin{tabular}{@{}c@{\hspace{3pt}}|l@{\hspace{2pt}}|@{\hspace{1pt}}ccc|cccccccc|c|cc@{}}
        \toprule
            \multicolumn{1}{@{}c@{\hspace{5pt}}}{} & & \multicolumn{3}{c|}{{Language Modeling}} & \multicolumn{9}{c|}{{Zero-Shot Commonsense Reasoning}} & \multicolumn{2}{c@{}}{{5-Shot}} \\
            \cmidrule(lr){3-5} \cmidrule(lr){6-14} \cmidrule(l){15-16}
            \multicolumn{1}{@{}c@{\hspace{5pt}}}{} & {Model} & {FW-Edu} & {Wiki.} & {LMB.} & {LMB.} & {PIQA} & {Hella.} & {Wino.} & {ARC-e} & {ARC-c} & {OBQA} & {SIQA} & {Avg.} & {Wino.} & {TQA} \\
            \multicolumn{1}{@{}c@{\hspace{5pt}}}{} & & NLL $\downarrow$ & ppl $\downarrow$ & ppl $\downarrow$ & acc $\uparrow$ & acc $\uparrow$ & \accn{} $\uparrow$ & acc $\uparrow$ & acc $\uparrow$ & \accn{} $\uparrow$ & acc $\uparrow$ & acc $\uparrow$ & acc $\uparrow$ & acc $\uparrow$ & acc $\uparrow$ \\
        \midrule
            \multirowrotate{5}{90}{\footnotesize 166M}
            & \tf & 2.4569 & 31.64 & 52.69 & 28.47 & \underline{65.67} & 35.95 & 50.99 & 55.35 & 26.28 & 20.20 & 36.39 & 39.91 & 49.96 & 4.74 \\
            & \gdntwo & 2.4466 & 30.84 & 39.89 & 30.60 & 64.74 & 37.02 & 50.28 & \tbf{56.61} & 26.45 & 20.40 & 37.92 & 40.50 & 50.12 & \tbf{5.80} \\
            & \ours (no $\vgamma$) & 2.4322 & 30.31 & \underline{34.87} & \underline{32.68} & 65.51 & \tbf{38.19} & \tbf{52.01} & \underline{56.48} & \tbf{27.82} & \underline{20.80} & \tbf{38.95} & \tbf{41.55} & \tbf{52.49} & 3.77 \\
            & \ours (static $\vgamma$) & \tbf{2.4273} & \tbf{29.83} & 40.98 & 30.58 & \tbf{66.43} & \underline{37.92} & \underline{51.78} & 55.30 & \underline{26.54} & \underline{20.80} & \underline{38.18} & \underline{40.94} & \underline{51.54} & 4.32 \\
            & \ours (ours) & \underline{2.4276} & \underline{30.28} & \tbf{34.45} & \tbf{32.72} & 63.93 & 37.84 & 51.70 & 54.88 & 25.85 & \tbf{21.80} & \underline{38.18} & 40.86 & 50.51 & \underline{5.32} \\
        \midrule
            \multirowrotate{5}{90}{\footnotesize 406M}
            & \tf & 2.2424 & 23.18 & 21.10 & 39.74 & 68.77 & 45.34 & 53.83 & 62.37 & 30.03 & 25.00 & \underline{39.71} & 45.60 & 52.41 & 9.06 \\
            & \gdntwo & 2.2354 & 22.90 & \underline{19.15} & \underline{40.64} & \tbf{69.26} & 45.51 & 53.04 & \underline{62.42} & 30.29 & \underline{25.40} & 39.10 & 45.71 & \underline{53.04} & 9.08 \\
            & \ours (no $\vgamma$) & 2.2230 & \tbf{22.26} & \tbf{18.11} & \tbf{41.55} & 68.34 & \underline{46.95} & \underline{53.99} & 62.12 & 30.89 & \tbf{26.80} & 39.51 & \tbf{46.27} & \tbf{53.12} & \underline{10.30} \\
            & \ours (static $\vgamma$) & \underline{2.2191} & 22.35 & 20.15 & 39.96 & \underline{69.04} & 46.50 & \tbf{54.06} & \tbf{64.60} & \tbf{32.17} & 24.60 & 38.18 & \underline{46.14} & 52.72 & \tbf{10.83} \\
            & \ours (ours) & \tbf{2.2174} & \underline{22.31} & 20.00 & 40.11 & 68.77 & \tbf{46.97} & 52.49 & 61.41 & \underline{31.66} & \underline{25.40} & \tbf{39.82} & 45.83 & 52.57 & 10.05 \\
        \midrule
            \multirowrotate{5}{90}{\footnotesize 834M}
            & \tf & 2.1186 & 19.46 & 14.79 & 44.77 & 70.78 & 52.04 & 55.56 & 66.04 & 32.59 & 26.80 & 39.36 & 48.49 & 55.17 & 11.85 \\
            & \gdntwo & 2.1060 & 19.17 & 13.49 & 46.54 & 71.00 & 52.48 & 55.41 & 66.96 & 35.24 & 28.00 & 39.00 & 49.33 & 54.70 & 13.40 \\
            & \ours (no $\vgamma$) & 2.0876 & \underline{18.46} & \underline{12.57} & 47.43 & 70.84 & \underline{53.91} & \tbf{58.25} & \tbf{69.32} & \underline{35.41} & \underline{28.20} & \tbf{40.89} & \tbf{50.53} & \tbf{58.17} & \tbf{14.34} \\
            & \ours (static $\vgamma$) & \underline{2.0869} & \tbf{18.42} & 12.64 & \underline{47.49} & \underline{71.22} & 53.80 & \underline{56.83} & \underline{68.69} & 35.32 & 27.80 & 39.97 & 50.14 & \underline{57.93} & 13.32 \\
            & \ours (ours) & \tbf{2.0862} & 18.58 & \tbf{11.88} & \tbf{49.00} & \tbf{71.38} & \tbf{53.96} & 56.27 & 67.09 & \tbf{36.43} & \tbf{28.80} & \underline{40.02} & \underline{50.37} & 57.62 & \underline{13.58} \\
        \midrule
            \multirowrotate{5}{90}{\footnotesize 1.34B}
            & \tf & 2.0335 & 17.17 & 10.30 & \tbf{51.80} & 71.27 & 56.96 & 57.85 & 71.04 & 36.60 & 28.40 & \tbf{42.48} & 52.05 & 57.77 & 14.56 \\
            & \gdntwo & 2.0177 & 16.80 & 10.38 & 50.92 & 72.36 & 57.91 & 58.72 & 70.71 & 38.74 & 30.40 & 39.92 & 52.46 & 59.91 & 16.48 \\
            & \ours (no $\vgamma$) & 2.0006 & 16.46 & 10.26 & 51.12 & \underline{73.01} & 59.12 & 58.48 & \tbf{73.23} & \underline{40.27} & \tbf{33.60} & \underline{41.45} & \underline{53.79} & \underline{60.14} & 16.52 \\
            & \ours (static $\vgamma$) & \underline{1.9994} & \tbf{16.35} & \underline{10.15} & 49.97 & 72.14 & \tbf{59.47} & \underline{59.35} & 72.94 & \tbf{41.13} & 31.80 & 40.38 & 53.40 & 58.48 & \underline{17.71} \\
            & \ours (ours) & \tbf{1.9973} & \underline{16.41} & \tbf{10.00} & \underline{51.58} & \tbf{73.56} & \underline{59.34} & \tbf{59.67} & \underline{73.15} & 39.16 & \underline{33.20} & 41.10 & \tbf{53.84} & \tbf{60.69} & \tbf{19.32} \\
        \bottomrule
        \end{tabular}%
    }
\end{table}

%% file: tables/tab2_abl.tex
\begin{table}[t]
    \centering
        \caption{%
        \textbf{Sensitivity-allocated gate ranks and dynamic readout match or outperform wider output-gate and uniform-rank allocations while preserving MXU alignment and FFN width.}
        For each configuration of gate bottleneck ranks and FFN width (shown at $d = 2{,}048$), we report FineWeb-Edu validation token NLL across all four model scales, alongside $1.34\text{B}$ WikiText (Wiki.) and LAMBADA (LMB.) word-level perplexity, 8-task zero-shot commonsense accuracy (LM-Eval), 6-task Based extractive recall accuracy (Based), and 5-shot TriviaQA accuracy (TQA) after $100\text{B}$ training tokens.
        %
    }
    \label{tab:abl}
    \resizebox{\textwidth}{!}{%
        \setlength{\tabcolsep}{4.0pt}%
        \renewcommand{\arraystretch}{1.08}%
        \begin{tabular}{@{}c@{\hspace{3pt}}|l@{\hspace{2pt}}|@{\hspace{2pt}}c@{\hspace{5pt}}c@{\hspace{5pt}}c@{\hspace{5pt}}c@{\hspace{5pt}}c@{\hspace{2pt}}|@{\hspace{2pt}}c@{\hspace{2pt}}|cccc|cc|ccc@{}}
            \toprule
            \multicolumn{1}{@{}c@{\hspace{5pt}}}{} & & \multicolumn{6}{c|}{{Configuration}} & \multicolumn{4}{c|}{{FineWeb-Edu Validation NLL $\downarrow$}} & \multicolumn{2}{c|}{{1.34B PPL $\downarrow$}} & \multicolumn{3}{c@{}}{{1.34B Accuracy $\uparrow$}} \\
            \cmidrule(lr){3-8} \cmidrule(lr){9-12} \cmidrule(lr){13-14} \cmidrule(l){15-17}
            \multicolumn{1}{@{}c@{\hspace{5pt}}}{} & {Model} & {$r_f$} & {$r_b$} & {$r_g$} & {$r_w$} & {$r_q$} & {$d_{\mathrm{ffn}}$} & {166M} & {406M} & {834M} & {1.34B} & {Wiki.} & {LMB.} & {LM-Eval} & {Based} & {TQA} \\
        \midrule
            \multirowrotate{2}{90}{\footnotesize Base}
            & \tf & -- & -- & -- & -- & -- & 8{,}192 & 2.4569 & 2.2424 & 2.1186 & 2.0335 & 17.17 & 10.30 & 52.05 & \tbf{45.92} & 14.56 \\
            & \gdntwo & 128 & $d$ & 128 & $d$ & -- & 5{,}888 & 2.4466 & 2.2354 & 2.1060 & 2.0177 & 16.80 & 10.38 & 52.46 & 32.08 & 16.48 \\
        \midrule
            \multirowrotate{3}{90}{\footnotesize Ours}
            & \ours (no $\vgamma$) & 64 & 64 & 128 & 256 & 0 & 7{,}680 & 2.4322 & 2.2230 & 2.0876 & 2.0006 & 16.46 & 10.26 & \underline{53.79} & 32.86 & 16.52 \\
            & \ours (static $\vgamma$) & 64 & 64 & 128 & 256 & 0 & 7{,}680 & \underline{2.4273} & 2.2191 & 2.0869 & 1.9994 & \underline{16.35} & 10.15 & 53.40 & 33.02 & 17.71 \\
            & \tbf{\ours (ours)} & 64 & 64 & 128 & 128 & 128 & 7{,}680 & 2.4276 & \tbf{2.2174} & \underline{2.0862} & \tbf{1.9973} & 16.41 & 10.00 & \tbf{53.84} & 34.60 & \tbf{19.32} \\
        \midrule
            \multirowrotate{3}{90}{\raisebox{0pt}[\height][0pt]{\footnotesize $r_g\!=\!256$}}
            & \ours (no $\vgamma$) & 64 & 64 & 256 & 256 & 0 & 7{,}552 & 2.4293 & 2.2202 & 2.0867 & 2.0009 & \tbf{16.31} & 10.27 & 52.47 & 32.47 & \underline{18.57} \\
            & \ours (static $\vgamma$) & 64 & 64 & 256 & 256 & 0 & 7{,}552 & 2.4285 & 2.2194 & 2.0873 & \underline{1.9988} & \underline{16.35} & \tbf{9.72} & 53.27 & 34.38 & 16.96 \\
            & \ours (dynamic $\vgamma_t$) & 64 & 64 & 256 & 256 & 128 & 7{,}424 & \tbf{2.4260} & \underline{2.2187} & \tbf{2.0845} & 2.0002 & 16.44 & 10.12 & 52.95 & 33.36 & 17.34 \\
        \midrule
            \multirowrotate{3}{90}{\footnotesize Uniform}
            & \ours (no $\vgamma$) & 128 & 128 & 128 & 128 & 0 & 7{,}680 & 2.4292 & 2.2207 & 2.0879 & 2.0001 & 16.42 & 10.28 & 52.69 & 34.18 & 16.67 \\
            & \ours (static $\vgamma$) & 128 & 128 & 128 & 128 & 0 & 7{,}680 & 2.4288 & 2.2205 & 2.0893 & 2.0001 & 16.51 & 10.22 & 52.99 & \underline{34.87} & 15.67 \\
            & \ours (dynamic $\vgamma_t$) & 128 & 128 & 128 & 128 & 128 & 7{,}552 & 2.4281 & 2.2213 & 2.0871 & 1.9999 & 16.40 & \underline{9.89} & 53.10 & 33.84 & 17.19 \\
        \bottomrule
        \end{tabular}%
    }
\end{table}

%% file: sections/6_analysis.tex
\section{Mechanistic Analysis}
\label{sec:mech}

We now evaluate our theoretical predictions from \S\ref{sec:theory:theory} directly on the $100\text{B}$-token checkpoints without retraining (protocols and supplementary tables are in \Cref{app:mech}).


\begin{figure}[!t]
    \centering
    \includegraphics[width=\textwidth]{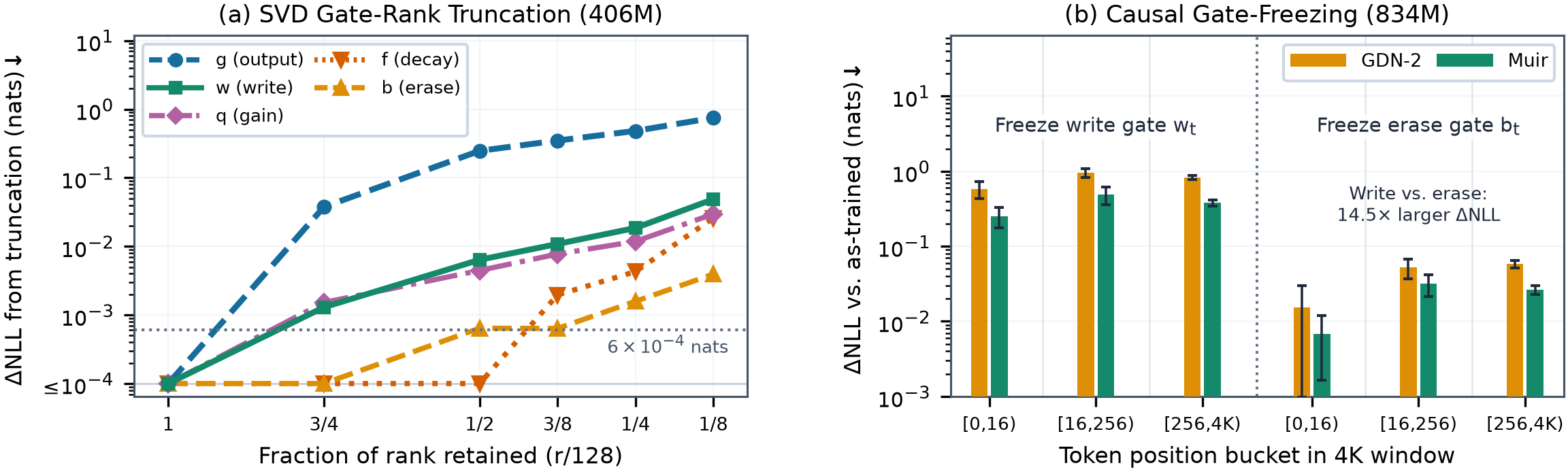}
    \caption{%
        \textbf{Trained models exhibit the predicted write-erase sensitivity asymmetry under both SVD rank truncation and causal gate freezing.}
        (a) Truncating each rank-$128$ gate projection of the uniform-rank \ours checkpoint via SVD (one gate at a time across all layers) shows that the decay ($f$) and erase ($b$) gates saturate by rank $64,$ whereas the output ($g$), write ($w$), and readout gain ($q$) gates require rank $128.$
        (b) Replacing the dynamic write gate with its token-averaged channel mean ($\vw_t \gets \bar\vw$) across all layers degrades validation NLL $14.5\times$ more than freezing the erase gate ($\vb_t \gets \bar\vb$) in both \gdntwo and \ours.%
    }
    \label{fig:mech_sens}
\end{figure}

\begin{figure}[!t]
  \centering
  \includegraphics[width=\textwidth]{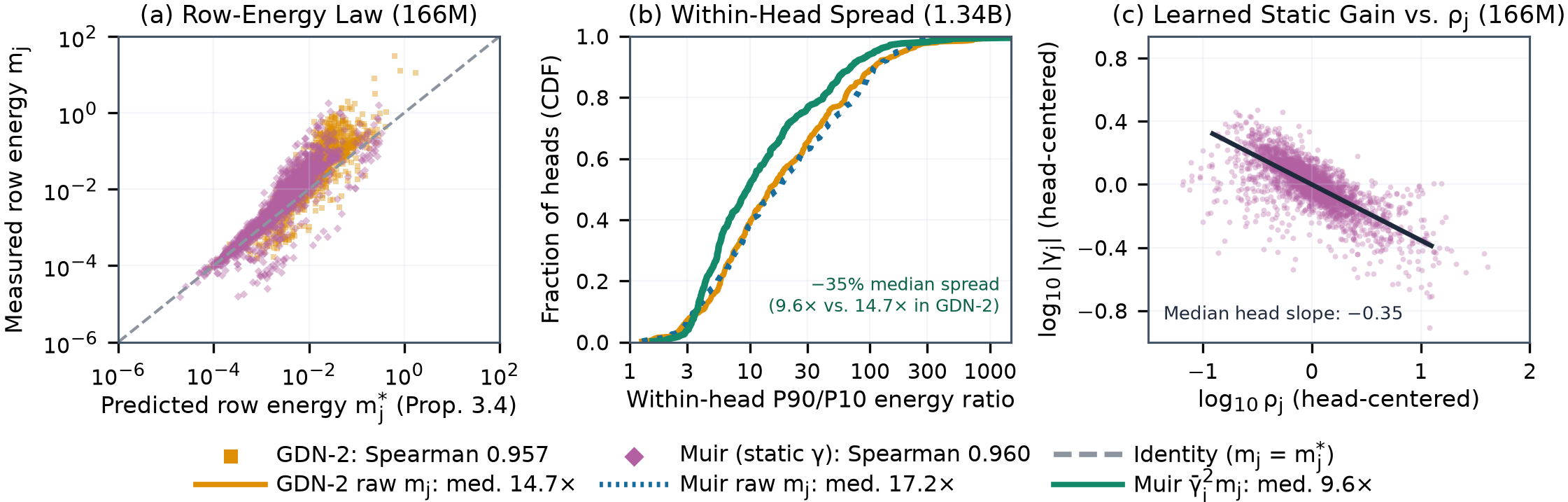}
  \caption{%
        \textbf{Stationary state row energies obey the closed-form law and are balanced by fractional power-law readout gains.}
        (a) Across state channels, the analytical fixed point $\evm_j^\ast$ \eqref{prop2:eq2} closely tracks measured stationary row energy $\evm_j = \E[\norm{\mS_{t,j,:}}_2^2]$ in both \gdntwo and \ours.
        (b) Across all heads, weighting state rows by the learned readout gain compresses the within-head $\mathrm{p90}/\mathrm{p10}$ energy spread by $35\%$ relative to \gdntwo.
        (c) Within each head, learned static gains $\abs{\evgamma_j}$ follow a fractional power law $\evgamma_j \propto \evrho_j^{-p}$ ($p = 0.35$) of stationary row RMS $\evrho_j,$ intermediate between unit readout ($p = 0$) and full whitening ($p = 1$).%
  }
  \label{fig:mech_row}
\end{figure}

\xhdr{Write-Erase Gate Sensitivity}

\Cref{fig:mech_sens} confirms the write-erase sensitivity asymmetry of \Cref{prop1} in trained models and explains why compressing $(f, b)$ to rank $64$ while keeping $(g, w, q)$ at rank $128$ in \eqref{eq:ours_proj} preserves model quality.
In \Figref{fig:mech_sens}{a}, truncating the singular values of $f$ or $b$ from rank $128$ to rank $64$ across all layers of the $406\text{M}$ uniform-rank \ours checkpoint increases validation NLL by at most $0.6\text{ mnats},$ whereas truncating $q,$ $w,$ or $g$ to rank $64$ increases validation NLL by $4.5,$ $6.4,$ and $247\text{ mnats},$ respectively.
\Figref{fig:mech_sens}{b} tests this asymmetry directly on the recurrence at $834\text{M}$: freezing $\vw_t$ degrades validation NLL across all three position buckets and increases overall NLL $14.5\times$ more than freezing $\vb_t$ in both \gdntwo ($+0.833$ vs.\ $+0.0574\text{ nats}$) and \ours ($+0.385$ vs.\ $+0.0266\text{ nats}$).
Consistent with \Cref{prop1:rem1}, erase perturbations have limited effect on the state because post-convolution $\silu$ keys concentrate their energy onto a small subspace, with a key participation ratio of only $7.43$ in \gdntwo and $5.98$ in \ours out of $d_k = 128$ channels (\Cref{tab:mech}).


\xhdr{Stationary Row-Energy Scaling}

\Cref{fig:mech_row} evaluates the stationary row-energy law (\Cref{prop2}) and learned readout scaling (\Cref{prop3:cor1}).
%
In \Figref{fig:mech_row}{a}, the closed-form prediction $\evm_j^\ast$ in \eqref{prop2:eq2} matches the empirical row energy $\evm_j$ across state channels at $166\text{M}$ with Spearman correlation $r_s = 0.957$ in \gdntwo and $0.960$ in \ours (static $\vgamma$).
Since the readout $\vo_t = (\vgamma_t \odot \vq_t)\mS_t$ scales each state row $j$ by $\evgamma_{t,j},$ weighting $\evm_j$ by the mean readout gain $\bar\evgamma_j = \E[\evgamma_{t,j}]$ in \ours reduces the median within-head $\mathrm{p90}/\mathrm{p10}$ spread of gain-weighted row energy $\bar\evgamma_j^2\evm_j$ from $14.7\times$ in \gdntwo to $9.6\times$ (\Figref{fig:mech_row}{b}, \Cref{tab:mech}).
Consistent with \Cref{prop3:cor1}, regressing $\log_{10}\abs{\evgamma_j}$ against $\log_{10}\evrho_j$ within each head (\Figref{fig:mech_row}{c}) yields a median log-log slope of $-0.35,$ which corresponds to a fractional power-law exponent $p = 0.35 \in (0, 1)$ in $\evgamma_j \propto \evrho_j^{-p}.$


\begin{figure}[!t]
    \centering
    \includegraphics[width=0.9\textwidth]{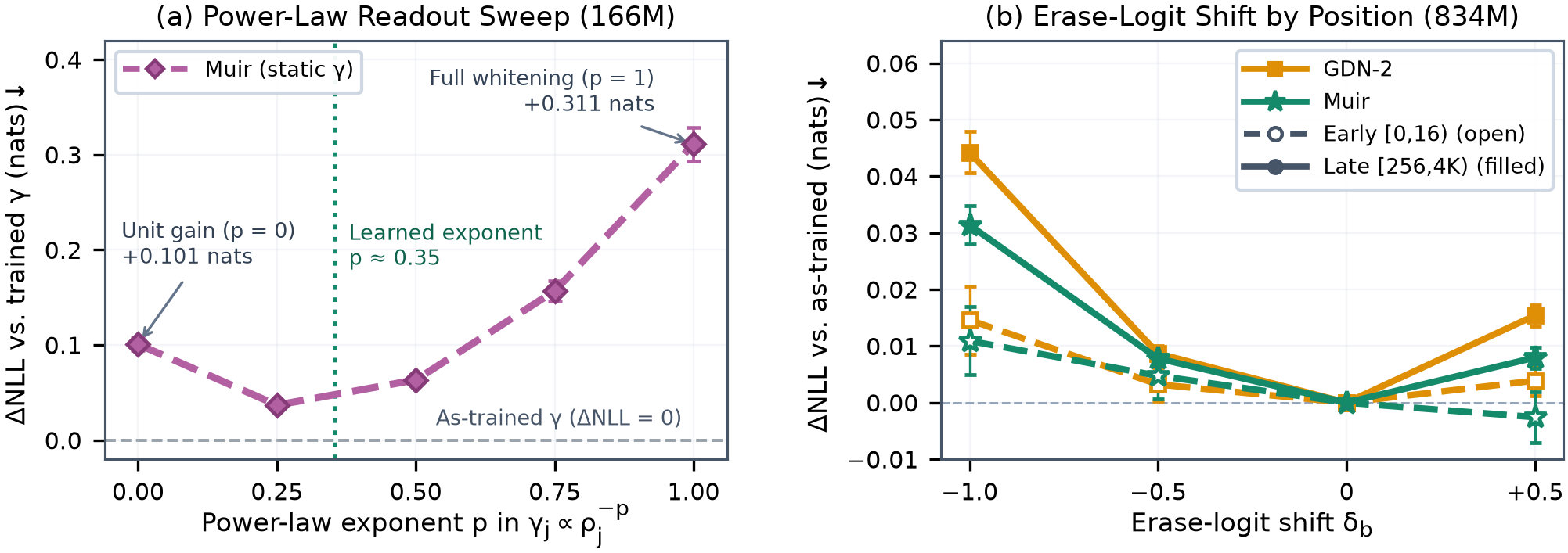}
    \caption{%
        \textbf{Readout-exponent sweeps and erase-logit shifts confirm the optimality of fractional power-law readout and the early-vs.-late erasure trade-off.}
        (a) Replacing the learned static gain in each head with an RMS-matched power law $\evgamma_j(p) \propto \evrho_j^{-p}$ yields a U-shaped loss curve minimized near the learned exponent ($p \approx 0.35$), outperforming both unit readout ($p = 0$) and full whitening ($p = 1$).
        %
        (b) Shifting the pre-activation erase logit ($\vz^b_t \gets \vz^b_t + \delta_b$) shows that weaker erasure degrades predictions across all positions, whereas stronger erasure leaves early tokens $[0, 16)$ nearly unchanged while degrading late tokens $[256, 4\text{K}).$\looseness=-1%
    }
    \label{fig:mech_cf}
\end{figure}

\input{tables/tab3_mech}

\xhdr{Counterfactual Readout and Erase Interventions}

In \Figref{fig:mech_cf}{a}, replacing the learned static gain at $166\text{M}$ with a pure power law of row RMS, $\evgamma_j(p) \propto \evrho_j^{-p},$ increases validation NLL by $+0.10\text{ nats}$ at $p = 0$ (where slow rows dominate readout) and $+0.31\text{ nats}$ at $p = 1$ (where noisy fast rows are over-amplified; \Cref{prop3:cor1:rem1}), but comes within $+0.037\text{ nats}$ of the unconstrained learned gain at $p = 0.25$ ($+0.063\text{ nats}$ at $p = 0.50$) near the fitted exponent $p = 0.35.$
%
In \Figref{fig:mech_cf}{b}, shifting the erase logit $\vz^b_t \gets \vz^b_t + \delta_b$ at $834\text{M}$ confirms the horizon trade-off of \Cref{prop4}: weakening erasure ($\delta_b = -1.0$) increases interference across all positions ($+0.031\text{--}0.044\text{ nats}$ at $[256, 4\text{K})$), whereas strengthening erasure ($\delta_b = +0.5$) leaves early tokens $[0, 16)$ within noise ($-0.003 \pm 0.005\text{ nats}$ in \ours) while degrading late tokens $[256, 4\text{K})$ ($+0.008\text{--}0.015\text{ nats}$) by overwriting stored context.
%
Consistent with \Cref{prop4} and \Cref{prop4:rem1}, suppressing slow-row interference at readout lets $1.34\text{B}$ \ours (\Cref{tab:mech}) operate with a $16.6\%$ lower parallel erase rate $\bar\evmu$ ($0.652$ vs.\ $0.782$) and $19.2\%$ lower adjacent self-erasure ($0.328$ vs.\ $0.406$), preserving $+16.1\%$ higher state spectral effective rank and a $59.6\%$ longer temporal lookback window ($49.3 \to 78.7$ tokens; \Cref{app:mech:intervene}).


\begin{figure}[!t]
    \centering
    \includegraphics[width=\textwidth]{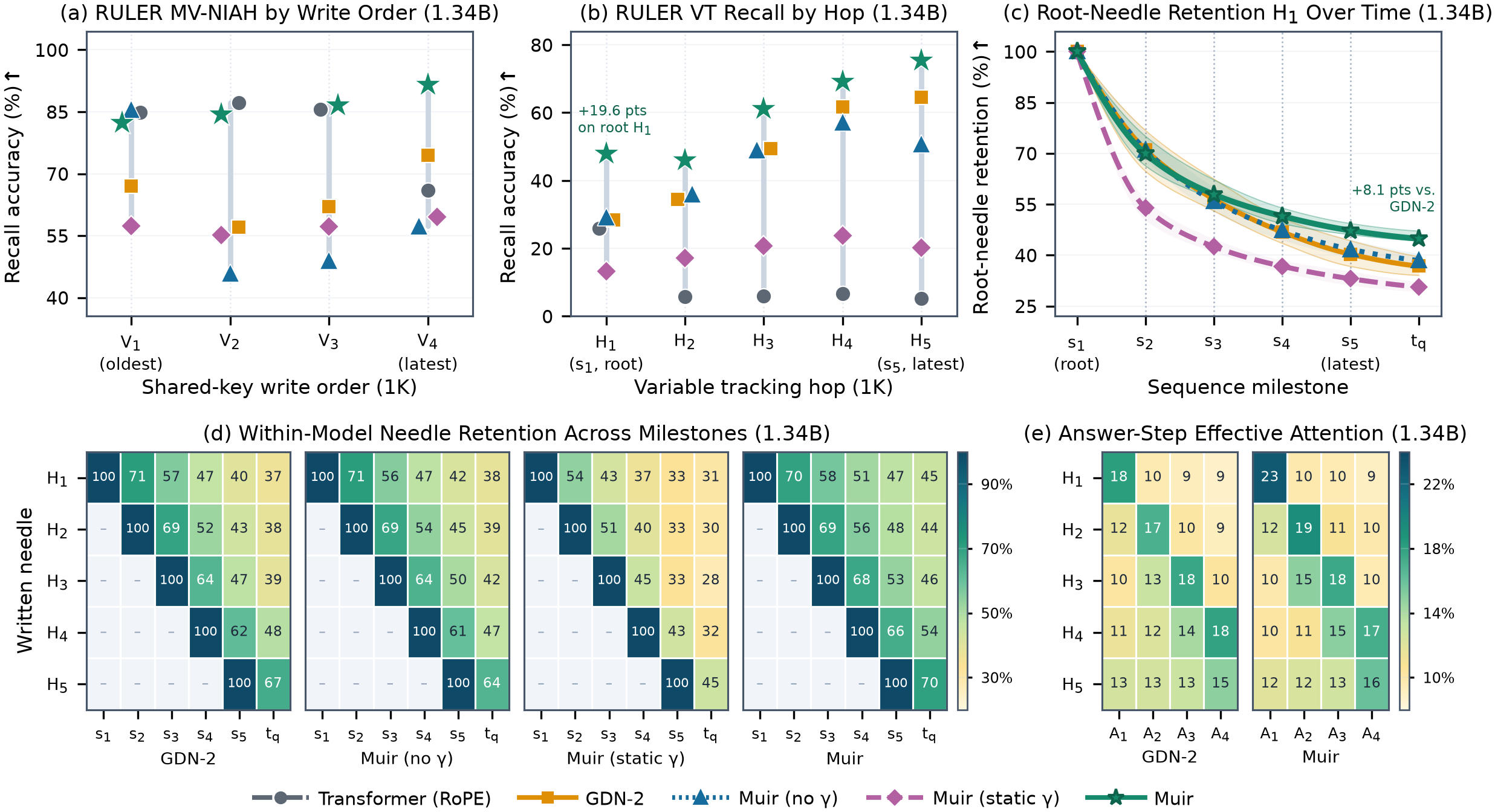}
    \caption{%
        \textbf{\ours preserves early variable bindings and routes readout to the target hop at each answer step.}
        (a) On 4-value MV-NIAH, \tf drops on the final write $V_4$ and unit-readout recurrences degrade after $V_1,$ whereas \ours remains accurate across $V_1\text{--}V_4.$
        (b) On 5-hop RULER VT, \tf recalls only the root assignment $H_1,$ whereas \ours maintains accuracy across all five hops $H_1\text{--}H_5.$
        %
        (c) Within-model root-needle ($H_1$) retention in $\mS_t$ from write milestone $s_1$ ($100\%$) to query milestone $t_{\mathrm{q}}$ across $16$ held-out prompts (interquartile bands across prompts).
        %
        (d) Within-model retention of written needles $H_1\text{--}H_5$ from each write ($100\%$) across sequence milestones $s_1\text{--}s_5$ to $t_{\mathrm{q}}.$
        %
        (e) Share of readout weight $\hat\vq_t\mS_t$ (over the haystack) on each needle $H_1\text{--}H_5$ across the four answer-generation steps $A_1\text{--}A_4$ that chain $H_1 \to \dotsb \to H_5.$%
    }
    \label{fig:mech_vt_interp}
\end{figure}

\xhdr{Multi-Value Recall and Multi-Hop Variable Tracking}

\Cref{fig:mech_vt_interp} connects our state-space diagnostics (\Cref{tab:mech}) to the 4-value MV-NIAH and 5-hop VT gains in \S\ref{sec:exp:results:recall} (details in \Cref{app:mech:vt}).
On MV-NIAH, \tf recalls $V_1\text{--}V_3,$ but query heads attend to already-emitted answer tokens on the fourth value $V_4$ (\Figref{fig:mech_vt_interp}{a}).
%
Consistent with latent multi-hop limits~\citep{Biran2024-fm,Yang2025-sq}, \tf resolves $H_1$ on VT but fails to chain $H_2\text{--}H_5$ in one pass (\Figref{fig:mech_vt_interp}{b}).
Recurrent models propagate bindings sequentially through $\mS_t,$ but repeated writes to the same state rows compete for memory.
In \gdntwo, heavy write-time erasure ($\bar\evmu = 0.782$) prematurely overwrites early bindings ($V_1\text{--}V_3$ and $H_1\text{--}H_3$).
Removing $\vgamma_t$ in \ours lets $V_1$ accumulate in slow rows that mask $V_2\text{--}V_4$ (\Cref{prop2}).
Static $\vgamma^{\mathrm{stat}}$ equalizes write order on MV-NIAH (\Cref{prop3:cor1}), but one fixed gain cannot match five hop distances (\Cref{prop3:rem1}).
%
%
Decomposing $\vo_t = \hat\vq_t\mS_t$ into per-needle write contributions, we track within-model retention of each hop write ($s_1\text{--}s_5$) in $\mS_t$ until query token $t_{\mathrm{q}}$ (\Figref{fig:mech_vt_interp}{c,d}) and how much weight $\hat\vq_t\mS_t$ places on each hop across answer steps $A_1\text{--}A_4$ (\Figref{fig:mech_vt_interp}{e}).
%
%
Lower write-time erasure (\Cref{prop4}) raises retention in \ours at $t_{\mathrm{q}}$ over \gdntwo from $37\%$ to $45\%$ on $H_1$ and from $46\%$ to $52\%$ on average across $H_1\text{--}H_5$ (\Figref{fig:mech_vt_interp}{c,d}).
Across $A_1\text{--}A_4,$ dynamic $\vgamma_t$ shifts the channel filter (\Cref{prop3}) so $\hat\vq_t\mS_t$ concentrates weight along the target diagonal $H_1\text{--}H_4,$ most notably on the oldest root hop ($18\%$ to $23\%$ at $A_1$; \Figref{fig:mech_vt_interp}{e}).

%% file: tables/tab3_mech.tex
\begin{table}[!t]
    \centering
    \caption{%
        \textbf{Trained \gdntwo and \ours checkpoints confirm the predicted write-erase sensitivity asymmetry, stationary row-energy law, and readout-induced erasure reduction.}
        Diagnostics are measured on held-out FineWeb-Edu validation sequences ($T = 4{,}096$) at $166\text{M}$ (gate sensitivity and stationary row-law fit) and $1.34\text{B}$ (within-head row-energy spread and write-time erasure).%
    }
    \label{tab:mech}
    \resizebox{\textwidth}{!}{%
        \setlength{\tabcolsep}{4.5pt}%
        \renewcommand{\arraystretch}{1.08}%
        \begin{tabular}{@{}c@{\hspace{3pt}}|l|l|c|cc|c|l@{}}
        \toprule
            \multicolumn{1}{@{}c@{\hspace{3pt}}}{} & {Diagnostic} & {Statistic} & {Scale} & {\gdntwo} & {\ours} & {Effect} & {Theory} \\
        \midrule
            \multirowrotate{2}{90}{\footnotesize Sens.}
            & Key participation ratio & $\bar{\eta}_{\mathrm{eff}} = 1/\E[\norm{\vk_t}_4^4]$ & 166M & 7.43 & 5.98 & $\ll\!128$ ch. & \Cref{prop1:rem1} \\
            & Write-erase sensitivity & Layer-mean $\E[\norm{\D_{\vz^w}\mS_t}_F^2] / \E[\norm{\D_{\vz^b}\mS_t}_F^2]$ & 166M & 13.2 & 13.3 & $13\times$ $w$ vs.\ $b$ & \Cref{prop1} \\
        \midrule
            \multirowrotate{3}{90}{\raisebox{0pt}[\height][0pt]{\footnotesize Energy}}
            & Stationary row-law fit & Median within-head $r_s(\evm_j, \evm_j^\ast)$ & 166M & 0.957 & 0.960 & $+\,0.003$ & \Cref{prop2} \\
            & Raw row-energy spread & Median within-head $\mathrm{p90}/\mathrm{p10}$ of $\evm_j$ & 1.34B & 14.7 & 17.2 & $+$17.0\% & \Cref{prop2} \\
            & Readout energy spread & Median within-head $\mathrm{p90}/\mathrm{p10}$ of $\bar\evgamma_j^2\evm_j$ & 1.34B & 14.7 & \tbf{9.6} & $-$34.7\% & \Cref{prop3:cor1} \\
        \midrule
            \multirowrotate{2}{90}{\footnotesize Erase}
            & Parallel erase rate & $\bar\evmu = \E[\inner{\vk_t}{\vb_t\odot\vk_t}]$ & 1.34B & 0.782 & \tbf{0.652} & $-$16.6\% & \Cref{prop4} \\
            & Adjacent self-erasure & $\E[\abs{\inner{\vk_{t-1}}{\vb_t\odot\vk_t}}]$ & 1.34B & 0.406 & \tbf{0.328} & $-$19.2\% & \Cref{prop4:rem1} \\
        \bottomrule
        \end{tabular}%
    }
\end{table}

%% file: sections/7_concl.tex
\section{Conclusion}
\label{sec:concl}

Channel-wise gates give delta-rule models fine-grained control over a fixed-size state, but its effective capacity depends on the state dynamics those gates induce.
In our trained models, key activation and unit normalization concentrate key energy onto roughly $5\%$ of channels per head, making the state $13\times$ more sensitive to write than to erase perturbations; \ours therefore fuses all gates into a low-rank projection with smaller decay and erase bottlenecks, widening the FFN by up to $30.4\%$ at fixed parameters and FLOPs.
Meanwhile, consistent with our theory, multi-timescale decay drives state row-energy variation across orders of magnitude, exceeding $100\times$ in heads mixing fast and slow channels.
%
Compared to \gdntwo, \ours reduces median readout-energy spread by $35\%$ and write-time erasure by $17\%,$ improves root-needle multi-hop retention from $37\%$ to $45\%,$ and lowers the median condition number of the chunkwise triangular solve by $24\%.$

Our results also reveal two remaining limitations.
First, under a fixed gate-rank budget, static and dynamic readout trade single-horizon retention against multi-lag adaptability.
%
On passkey lookup, a static filter suffices for a single query lag and preserves write rank for near-perfect recall to $1\text{M}$ tokens, whereas halving write rank to fund dynamic readout limits single-needle retention beyond $32\text{K}.$
Conversely, static readout cannot match varying lags on multi-hop and extractive tasks, and drifts on continuous text as slow-row energy accumulates, whereas dynamic readout adapts per token to resolve multi-lag queries and hold flat perplexity across $10\text{M}$-token streams.
Second, while our pure-JAX kernel matches \gdntwo throughput and trains $2.74\times$ faster than \tf at $32\text{K},$ Splash Attention remains faster at short contexts ($<\!11\text{K}$) because its Pallas kernel fuses masking and block contractions in VMEM while XLA materializes chunk temporaries in HBM.

These limitations suggest several directions for future work.
Parameterizing dynamic readout through our analytical matched filter by predicting one lag per head and tracking expected row energy could unify distant passkey recall, multi-lag routing, and stream stability without sacrificing write rank.
Alternatively, pairing \emph{sliding-window} attention with static-readout \ours would resolve this tension architecturally: local attention copies nearby spans exactly in constant memory, freeing the recurrent state to keep full write rank for distant lookup without periodic global attention or a growing KV cache.
At the kernel level, fusing leaf Neumann inversion, the boundary-pivoted merge tree, and the cross-chunk state scan into a single Pallas kernel would keep chunk temporaries in VMEM and narrow the short-context throughput gap.
Beyond gated delta recurrences, sensitivity-allocated gating and matched-filter readout extend to matrix-state test-time optimization models whose channels decay or update at different rates under unweighted queries.

%% file: sections/acks.tex
\section*{Acknowledgments}

We thank Amanda~Tomlinson, Burkay~Donderici, Nick~Tombari, and Xinling~Yu for helpful discussions and feedback throughout this work.
We also thank Kostia~Liepieshov for early work on numerically stable matrix inversion that informed our chunkwise kernel.
Finally, we are especially grateful to Jake~Ades for dedicated hardware and infrastructure support that made our large-scale training runs possible.

%% file: apps/a_related.tex
\clearpage
\section{Additional Related Work}
\label{sec:related}

\xhdr{Linear Attention, State-Space Models, and Fast-Weight Memories}

Replacing the $\gO(T)$ key-value cache and $\gO(T^2)$ attention matrix of softmax attention~\citep{Vaswani2017-el} with a fixed-size associative state matrix has roots in Hopfield networks~\citep{Hopfield1982-fm} and fast-weight controllers~\citep{Schmidhuber1992-di,Ba2016-xn}.
Kernelized linear transformers~\citep{Katharopoulos2020-km,Choromanski2020-tg,Pham2013-mi,Kacham2024-hx,Ghriss2026-tj} factorize the attention kernel into feature maps so that causal sequence mixing reduces to an additive outer-product recurrence $\mS_t = \mS_{t-1} + \vk_t\tr\vv_t,$ but purely additive accumulation suffers from unbounded state growth and associative cross-talk once sequence length exceeds the key dimension~\citep{Schlag2021-pg,Arora2023-tw,Jelassi2024-ns,Arora2024-lm,Arora2024-at}.
Structured state-space models (SSMs), long-convolution architectures, and fixed-decay linear recurrences such as S4~\citep{Gu2021-lj}, BiGS~\citep{Wang2023-sw}, H3~\citep{Fu2022-hd}, Hyena~\citep{Poli2023-yd}, LRU~\citep{Orvieto2023-hc}, RWKV~\citep{Peng2023-ni}, and RetNet~\citep{Sun2023-af} introduce data-independent exponential decay across state channels or heads, while selective SSMs and gated linear recurrences, including Mamba~\citep{Gu2023-wx}, Mamba-2~\citep{Dao2024-cx}, Gated Attention Unit (FLASH; \citealp{Hua2022-di}), Griffin~\citep{De2024-hf}, HGRN~\citep{Qin2023-us}, HGRN2~\citep{Qin2024-xm}, xLSTM~\citep{Beck2024-zr}, Gated Linear Attention (GLA; \citealp{Yang2023-pq}), Gated Slot Attention (GSA; \citealp{Zhang2024-oz}), ReGLA~\citep{Lu2025-bd}, and Mamba-3~\citep{Lahoti2026-fc}, make decay gates data-dependent so the model can selectively forget stale context~\citep{van-der-Westhuizen2018-xq,Sun2025-qs}.
Although data-dependent exponential decay bounds the total state norm, diagonal decay scales an entire state channel uniformly and cannot selectively erase a single key-value association without flushing other associations stored along the same channel~\citep{Yang2024-ja,Yang2025-dr}.


\xhdr{Delta-Rule Recurrences and Online Test-Time Optimization}

To support targeted associative updates, Fast Weight Programmers~\citep{Schlag2021-pg} and DeltaNet~\citep{Yang2024-ja} replace additive writes with the Widrow-Hoff delta rule~\citep{Widrow1960-rx}, which subtracts the current state prediction $\vk_t\mS_{t-1}$ along key $\vk_t$ prior to writing the new value.
%
Gated DeltaNet~\citep{Yang2025-dr} combines scalar exponential decay with a scalar Householder-like delta transition; more broadly, selective SSMs and gated linear recurrences are widely interleaved with softmax attention in hybrid sequence models~\citep{Ren2024-ob,Dong2025-xl,Lenz2025-ar,MiniMax2025-js,Kimi-Team2025-rg,Qiu2026-af}.
%
Subsequent extensions introduce vector or matrix-valued erasure and decay in RWKV-7~\citep{Peng2025-vq}, Kimi Linear (KDA; \citealp{Kimi-Team2025-rg}), and \gdntwo~\citep{Hatamizadeh2026-cw}, cross-layer value and residual routing in \citet{Cerruti2026-wd}, multi-step or negative-eigenvalue state transitions in \citet{Grazzi2024-rh}, DeltaProduct~\citep{Siems2025-fj}, and Comba~\citep{Hu2025-lj}, non-linear matrix-state transitions in $\mathrm{M}^2$RNN~\citep{Mishra2026-ow}, sparse slot-addressed state expansion in Sparse Delta Memory~\citep{Cabannes2026-ct}, and hierarchical or higher-order recurrent states in Log-Linear Attention~\citep{Guo2026-du} and Triadic Linear Attention~\citep{Sieberling2026-qn}.
A parallel line of work interprets these recurrences as online test-time optimization of an internal regression or associative memory objective: Test-Time Training (TTT; \citealp{Sun2024-ih,Zhang2025-eb,Liu2026-sb}), Longhorn~\citep{Liu2024-pp}, Titans~\citep{Behrouz2025-id}, ATLAS~\citep{Behrouz2025-dg}, Miras~\citep{Behrouz2025-ys}, Test-Time Regression~\citep{Wang2025-iq}, MesaNet~\citep{von-Oswald2025-ks}, Lattice~\citep{Karami2025-zz}, Gated KalmaNet~\citep{Peng2026-zt}, Kaczmarz Linear Attention~\citep{Zou2026-od}, and Momentum Delta Networks~\citep{Huang2026-rc} derive recurrent state updates as single-step gradient, momentum, Kaczmarz, Newton, or Kalman updates on a key-value reconstruction loss (see \Cref{app:optim} for the optimization lens).
%
Across these architectures, KDA~\citep{Kimi-Team2025-rg} and RWKV-7~\citep{Peng2025-vq} retain a $4d$ FFN by avoiding dense $d \times d$ recurrence gates: KDA restricts key erasure and value writes to a shared per-head scalar alongside low-rank decay and output gates, while RWKV-7 uses low-rank decay, removal, value-residual, and output gates with decoupled removal and replacement keys.
%
By contrast, \gdntwo~\citep{Hatamizadeh2026-cw} decouples channel-wise erasure and writes on a shared key via dense $d \times d$ projections at the expense of FFN width~\citep{Shazeer2020-lf,Geva2021-vm}.
%
For state readout, whereas query-key normalized attention~\citep{Hua2022-di,Dehghani2023-iu} and diagonal SSMs~\citep{Lahoti2026-fc} apply learned per-channel gains to both queries and keys after normalization, delta-rule recurrences require unit-norm keys ($\norm{\vk_t}_2 = 1$) to keep rank-one erasure contractive and therefore evaluate the unweighted product $\vo_t = \vq_t\mS_t$ (or a scalar per-head rescaling; \citealp{Hu2025-lj}) after parameter-free $\ell_2$ normalization~\citep{Henry2020-es,Schlag2021-pg,Yang2024-ja} and post-readout gating~\citep{Qiu2025-vl}.
Recent test-time optimization models instead precondition queries via a running key-covariance state~\citep{von-Oswald2025-ks,Peng2026-zt,Le2026-up} or orthogonalize $\mS_t$ via Newton-Schulz iterations during training~\citep{Aquino-Michaels2026-ks}.
Our analysis in \S\ref{sec:theory} shows that state updates are much more sensitive to write-gate than to erase-gate perturbations (\Cref{prop1}), allowing all five gates to share a low-rank bottleneck, and that a diagonal matched-filter readout $\vo_t = (\vgamma_t\odot\vq_t)\mS_t$ corrects the row-energy imbalance of channel-wise decay (\Cref{prop2,prop3,prop4}) without covariance state or matrix orthogonalization.


\xhdr{Hardware-Efficient Chunkwise Algorithms}

Executing token-level matrix recurrences efficiently on modern accelerators requires reformulating sequential outer products into block-parallel matrix multiplications~\citep{Blelloch1989-ma,Dao2022-vw,Dao2023-ln,Shah2024-qo}.
For diagonal decay recurrences, FLASH~\citep{Hua2022-di}, RetNet~\citep{Sun2023-af}, GLA~\citep{Yang2023-pq}, Lightning Attention-2~\citep{Qin2024-lt}, Mamba-2~\citep{Dao2024-cx,Santoni2026-et}, FLA~\citep{Yang2024-vn}, and Tiled Flash Linear Attention~\citep{Beck2025-lk} partition the sequence into chunks of length $C,$ compute within-chunk outputs via masked matrix-matrix multiplications, and propagate the state across chunk boundaries.
When the state transition includes rank-one key erasure $\I - \vk_t\tr\hat\vk_t,$ unrolling a chunk of $C$ transitions yields a product of $C$ rank-one perturbations.
DeltaNet~\citep{Yang2024-ja} and Gated DeltaNet~\citep{Yang2025-dr} compact this product using the classical WY representation of Householder transformations~\citep{Bischof1987-cw,Schreiber1989-fu,Joffrain2006-ct}, which reduces within-chunk parallelization to solving a $C \times C$ unit lower-triangular system $\mT = \I + \mL_{\mathrm{strict}}.$
%
On GPUs, FLA~\citep{Yang2024-vn} and KDA~\citep{Kimi-Team2025-rg} subdivide each $C = 64$ chunk into $16$-token sub-chunks~\citep{Yang2023-pq}, compute decays relative to sub-chunk boundaries instead of dividing by full-chunk prefixes, invert the four diagonal $16\times 16$ blocks sequentially in SRAM, and merge off-diagonal $16\times 16$ blocks on TensorCores.
%
On TPUs, however, the matrix multiply unit (MXU) is a $256\times 256$ systolic array while the vector processing unit (VPU) is over $60\times$ slower in \texttt{bfloat16}~\citep{Austin2025-yc,Jiang2026-cm}, so $16\times 16$ tiles under-utilize the MXU and sequential leaf substitution stalls on the VPU.
%
Concurrent work on FLAT~\citep{Mishra2026-ec} provides hand-tuned Pallas TPU kernels for diagonal-decay recurrences (Linear Attention, Mamba-2, GLA) without the delta-rule triangular solve; FlyWheel~\citep{Bao2026-ul} develops Pallas TPU kernels for softmax attention and for Gated DeltaNet inference via MXU triangular inversion; and \citet{Sobczyk2026-lr} benchmark matrix-product-rich triangular inversion (including hybrid leaf-Neumann and recursive block-merge forward kernels) on an already-materialized $\mT$ on GPUs and NPUs.
%
In \S\ref{sec:method:kernel}, rather than materializing $\mT$ upfront via prefix division, we form cross-blocks via boundary-pivoted log-decay splitting at each merge step of recursive block-Neumann inversion~\citep{Heller1974-mg,Orcutt1974-js,Heller1978-ni,Bini1994-bt,Sobczyk2026-lr} and differentiate the solve in two full-chunk GEMMs with on-the-fly VMEM pivot recomputation, routing $98\%$ of $C = 256$ training FLOPs onto $256\times 256$ MXU contractions.

%% file: apps/b_notation.tex
\clearpage
\section{Summary of Notation}
\label{app:notation}

\input{tables/tab4_notation}

%% file: tables/tab4_notation.tex
\begin{table}[!ht]
    \centering
    \caption{%
        \textbf{Summary of notation.}
        All vectors are row vectors; hats ($\hat{\cdot}$) denote gate-modulated vectors or decayed states, bars ($\bar{\cdot}$) denote expectations or empirical means, and asterisks (${}^\ast$) denote stationary fixed points or optimal solutions.%
    }
    \label{tab:notation}
    \resizebox{\textwidth}{!}{%
        \setlength{\tabcolsep}{3pt}%
        \renewcommand{\arraystretch}{1.01}%
        \begin{tabular}{@{}ll|l@{}}
        \toprule
        {Symbol} & & {Definition} \\
        \midrule
            $L$ & $\mathbb{N}$ & Number of residual layers \\
            $d$ & $\mathbb{N}$ & Model residual-stream width \\
            $d_{\mathrm{ffn}}$ & $\mathbb{N}$ & FFN intermediate width ($4d$ in \tf, $3d - 256$ in \gdntwo, $4d - 512$ in \ours) \\
            $d_k, d_v$ & $\mathbb{N}$ & Per-head key and value dimensions ($d_k = d_v = 128$) \\
            $N_h$ & $\mathbb{N}$ & Number of attention heads ($N_h = d/d_k$) \\
            $T, C$ & $\mathbb{N}$ & Sequence length and chunk length ($C = 256$) \\
            $N_c$ & $\mathbb{N}$ & Number of chunks per sequence ($N_c = T/C$) \\
            $r_f, r_b, r_g, r_w, r_q$ & $\mathbb{N}$ & Bottleneck ranks of decay, erase, output, write, and query-gain projections \\
            $R_{\mathrm{tot}}$ & $\mathbb{N}$ & Total fused gate rank $R_{\mathrm{tot}} \defas r_f + r_b + r_g + r_w + r_q$ ($R_{\mathrm{tot}} = 512$ in \ours) \\
            $n_0$ & $\mathbb{N}$ & Diagonal leaf block size ($n_0 = 8$) \\
            $r$ & $\mathbb{N}$ & Binary merge-tree depth ($r = \log_2(C/n_0) = 5$) \\
        \midrule
            $\vx_t$ & $\Rvec{d}$ & Token embedding entering a sequence-mixing block at timestep $t$ \\
            $\vq_t, \vk_t$ & $\bS{d_k-1}$ & $\ell_2$-normalized per-head query and key vectors ($\norm{\vq_t}_2 = \norm{\vk_t}_2 = 1$) \\
            $\vv_t$ & $\Rvec{d_v}$ & Per-head value vector after $\silu$ activation \\
            $\valpha_t$ & $(0, 1)^{1\times d_k}$ & Channel-wise exponential decay (retention) gate \\
            $\vb_t$ & $(0, 2)^{1\times d_k}$ & Channel-wise erase gate $\vb_t = 2\sigmoid(\vz^b_t)$ \\
            $\vw_t$ & $(0, 1)^{1\times d_v}$ & Channel-wise write gate $\vw_t = \sigmoid(\vz^w_t)$ \\
            $\vg_t$ & $\Rvec{d_v}$ & Pre-activation output gate ($\silu(\vg_t)$ modulates $\rms(\vo_t)$) \\
            $\vgamma^{\mathrm{stat}}$ & $\Rvec{d_k}$ & Learnable per-head static query-gain vector (initialized to $\vone_{d_k}$) \\
            $\vgamma_t$ & $\Rvec{d_k}$ & Dynamic query-readout gain $\vgamma_t = \vgamma^{\mathrm{stat}} \odot 2\sigmoid(\vz^q_t)$ \\
            $\hat\vq_t, \hat\vk_t$ & $\Rvec{d_k}$ & Gain-modulated query $\hat\vq_t \defas \vgamma_t \odot \vq_t$ and erase-gated key $\hat\vk_t \defas \vb_t \odot \vk_t$ \\
            $\hat\vv_t$ & $\Rvec{d_v}$ & Write-gated value vector $\hat\vv_t \defas \vw_t \odot \vv_t$ \\
            $\mS_t$ & $\R{d_k\times d_v}$ & Per-head recurrent associative state matrix at timestep $t$ \\
            $\hat\mS_t$ & $\R{d_k\times d_v}$ & Decayed prior state $\hat\mS_t \defas \Diag(\valpha_t)\mS_{t-1}$ \\
            $\tilde\vv_t$ & $\Rvec{d_v}$ & Effective delta write vector $\tilde\vv_t \defas \hat\vv_t - \hat\vk_t\hat\mS_t$ \\
            $\vo_t$ & $\Rvec{d_v}$ & Per-head state readout $\vo_t = \hat\vq_t\mS_t$ (or $\vq_t\mS_t$ under unit gain) \\
        \midrule
            $\evmu_t$ & $(0, 2)$ & Parallel erase strength $\evmu_t \defas \inner{\vk_t}{\hat\vk_t} = \inner{\vb_t}{\vk_t^{\odot 2}}$ along $\vk_t$ (Prop.~\ref{prop1}) \\
            $\vs_t$ & $\Rvec{d_k}$ & Orthogonal erase residual $\vs_t \defas \hat\vk_t - \evmu_t\vk_t$ ($\inner{\vs_t}{\vk_t} = 0$) (Prop.~\ref{prop1}) \\
            $\eta_{\mathrm{eff}}(\vk_t)$ & $[1, d_k]$ & Per-token key participation ratio $\eta_{\mathrm{eff}}(\vk_t) \defas 1/\norm{\vk_t}_4^4$ (Prop.~\ref{prop1}) \\
            $\evc_j$ & $(0, 1)$ & Expected channel key energy $\evc_j \defas \E[\evk_{t,j}^2]$ ($\sum_{j=1}^{d_k} \evc_j = 1$) \\
            $\evlambda_j$ & $(0, 1)$ & Mean squared channel decay factor $\evlambda_j \defas \E[\evalpha_{t,j}^2]$ \\
            $\bar\evb_j$ & $(0, 2)$ & Expected channel erase rate $\bar\evb_j \defas \E[\evb_{t,j}]$ \\
            $\evsigma_v^2$ & $(0, \infty)$ & Expected gated value energy $\evsigma_v^2 \defas \E[\norm{\hat\vv_t}_2^2]$ \\
            $\tau_{\mathrm{eff},j}(\bar\evb_j)$ & $(1/2, \infty)$ & Effective channel memory horizon $(1 - \evlambda_j + 2\evlambda_j\bar\evb_j\evc_j)\inv$ \eqref{prop2:eq2} \\
            $\evm_{t,j}$ & $[0, \infty)$ & Expected state-row energy $\evm_{t,j} \defas \E[\norm{\mS_{t,j,:}}_2^2]$ at timestep $t$ \eqref{prop2:eq1} \\
            $\evm_j^\ast, \evrho_j$ & $[0, \infty)$ & Stationary row energy $\evm_j^\ast \defas \lim_{t\to\infty}\evm_{t,j} = \evc_j\evsigma_v^2\,\tau_{\mathrm{eff},j}(\bar\evb_j)$ and RMS $\evrho_j \defas \sqrt{\smash{\evm_j^\ast}\vphantom{\evm}}$ \eqref{prop2:eq2} \\
            $\eva_{\ell,j}, \bar\eva_{\ell,j}$ & $(-1, 1), (0, 1)$ & Lag-$\ell$ cumulative retention $\prod_{s=t-\ell+1}^t\evalpha_{s,j}(1-\evb_{s,j}\evk_{s,j}^2)$ and its mean $\E[\eva_{\ell,j}]$ (Prop.~\ref{prop3}) \\
            $\evgamma_j^\ast(\ell)$ & $(0, \infty)$ & Lag-$\ell$ matched-filter readout gain $\evgamma_j^\ast(\ell) = \kappa_\ell\,\bar\eva_{\ell,j}/\evm_j^\ast \propto \bar\eva_{\ell,j}/\evrho_j^2$ \eqref{prop3:eq2} \\
            $g_H(\evlambda_j)$ & $(0, 1)$ & Geometric-horizon signal survival $g_H(\evlambda_j) \defas \E_\ell[\evlambda_j^{\ell\sdiv 2}]$ under mean lag $H > 1$ (Cor.~\ref{prop3:cor1}) \\
            $\gI(\bar\evb, \vgamma)$ & $(0, \infty)$ & Normalized horizon-averaged interference-to-signal ratio (Prop.~\ref{prop4}) \\
            $\bar\evb^\ast(\Phi)$ & $(0, 2)$ & Equilibrium uniform erase rate under interference multiplier $\Phi > 0$ (Prop.~\ref{prop4}) \\
        \midrule
            $\mLambda_t$ & $(-\infty, 0]^{1\times d_k}$ & Within-chunk cumulative log-decay $\mLambda_t \defas \sum_{r=1}^t\log\valpha_r$ (rows of $\mLambda \in \R{C\times d_k}$) \\
            $\vpi_t, \vpi_{t,s}$ & $(0, 1]^{1\times d_k}$ & Chunk-prefix decay $\vpi_t \defas \exp(\mLambda_t)$ and pairwise relative decay $\vpi_{t,s} \defas \exp(\mLambda_t-\mLambda_s)$ ($t \ge s$) \\
            $\mL_{\mathrm{strict}}$ & $\R{C\times C}$ & Strictly lower-triangular coupling matrix $(\mL_{\mathrm{strict}})_{t,s} = \inner{\hat\vk_t}{\vpi_{t,s}\odot\vk_s}$ ($t > s$) \\
            $\mT$ & $\R{C\times C}$ & Unit lower-triangular chunk system matrix $\mT \defas \I + \mL_{\mathrm{strict}}$ \\
            $\mM$ & $\R{C\times C}$ & Lower-triangular causal readout matrix $\emM_{t,s} = \inner{\hat\vq_t}{\vpi_{t,s}\odot\vk_s}$ ($t \ge s$) \\
            $\hat\mK_{\odot\pi}, \hat\mQ_{\odot\pi}, \tilde\mK$ & $\R{C\times d_k}$ & Prefix-decayed keys $\vpi_t\odot\hat\vk_t,$ queries $\vpi_t\odot\hat\vq_t,$ and suffix-decayed keys $\vpi_{C,s}\odot\vk_s$ \\
            $\mU$ & $\R{C\times d_v}$ & Solved chunk value operator $\mU \defas \mT\inv\hat\mV$ (where $\hat\mV$ has rows $\hat\vv_t$) \\
            $\mY$ & $\R{C\times d_k}$ & Solved chunk key operator $\mY \defas \mT\inv\hat\mK_{\odot\pi}$ \\
            $\tilde\mV$ & $\R{C\times d_v}$ & Stacked effective chunk writes $\tilde\mV = \mU - \mY\mS_0$ (with rows $\tilde\vv_t$) \\
            $\mS_0, \mS_C$ & $\R{d_k\times d_v}$ & Chunk-initial state and end-of-chunk state $\mS_C = (\Diag(\vpi_C) - \tilde\mK\tr\mY)\mS_0 + \tilde\mK\tr\mU$ \eqref{eq:chunk_state_out} \\
            $\mO$ & $\R{C\times d_v}$ & Chunk output matrix $\mO = (\hat\mQ_{\odot\pi} - \mM\mY)\mS_0 + \mM\mU$ \eqref{eq:chunk_state_out} \\
        \bottomrule
        \end{tabular}%
    }
\end{table}

%% file: apps/c_optim.tex
\clearpage
\section{Online Regression Formulation}
\label{app:optim}

Here, we show how the state update and readout of \gdntwo and \ours arise from a preconditioned online regression objective~\citep{Schlag2021-pg,Yang2024-ja,Sun2024-ih,Liu2024-pp,Yang2025-dr,Behrouz2025-ys,Wang2025-iq,Hatamizadeh2026-cw}, and summarize how \ours compares with prior linear-time sequence models (\Cref{tab:optim_taxonomy}).


\xhdr{Online Associative Memory Update}

Consider a single head maintaining a linear associative memory $\mS \in \R{d_k \times d_v}$ that maps keys $\vk \in \bS{d_k-1}$ to values $\vv \in \Rvec{d_v}$ via $\vk \mapsto \vk\mS.$
At timestep $t,$ updating $\mS_{t-1} \to \mS_t$ upon observing $(\vk_t, \vv_t)$ proceeds in two stages.
First, before incorporating the new observation, the prior state $\mS_{t-1}$ is contracted row-wise via the channel-wise decay operator $\hat\mS_t \defas \Diag(\valpha_t)\mS_{t-1},$ which minimizes the row-weighted Frobenius proximity objective $\norm{\mS - \mS_{t-1}}_F^2 + \sum_{j=1}^{d_k}(\evalpha_{t,j}\inv - 1)\norm{\mS_{j,:}}_2^2.$
Second, given the decayed state $\hat\mS_t,$ consider the gated key-value reconstruction loss
\[
    \gL_t(\mS) \defas \frac{1}{2}\norm{\hat\vk_t\mS - \hat\vv_t}_2^2,
    \qquad\text{where}\qquad
    \hat\vk_t \defas \vb_t \odot \vk_t,
    \quad
    \hat\vv_t \defas \vw_t \odot \vv_t,
\]
whose Euclidean gradient with respect to $\mS$ is $\nabla_{\mS}\gL_t(\mS) = \hat\vk_t\tr(\hat\vk_t\mS - \hat\vv_t).$
Applying a left row preconditioner $\mP_t \defas \Diag(\vb_t)\inv$ at $\mS = \hat\mS_t$ so that the update direction aligns with the unit key $\vk_t = \hat\vk_t\Diag(\vb_t)\inv$ yields the single-step update
\[
    \mS_t = \hat\mS_t - \mP_t\nabla_{\mS}\gL_t(\hat\mS_t)
    = \hat\mS_t - \vk_t\tr\big(\hat\vk_t\hat\mS_t - \hat\vv_t\big)
    = \big(\I - \vk_t\tr(\vb_t \odot \vk_t)\big)\Diag(\valpha_t)\mS_{t-1} + \vk_t\tr(\vw_t \odot \vv_t),
\]
which recovers the state update in \gdntwo \eqref{eq:gdn2} and \ours \eqref{eq:ours}.


\xhdr{Matched-Filter Readout}

In standard delta-rule architectures (DeltaNet, Gated DeltaNet, RWKV-7, KDA, \gdntwo), querying $\mS_t$ with $\vq_t$ evaluates the unweighted readout $\vo_t = \vq_t\mS_t = \sum_{j=1}^{d_k}\evq_{t,j}\mS_{t,j,:}.$
However, because each row $j \in \{1, \dotsc, d_k\}$ evolves under its own decay factor $\evalpha_{t,j}$ and erase gate $\evb_{t,j},$ the rows of $\mS_t$ accumulate multi-write interference with unequal stationary variance $\evm_j^\ast = \E[\norm{\mS_{t,j,:}}_2^2]$ (\Cref{prop2}).
When querying a target association written $\ell$ timesteps earlier, the $j$-th row carries signal amplitude proportional to $\evlambda_j^{\ell\sdiv 2}$ and noise variance $\evm_j^\ast.$
%
By \Cref{prop3}, the minimum-MSE linear estimator weights each row $j$ by its signal-to-noise ratio $\evgamma_{t,j} \propto \bar\eva_{\ell,j}/\evm_j^\ast \approx \evlambda_j^{\ell\sdiv 2}/\evm_j^\ast,$ which is equivalent to querying $\mS_t$ with the gain-modulated query $\hat\vq_t = \vgamma_t \odot \vq_t$ in \eqref{eq:ours}:
\[
    \vo_t = \hat\vq_t\mS_t = \big(\vgamma_t \odot \vq_t\big)\mS_t = \vq_t\Diag(\vgamma_t)\mS_t.
\]
%
%
As summarized in \Cref{tab:optim_taxonomy}, prior delta-rule models restrict state decay and key erasure to per-head scalars (DeltaNet, Gated DeltaNet), pair channel-wise decay with scalar erasure (KDA), use low-rank decay and removal gates on decoupled removal and replacement keys (RWKV-7), or use dense $d \times d$ projections for channel-wise erasure and write gating (\gdntwo), all with unweighted readout $\vq_t\mS_t.$
By contrast, \ours pairs channel-wise decay, erasure, and write gating on a shared key with gain-modulated readout $(\vgamma_t \odot \vq_t)\mS_t$ while parameterizing all five channel-wise gates via a fused low-rank projection.

\input{tables/tab5_optim}

%% file: tables/tab5_optim.tex
\begin{table}[!t]
    \centering
    \caption{%
        \textbf{Comparison of per-head state recurrence, query readout, and gate projection parameterizations across linear-time sequence models.}
        %
        For each model, we list the per-head decay, key-erasure, and value-write factors, the query readout, and whether each learned gate projection across heads is a per-head scalar ($d \to N_h$), low-rank bottleneck ($d \to r \to d$), or dense matrix ($d \to d$).
        ($^{\dagger}$RetNet fixes its per-head decay scalar $\evalpha$ without an input projection; $^{\ddagger}$RWKV-7 combines decay and removal additively, writes with a decoupled replacement key $\tilde\vk_t\tr\vv_t,$ and includes a low-rank cross-layer value-residual gate.)%
    }
    \label{tab:optim_taxonomy}
    \resizebox{\textwidth}{!}{%
        \setlength{\tabcolsep}{5pt}%
        \renewcommand{\arraystretch}{1.15}%
        \begin{tabular}{@{}l@{\hspace{8pt}}|cccc|ccc@{}}
        \toprule
            & \multicolumn{4}{c|}{{Per-Head Recurrence and Readout}} & \multicolumn{3}{c@{}}{{Gate Projections}} \\
            \cmidrule(lr){2-5} \cmidrule(l){6-8}
            {Model} & {Decay} & {Erase} & {Write} & {Readout} & {Scalar} & {Low-Rank} & {Dense} \\
        \midrule
            Lin.\ Attn.~\citep{Katharopoulos2020-km} & $\I$ & $\I$ & $\vv_t$ & $\vq_t\mS_t$ & -- & -- & -- \\
            RetNet~\citep{Sun2023-af} & $\evalpha\I$ & $\I$ & $\vv_t$ & $\vq_t\mS_t$ & --\rlap{$^{\dagger}$} & -- & $\vg_t$ \\
            Mamba-2~\citep{Dao2024-cx} & $\evalpha_t\I$ & $\I$ & $\evbeta_t\vv_t$ & $\vq_t\mS_t$ & $\evalpha_t, \evbeta_t$ & -- & $\vg_t$ \\
            GLA~\citep{Yang2023-pq} & $\Diag(\valpha_t)$ & $\I$ & $\vv_t$ & $\vq_t\mS_t$ & -- & $\valpha_t$ & $\vg_t$ \\
            DeltaNet~\citep{Yang2024-ja} & $\I$ & $\I - \evbeta_t\vk_t\tr\vk_t$ & $\evbeta_t\vv_t$ & $\vq_t\mS_t$ & $\evbeta_t$ & -- & $\vg_t$ \\
            RWKV-7~\citep{Peng2025-vq}$^{\ddagger}$ & \multicolumn{2}{c}{$\Diag(\valpha_t) - (\vb_t \odot \vk_t)\tr\vk_t$} & $\vv_t\rlap{$^{\ddagger}$}$ & $\vq_t\mS_t$ & -- & $\valpha_t, \vb_t, \vg_t$ & -- \\
            Gated DeltaNet~\citep{Yang2025-dr} & $\evalpha_t\I$ & $\I - \evbeta_t\vk_t\tr\vk_t$ & $\evbeta_t\vv_t$ & $\vq_t\mS_t$ & $\evalpha_t, \evbeta_t$ & -- & $\vg_t$ \\
            KDA~\citep{Kimi-Team2025-rg} & $\Diag(\valpha_t)$ & $\I - \evbeta_t\vk_t\tr\vk_t$ & $\evbeta_t\vv_t$ & $\vq_t\mS_t$ & $\evbeta_t$ & $\valpha_t, \vg_t$ & -- \\
            \gdntwo~\citep{Hatamizadeh2026-cw} & $\Diag(\valpha_t)$ & $\I - \vk_t\tr(\vb_t \odot \vk_t)$ & $\vw_t \odot \vv_t$ & $\vq_t\mS_t$ & -- & $\valpha_t, \vg_t$ & $\vb_t, \vw_t$ \\
            \midrule
            \tbf{\ours (ours)} & $\Diag(\valpha_t)$ & $\I - \vk_t\tr(\vb_t \odot \vk_t)$ & $\vw_t \odot \vv_t$ & $(\vgamma_t \odot \vq_t)\mS_t$ & -- & $\valpha_t, \vb_t, \vg_t, \vw_t, \vgamma_t$ & -- \\
        \bottomrule
        \end{tabular}%
    }
\end{table}

%% file: apps/d_theory_v2.tex
\clearpage
\section{Mathematical Proofs}
\label{app:theory}

\subsection{First-Order Gate Sensitivity and Spectral Norm of Oblique Erase Transitions}
\label{app:theory:prop1}

\begin{proof}[Proof of \Cref{prop1}]
    (i) Because $\hat\mS_t$ does not depend on $(\vb_t,\vw_t),$ the update \eqref{eq:gdn2} is affine in $\vb_t$ and in $\vw_t$ separately.
    For any $\eps\neq 0,$ $\eps\inv(\mS_t(\vb_t+\eps\delta\vb,\vw_t)-\mS_t(\vb_t,\vw_t))=-\vk_t\tr\big((\delta\vb\odot\vk_t)\hat\mS_t\big)$ and $\eps\inv(\mS_t(\vb_t,\vw_t+\eps\delta\vw)-\mS_t(\vb_t,\vw_t))=\vk_t\tr(\delta\vw\odot\vv_t),$ which gives \eqref{prop1:eq1}.

    (ii) For any $\vu\in\Rvec{d_k}$ and $\vr\in\Rvec{d_v},$ the rank-one outer product $\vu\tr\vr\in\R{d_k\times d_v}$ satisfies $\norm{\vu\tr\vr}_F^2 = \Tr(\vr\tr\vu\vu\tr\vr) = \norm{\vu}_2^2\norm{\vr}_2^2.$
    Applying this to \eqref{prop1:eq1} with $\vu=\vk_t$ and $\norm{\vk_t}_2^2=1$ gives $\norm{\D_{\vb}\mS_t[\delta\vb]}_F^2 = \norm{\vr_b}_2^2$ and $\norm{\D_{\vw}\mS_t[\delta\vw]}_F^2 = \norm{\vr_w}_2^2,$ where $\vr_b\defas(\delta\vb\odot\vk_t)\hat\mS_t = \sum_{j=1}^{d_k}\delta\evb_j\evk_{t,j}\hat\mS_{t,j,:}$ and $\vr_w\defas\delta\vw\odot\vv_t.$ 
    Taking expectations over $\delta\vb\sim\gN(\vzero,\I_{d_k})$ and $\delta\vw\sim\gN(\vzero,\I_{d_v})$ using $\E[\delta\evb_i\delta\evb_j]=\delta_{ij}$ and $\E[\delta\evw_j^2]=1$ yields \eqref{prop1:eq2}:
    \[
        \E_{\delta\vb}\!\left[\norm{\vr_b}_2^2\right] = \sum_{i,j=1}^{d_k}\E[\delta\evb_i\delta\evb_j]\,\evk_{t,i}\evk_{t,j}\inner{\hat\mS_{t,i,:}}{\hat\mS_{t,j,:}} = \sum_{j=1}^{d_k}\evk_{t,j}^2\norm{\hat\mS_{t,j,:}}_2^2,\quad
        \E_{\delta\vw}\!\left[\norm{\vr_w}_2^2\right] = \sum_{j=1}^{d_v}\E[\delta\evw_j^2]\,\evv_{t,j}^2 = \norm{\vv_t}_2^2.
    \]
    (iii) Since $\inner{\vs_t}{\vk_t}=0,$ $\sum_{j=1}^{d_k}\evk_{t,j}^2=1,$ and $\sum_{j=1}^{d_k}\evk_{t,j}^2\evb_{t,j}=\evmu_t,$ the Pythagorean theorem gives
    \[
        \norm{\vs_t}_2^2 = \norm{\hat\vk_t}_2^2 - \evmu_t^2 = \sum_{j=1}^{d_k}\evk_{t,j}^2\evb_{t,j}^2 - \evmu_t^2 = \sum_{j=1}^{d_k}\evk_{t,j}^2\big(\evb_{t,j}-\evmu_t\big)^2.
    \]
    Differentiating $\evmu_t=\sum_{j=1}^{d_k}\evb_{t,j}\evk_{t,j}^2$ with respect to $\vb_t$ gives $\nabla_{\vb}\evmu_t=\vk_t^{\odot 2},$ so $\norm{\nabla_{\vb}\evmu_t}_2^2=\sum_{j=1}^{d_k}\evk_{t,j}^4=\norm{\vk_t}_4^4=\eta_{\mathrm{eff}}(\vk_t)\inv.$
    Since each $\evk_{t,j}^2\in[0,1],$ we have $\evk_{t,j}^4\le\evk_{t,j}^2$ and summing over $j$ gives $\norm{\vk_t}_4^4\le 1$ (with equality if and only if $\vk_t$ is $1$-sparse).
    Conversely, Cauchy-Schwarz applied to $\vk_t^{\odot 2}$ and $\vone_{d_k}$ yields $1=\inner{\vk_t^{\odot 2}}{\vone_{d_k}}^2\le d_k\norm{\vk_t}_4^4$ (with equality if and only if $|\evk_{t,j}|=d_k^{-1\sdiv 2}$ for all $j$), which completes the proof.
\end{proof}

In Proposition~\ref{prop1:item3} and \Cref{prop1:rem2}, we decompose the erase-gated key $\hat\vk_t \defas \vb_t \odot \vk_t \in \Rvec{d_k}$ relative to the unit key $\vk_t \in \bS{d_k-1}$ ($d_k \ge 2$) into its parallel component $\evmu_t\vk_t$ and orthogonal residual $\vs_t \defas \hat\vk_t - \evmu_t\vk_t$ as
\[
    \hat\vk_t = \evmu_t\vk_t + \vs_t = \evmu_t\vk_t + \evs_t\vu_t,
    \qquad
    \evmu_t \defas \inner{\vk_t}{\vb_t \odot \vk_t} \in (0, 2),
    \qquad
    \evs_t \defas \norm{\vs_t}_2 \ge 0,
\]
where $\inner{\vk_t}{\vu_t} = 0$ and $\norm{\vu_t}_2 = 1$ (with $\vu_t \defas \vs_t/\evs_t$ whenever $\evs_t > 0,$ and any unit vector orthogonal to $\vk_t$ when $\evs_t = 0$).
The following lemma establishes the exact spectral norm and lower bound of the rank-one erase transition matrix $\mE_t \defas \I - \vk_t\tr\hat\vk_t$ cited in \Cref{prop1:rem2}.

\begin{lemma}[Spectral norm of oblique rank-one erase]
\label{lem1}
    Let $\vk_t \in \bS{d_k-1}$ ($d_k \ge 2$) and $\vb_t \in (0, 2)^{1\times d_k}$ with $\evmu_t \in (0, 2),$ $\evs_t = \norm{\vs_t}_2 \ge 0,$ and $\delta_t \defas \evmu_t(2 - \evmu_t) \in (0, 1].$
    Then the erase transition matrix $\mE_t \defas \I - \vk_t\tr(\vb_t \odot \vk_t)$ satisfies
    \begin{equation}
    \label{eq:lem1_norm}
        \norm{\mE_t}_2
        = \sqrt{1 + \frac{\evs_t^2 - \delta_t + \sqrt{(\evs_t^2 - \delta_t)^2 + 4\evs_t^2}}{2}}
        \ge \sqrt{1 + \evs_t^2},
    \end{equation}
    with the small-$\evs_t$ expansion
    \[
        \norm{\mE_t}_2 = 1 + \frac{\evs_t^2}{2\delta_t} + \gO(\evs_t^4) = 1 + \frac{\norm{\vs_t}_2^2}{2\evmu_t(2 - \evmu_t)} + \gO(\norm{\vs_t}_2^4).
    \]
    In particular, $\norm{\mE_t}_2 = 1$ if and only if $\vs_t = \vzero$ (uniform erase $\vb_t = \evmu_t\vone$ on the support of $\vk_t$), and $\norm{\mE_t}_2 \ge \sqrt{1 + \norm{\vs_t}_2^2} > 1$ whenever $\vs_t \neq \vzero.$
\end{lemma}

\begin{proof}
    For any vector $\vz \in \mathrm{span}(\vk_t, \vu_t)^\perp,$ we have $\inner{\vk_t}{\vz} = 0$ and $\inner{\hat\vk_t}{\vz} = 0,$ so $\vz\mE_t = \vz$ and $\mE_t\vz\tr = \vz\tr.$
    Thus $\mE_t$ acts as the identity on the $(d_k - 2)$-dimensional orthogonal complement of $\mathrm{span}(\vk_t, \vu_t),$ with all $(d_k - 2)$ corresponding singular values equal to $1.$
    On the two-dimensional invariant subspace $\mathrm{span}(\vk_t, \vu_t)$ with orthonormal basis $(\vk_t, \vu_t),$ we have
    \[
        \vk_t\mE_t = \vk_t - \underbrace{\inner{\vk_t}{\vk_t}}_{=1}\big(\evmu_t\vk_t + \evs_t\vu_t\big) = (1 - \evmu_t)\vk_t - \evs_t\vu_t,
        \qquad
        \vu_t\mE_t = \vu_t - \underbrace{\inner{\vu_t}{\vk_t}}_{=0}\hat\vk_t = \vu_t.
    \]
    Hence the restriction of $\mE_t$ to $\mathrm{span}(\vk_t, \vu_t)$ has the $2 \times 2$ matrix representation and Gram matrix
    \[
        \mE_{t\mid 2} = \begin{bmatrix} 1 - \evmu_t & -\evs_t \\ 0 & 1 \end{bmatrix},
        \qquad
        \mG_t \defas \mE_{t\mid 2}\mE_{t\mid 2}\tr = \begin{bmatrix} (1 - \evmu_t)^2 + \evs_t^2 & -\evs_t \\ -\evs_t & 1 \end{bmatrix}.
    \]
    Since $(1 - \evmu_t)^2 = 1 - \evmu_t(2 - \evmu_t) = 1 - \delta_t,$ the trace and determinant of $\mG_t$ are
    \[
        \Tr\mG_t = 2 + \evs_t^2 - \delta_t,
        \qquad
        \det\mG_t = (1 - \evmu_t)^2 = 1 - \delta_t.
    \]
    Solving the characteristic polynomial $\nu^2 - (\Tr\mG_t)\nu + \det\mG_t = 0$ for the larger eigenvalue $\nu_{\max}(\mG_t)$ and taking its square root $\norm{\mE_t}_2 = \sqrt{\nu_{\max}(\mG_t)}$ yields the exact equality in \eqref{eq:lem1_norm}.
    %
    Because $\delta_t \in (0, 1],$ we have $\sqrt{(\evs_t^2 - \delta_t)^2 + 4\evs_t^2} = \sqrt{(\evs_t^2 + \delta_t)^2 + 4(1 - \delta_t)\evs_t^2} \ge \evs_t^2 + \delta_t,$ so $\nu_{\max}(\mG_t) \ge 1 + \evs_t^2,$ which establishes the lower bound $\norm{\mE_t}_2 \ge \sqrt{1 + \evs_t^2}.$
    Finally, expanding the inner square root as $\evs_t \to 0$ gives
    \[
        \sqrt{(\delta_t - \evs_t^2)^2 + 4\evs_t^2}
        = \sqrt{\delta_t^2 + 2(2 - \delta_t)\evs_t^2 + \evs_t^4}
        = \delta_t - \evs_t^2 + \frac{2\evs_t^2}{\delta_t} + \gO(\evs_t^4),
    \]
    which implies $\nu_{\max}(\mG_t) = 1 + \evs_t^2/\delta_t + \gO(\evs_t^4)$ and $\norm{\mE_t}_2 = 1 + \evs_t^2/(2\delta_t) + \gO(\evs_t^4).$
\end{proof}


\subsection{Stationary Row-Energy Law and Fourth-Moment Expansion}
\label{app:theory:prop2}

\begin{proof}[Proof of \Cref{prop2}]
    From \eqref{eq:gdn2}, the $j$-th row of $\mS_t = \hat\mS_t - \vk_t\tr(\hat\vk_t\hat\mS_t) + \vk_t\tr\hat\vv_t$ decomposes as $\mS_{t,j,:}=\va_j-\vr_j+\vw_j,$ where $\va_j\defas\evalpha_{t,j}\mS_{t-1,j,:}$ is the decayed prior row, $\vw_j\defas\evk_{t,j}\hat\vv_t$ is the new write row, and $\vr_j\defas\evk_{t,j}\sum_{i=1}^{d_k}\evb_{t,i}\evk_{t,i}\evalpha_{t,i}\mS_{t-1,i,:}$ is the key-erasure row.
    (Note that if $\mS_0 = \mZero$ and key coordinates have zero odd cross-moments $\E[\evk_{t,j}\evk_{t,j'}\evk_{t,i}^2] = 0$ for $j \neq j',$ taking the inner product $\E[\inner{\mS_{t,j,:}}{\mS_{t,j',:}}]$ preserves row orthogonality $\E[\inner{\mS_{t,j,:}}{\mS_{t,j',:}}] = \delta_{jj'}\evm_{t,j}$ inductively for all $t \ge 0.$)
    Because $\E[\hat\vv_t]=\vzero$ and $\hat\vv_t$ is independent of $(\vk_t,\valpha_t,\vb_t,\mS_{t-1}),$ we have $\E[\inner{\va_j-\vr_j}{\vw_j}]=0,$ and expanding $\norm{\mS_{t,j,:}}_2^2$ in expectation yields
    \[
        \evm_{t,j} = \E[\norm{\va_j}_2^2] - 2\E[\inner{\va_j}{\vr_j}] + \E[\norm{\vr_j}_2^2] + \E[\norm{\vw_j}_2^2].
    \]
    Define the exact fourth-order key moments $\kappa_{ji} \defas \E[\evk_{t,j}^2\evk_{t,i}^2]$ for $i, j \in \{1, \dotsc, d_k\}.$
    Because $\norm{\vk_t}_2^2 = \sum_{i=1}^{d_k}\evk_{t,i}^2 = 1$ identically at every timestep $t,$ summing $\kappa_{ji}$ over $i$ gives the exact marginal identity
    \[
        \sum_{i=1}^{d_k}\kappa_{ji} = \E\!\left[\evk_{t,j}^2\sum_{i=1}^{d_k}\evk_{t,i}^2\right] = \E[\evk_{t,j}^2] = \evc_j,
        \qquad
        \text{so that}\quad
        0 \le \kappa_{jj} \le \evc_j
        \quad\text{and}\quad
        \sum_{i \ne j}\kappa_{ji} = \evc_j - \kappa_{jj}.
    \]
    Conditioning on $\mS_{t-1}$ and applying mutual independence of $(\vk_t,\hat\vv_t,\valpha_t,\vb_t)$ together with $\E[\evk_{t,i}\evk_{t,j}]=\delta_{ij}\evc_j$ and $\E[\inner{\mS_{t-1,i,:}}{\mS_{t-1,i',:}}]=\delta_{ii'}\evm_{t-1,i}$ (which eliminates all off-diagonal cross-row inner products $i\neq j$ in $\E[\inner{\va_j}{\vr_j}]$ and $i\neq i'$ in $\E[\norm{\vr_j}_2^2]$), the four terms evaluate exactly to
    \begin{align*}
        \E[\norm{\va_j}_2^2] &= \E[\evalpha_{t,j}^2]\,\E[\norm{\mS_{t-1,j,:}}_2^2] = \evlambda_j\evm_{t-1,j},\qquad \E[\norm{\vw_j}_2^2] = \E[\evk_{t,j}^2]\,\E[\norm{\hat\vv_t}_2^2] = \evc_j\evsigma_v^2,\\
        -2\E[\inner{\va_j}{\vr_j}] &= -2\sum_{i=1}^{d_k}\E[\evalpha_{t,j}\evalpha_{t,i}\evb_{t,i}\evk_{t,j}\evk_{t,i}]\,\E[\inner{\mS_{t-1,j,:}}{\mS_{t-1,i,:}}] = -2\evlambda_j\bar\evb_j\evc_j\evm_{t-1,j}, \\
        \E[\norm{\vr_j}_2^2] &= \sum_{i,i'=1}^{d_k}\E[\evk_{t,j}^2\evk_{t,i}\evk_{t,i'}\evb_{t,i}\evb_{t,i'}\evalpha_{t,i}\evalpha_{t,i'}]\,\E[\inner{\mS_{t-1,i,:}}{\mS_{t-1,i',:}}] = \sum_{i=1}^{d_k}\E[\evb_{t,i}^2]\kappa_{ji}\evlambda_i\evm_{t-1,i}.
    \end{align*}
    Summing these four expressions yields the exact fourth-moment recurrence
    \[
        \evm_{t,j}
        = \evlambda_j\big(1 - 2\bar\evb_j\evc_j + \E[\evb_{t,j}^2]\kappa_{jj}\big)\evm_{t-1,j}
        + \sum_{i \ne j}\E[\evb_{t,i}^2]\kappa_{ji}\evlambda_i\evm_{t-1,i}
        + \evc_j\evsigma_v^2.
    \]
    Applying the second-order closures $\kappa_{ji} \approx \evc_j\evc_i$ and $\E[\evb_{t,i}^2] \approx \bar\evb_i^2$ (while noting that on the diagonal $i=j,$ $\E[\evb_{t,j}^2]\kappa_{jj} = \gO(\evc_j^2)$ is a second-order correction to $2\bar\evb_j\evc_j$) gives \eqref{prop2:eq1}.
    %
    Dropping the cross-row coupling term reduces \eqref{prop2:eq1} to the scalar recurrence $\evm_{t,j}=\evlambda_j(1-2\bar\evb_j\evc_j)\evm_{t-1,j}+\evc_j\evsigma_v^2.$
    Since $\evlambda_j\in(0,1)$ and $\bar\evb_j\evc_j\in(0,1)$ imply $|\evlambda_j(1-2\bar\evb_j\evc_j)|<1,$ this recurrence is a strict contraction and converges as $t\to\infty$ to the unique fixed point $\evm_{t,j}=\evm_{t-1,j}=\evm_j^\ast$ satisfying $(1-\evlambda_j+2\evlambda_j\bar\evb_j\evc_j)\evm_j^\ast=\evc_j\evsigma_v^2,$ which gives \eqref{prop2:eq2}.
    Moreover, in \eqref{prop2:eq1}, the cross-row sum $\Sigma_{\mathrm{cross}} \defas \sum_{i=1}^{d_k}\bar\evb_i^2\evc_i\evlambda_i\evm_i^\ast$ is independent of $j$ and enters through $\evc_j(\evsigma_v^2 + \Sigma_{\mathrm{cross}}),$ so retaining it at stationarity simply rescales $\evsigma_v^2 \to \tilde\evsigma_v^2 \defas \evsigma_v^2 / \big(1 - \sum_{i=1}^{d_k}\bar\evb_i^2\evc_i^2\evlambda_i\,\tau_{\mathrm{eff},i}(\bar\evb_i)\big) > 0$ (where the denominator is strictly positive since $\bar\evb_i^2\evc_i^2\evlambda_i\,\tau_{\mathrm{eff},i}(\bar\evb_i) < \frac{1}{2}\bar\evb_i\evc_i$ and $\frac{1}{2}\sum_{i=1}^{d_k}\bar\evb_i\evc_i < 1$) while preserving the exact relative channel profile $\evm_j^\ast \propto \evc_j\,\tau_{\mathrm{eff},j}(\bar\evb_j)$ and the matched-filter direction in \Cref{prop3}.
\end{proof}


\subsection{Lag-Dependent Matched-Filter Readout}
\label{app:theory:prop3}

\begin{proof}[Proof of \Cref{prop3}]
    (i) Substituting $\mS_{t,j,:}=\eva_{\ell,j}\evk_{t-\ell,j}\hat\vv_{t-\ell}+\vxi_{t,j}$ and $\evq_{t,j}=\evk_{t-\ell,j}$ into $\vo_t(\vgamma)$ gives
    \[
        \vo_t(\vgamma)-\hat\vv_{t-\ell} = \left(\sum_{j=1}^{d_k}\evgamma_j\evk_{t-\ell,j}^2\eva_{\ell,j}-1\right)\hat\vv_{t-\ell} + \sum_{j=1}^{d_k}\evgamma_j\evk_{t-\ell,j}\vxi_{t,j}.
    \]
    Because $\hat\vv_{t-\ell}$ is zero-mean with $\E[\norm{\hat\vv_{t-\ell}}_2^2]=\evsigma_v^2,$ independent of $(\vk_{t-\ell},\eva_{\ell,:}),$ and uncorrelated with $\vxi_{t,:},$ the cross term between the signal residual and the interference sum vanishes in expectation.
    Applying $\E[\evk_{t-\ell,i}^2\evk_{t-\ell,j}^2]\approx\evc_i\evc_j$ and $\E[\eva_{\ell,i}\eva_{\ell,j}]\approx\bar\eva_{\ell,i}\bar\eva_{\ell,j}$ to the squared signal term and $\E[\inner{\vxi_{t,i}}{\vxi_{t,j}}\mid\vk_{t-\ell}]=\delta_{ij}\evm_j^\ast$ to the squared interference term yields \eqref{prop3:eq1}.

    (ii) Under the constraint $\sum_{j=1}^{d_k}\evgamma_j\evc_j\bar\eva_{\ell,j}=1,$ let $\evu_j\defas\evgamma_j\sqrt{\smash{\evc_j\evm_j^\ast}\vphantom{\evm}}$ and $\evv_j\defas\sqrt{\smash{\evc_j/\evm_j^\ast}\vphantom{\evm}}\,\bar\eva_{\ell,j}.$
    By Cauchy-Schwarz,
    \[
        1 = \left(\sum_{j=1}^{d_k}\evu_j\evv_j\right)^{\!2}
        \le \left(\sum_{j=1}^{d_k}\evu_j^2\right)\!\left(\sum_{j=1}^{d_k}\evv_j^2\right)
        = \left(\sum_{j=1}^{d_k}\evgamma_j^2\evc_j\evm_j^\ast\right)\!\left(\sum_{j=1}^{d_k}\frac{\evc_j\bar\eva_{\ell,j}^2}{\evm_j^\ast}\right),
    \]
    with equality if and only if $\evu_j\propto\evv_j,$ i.e., $\evgamma_j^\ast(\ell)\propto\bar\eva_{\ell,j}/\evm_j^\ast.$
    Setting $\partial\gL_\ell(\vgamma)/\partial\evgamma_j=0$ in \eqref{prop3:eq1} gives $\evgamma_j^\ast(\ell)=\kappa_\ell\,\bar\eva_{\ell,j}/\evm_j^\ast$ with $\kappa_\ell=\evsigma_v^2\big(1+\evsigma_v^2\sum_{i=1}^{d_k}\evc_i\bar\eva_{\ell,i}^2/\evm_i^\ast\big)\inv,$ which establishes \eqref{prop3:eq2}.
\end{proof}


\subsection{Geometric Horizon Survival Profile and Fractional Power Law}
\label{app:theory:prop3:cor1}

\begin{proof}[Proof of \Cref{prop3:cor1}]
    Under the geometric lag distribution $\mathbb{P}(\ell = k) = (1 - \evtheta)\evtheta^{k-1}$ ($k \in \{1, 2, \dotsc\}$) of mean horizon $H = \E[\ell] = (1 - \evtheta)\inv > 1,$ we have $\evtheta = 1 - H\inv \in (0, 1).$
    Because $\evlambda_j \in (0, 1)$ and $\evtheta \in (0, 1),$ the product $\evtheta\sqrt{\evlambda_j} < 1,$ so the geometric series for $g_H(\evlambda_j) \defas \E_\ell[\evlambda_j^{\ell\sdiv 2}]$ converges absolutely:
    \[
        g_H(\evlambda_j)
        = \sum_{k=1}^\infty (1 - \evtheta)\evtheta^{k-1}\evlambda_j^{k\sdiv 2}
        = (1 - \evtheta)\sqrt{\evlambda_j}\sum_{m=0}^\infty \big(\evtheta\sqrt{\evlambda_j}\big)^m
        = \frac{(1 - \evtheta)\sqrt{\evlambda_j}}{1 - \evtheta\sqrt{\evlambda_j}}.
    \]
    Substituting $1 - \evtheta = H\inv$ and $\evtheta = (H - 1)H\inv$ and multiplying numerator and denominator by $H$ gives
    \[
        g_H(\evlambda_j)
        = \frac{\sqrt{\evlambda_j}}{H - (H - 1)\sqrt{\evlambda_j}}
        = \frac{\sqrt{\evlambda_j}}{1 + (H - 1)\big(1 - \sqrt{\evlambda_j}\big)}.
    \]
    Applying Proposition~\ref{prop3:item2} under the horizon-averaged constraint $\sum_{j=1}^{d_k}\evgamma_j\evc_j g_H(\evlambda_j)=1$ yields $\evgamma_j^{\mathrm{stat}}\propto g_H(\evlambda_j)/\evm_j^\ast = g_H(\evlambda_j)/\evrho_j^2.$
    In the decay-dominated regime $1-\evlambda_j=\tau_{\mathrm{eff},j}\inv\ll 1,$ substituting the first-order expansion $1-\sqrt{\evlambda_j}\approx(2\tau_{\mathrm{eff},j})\inv$ into $g_H(\evlambda_j)$ and differentiating $\log\evgamma_j^{\mathrm{stat}}=\log g_H(\evlambda_j)-\log\tau_{\mathrm{eff},j}+\mathrm{const}$ with respect to $\log\evrho_j$ ($\mathrm{d}\log\tau_{\mathrm{eff},j}/\mathrm{d}\log\evrho_j=2$) gives
    \[
        g_H(\evlambda_j) \approx \frac{2\tau_{\mathrm{eff},j}}{2\tau_{\mathrm{eff},j}+H-1},\qquad
        -\frac{\mathrm{d}\log\evgamma_j^{\mathrm{stat}}}{\mathrm{d}\log\evrho_j}
        = 2\!\left(1-\frac{\mathrm{d}\log g_H(\evlambda_j)}{\mathrm{d}\log\tau_{\mathrm{eff},j}}\right)
        \approx \frac{4\tau_{\mathrm{eff},j}}{2\tau_{\mathrm{eff},j}+H-1},
    \]
    which lies strictly inside $(0,1)$ whenever $\tau_{\mathrm{eff},j}<(H-1)/2,$ completing the proof.
\end{proof}

To extend the log-slope derivation of \Cref{prop3:cor1} beyond the pure decay-dominated limit by retaining the key-erasure floor $2\evlambda_j\bar\evb\evc > 0$ at fixed channel key energy $\evc_j = \evc$ and erase rate $\bar\evb_j = \bar\evb,$ write $\epsilon_j \defas 1 - \evlambda_j \in (0, 1)$ and retain the per-step target-erasure factor $(1 - \bar\evb\evc)$ in $\bar\eva_{\ell,j} \approx \bar\eva_{1,j}^\ell$ with $\bar\eva_{1,j} \defas \sqrt{\evlambda_j}(1 - \bar\evb\evc)$ (\Cref{prop3}).
%
For $\epsilon_j \ll 1$ and $\bar\evb\evc \ll 1,$ we have $1 - \bar\eva_{1,j} \approx (\epsilon_j + 2\bar\evb\evc)/2 \approx (2\tau_{\mathrm{eff},j})\inv$ from $\tau_{\mathrm{eff},j} = (\epsilon_j + 2\evlambda_j\bar\evb\evc)\inv \approx (\epsilon_j + 2\bar\evb\evc)\inv$ in \eqref{prop2:eq2}, so the horizon-averaged signal survival $\tilde{g}_H(\bar\eva_{1,j}) \defas \E_\ell[\bar\eva_{1,j}^\ell] = \bar\eva_{1,j} / \big(1 + (H - 1)(1 - \bar\eva_{1,j})\big)$ takes the same functional form in $\tau_{\mathrm{eff},j}$:
\[
    \tau_{\mathrm{eff},j} \approx (\epsilon_j + 2\bar\evb\evc)\inv,
    \qquad
    \tilde{g}_H(\bar\eva_{1,j}) \approx \frac{2\tau_{\mathrm{eff},j}}{2\tau_{\mathrm{eff},j} + H - 1}.
\]
%
Because $\evm_j^\ast = \evrho_j^2 = \evc\evsigma_v^2\,\tau_{\mathrm{eff},j}$ at fixed $\evc,$ we have $\mathrm{d}\log\tau_{\mathrm{eff},j}/\mathrm{d}\log\evrho_j = 2,$ and differentiating $\log\evgamma_j^{\mathrm{stat}} = \log \tilde{g}_H(\bar\eva_{1,j}) - \log\tau_{\mathrm{eff},j} + \mathrm{const}$ with respect to $\log\evrho_j$ via the chain rule yields the exact same local power-law exponent for all $\bar\evb\evc > 0$:
\[
    p(\tau_{\mathrm{eff},j}) \defas -\frac{\mathrm{d}\log\evgamma_j^{\mathrm{stat}}}{\mathrm{d}\log\evrho_j}
    = 2\left(1 - \frac{\mathrm{d}\log \tilde{g}_H(\bar\eva_{1,j})}{\mathrm{d}\log\tau_{\mathrm{eff},j}}\right)
    \approx \frac{4\tau_{\mathrm{eff},j}}{2\tau_{\mathrm{eff},j} + H - 1} \in (0, 2),
\]
%
which lies strictly in $(0, 1)$ across all channels whose effective memory horizon satisfies $\tau_{\mathrm{eff},j} < (H - 1)/2.$


\subsection{Erase-Readout Substitution}
\label{app:theory:prop4}

\begin{proof}[Proof of \Cref{prop4}]
    For each channel $j,$ differentiating $\tau_{\mathrm{eff},j}(\bar\evb)=(1-\evlambda_j+2\evlambda_j\bar\evb\evc_j)\inv$ with respect to $\bar\evb$ yields
    \[
        \tau_{\mathrm{eff},j}'(\bar\evb) = -\frac{2\evlambda_j\evc_j}{(1-\evlambda_j+2\evlambda_j\bar\evb\evc_j)^2} < 0,\qquad
        \tau_{\mathrm{eff},j}''(\bar\evb) = \frac{8\evlambda_j^2\evc_j^2}{(1-\evlambda_j+2\evlambda_j\bar\evb\evc_j)^3} > 0,
    \]
    so $\bar\tau(\bar\evb)=S(\vone)^{-2}\sum_{j=1}^{d_k}\evc_j^2\tau_{\mathrm{eff},j}(\bar\evb)$ is strictly decreasing ($\bar\tau'(\bar\evb)<0$) and strictly convex ($\bar\tau''(\bar\evb)>0$) on $(0,2).$
    By \Cref{prop3:cor1}, among all $\vgamma$ satisfying $S(\vgamma)=S(\vone),$ the static matched filter $\evgamma_j^\ast\propto g_H(\evlambda_j)/\evm_j^\ast(\bar\evb)$ strictly minimizes the numerator of $\gI(\bar\evb,\vgamma)$ whenever $\evm_j^\ast(\bar\evb)/g_H(\evlambda_j)$ varies across $j,$ which gives $\Phi^\ast=\gI(\bar\evb,\vgamma^\ast)/\bar\tau(\bar\evb)<1.$
    At an interior minimizer $\bar\evb^\ast(\Phi),$ the first-order optimality condition is $\Phi\,\bar\tau'(\bar\evb^\ast(\Phi))+B'(\bar\evb^\ast(\Phi))=0.$
    Implicitly differentiating with respect to $\Phi>0$ gives
    \[
        \frac{\mathrm{d}\bar\evb^\ast(\Phi)}{\mathrm{d}\Phi} = -\frac{\bar\tau'(\bar\evb^\ast(\Phi))}{\Phi\,\bar\tau''(\bar\evb^\ast(\Phi))+B''(\bar\evb^\ast(\Phi))} > 0,
    \]
    since $\bar\tau'(\bar\evb^\ast)<0,$ $\bar\tau''(\bar\evb^\ast)>0,$ and $B''(\bar\evb^\ast)\ge 0.$
    Because $\bar\evb^\ast(\Phi)$ is strictly increasing in $\Phi$ and $\Phi^\ast<1,$ we obtain $\bar\evb^\ast(\Phi^\ast)<\bar\evb^\ast(1).$
    Since $\tau_{\mathrm{eff},j}(\bar\evb)$ is strictly decreasing in $\bar\evb$ for every $j,$ the inequality $\tau_{\mathrm{eff},j}(\bar\evb^\ast(\Phi^\ast))>\tau_{\mathrm{eff},j}(\bar\evb^\ast(1))$ follows immediately, which completes the proof.
\end{proof}

%% file: apps/e_toy.tex
\clearpage
\section{Synthetic Memory Experiment Details}
\label{app:toy}

This appendix provides the complete experimental protocols and numerical results for the two controlled memory tasks in \S\ref{sec:theory:exp}.


\subsection{Single-Head Stationary Retrieval}
\label{app:toy:task1}

We simulate a single recurrent head with $d_k = d_v = 64$ under the stationary stream conditions of \S\ref{sec:theory:theory}.
To model the anisotropic, heavy-tailed channel key energy after the $\silu$ nonlinearity in trained language models, we set power-law weights $\tilde\evc_j \propto (65 - j)^{-0.25}$ for $j \in \{1, \dotsc, 64\}$ with $\sum_{j=1}^{64}\tilde\evc_j = 1.$
At each timestep $t,$ we draw $\evg_{t,j} \sim \gN(0, 1)$ i.i.d., form post-$\silu$-like coordinates $\evu_{t,j} = \sqrt{\tilde\evc_j}\,(\sign(\evg_{t,j})\abs{\evg_{t,j}}^{1.6} + 0.22),$ and normalize $\vk_t = \vu_t / \norm{\vu_t}_2 \in \bS{63},$ yielding mean key participation ratio $\E[\eta_{\mathrm{eff}}(\vk_t)] = \E[1/\norm{\vk_t}_4^4] = 12.24$ and channel second moments $\evc_j = \E[\evk_{t,j}^2].$
Values are drawn i.i.d.\ as $\vv_t \sim \gN(\vzero, d_v\inv\I_{d_v})$ so that $\evsigma_v^2 = \E[\norm{\vv_t}_2^2] = 1.$
The decay profile $\valpha \in (0, 1)^{1\times 64}$ uses retention timescales $\tau_j$ log-spaced over $[1.2, 600]$ ($\evalpha_j = \exp(-1/\tau_j) \in [0.435, 0.9983],$ with $\evlambda_j = \evalpha_j^2$), and the state evolves as $\mS_t = (\I_{64} - \evmu\vk_t\tr\vk_t)\Diag(\valpha)\mS_{t-1} + \vk_t\tr\vv_t$ over $3{,}000$ burn-in and $30{,}000$ evaluation steps across $20$ random seeds.

At each evaluation step $t,$ we sample a retrieval lag $\ell_t \sim p(\ell),$ query the state with $\vq_t = \vk_{t-\ell_t}$ and readout gain $\vgamma \in \Rvec{64}$ to form $\vo_t(\vgamma) = (\vgamma \odot \vk_{t-\ell_t})\mS_t,$ and measure normalized retrieval MSE $\E[\norm{\vo_t(\vgamma) - \vv_{t-\ell_t}}_2^2]/\evsigma_v^2.$
For the geometric-horizon sweep across $\evmu \in \{0.1, 0.2, \dotsc, 1.9\}$ and $H \in \{10, 27, 81\},$ lags follow $p(\ell) \propto (1 - q)^{\ell - 6}$ on $\ell \in \{6, \dotsc, 1{,}024\}$ with $q = 1/(H - 5)$ (mean $H$); for the bimodal mixture at $\evmu = \evmu_{\mathrm{unit}}^\ast = 0.60,$ we draw $\ell_t \in \{2, 80\}$ with equal probability.
Because each analytical profile $\tilde\vgamma \in \Rvec{64}$ specifies only the relative weighting across channels up to the global shrinkage scalar $\kappa_\ell$ in \eqref{prop3:eq2}, we multiply $\tilde\vgamma$ by $\kappa^\ast = \argmin_{\kappa \in \R{}}\E[\norm{\kappa\vo_t(\tilde\vgamma) - \vv_{t-\ell_t}}_2^2]$ fitted on the even-$t$ evaluation steps ($15{,}000$ steps) and evaluated without refitting on the disjoint odd-$t$ split ($15{,}000$ steps); the unconstrained $64$-parameter least-squares static oracle $\vgamma^{\mathrm{LS}}$ and lag-conditioned oracle $\vgamma^{\mathrm{LS}}(\ell)$ are fitted and evaluated on the same disjoint split.
\Cref{tab:toy_task1} reports the complete stationary row-energy verification, geometric-horizon retrieval sweep across $H \in \{10, 27, 81\},$ and bimodal two-lag mixture ($\ell_t \in \{2, 80\}$) results.

\input{tables/tab6_toy_task1}


\subsection{Learned Multi-Query Associative Recall}
\label{app:toy:task2}

We train a single-layer, two-head gated delta model ($N_h = 2,$ $d_k = d_v = 32,$ $N_h d_k = 64,$ residual width $d = 128$) end-to-end on streaming multi-query associative recall~\citep{Arora2023-tw} with overwriting writes.
Each sequence consists of $T = 256$ write/read events (each encoded as a single composite input vector) over a vocabulary of $K = 256$ keys and $V = 64$ values.
For each sequence, we sample $K_{\mathrm{live}} = 96$ active keys ($1.5\times$ the total state rank $N_h d_k = 64$) and assign them $\zipf(1.0)$ occurrence probabilities.
Each timestep $t \in \{1, \dotsc, 256\}$ is independently chosen as a write event with probability $0.7$ (binding a sampled active key to a uniformly drawn value in $\{1, \dotsc, V\}$ and overwriting any previous value stored for that key) or a read event with probability $0.3$ (querying a previously written active key at event lag $\ell_t = t - t_{\mathrm{last}} \in [1, 256)$ since its most recent write).

To require subspace-selective value gating amid background residual-stream activity, we partition the $V = 64$ values evenly across four orthogonal $d_{\mathrm{sub}}$-dimensional subspaces $\{\mQ_m\}_{m=1}^4$ ($d_{\mathrm{sub}} = 16,$ $\mQ_m \in \R{16\times 128},$ $\mQ_m\mQ_{m'}\tr = \delta_{m,m'}\I_{16},$ spanning $64$ dimensions of $\Rvec{128}$).
On write steps, value $v \in \{1, \dotsc, V\}$ activates its assigned subspace $m(v) \in \{1, \dotsc, 4\}$ with $\ve_v^{\mathrm{sub}}\mQ_{m\suppar{v}} + \ve_{m\suppar{v}}^{\mathrm{cat}},$ while inactive subspaces $m \neq m(v)$ carry scaled Gaussian noise $\evsigma_{\mathrm{bg}}\sqrt{d/d_{\mathrm{sub}}}\,\vz_{t,m}\mQ_m$ ($\vz_{t,m} \sim \gN(\vzero, \I_{16}),$ $\evsigma_{\mathrm{bg}} = 0.3$); on read steps, all four subspaces carry only this background noise.
Adding the key and write/read role embeddings forms the pre-RMS input ($\mE_{\mathrm{key}} \in \R{256\times 128},$ $\mE_{\mathrm{role}} \in \R{2\times 128},$ and $\mE_{\mathrm{cat}} \in \R{4\times 128}$ initialized from $\gN(0, 1),$ and $\mE_{\mathrm{val}}^{\mathrm{sub}} \in \R{64\times 16}$ from $\gN(0, d/d_{\mathrm{sub}})$ so $\E[\norm{\ve_v^{\mathrm{sub}}\mQ_{m(v)}}_2^2] = d$), while queries and keys apply $\silu$ before $\ell_2$ normalization.
All linear projections ($\mW_q, \mW_k, \mW_v \in \R{128\times 64},$ $\mW_{\mathrm{out}} \in \R{64\times 64},$ dense $\mW_b, \mW_w,$ and bottleneck factors $\mW_{m\downarrow}, \mW_{m\uparrow}$) use fan-in initialization $\gN(0, d_{\mathrm{in}}\inv)$ with zero gate biases ($\vc_b = \vc_w = \vzero$), except the rank-$8$ decay gate ($\mW_{f\uparrow} \sim \gN(0, 0.01 r_f\inv),$ $\exp(\va_{\log}) \sim \mathrm{Unif}(1, 16),$ $\softplus(\vc_f) \sim \mathrm{LogUnif}(10^{-3}, 10^{-1})$) and readout gain ($r_q = 8,$ $\mW_{q\uparrow} = \mZero,$ $\vc_q = \vzero,$ so $\vgamma_t = \vone$ initially).
All models are trained for $1{,}000$ steps at batch size $32$ using Adam (peak $\text{LR} = 4\times 10^{-3}$ with $100$-step linear warmup and cosine decay to $8\times 10^{-5},$ global gradient-norm clip $1.0$) across $20$ random seeds and evaluated on $256$ held-out sequences ($\approx 19{,}700$ read queries per seed) partitioned into short ($\ell_t \in [8, 32)$), medium ($\ell_t \in [32, 128)$), and long ($\ell_t \in [128, 256)$) event-lag buckets.

\Cref{tab:toy_task2} reports all $16$ gate-rank and readout-gain configurations, alongside the token-dependent erase-logit variance fraction on write steps (\Cref{prop1:rem2}) and mean terminal row energy.

\input{tables/tab7_toy_task2}

%% file: tables/tab6_toy_task1.tex
\begin{table}[!t]
    \centering
    \caption{%
        \textbf{Single-head stationary row-energy verification, geometric-horizon retrieval sweep, and bimodal two-lag readout comparison.}
        All results are reported as $\text{mean} \pm \text{SD}$ across $20$ random seeds for a single recurrent head with $d_k = d_v = 64.$
        (Top) Verification of the closed-form row-energy law at $\evmu = 0.60.$
        (Middle) Optimal erase strength $\evmu^\ast$ and normalized retrieval MSE under geometric lag distributions of mean $H \in \{10, 27, 81\}.$
        (Bottom) Retrieval MSE on short ($\ell = 2$), long ($\ell = 80$), and mixed ($50/50$) queries at $\evmu = 0.60,$ along with the Spearman correlation $r_s(\vgamma, \valpha)$ between readout gain $\vgamma$ and the channel decay factor $\valpha.$\looseness=-1%
    }
    \label{tab:toy_task1}
    \resizebox{\textwidth}{!}{%
        \setlength{\tabcolsep}{2.4pt}%
        \renewcommand{\arraystretch}{1.08}%
        \begin{tabular}{@{}c@{\hspace{3pt}}|l|ccc|ccc@{}}
        \toprule
            \multicolumn{1}{@{}c@{\hspace{5pt}}}{} & {Row-Energy Law} & {Key Rank} & {Max / Min} & {RMS Spread} & {Energy} & {Spearman} & {Log-Pearson} \\
            \multicolumn{1}{@{}c@{\hspace{5pt}}}{} & {Diagnostic ($\evmu = 0.60$)} & {$\E[\eta_{\mathrm{eff}}(\vk_t)]$} & {$\max_j \evm_j / \min_j \evm_j$} & {$\mathrm{p90}/\mathrm{p10}(\evrho_j)$} & {$\mathrm{p90}/\mathrm{p10}$} & {$r_s(\evm_j, \evm_j^\ast)$} & {$r_{\log}$} \\
            \midrule
            & Stationary row energy $\evm_j$ & 12.24 & 52.4$\times$ & 5.69$\times$ & 32.3$\times$ & 1.0000 & 1.0000 \\
        \midrule
        \midrule
            \multicolumn{1}{@{}c@{\hspace{5pt}}}{} & & \multicolumn{3}{c|}{{Normalized Retrieval MSE $\downarrow$}} & \multicolumn{3}{c@{}}{{Optimal Erase $\evmu^\ast$}} \\
            \cmidrule(lr){3-5} \cmidrule(l){6-8}
            \multicolumn{1}{@{}c@{\hspace{5pt}}}{} & {Geometric Horizon Readout $\vgamma$} & {Horizon $H\!=\!10$} & {Horizon $H\!=\!27$} & {Horizon $H\!=\!81$} & {$H\!=\!10$} & {$H\!=\!27$} & {$H\!=\!81$} \\
        \midrule
            \multirowrotate{5}{90}{\footnotesize Geometric}
            & Unit gain $\evgamma_j = 1$ (\gdntwo) & 0.3599$\,\pm\,$0.0019 & 0.6389$\,\pm\,$0.0017 & 0.8607$\,\pm\,$0.0017 & 0.90 & 0.60 & 0.30 \\
            & Full whitening $\evgamma_j \propto \evrho_j\inv$ ($p = 1$) & 0.4068$\,\pm\,$0.0025 & 0.6802$\,\pm\,$0.0021 & 0.8817$\,\pm\,$0.0016 & 0.80 & 0.50 & 0.20 \\
            & Half-power $\evgamma_j \propto \evrho_j^{-1\sdiv 2}$ ($p = 1/2$) & 0.3577$\,\pm\,$0.0020 & 0.6372$\,\pm\,$0.0018 & 0.8578$\,\pm\,$0.0017 & 0.80 & 0.50 & 0.20 \\
            & \Cref{prop3:cor1} $\evgamma_j^{\mathrm{stat}} \propto g_H(\evlambda_j)/\evm_j^\ast$ & \underline{0.3559$\,\pm\,$0.0021} & \underline{0.6205$\,\pm\,$0.0019} & \underline{0.8484$\,\pm\,$0.0017} & 0.80 & 0.50 & 0.20 \\
            & Static oracle $\vgamma^{\mathrm{LS}}$ ($64$ params) & \tbf{0.3398$\,\pm\,$0.0020} & \tbf{0.6192$\,\pm\,$0.0018} & \tbf{0.8483$\,\pm\,$0.0018} & 0.90 & 0.50 & 0.20 \\
        \midrule
        \midrule
            \multicolumn{1}{@{}c@{\hspace{5pt}}}{} & & \multicolumn{3}{c|}{{Normalized Retrieval MSE $\downarrow$}} & \multicolumn{3}{c@{}}{{Params \& $r_s(\vgamma, \valpha)$}} \\
            \cmidrule(lr){3-5} \cmidrule(l){6-8}
            \multicolumn{1}{@{}c@{\hspace{5pt}}}{} & {Bimodal Readout ($\ell \in \{2, 80\}$)} & {Short ($\ell\!=\!2$)} & {Long ($\ell\!=\!80$)} & {Mixed} & {Params} & {$\ell\!=\!2$} & {$\ell\!=\!80$} \\
        \midrule
            \multirowrotate{4}{90}{\footnotesize Bimodal}
            & Unit gain $\evgamma_j = 1$ (\gdntwo) & 0.1831$\,\pm\,$0.0005 & 1.0487$\,\pm\,$0.0011 & 0.6159$\,\pm\,$0.0007 & 1 scalar & -- & -- \\
            & Shared static oracle $\vgamma^{\mathrm{LS}}$ & \underline{0.1096$\,\pm\,$0.0006} & 1.0401$\,\pm\,$0.0010 & 0.5749$\,\pm\,$0.0006 & 64 static & $-$0.98 & $-$0.98 \\
            & \Cref{prop3} $\evgamma_j^\ast(\ell) \propto \evlambda_j^{\ell\sdiv 2}/\evm_j^\ast$ & 0.1617$\,\pm\,$0.0007 & \underline{0.9576$\,\pm\,$0.0004} & \underline{0.5597$\,\pm\,$0.0005} & 1 / lag & $-$0.97 & $+$0.98 \\
            & Lag-matched oracle $\vgamma^{\mathrm{LS}}(\ell)$ & \tbf{0.0879$\,\pm\,$0.0005} & \tbf{0.9561$\,\pm\,$0.0004} & \tbf{0.5220$\,\pm\,$0.0003} & 64 / lag & $-$0.98 & $+$0.83 \\
        \bottomrule
        \end{tabular}%
    }
\end{table}

%% file: tables/tab7_toy_task2.tex
\begin{table}[!t]
    \centering
    \caption{%
        \textbf{Learned multi-query associative recall accuracy, erase-logit token variance, and state row energy across gate-rank and readout-gain configurations.}
        All results are evaluated on a two-head recurrent model with $d_k = d_v = 32$ and residual width $d = 128$ across $20$ random seeds (mean $\pm$ SD; dense gates have full head-channel rank $N_h d_k = 64$).
        We report read accuracy overall and by event lag across single-gate rank sweeps and readout-gain variants with low-rank $(b, w)$ at $r_b = r_w = 4$ ($r_q = 8$ for dynamic $\vgamma_t$), along with the token-dependent fraction of projected erase-logit energy on write steps $\E[\norm{\tilde\vz^b_t - \bar\vz^b}_2^2]/\E[\norm{\tilde\vz^b_t}_2^2]$ ($\tilde\vz^b_t = \vz^b_t - \vc_b$) and mean terminal row energy $\bar\evm_T = (N_h d_k)\inv\E[\norm{\mS_T}_F^2].$%
    }
    \label{tab:toy_task2}
    \resizebox{\textwidth}{!}{%
        \setlength{\tabcolsep}{4pt}%
        \renewcommand{\arraystretch}{1.03}%
        \begin{tabular}{@{}c@{\hspace{3pt}}|l@{\hspace{2pt}}|cc@{\hspace{3pt}}c@{\hspace{1pt}}|cccc|@{\hspace{3pt}}c@{}c@{}}
        \toprule
            \multicolumn{1}{@{}c@{\hspace{5pt}}}{} & & \multicolumn{3}{c|}{{Gate Config}} & \multicolumn{4}{c|}{{Associative Recall Accuracy (\%) $\uparrow$}} & \multicolumn{2}{c@{}}{{Diagnostics}} \\
            \cmidrule(lr){3-5} \cmidrule(lr){6-9} \cmidrule(l){10-11}
            \multicolumn{1}{@{}c@{\hspace{5pt}}}{} & {Model} & {$r_b$} & {$r_w$} & {$\vgamma_t$} & {Overall} & {Short} & {Medium} & {Long} & {Erase} & {Row Ener.} \\
            \multicolumn{1}{@{}c@{\hspace{5pt}}}{} & & & & & {acc $\uparrow$} & {$[8,32)$} & {$[32,128)$} & {$[128,256)$} & {tok var\,$\downarrow$} & {$\bar\evm_T$\,$\downarrow$} \\
        \midrule
            & Dense $(b, w)$ (\gdntwo) & 64 & 64 & $\vone$ & 90.40$\,\pm\,$3.79 & 96.29$\,\pm\,$2.77 & 77.32$\,\pm\,$8.49 & 33.68$\,\pm\,$9.47 & 27.9\% & 1{,}593.0 \\
        \midrule
            \multirowrotate{6}{90}{\footnotesize Erase $r_b$}
            & Erase rank $r_b = 1$ & 1 & 64 & $\vone$ & 91.28$\,\pm\,$3.26 & 97.06$\,\pm\,$2.10 & 79.01$\,\pm\,$8.12 & 35.77$\,\pm\,$9.03 & 8.4\% & 1{,}451.0 \\
            & Erase rank $r_b = 2$ & 2 & 64 & $\vone$ & 91.33$\,\pm\,$2.91 & 97.04$\,\pm\,$1.68 & 79.30$\,\pm\,$7.46 & 35.76$\,\pm\,$9.45 & \tbf{3.0\%} & 1{,}436.1 \\
            & Erase rank $r_b = 4$ & 4 & 64 & $\vone$ & 91.51$\,\pm\,$3.37 & 97.03$\,\pm\,$2.36 & 80.05$\,\pm\,$7.91 & 36.91$\,\pm\,$9.07 & 4.0\% & 1{,}446.3 \\
            & Erase rank $r_b = 8$ & 8 & 64 & $\vone$ & 91.23$\,\pm\,$3.65 & 96.85$\,\pm\,$2.46 & 79.28$\,\pm\,$8.81 & 36.03$\,\pm\,$9.55 & 6.7\% & 1{,}500.9 \\
            & Erase rank $r_b = 16$ & 16 & 64 & $\vone$ & \tbf{91.97$\,\pm\,$2.72} & \tbf{97.36$\,\pm\,$1.62} & \tbf{80.98$\,\pm\,$6.84} & \tbf{38.08$\,\pm\,$8.57} & 8.9\% & 1{,}601.3 \\
            & Erase rank $r_b = 32$ & 32 & 64 & $\vone$ & \underline{91.71$\,\pm\,$2.69} & \underline{97.23$\,\pm\,$1.77} & \underline{80.39$\,\pm\,$6.43} & \underline{37.39$\,\pm\,$8.02} & 15.6\% & 1{,}561.9 \\
        \midrule
            \multirowrotate{6}{90}{\footnotesize Write $r_w$}
            & Write rank $r_w = 1$ & 64 & 1 & $\vone$ & 89.55$\,\pm\,$4.99 & 95.62$\,\pm\,$4.14 & 75.45$\,\pm\,$10.86 & 31.74$\,\pm\,$10.15 & 26.8\% & 865.7 \\
            & Write rank $r_w = 2$ & 64 & 2 & $\vone$ & 90.06$\,\pm\,$3.87 & 96.06$\,\pm\,$3.06 & 76.49$\,\pm\,$8.51 & 32.52$\,\pm\,$8.66 & 27.5\% & 1{,}053.2 \\
            & Write rank $r_w = 4$ & 64 & 4 & $\vone$ & 89.38$\,\pm\,$3.99 & 95.66$\,\pm\,$3.03 & 74.51$\,\pm\,$9.13 & 30.99$\,\pm\,$9.46 & 28.9\% & 1{,}102.0 \\
            & Write rank $r_w = 8$ & 64 & 8 & $\vone$ & 88.97$\,\pm\,$6.17 & 95.17$\,\pm\,$5.25 & 74.10$\,\pm\,$13.03 & 30.87$\,\pm\,$10.31 & 27.8\% & 1{,}142.6 \\
            & Write rank $r_w = 16$ & 64 & 16 & $\vone$ & 90.70$\,\pm\,$2.71 & 96.64$\,\pm\,$1.77 & 77.85$\,\pm\,$6.61 & 33.15$\,\pm\,$7.80 & 27.9\% & 1{,}447.7 \\
            & Write rank $r_w = 32$ & 64 & 32 & $\vone$ & 91.37$\,\pm\,$4.16 & 96.86$\,\pm\,$3.21 & 79.80$\,\pm\,$9.43 & 36.68$\,\pm\,$8.68 & 22.5\% & 1{,}463.3 \\
        \midrule
            \multirowrotate{3}{90}{\footnotesize Gain $\vgamma$}
            & Low-rank $(b, w),$ no $\vgamma$ & 4 & 4 & $\vone$ & 89.59$\,\pm\,$4.87 & 95.79$\,\pm\,$4.10 & 75.10$\,\pm\,$10.61 & 31.50$\,\pm\,$9.10 & 4.3\% & 1{,}047.8 \\
            & Low-rank $(b, w),$ stat.\ $\vgamma$ & 4 & 4 & $2\sigmoid(\vc_q)$ & 90.40$\,\pm\,$4.30 & 96.40$\,\pm\,$3.38 & 77.10$\,\pm\,$9.62 & 33.54$\,\pm\,$8.98 & \underline{3.9\%} & \underline{741.5} \\
            & Low-rank $(b, w),$ dyn.\ $\vgamma_t$ & 4 & 4 & $2\sigmoid(\vz^q_t)$ & 91.35$\,\pm\,$2.44 & 97.00$\,\pm\,$1.56 & 79.41$\,\pm\,$5.93 & 35.68$\,\pm\,$7.18 & \underline{3.9\%} & \tbf{569.2} \\
        \bottomrule
        \end{tabular}%
    }
\end{table}

%% file: apps/f_kernel.tex
\clearpage
\section{Chunkwise Kernel Analysis}
\label{app:kernel}

\input{tables/tab8_flops}

\subsection{MXU and VPU FLOP Breakdown}
\label{app:kernel:flops}

\Cref{tab:kernel_flops} compares the per-token, per-head MXU and VPU operation counts of our boundary-pivoted chunkwise kernel (\S\ref{sec:method:kernel}) against naive chunkwise execution, partitioning $\mT$ and $\mM$ into the $N_{\mathrm{leaf}} = C/n_0$ diagonal blocks of size $n_0 \times n_0$ and the off-diagonal blocks across the $r = \log_2(C/n_0)$ merge levels.
Naive chunkwise execution evaluates the off-diagonal blocks via the 3D decayed-key tensor on the VPU and solves $\mT\mZ = \begin{bmatrix}\hat\mV & \hat\mK_{\odot\pi}\end{bmatrix}$ via sequential forward and backward substitution on the VPU.
In our kernel, at merge level $m \in \{0, \dotsc, r-1\}$ ($n_m = n_0 2^m,$ $N_m = C/(2n_m)$ merges per chunk), forming the pivoted cross-blocks $(\mT_{\mathrm{cross}}, \mM_{\mathrm{cross}})$ costs $2n_m d_k$ MXU FLOPs per token (with $\vpi_\gR, \vpi_\gL$ evaluated on the VPU) and merging $\mT_{\mathrm{blk}}\inv$ via \eqref{eq:block_inv} costs $2n_m^2$ MXU FLOPs per token, which sum over $\sum_{m=0}^{r-1} n_m = C - n_0$ and $\sum_{m=0}^{r-1} n_m^2 = \frac{1}{3}(C^2 - n_0^2)$ in the forward pass while the backward pass reuses the saved $\mT\inv$ directly.


\subsection{Chunkwise Backward Pass}
\label{app:kernel:bwd}

Given upstream gradients $\mathrm{d}\mO \in \R{C\times d_v}$ for each chunk and $\mathrm{d}\mS_{N_c} \in \R{d_k\times d_v}$ at the end of the sequence (where $\mathrm{d}\mX \defas \partial\gJ/\partial\mX$ denotes the gradient of the scalar loss $\gJ$ with respect to $\mX$), the backward pass proceeds in reverse order through the chunk readout and state scan, the triangular solve, and the boundary-pivoted merge tree.


\xhdr{Reverse Cross-Chunk State Scan and Readout Gradients}

In the forward pass \eqref{eq:chunk_state_out}, each chunk advances $\mS_0 \to (\mO, \mS_C)$ via $\mO = (\hat\mQ_{\odot\pi} - \mM\mY)\mS_0 + \mM\mU$ and $\mS_C = (\Diag(\vpi_C) - \tilde\mK\tr\mY)\mS_0 + \tilde\mK\tr\mU.$
Running the backward scan in reverse chunk order $n = N_c, \dotsc, 1$ from terminal state gradient $\mathrm{d}\mS_C$ and chunk output gradient $\mathrm{d}\mO$ computes
\[
    \mathrm{d}\mS_0 = \hat\mQ_{\odot\pi}\tr\mathrm{d}\mO + \Diag(\vpi_C)\mathrm{d}\mS_C - \mY\tr(\mM\tr\mathrm{d}\mO + \tilde\mK\,\mathrm{d}\mS_C)
\]
entirely in VMEM, passing $\mathrm{d}\mS_0$ as $\mathrm{d}\mS_C$ to chunk $n - 1.$
Once $\mS_0$ and $\mathrm{d}\mS_C$ are available for all chunks, the gradients with respect to $(\mM, \tilde\mK, \mU, \mY, \hat\mQ_{\odot\pi})$ and $\vpi_C$ (with $\tilde\mV = \mU - \mY\mS_0$) are computed across all chunks in parallel via packed MXU matrix multiplications:
\begin{align*}
    \mathrm{d}\mM
    &= \tril(\mathrm{d}\mO\,\tilde\mV\tr),
    &
    \mathrm{d}\tilde\mK
    &= \tilde\mV\,\mathrm{d}\mS_C\tr,
    &
    \mathrm{d}\mU
    &= \mM\tr\mathrm{d}\mO + \tilde\mK\,\mathrm{d}\mS_C,
    \\
    \mathrm{d}\mY
    &= -\mathrm{d}\mU\,\mS_0\tr,
    &
    \mathrm{d}\hat\mQ_{\odot\pi}
    &= \mathrm{d}\mO\,\mS_0\tr,
    &
    \mathrm{d}\vpi_C
    &= \vone_{d_v}\big(\mathrm{d}\mS_C \odot \mS_0\big)\tr.
\end{align*}
Finally, differentiating the forward solve $\mZ = \mT\inv[\hat\mV\;\,\hat\mK_{\odot\pi}]$ (where $\mZ = [\mU\;\,\mY]$ and $\mathrm{d}\mZ = [\mathrm{d}\mU\;\,\mathrm{d}\mY]$) yields
\[
    \begin{bmatrix}\mathrm{d}\hat\mV & \mathrm{d}\hat\mK_{\odot\pi}\end{bmatrix}
    = \mT^{-\top}\mathrm{d}\mZ,
    \qquad
    \mathrm{d}\mL_{\mathrm{strict}}
    = -\stril\!\Big(\mathrm{d}\hat\mV\,\mU\tr + \mathrm{d}\hat\mK_{\odot\pi}\mY\tr\Big),
\]
where $\mT^{-\top}\mathrm{d}\mZ$ is evaluated in a single $256\times 256$ MXU matrix multiplication per chunk using the saved $\mT\inv.$


\input{tables/tab9-10_throughput-stability}

\xhdr{Sub-Block Merge Tree and Log-Decay Gradients}

At each merge level $m \in \{0, \dotsc, r-1\}$ with adjacent child intervals $\gL$ and $\gR$ of length $n_m = n_0 2^m$ pivoted at $p = \mathrm{last}(\gL),$ slicing $\mathrm{d}\mT_{\mathrm{cross}} = (\mathrm{d}\mL_{\mathrm{strict}})_{\gR,\gL}$ and $\mathrm{d}\mM_{\mathrm{cross}} = (\mathrm{d}\mM)_{\gR,\gL}$ and contracting via the stacked MXU equations in \S\ref{sec:method:kernel} produces $\mathrm{d}(\hat\mK_\gR \odot \vpi_\gR),$ $\mathrm{d}(\hat\mQ_\gR \odot \vpi_\gR),$ and $\mathrm{d}(\mK_\gL \odot \vpi_\gL).$
Because $\vpi_{\gR,t} = \exp(\mLambda_t - \mLambda_p)$ for $t \in \gR$ and $\vpi_{\gL,s} = \exp(\mLambda_p - \mLambda_s)$ for $s \in \gL,$ the elementwise chain rule accumulates $\mathrm{d}\hat\mK_\gR \mathrel{+}= \mathrm{d}(\hat\mK_\gR \odot \vpi_\gR) \odot \vpi_\gR,$ $\mathrm{d}\hat\mQ_\gR \mathrel{+}= \mathrm{d}(\hat\mQ_\gR \odot \vpi_\gR) \odot \vpi_\gR,$ and $\mathrm{d}\mK_\gL \mathrel{+}= \mathrm{d}(\mK_\gL \odot \vpi_\gL) \odot \vpi_\gL$ alongside the log-decay gradients $\mathrm{d}\mLambda$:
\begin{align*}
    \mathrm{d}\vpi_{\gR,t}
    &\defas \mathrm{d}(\hat\mK_\gR \odot \vpi_\gR)_t \odot \hat\vk_t + \mathrm{d}(\hat\mQ_\gR \odot \vpi_\gR)_t \odot \hat\vq_t,
    &
    \mathrm{d}\vpi_{\gL,s}
    &\defas \mathrm{d}(\mK_\gL \odot \vpi_\gL)_s \odot \vk_s,
    \\
    \mathrm{d}\mLambda_t
    &\mathrel{+}= \mathrm{d}\vpi_{\gR,t} \odot \vpi_{\gR,t} \quad (t \in \gR),
    &
    \mathrm{d}\mLambda_s
    &\mathrel{-}= \mathrm{d}\vpi_{\gL,s} \odot \vpi_{\gL,s} \quad (s \in \gL),
\end{align*}
with the pivot row $p = \mathrm{last}(\gL)$ receiving the net boundary flux $\mathrm{d}\mLambda_p \mathrel{+}= \sum_{s\in\gL}\mathrm{d}\vpi_{\gL,s}\odot\vpi_{\gL,s} - \sum_{t\in\gR}\mathrm{d}\vpi_{\gR,t}\odot\vpi_{\gR,t}.$
Finally, adding the prefix/suffix contributions from $\hat\mK_{\odot\pi}, \hat\mQ_{\odot\pi}, \tilde\mK,$ and $\vpi_C$ into $(\mathrm{d}\hat\mQ, \mathrm{d}\mK, \mathrm{d}\hat\mK, \mathrm{d}\mLambda)$ and multiplying by the transposed lower-triangular all-ones matrix on the MXU converts $\mathrm{d}\mLambda$ into $\mathrm{d}\log\mAlpha = \tril(\vone_{C\times C})\tr\mathrm{d}\mLambda = \triu(\vone_{C\times C})\mathrm{d}\mLambda$ (with rows $\mathrm{d}\log\valpha_t = \sum_{u=t}^C \mathrm{d}\mLambda_u$;\; see \Cref{alg:chunk_fwd,alg:chunk_bwd}, reference JAX implementation in \Cref{app:code}).

\begin{algorithm}[!t]
\caption{Boundary-Pivoted Chunkwise Forward Pass}
\label{alg:chunk_fwd}
\begin{algorithmic}[1]
    \Require $\hat\mQ, \mK, \hat\mK \in \R{N_c \times C \times d_k},$ $\hat\mV \in \R{N_c \times C \times d_v},$ $\log\mAlpha \in \R{N_c \times C \times d_k}$ ($C = 256,$ $n_0 = 8,$ $r = \log_2(C/n_0) = 5$)
    \Ensure $\mO \in \R{N_c \times C \times d_v}$
    \State Compute $\mLambda = \tril(\vone_{C\times C})\,\log\mAlpha,$ prefix decay $\vpi = \exp(\mLambda),$ suffix decay $\tilde\vpi$ with rows $\vpi_{C,s} = \exp(\mLambda_C - \mLambda_s)$
    \State On $2^r$ leaves of size $n_0,$ compute $\vpi_{t,s} = \exp(\mLambda_t - \mLambda_s)$ ($t \ge s$), $(\mL_{\mathrm{leaf}})_{t,s} = \inner{\hat\vk_t}{\vpi_{t,s} \odot \vk_s}$ ($t > s$), and $(\mM_{\mathrm{leaf}})_{t,s} = \inner{\hat\vq_t}{\vpi_{t,s} \odot \vk_s}$ ($t \ge s$)
    \State Invert diagonal leaf blocks $\mT_{\mathrm{leaf}}\inv = (\I_{n_0} + \mL_{\mathrm{leaf}})\inv = \prod_{j=0}^{\log_2 n_0 - 1}(\I_{n_0} + (-\mL_{\mathrm{leaf}})^{2^j})$
    \For{merge level $m = 0, \dotsc, r - 1$ (adjacent child intervals $\gL, \gR$ of size $n_m = n_0 2^m,$ pivot $p = \mathrm{last}(\gL)$)}
        \State Compute boundary-pivot decays $\vpi_\gR = \exp(\mLambda_\gR - \mLambda_p)$ and $\vpi_\gL = \exp(\mLambda_p - \mLambda_\gL)$
        \State Compute off-diagonal blocks $\mT_{\mathrm{cross}} = (\hat\mK_\gR \odot \vpi_\gR)(\mK_\gL \odot \vpi_\gL)\tr$ and $\mM_{\mathrm{cross}} = (\hat\mQ_\gR \odot \vpi_\gR)(\mK_\gL \odot \vpi_\gL)\tr$
        \State Merge off-diagonal inverse block $(\mT_{\mathrm{blk}}\inv)_{\gR,\gL} = -\mT_\gR\inv\mT_{\mathrm{cross}}\mT_\gL\inv$ \eqref{eq:block_inv}
    \EndFor
    \State Compute $\hat\mK_{\odot\pi} = \vpi \odot \hat\mK,$ $\hat\mQ_{\odot\pi} = \vpi \odot \hat\mQ,$ and $\tilde\mK = \tilde\vpi \odot \mK$ (with rows $\tilde\mK_{s,:} = \vpi_{C,s} \odot \vk_s$)
    \State Compute $\mZ = [\mU\;\, \mY] = \mT\inv[\hat\mV\;\, \hat\mK_{\odot\pi}]$
    \State Compute $\mP_{\mathrm{read}} = [\hat\mQ_{\odot\pi} - \mM\mY;\; \Diag(\vpi_C) - \tilde\mK\tr\mY]$ and $\mP_{\mathrm{bias}} = [\mM\mU;\; \tilde\mK\tr\mU]$ \eqref{eq:chunk_state_out}
    \State Initialize recurrent state $\mS = \mZero \in \R{d_k \times d_v}$ in VMEM
    \For{$n = 1, \dotsc, N_c$}
        \State Save chunk-initial state $\mS_{0,[n]} = \mS$ and update $[\mO_{[n]};\; \mS] \gets \mP_{\mathrm{read},[n]}\mS + \mP_{\mathrm{bias},[n]}$
    \EndFor
    \State \Return $\mO = \{\mO_{[1]}, \dotsc, \mO_{[N_c]}\}$
\end{algorithmic}
\end{algorithm}

\begin{algorithm}[!t]
\caption{Boundary-Pivoted Chunkwise Custom Backward Pass}
\label{alg:chunk_bwd}
\begin{algorithmic}[1]
    \Require Saved $\mT\inv, \mM, \mZ = [\mU\;\, \mY],$ chunk-initial states $\mS_0 = \{\mS_{0,[n]}\}_{n=1}^{N_c},$ inputs $(\hat\mQ, \mK, \hat\mK, \log\mAlpha),$ and $\mathrm{d}\mO$
    \Ensure $\mathrm{d}\hat\mQ, \mathrm{d}\mK, \mathrm{d}\hat\mK, \mathrm{d}\hat\mV, \mathrm{d}\log\mAlpha$
    \State Compute $\mLambda = \tril(\vone_{C\times C})\log\mAlpha,$ $\vpi = \exp(\mLambda),$ $\hat\mQ_{\odot\pi} = \vpi \odot \hat\mQ,$ and $\tilde\mK$ with rows $\vpi_{C,s} \odot \vk_s$ (with $\vpi_C = \vpi_{C,:}$)
    \State Initialize terminal state gradient $\mathrm{d}\mS = \mZero \in \R{d_k \times d_v}$ in VMEM
    \For{$n = N_c, \dotsc, 1$}
        \State Save $\mathrm{d}\mS_{C,[n]} = \mathrm{d}\mS$ and update $\mathrm{d}\mS \gets \hat\mQ_{\odot\pi,[n]}\tr\mathrm{d}\mO_{[n]} + \Diag(\vpi_{C,[n]})\mathrm{d}\mS - \mY_{[n]}\tr(\mM_{[n]}\tr\mathrm{d}\mO_{[n]} + \tilde\mK_{[n]}\mathrm{d}\mS)$
    \EndFor
    \State Compute $\tilde\mV = \mU - \mY\mS_0,$ $\mathrm{d}\mM = \tril(\mathrm{d}\mO\,\tilde\mV\tr),$ and $\mathrm{d}\tilde\mK = \tilde\mV\,\mathrm{d}\mS_C\tr$
    \State Compute $\mathrm{d}\mU = \mM\tr\mathrm{d}\mO + \tilde\mK\,\mathrm{d}\mS_C,$ $\mathrm{d}\mY = -\mathrm{d}\mU\,\mS_0\tr,$ $\mathrm{d}\hat\mQ_{\odot\pi} = \mathrm{d}\mO\,\mS_0\tr,$ and $\mathrm{d}\vpi_C = \vone_{d_v}(\mathrm{d}\mS_C \odot \mS_0)\tr$
    \State Compute $[\mathrm{d}\hat\mV\;\, \mathrm{d}\hat\mK_{\odot\pi}] = \mT^{-\top}[\mathrm{d}\mU\;\, \mathrm{d}\mY]$ and $\mathrm{d}\mL_{\mathrm{strict}} = -\stril(\mathrm{d}\hat\mV\,\mU\tr + \mathrm{d}\hat\mK_{\odot\pi}\mY\tr)$
    \State Differentiate $2^r$ diagonal leaves $\mathrm{d}\mL_{\mathrm{leaf}} = (\mathrm{d}\mL_{\mathrm{strict}})_{\mathrm{leaf}}$ and $\mathrm{d}\mM_{\mathrm{leaf}} = (\mathrm{d}\mM)_{\mathrm{leaf}}$ into $(\mathrm{d}\hat\mQ, \mathrm{d}\mK, \mathrm{d}\hat\mK, \mathrm{d}\mLambda)$
    \For{merge level $m = 0, \dotsc, r - 1$ (adjacent child intervals $\gL, \gR$ of size $n_m = n_0 2^m,$ pivot $p = \mathrm{last}(\gL)$)}
        \State Recompute boundary-pivot decays $\vpi_\gR = \exp(\mLambda_\gR - \mLambda_p)$ and $\vpi_\gL = \exp(\mLambda_p - \mLambda_\gL)$
        \State Slice $\mathrm{d}\mT_{\mathrm{cross}} = (\mathrm{d}\mL_{\mathrm{strict}})_{\gR,\gL}$ and $\mathrm{d}\mM_{\mathrm{cross}} = (\mathrm{d}\mM)_{\gR,\gL}$
        \State Compute $\mathrm{d}(\hat\mK_\gR \odot \vpi_\gR) = \mathrm{d}\mT_{\mathrm{cross}}(\mK_\gL \odot \vpi_\gL)$ and $\mathrm{d}(\hat\mQ_\gR \odot \vpi_\gR) = \mathrm{d}\mM_{\mathrm{cross}}(\mK_\gL \odot \vpi_\gL)$
        \State Compute $\mathrm{d}(\mK_\gL \odot \vpi_\gL) = \mathrm{d}\mT_{\mathrm{cross}}\tr(\hat\mK_\gR \odot \vpi_\gR) + \mathrm{d}\mM_{\mathrm{cross}}\tr(\hat\mQ_\gR \odot \vpi_\gR)$
        \State Accumulate chain-rule gradients through $(\vpi_\gR, \vpi_\gL)$ into $(\mathrm{d}\hat\mQ_\gR, \mathrm{d}\hat\mK_\gR, \mathrm{d}\mK_\gL)$ and $(\mathrm{d}\mLambda_\gR, \mathrm{d}\mLambda_\gL, \mathrm{d}\mLambda_p)$
    \EndFor
    \State Accumulate prefix/suffix decay gradients from $(\mathrm{d}\hat\mK_{\odot\pi}, \mathrm{d}\hat\mQ_{\odot\pi}, \mathrm{d}\tilde\mK, \mathrm{d}\vpi_C)$ into $(\mathrm{d}\hat\mQ, \mathrm{d}\mK, \mathrm{d}\hat\mK, \mathrm{d}\mLambda)$
    \State Compute $\mathrm{d}\log\mAlpha = \tril(\vone_{C\times C})\tr\mathrm{d}\mLambda = \triu(\vone_{C\times C})\mathrm{d}\mLambda$
    \State \Return $\mathrm{d}\hat\mQ, \mathrm{d}\mK, \mathrm{d}\hat\mK, \mathrm{d}\hat\mV, \mathrm{d}\log\mAlpha$
\end{algorithmic}
\end{algorithm}


\subsection{Training Throughput and Numerical Stability}
\label{app:kernel:perf}

\Cref{tab:kernel_tput} reports the complete $1.34\text{B}$ training throughput ($\mathrm{p50}$ over $100$-step training intervals), speedup ratios, compiled XLA~\citep{Sabne2020-xj} FLOPs, achieved model FLOP utilization (MFU), and HBM traffic for sequence lengths $T \in \{2\text{K}, 4\text{K}, 8\text{K}, 16\text{K}, 32\text{K}\}$ on $8$ TPU7x chips at a fixed global batch size $B\times T = 524{,}288\text{ tokens/step}.$
Across all \tf runs, TPU Pallas Splash Attention uses $512$-token \texttt{bfloat16} query and key-value VMEM tiles, with $128$-token key-value compute sub-blocks in the forward and key-value backward kernels.
\Cref{tab:kernel_num} reports the forward and backward relative errors against an exact \texttt{float64} sequential reference and the empirical condition number $\kappa_2(\mT)$ across synthetic stress regimes and $100\text{B}$-token trained checkpoints.

%% file: tables/tab8_flops.tex
\begin{table}[!t]
    \centering
    \caption{%
        \textbf{Per-token, per-head FLOP breakdown of the boundary-pivoted chunkwise kernel and naive chunkwise execution.}
        Stage rows report leading per-token MXU and VPU operation counts (given cumulative log-decay $\mLambda$) as a function of chunk length $C,$ head dimensions $(d_k, d_v),$ leaf size $n_0,$ and tree depth $r = \log_2(C/n_0).$
        MXU shares and totals are evaluated at $C = 256,$ $d_k = d_v = 128,$ and $n_0 = 8$ ($r = 5$).%
    }
    \label{tab:kernel_flops}
    \resizebox{\textwidth}{!}{%
        \setlength{\tabcolsep}{2pt}%
        \renewcommand{\arraystretch}{1.12}%
        \begin{tabular}{@{}l|cc|c|cc|c@{}}
        \toprule
            & \multicolumn{3}{c|}{{Naive Chunkwise}} & \multicolumn{3}{c@{}}{{Boundary-Pivoted Kernel (Ours)}} \\
            \cmidrule(lr){2-4} \cmidrule(l){5-7}
            {Kernel Stage} & {MXU} & {VPU} & {MXU} & {MXU} & {VPU} & {MXU} \\
            & {FLOPs} & {FLOPs} & {Share} & {FLOPs} & {FLOPs} & {Share} \\
        \midrule
            Diag.\ blocks ($n_0$), $\mT_{\mathrm{leaf}}\inv$ \eqref{eq:neumann} & $0$ & $\gO(n_0 d_k)$ & 0.0\% & $4n_0^2(\log_2 n_0\!-\!1)$ & $\gO(n_0 d_k)$ & 6.3\% \\
            Off-diag.\ blocks ($\mT_{\mathrm{cross}}, \mM_{\mathrm{cross}}$) & $0$ & $7(C\!-\!n_0)d_k$ & 0.0\% & $2(C\!-\!n_0)d_k$ & $\gO(r d_k)$ & 96.6\% \\
            $2\!\times\!2$ block-tri.\ merges \eqref{eq:block_inv} & $0$ & $0$ & -- & $\tfrac{2}{3}(C^2\!-\!n_0^2)$ & $0$ & 100.0\% \\
            Tri.\ solve $\mT\mZ = [\hat\mV\;\,\hat\mK_{\odot\pi}]$ & $0$ & $C(d_k\!+\!d_v)$ & 0.0\% & $2C(d_k\!+\!d_v)$ & $0$ & 100.0\% \\
            Chunk readout, state scan \eqref{eq:chunk_state_out} & $\tfrac{2}{C}(C\!+\!d_k)^2 d_v\!+\!2(C\!+\!d_k)d_k$ & $\gO(d_k)$ & 99.5\% & $\tfrac{2}{C}(C\!+\!d_k)^2 d_v\!+\!2(C\!+\!d_k)d_k$ & $\gO(d_k)$ & 99.5\% \\
        \midrule
            {Fwd pass subtotal} & 245{,}760 & 296{,}576 & 45.3\% & \tbf{484{,}480} & \tbf{11{,}072} & \tbf{97.8\%} \\
        \midrule
            Rev.\ state scan \& chunk grads & $\tfrac{4}{C}(C\!+\!d_k)^2 d_v\!+\!4(C\!+\!d_k)d_k$ & $\gO(d_k)$ & 99.5\% & $\tfrac{4}{C}(C\!+\!d_k)^2 d_v\!+\!4(C\!+\!d_k)d_k$ & $\gO(d_k)$ & 99.5\% \\
            Tri.\ solve grads $(\mathrm{d}\mZ, \mathrm{d}\mL_{\mathrm{strict}})$ & $2C(d_k\!+\!d_v)$ & $C(d_k\!+\!d_v)$ & 66.7\% & $4C(d_k\!+\!d_v)$ & $0$ & 100.0\% \\
            Off-diag.\ grads $(\mathrm{d}\mT_{\mathrm{cross}}, \mathrm{d}\mM_{\mathrm{cross}})$ & $0$ & $11(C\!-\!n_0)d_k$ & 0.0\% & $4(C\!-\!n_0)d_k$ & $\gO(r d_k)$ & 96.4\% \\
            Diag.\ leaf grads $(\mathrm{d}\mL_{\mathrm{leaf}}, \mathrm{d}\mM_{\mathrm{leaf}})$ & $0$ & $\gO(n_0 d_k)$ & 0.0\% & $0$ & $\gO(n_0 d_k)$ & 0.0\% \\
        \midrule
            {Bwd pass subtotal} & 622{,}592 & 428{,}544 & 59.2\% & \tbf{880{,}640} & \tbf{18{,}624} & \tbf{97.9\%} \\
        \midrule
            {Full fwd+bwd total} & 868{,}352 & 725{,}120 & 54.5\% & \tbf{1{,}365{,}120} & \tbf{29{,}696} & \tbf{97.9\%} \\
        \bottomrule
        \end{tabular}%
    }
\end{table}

%% file: tables/tab9-10_throughput-stability.tex
\begin{table}[!t]
    \centering
    \caption{%
        \textbf{End-to-end training throughput, compiled XLA FLOPs, and HBM traffic across sequence lengths at $\boldsymbol{1.34\text{B}}$ scale.}
        All measurements are collected on $8$ TPU7x chips over $100$-step pure-training intervals at a fixed global batch size of $B\times T = 524{,}288\text{ tokens/step},$ comparing iso-parameter \ours ($d_{\mathrm{ffn}} = 7{,}680$) and matched-FFN \ours ($d_{\mathrm{ffn}} = 5{,}888$) against \tf (TPU Pallas Splash Attention) and \gdntwo.
        (XLA profile metrics are reported per TensorCore device across the $16$ TensorCores of the $8$ TPU7x chips; compiled TFLOP/dev includes activation rematerialization, whereas MFU counts non-rematerialized model FLOPs.
        Bottom-row XLA profile ratios are evaluated at $T = 32\text{K}.$)%
    }
    \label{tab:kernel_tput}
    \resizebox{\textwidth}{!}{%
        \setlength{\tabcolsep}{3pt}%
        \renewcommand{\arraystretch}{1.08}%
        \begin{tabular}{@{}c@{\hspace{3pt}}|l|ccccc|ccc@{}}
        \toprule
        \multicolumn{1}{@{}c@{\hspace{5pt}}}{} & & \multicolumn{5}{c|}{{Throughput ($\text{K tokens/s}$)}} & \multicolumn{3}{c@{}}{{XLA Profile ($4\text{K} \to 32\text{K}$)}} \\
            \cmidrule(lr){3-7} \cmidrule(l){8-10}
            \multicolumn{1}{@{}c@{\hspace{5pt}}}{} & {Model ($1.34\text{B}$)} & {$T\!=\!2\text{K}$} & {$T\!=\!4\text{K}$} & {$T\!=\!8\text{K}$} & {$T\!=\!16\text{K}$} & {$T\!=\!32\text{K}$} & {Compiled} & {Achieved} & {HBM Traffic} \\
            \multicolumn{1}{@{}c@{\hspace{5pt}}}{} & & {($B\!=\!256$)} & {($B\!=\!128$)} & {($B\!=\!64$)} & {($B\!=\!32$)} & {($B\!=\!16$)} & {TFLOP/dev} & {MFU ($\%$)} & {(GB/dev)} \\
            \midrule
            \multirowrotate{4}{90}{\footnotesize $\mathrm{p50}$}
            & \tf ($8{,}192$) & \tbf{523.5} & \tbf{396.9} & \tbf{267.9} & 163.6 & 93.6 & 348$\,\to\,$652 & 18.2$\,\to\,$8.6 & 613$\,\to\,$750 \\
            & \gdntwo ($5{,}888$) & 179.6 & 179.3 & 179.1 & 264.3 & 253.5 & 330$\,\to\,$330 & 7.1$\,\to\,$10.1 & 4{,}478$\,\to\,$4{,}148 \\
            & \ours (matched, $5{,}888$) & \underline{183.4} & \underline{183.2} & \underline{182.6} & \tbf{276.8} & \tbf{261.8} & 285$\,\to\,$285 & 6.2$\,\to\,$8.9 & 4{,}413$\,\to\,$4{,}058 \\
            & \tbf{\ours (ours; iso, $\boldsymbol{7{,}680}$)} & 179.5 & 179.4 & 175.6 & \underline{268.5} & \underline{256.1} & 324$\,\to\,$324 & 7.1$\,\to\,$10.2 & 4{,}445$\,\to\,$4{,}089 \\
        \midrule
            \multirowrotate{3}{90}{\footnotesize Ratio}
            & \ours vs.\ \gdntwo (iso) & 0.999\texttimes{} & \tbf{1.001\texttimes{}} & 0.980\texttimes{} & \tbf{1.016\texttimes{}} & \tbf{1.010\texttimes{}} & $-$1.8\% & $+$0.1 pt & $-$1.4\% \\
            & \ours vs.\ \gdntwo (matched) & \tbf{1.021\texttimes{}} & \tbf{1.022\texttimes{}} & \tbf{1.020\texttimes{}} & \tbf{1.047\texttimes{}} & \tbf{1.033\texttimes{}} & $-$13.6\% & $-$1.2 pt & $-$2.2\% \\
            & \ours vs.\ \tf & 0.343\texttimes{} & 0.452\texttimes{} & 0.655\texttimes{} & \tbf{1.641\texttimes{}} & \tbf{2.736\texttimes{}} & $-$50.3\% & $+$1.6 pt & 5.45\texttimes{} \\
        \bottomrule
        \end{tabular}%
    }

\vspace{1em}

    \caption{%
        \textbf{Numerical accuracy and unit lower-triangular conditioning of our boundary-pivoted chunkwise kernel.}
        (Top) Forward ($T = 4{,}096, C = 256, d_k = d_v = 128$) and single-layer backward pass ($T = 128, C = 64, d_k = d_v = 16$) relative $\ell_2$ errors against an exact \texttt{float64} sequential reference.
        %
        (Bottom) Condition numbers $\kappa_2(\mT)$ and triangular-solve relative errors ($\mZ = \mT\inv[\hat\mV\;\,\hat\mK_{\odot\pi}]$; \texttt{bf16} block-Neumann vs.\ sequential \texttt{fp32} forward substitution) across $1{,}536$ chunk systems ($C = 256$) sampled from $166\text{M}$ $100\text{B}$-token checkpoints and synthetic stress regimes.%
    }
    \label{tab:kernel_num}
    \resizebox{\textwidth}{!}{%
        \setlength{\tabcolsep}{2pt}%
        \renewcommand{\arraystretch}{1.08}%
        \begin{tabular}{@{}c@{\hspace{3pt}}|l|cc|cccccc@{}}
        \toprule
        \multicolumn{1}{@{}c@{\hspace{5pt}}}{} & & \multicolumn{2}{c|}{{Forward Error $\downarrow$}} & \multicolumn{6}{c@{}}{{Backward Gradient Relative $\ell_2$ Error $\downarrow$}} \\
            \cmidrule(lr){3-4} \cmidrule(l){5-10}
            \multicolumn{1}{@{}c@{\hspace{5pt}}}{} & {Kernel Implementation} & {Out $\mO$} & {State $\mS_T$} & {$\mathrm{d}\vq$} & {$\mathrm{d}\vk$} & {$\mathrm{d}\vv$} & {$\mathrm{d}\vb$} & {$\mathrm{d}\vw$} & {$\mathrm{d}\log\valpha$} \\
        \midrule
        \multirowrotate{3}{90}{\footnotesize Error}
            & Naive unpivoted & \texttt{NaN} & \texttt{NaN} & \texttt{NaN} & \texttt{NaN} & \texttt{NaN} & \texttt{NaN} & \texttt{NaN} & \texttt{NaN} \\
            & Boundary-pivoted \texttt{bf16} & $7.62\!\times\!10^{-3}$ & $6.72\!\times\!10^{-3}$ & $5.76\!\times\!10^{-3}$ & $7.15\!\times\!10^{-3}$ & $5.81\!\times\!10^{-3}$ & $8.14\!\times\!10^{-3}$ & $5.65\!\times\!10^{-3}$ & $9.14\!\times\!10^{-3}$ \\
            & Boundary-pivoted \texttt{fp32} & $6.99\!\times\!10^{-7}$ & $8.12\!\times\!10^{-7}$ & $2.74\!\times\!10^{-7}$ & $4.04\!\times\!10^{-7}$ & $3.49\!\times\!10^{-7}$ & $5.04\!\times\!10^{-7}$ & $3.53\!\times\!10^{-7}$ & $6.43\!\times\!10^{-7}$ \\
        \midrule
        \midrule
        \multicolumn{1}{@{}c@{\hspace{5pt}}}{} & & \multicolumn{2}{c|}{{Regime Bounds}} & \multicolumn{4}{c|}{{Condition Number $\kappa_2(\mT)$}} & \multicolumn{2}{c@{}}{{Solve Error $\mZ$ $\downarrow$}} \\
            \cmidrule(lr){3-4} \cmidrule(lr){5-8} \cmidrule(l){9-10}
            \multicolumn{1}{@{}c@{\hspace{5pt}}}{} & {Conditioning Regime} & {Decay $\evalpha_{t,j}$} & {Erase $\bar\evmu$} & {$\mathrm{p50}$} & {$\mathrm{p90}$} & {$\mathrm{p99}$} & \multicolumn{1}{c|}{{Max}} & {\texttt{bf16}} & {\texttt{fp32}} \\
            \midrule
            \multirowrotate{2}{90}{\footnotesize Ckpt}
            & \gdntwo ($100\text{B}$) & $[0.083, 0.9992]$ & $0.65$ & 4.23 & 11.13 & 17.75 & \multicolumn{1}{c|}{21.74} & $1.85\!\times\!10^{-3}$ & $5.45\!\times\!10^{-8}$ \\
            & \tbf{\ours (static $\boldsymbol{\vgamma},$ 100B)} & $[0.048, 0.9995]$ & $0.57$ & 3.20 & 7.73 & 20.69 & \multicolumn{1}{c|}{39.18} & $1.80\!\times\!10^{-3}$ & $4.58\!\times\!10^{-8}$ \\
        \midrule
        \multirowrotate{4}{90}{\footnotesize Synth.}
            & Synth.\ mild & $[0.069, 0.9992]$ & $1.12$ & 6.13 & 43.74 & 45.79 & \multicolumn{1}{c|}{47.00} & $2.11\!\times\!10^{-3}$ & $5.95\!\times\!10^{-8}$ \\
            & Synth.\ moderate & $[0.050, 0.9990]$ & $1.12$ & 11.46 & 64.74 & 67.20 & \multicolumn{1}{c|}{68.01} & $2.63\!\times\!10^{-3}$ & $7.65\!\times\!10^{-8}$ \\
            & Synth.\ coherent keys & $[0.985, 0.9992]$ & $1.12$ & 158.6 & 161.8 & 164.0 & \multicolumn{1}{c|}{164.2} & $1.19\!\times\!10^{-2}$ & $4.02\!\times\!10^{-7}$ \\
            & Synth.\ adversarial & $[0.988, 0.9993]$ & $1.99$ & 11{,}662 & 12{,}396 & 12{,}818 & \multicolumn{1}{c|}{12{,}865} & $5.23\!\times\!10^{0}$ & $5.79\!\times\!10^{-6}$ \\
        \bottomrule
        \end{tabular}%
    }
\end{table}

%% file: apps/g_train.tex
\clearpage
\section{Training Setup and Model Configurations}
\label{app:train}

\subsection{Pretraining Setup}
\label{app:train:pretrain}

Across all four model scales (\Cref{tab:train_configs}), every layer allocates $4d^2$ parameters to the core query, key, value, and output projections ($\mW_q, \mW_k, \mW_v, \mW_o \in \R{d\times d}$) and $8d^2$ parameters between the gate projections and the two-matrix $\gelu$ FFN ($2d\,d_{\mathrm{ffn}}$ parameters), yielding an exact per-layer non-embedding budget of $12d^2$:
\[
    \text{\tf:}\;\; d_{\mathrm{ffn}} = 4d;\qquad
    \text{\gdntwo:}\;\; d_{\mathrm{ffn}} = 3d - 256;\qquad
    \text{\ours:}\;\; d_{\mathrm{ffn}} = 4d - R_{\mathrm{tot}} = 4d - 512.
\]
All recurrent models include a four-token depthwise causal convolution over the concatenated $(q, k, v)$ channels ($4 \times 3d$ parameters per layer) and per-head $\rms$ normalization prior to the output gate $\silu(\vg_t).$

Adapting the decay parameterization of \citet{Yang2025-dr} and \citet{Hatamizadeh2026-cw}, we stratify the $N_h$ heads across four quartiles $k \in \{1, 2, 3, 4\}$ to reduce initialization variance at fixed seed:\footnote{%
    The $(64, 64, 128, 256, 0)$ no-$\vgamma$ runs used unstratified $[1, 16]$ and $[0.001, 0.1]$ initialization; since $a_{\log}$ and $\vc_f$ are freely learned over $100\text{B}$ tokens, they match the stratified $(128, 128, 128, 128, 0)$ no-$\vgamma$ runs in \Cref{tab:abl} within $0.5\text{ mnats}$ at $834\text{M}$ and $1.34\text{B}.$%
}
\[
    \exp(a_{\log}) \sim \mathrm{Unif}\big(2^{k-1},\, 2^k\big);\qquad
    \softplus(\evc_{f,j}) \sim \mathrm{LogUnif}(\tau_{k-1},\, \tau_k),\;\;
    \tau_{0:4} = (0.001, 0.004, 0.015, 0.05, 0.15).
\]
The remaining up-projection biases ($\vc_g$ inherited from \gdntwo, and $\vc_b, \vc_w, \vc_q$ newly added in \ours) are initialized to $\vzero.$
In \ours, the dynamic readout up-projection is initialized to $\mW_{q\uparrow} = \mZero$ and the static readout gain to $\vgamma^{\mathrm{stat}} = \vone$ so that $2\sigmoid(\vz^q_t) = \vone$ and $\vgamma_t = \vone$ at initialization (with zero weight decay on $a_{\log},$ additive biases $\vc_m,$ and $\vgamma^{\mathrm{stat}}$).
All pretraining runs across all scales use a fixed random seed of $0$ for both model initialization and deterministic data-shard ordering, and all models and the boundary-pivoted chunkwise kernel are implemented in JAX~\citep{Bradbury2018-ab} and compiled with XLA~\citep{Sabne2020-xj}.

\input{tables/tab11_train}


\subsection{Long-Context Cooldown and Position Extrapolation}
\label{app:train:longctx}

For the $64\text{K}$ continual cooldown at $1.34\text{B}$ parameters, we initialize model weights from the $100\text{B}$-token checkpoints with fresh optimizer state and train for $300$ steps at sequence length $T = 65{,}536$ and global batch size $1{,}048{,}576\text{ tokens/step}$ ($16$ sequences sharded across $16$ TPU7x TensorCores via fully sharded data parallelism, FSDP), totaling $314{,}572{,}800$ tokens ($0.315\text{B},$ or $0.3\%$ of pretraining).
Following \citet{Fu2024-bl}, the cooldown stream is drawn from FineWeb-Edu shards disjoint from both pretraining and evaluation data, using a $70/30$ token mixture of intact long documents ($\ge\!16{,}384$ tokens) and standard documents ($<\!16{,}384$ tokens) without synthetic retrieval examples.
Optimization uses AdamW ($\beta_1 = 0.9,$ $\beta_2 = 0.999,$ weight decay $0.1,$ gradient norm clip $1.0$) with a $30$-step ($10\%$) linear warmup to $10^{-4}$ and cosine decay to $10^{-5}.$

Recurrent models keep identical parameterization and hyperparameters across context lengths, whereas \tf raises its RoPE base from $\theta = 10^4$ to $5\times 10^5$ during cooldown \citep{Xiong2024-rg}.
At evaluation lengths $T > T_0$ beyond the active training window ($T_0 = 4{,}096$ zero-shot and $T_0 = 65{,}536$ post-cooldown), our \tf baseline applies training-free Dynamic YaRN \citep{Peng2024-tz} with scale factor $s(T) = T / T_0,$ NTK-by-parts rotation thresholds $(\beta_{\mathrm{slow}}, \beta_{\mathrm{fast}}) = (1, 32)$ across rotary dimension pairs, and query-key rotary amplitude scale $0.1\ln s(T) + 1.$

%% file: tables/tab11_train.tex
\begin{table}[!t]
    \centering
    \caption{%
        \textbf{Architectural specifications and parameter accounting across all four iso-parameter scales ($\boldsymbol{166\text{M}}$ to $\boldsymbol{1.34\text{B}}$).}
        All models at a given scale share depth $L,$ residual width $d,$ head dimensions $d_k = d_v = 128,$ head count $N_h = d/128$ (so all recurrent models are strictly iso-state with $N_h d_k d_v = 128d$ state elements per layer), vocabulary size $V = 32{,}000$ (untied input/output embeddings), per-layer non-embedding budget $12d^2,$ global batch size $524{,}288\text{ tokens/step}$ ($B = 128,$ $T = 4{,}096$), and $100\text{B}$ training tokens.%
    }
    \label{tab:train_configs}
    \resizebox{0.96\textwidth}{!}{%
        \setlength{\tabcolsep}{5pt}%
        \renewcommand{\arraystretch}{1.08}%
        \begin{tabular}{@{}c@{\hspace{4pt}}|l|cccc|cccccc|ccc@{}}
            \toprule
            \multicolumn{1}{@{}c@{\hspace{5pt}}}{} & & \multicolumn{4}{c|}{{Configuration}} & \multicolumn{6}{c|}{{Gate Ranks \& FFN Width}} & \multicolumn{3}{c@{}}{{Param Breakdown (M)}} \\
            \cmidrule(lr){3-6} \cmidrule(lr){7-12} \cmidrule(l){13-15}
            \multicolumn{1}{@{}c@{\hspace{5pt}}}{} & {Model} & {$L$} & {$d$} & {$N_h$} & {$d_k=d_v$} & {$r_f$} & {$r_b$} & {$r_g$} & {$r_w$} & {$r_q$} & {$d_{\mathrm{ffn}}$} & {Mixer} & {FFN} & {Total} \\
            \multicolumn{1}{@{}c@{\hspace{5pt}}}{} & & & & & & & & & & & & {+ Gates} & {Params} & {Params} \\
            \midrule
            \multirowrotate{5}{90}{\footnotesize 166M}
            & \tf & 8 & 1{,}024 & 8 & 128 & -- & -- & -- & -- & -- & 4{,}096 & 33.55 & 67.11 & 166.20 \\
            & \gdntwo & 8 & 1{,}024 & 8 & 128 & 128 & 1{,}024 & 128 & 1{,}024 & -- & 2{,}816 & 54.53 & 46.14 & 166.20 \\
            & \ours (no $\vgamma$) & 8 & 1{,}024 & 8 & 128 & 64 & 64 & 128 & 256 & 0 & 3{,}584 & 41.94 & 58.72 & 166.20 \\
            & \ours (static $\vgamma$) & 8 & 1{,}024 & 8 & 128 & 64 & 64 & 128 & 256 & 0 & 3{,}584 & 41.95 & 58.72 & 166.21 \\
            & \tbf{\ours (ours)} & 8 & 1{,}024 & 8 & 128 & 64 & 64 & 128 & 128 & 128 & 3{,}584 & 41.95 & 58.72 & 166.21 \\
            \midrule
            \multirowrotate{5}{90}{\footnotesize 406M}
            & \tf & 11 & 1{,}536 & 12 & 128 & -- & -- & -- & -- & -- & 6{,}144 & 103.81 & 207.62 & 409.73 \\
            & \gdntwo & 11 & 1{,}536 & 12 & 128 & 128 & 1{,}536 & 128 & 1{,}536 & -- & 4{,}352 & 164.36 & 147.06 & 409.73 \\
            & \ours (no $\vgamma$) & 11 & 1{,}536 & 12 & 128 & 64 & 64 & 128 & 256 & 0 & 5{,}632 & 121.11 & 190.32 & 409.73 \\
            & \ours (static $\vgamma$) & 11 & 1{,}536 & 12 & 128 & 64 & 64 & 128 & 256 & 0 & 5{,}632 & 121.13 & 190.32 & 409.75 \\
            & \tbf{\ours (ours)} & 11 & 1{,}536 & 12 & 128 & 64 & 64 & 128 & 128 & 128 & 5{,}632 & 121.13 & 190.32 & 409.75 \\
            \midrule
            \multirowrotate{5}{90}{\footnotesize 834M}
            & \tf & 14 & 2{,}048 & 16 & 128 & -- & -- & -- & -- & -- & 8{,}192 & 234.88 & 469.76 & 835.72 \\
            & \gdntwo & 14 & 2{,}048 & 16 & 128 & 128 & 2{,}048 & 128 & 2{,}048 & -- & 5{,}888 & 367.00 & 337.64 & 835.72 \\
            & \ours (no $\vgamma$) & 14 & 2{,}048 & 16 & 128 & 64 & 64 & 128 & 256 & 0 & 7{,}680 & 264.24 & 440.40 & 835.72 \\
            & \ours (static $\vgamma$) & 14 & 2{,}048 & 16 & 128 & 64 & 64 & 128 & 256 & 0 & 7{,}680 & 264.27 & 440.40 & 835.74 \\
            & \tbf{\ours (ours)} & 14 & 2{,}048 & 16 & 128 & 64 & 64 & 128 & 128 & 128 & 7{,}680 & 264.27 & 440.40 & 835.74 \\
            \midrule
            \multirowrotate{5}{90}{\footnotesize 1.34B}
            & \tf & 24 & 2{,}048 & 16 & 128 & -- & -- & -- & -- & -- & 8{,}192 & 402.65 & 805.31 & 1{,}339.03 \\
            & \gdntwo & 24 & 2{,}048 & 16 & 128 & 128 & 2{,}048 & 128 & 2{,}048 & -- & 5{,}888 & 629.15 & 578.81 & 1{,}339.03 \\
            & \ours (no $\vgamma$) & 24 & 2{,}048 & 16 & 128 & 64 & 64 & 128 & 256 & 0 & 7{,}680 & 452.98 & 754.97 & 1{,}339.03 \\
            & \ours (static $\vgamma$) & 24 & 2{,}048 & 16 & 128 & 64 & 64 & 128 & 256 & 0 & 7{,}680 & 453.03 & 754.97 & 1{,}339.08 \\
            & \tbf{\ours (ours)} & 24 & 2{,}048 & 16 & 128 & 64 & 64 & 128 & 128 & 128 & 7{,}680 & 453.03 & 754.97 & 1{,}339.08 \\
            \bottomrule
        \end{tabular}%
    }
\end{table}

%% file: apps/h_eval.tex
\clearpage
\section{Evaluation Protocols and Supplementary Results}
\label{app:eval}


\subsection{Language Modeling and Commonsense Reasoning}
\label{app:eval:lm}

All perplexity and zero-shot commonsense evaluations in \Cref{tab:main_lm,tab:abl} are executed via the \texttt{lm-evaluation-harness} (\texttt{v0.4.11}; \citealp{Gao2024-jc}) with \texttt{float32} output logits, rolling log-likelihood context window $2{,}048$ (and $4{,}096$ in \Cref{tab:eval_lm_supp}) for WikiText, and length-normalized accuracy (\accn{}) for HellaSwag and ARC-Challenge~\citep{Yang2025-dr,Lahoti2026-fc}.
As noted in \S\ref{sec:exp:setup}, the BoolQ validation split ($3{,}270$ binary yes/no passages) has a $62.2\%$ majority-class (``yes'') baseline~\citep{Clark2019-zw}.
For language models below $1.5\text{B}$ parameters trained on $100\text{B}$ tokens, log-likelihood margins between ``yes'' and ``no'' cluster near zero, so small shifts in decision threshold flip hundreds of predictions and induce $5\text{--}8$ percentage points of single-checkpoint variance that is uncorrelated with held-out validation NLL~\citep{Madaan2024-bx,Miller2024-aw,Heineman2025-ji}.
Complementing the FineWeb-Edu validation NLL, $2\text{K}$ WikiText and LAMBADA perplexities, and 8-task zero-shot commonsense average in \Cref{tab:main_lm}, \Cref{tab:eval_lm_supp} reports FineWeb-Edu validation token perplexity, full-context ($T = 4{,}096$) WikiText word perplexity, zero-shot BoolQ accuracy, and the 9-task zero-shot average including BoolQ (``Avg.\ (9)'') across all four scales.

\input{tables/tab12_lm_suppl}


\subsection{Associative Recall}
\label{app:eval:recall}

\xhdr{RULER Benchmark Protocol}

For synthetic associative retrieval on RULER \citep{Hsieh2024-ji}, we evaluate six tasks across context lengths $T \in \{1\text{K}, 2\text{K}, 4\text{K}, 8\text{K}\}$ ($N = 500$ examples per task and context length, generated with fixed random seed $42$ and tokenized with the Llama-2 vocabulary, reserving $64$ tokens for generation so each input prompt contains at most $T - 64$ tokens).
For each task and context length, target needles are inserted across $40$ uniformly spaced relative depth positions $d \in [0, 1]$ from the start ($0\%$) to the end ($100\%$) of the haystack.

\input{tables/tab13_niah_suppl}

On single-needle retrieval (S-NIAH-1/2/3), each example embeds a single needle of the form \texttt{"One of the special magic \{type\} for \{key\} is: \{value\}."} inside a background haystack of repetitive noise sentences (on S-NIAH-1) or concatenated Paul Graham essays (on S-NIAH-2 and S-NIAH-3).
Here, \texttt{\{key\}} is a random two-word hyphenated string (e.g., \texttt{"wandering-age"}) and \texttt{(\{type\}, \{value\})} is \texttt{("number", 7-digit integer)} on S-NIAH-1/2 and \texttt{("uuid", 36-char UUID)} on S-NIAH-3, formatted with the prompt template:
\begin{quote}\ttfamily
    A special magic \{type\} is hidden within the following text. Make sure to memorize it. I will quiz you about the \{type\} afterwards.\\
    \{haystack\_with\_needle\}\\
    What is the special magic \{type\} for \{key\} mentioned in the provided text? The special magic \{type\} for \{key\} mentioned in the provided text is:
\end{quote}

On multi-value retrieval (MV-NIAH), four distinct $7$-digit integer values $(V_1, V_2, V_3, V_4)$ bound to a single shared key are inserted in chronological write order inside a Paul Graham essay haystack (\texttt{"One of the special magic numbers for \{key\} is: \{$V_m$\}."}) and queried together:
\begin{quote}\ttfamily
    Some special magic numbers are hidden within the following text. Make sure to memorize it. I will quiz you about the numbers afterwards.\\
    \{haystack\_with\_needles\}\\
    What are all the special magic numbers for \{key\} mentioned in the provided text? The special magic numbers for \{key\} mentioned in the provided text are:
\end{quote}

On multi-query retrieval (MQ-NIAH), four distinct keys $K_1, K_2, K_3, K_4$ paired with $7$-digit integer values $V_1, V_2, V_3, V_4$ are embedded at four positions inside a Paul Graham essay haystack and queried jointly:
\begin{quote}\ttfamily
    Some special magic numbers are hidden within the following text. Make sure to memorize it. I will quiz you about the numbers afterwards.\\
    \{haystack\_with\_needles\}\\
    What are all the special magic numbers for \{$K_1$\}, \{$K_2$\}, \{$K_3$\}, and \{$K_4$\} mentioned in the provided text? The special magic numbers for \{$K_1$\}, \{$K_2$\}, \{$K_3$\}, and \{$K_4$\} mentioned in the provided text are:
\end{quote}

On variable tracking (VT), following \citet{Hsieh2024-ji}, each prompt prepends a fixed one-shot demonstration of a $5$-hop variable-binding chain in noise text, followed by a test passage of repetitive noise sentences containing a target chain of $5$ random $5$-letter uppercase variable names $(\texttt{VAR\_1}, \dotsc, \texttt{VAR\_5})$ bound sequentially to a $5$-digit integer $V$ across five hop statements $H_1\text{--}H_5$ at token positions $s_1 < \dotsb < s_5$ (\texttt{"VAR\_1 = $V$"}, \texttt{"VAR\_2 = VAR\_1"}$,\dotsc,\,$\texttt{"VAR\_5 = VAR\_4"}) before the query token $t_{\mathrm{q}}$:
\begin{quote}\ttfamily
    Memorize and track the chain(s) of variable assignment hidden in the following text.\\
    \{one\_shot\_demonstration\}\\
    \{haystack\_with\_variable\_assignments\}\\
    Question: Find all variables that are assigned the value \{$V$\} in the text above. Answer: According to the chain(s) of variable assignment in the text above, 5 variables are assigned the value \{$V$\}, they are:
\end{quote}

\input{tables/tab14_based_suppl}


\xxhdr{Decoding, Scoring, and Depth-Resolved Analysis}
Under greedy decoding, following the official RULER \texttt{string\_match\_all} metric \citep{Hsieh2024-ji}, an example with reference target list $\gY = (y_1, \dotsc, y_M)$ ($M = 1$ on S-NIAH-1/2/3, $M = 4$ on MV-NIAH and MQ-NIAH, and $M = 5$ on VT) and generated continuation $\hat{s}$ receives fractional recall score
\[
    \mathrm{Score}(\hat{s}, \gY) = \frac{1}{M}\sum_{m=1}^{M} \mathbb{I}\!\left[y_m \in \hat{s}\right] \times 100\%.
\]
For the depth-by-length heatmap in \Figref{fig:recall}{e}, each target needle $y_m$ is assigned to one of five $20\%$ relative depth quintiles ($\{0\text{--}20\%, 20\text{--}40\%, 40\text{--}60\%, 60\text{--}80\%, 80\text{--}100\%\}$) according to the character offset of its insertion statement within the haystack passage (excluding the one-shot header on VT and the trailing question suffix), and each cell reports the unweighted mean across the plotted tasks at that context length ($6$ tasks at $1\text{K}\text{--}4\text{K}$ and the $4$ single-needle and variable-tracking tasks at $8\text{K}$).
In \Figref{fig:mech_vt_interp}{a,b}, we average $\mathbb{I}[y_m \in \hat{s}]$ for each target index $m \in \{1, \dotsc, M\}$ separately across the $N = 500$ examples to compute per-write ($V_1\text{--}V_4$) and per-hop ($H_1\text{--}H_5$) recall accuracies.

\Cref{tab:eval_ruler} reports the complete per-task and per-length results across all six RULER tasks at $1.34\text{B}$ parameters, with all models (including our \gdntwo baseline) pretrained from scratch and evaluated under the same pipeline (\S\ref{sec:exp:setup}).
While our re-trained $1.34\text{B}$ \gdntwo baseline closely matches the language modeling perplexities and zero-shot accuracies in \citet{Hatamizadeh2026-cw}, its RULER scores differ in several cells from Table~3 of \citet{Hatamizadeh2026-cw} for three reasons.
First, we evaluate S-NIAH-1/2/3 with \texttt{lm-evaluation-harness} (\texttt{v0.4.11}), which constructs different essay haystacks and prompt lengths than the standalone \texttt{NVIDIA/RULER} scripts (for example, on S-NIAH-3 at $1\text{K},$ our same $1.34\text{B}$ \gdntwo checkpoint scores $77.4\%$ under \texttt{lm-evaluation-harness} vs.\ $84.6\%$ under \texttt{NVIDIA/RULER}).
Second, we include a trailing colon in the answer prefix (\texttt{"is:"}/\texttt{"are:"}) across all tasks to match the needle format inside the haystack.
Third, whereas in-distribution recall at $1\text{K}$ and $2\text{K}$ agrees within $0.8\%$ across both studies ($99.2\%$ to $100.0\%$ on S-NIAH-1/2), zero-shot length extrapolation fluctuates substantially across runs and scales: on S-NIAH-1 at $8\text{K},$ our $406\text{M}$ and $834\text{M}$ \gdntwo checkpoints reach $99.0\%$ and $88.8\%$ while our $1.34\text{B}$ checkpoint drops to $40.8\%,$ whereas on S-NIAH-2 at $8\text{K}$ our $1.34\text{B}$ checkpoint scores higher than \citet{Hatamizadeh2026-cw} ($51.6\%$ vs.\ $39.2\%$).


\input{tables/tab15_longctx_suppl}

\xhdr{Based/JRT Extractive Recall Protocol}

For real-world extractive recall on Based/JRT~\citep{Arora2024-lm,Arora2024-at}, \Cref{tab:eval_based} reports per-task accuracy across all four scales ($12{,}077$ total validation examples across SWDE, FDA, SQuAD, TriviaQA, Natural Questions, and DROP) under the official $2\text{K}$ context-cut protocol, which truncates passages to a $2{,}048$-token window centered around the target answer span following \citet{Yang2025-dr}, alongside the 6-task average under two additional context-truncation protocols: the $1\text{K}$ context-cut protocol ($1{,}024$-token window) and uncut $2\text{K}$ left-truncation (each raw document is left-truncated to fit the $2{,}048$-token window, reserving $48$ tokens for generation).


\input{tables/tab16_longctx_10m_suppl}
\input{tables/tab17_longctx_niah_suppl}

\subsection{Length Extrapolation}
\label{app:eval:longctx}

\xhdr{Language-Modeling Extrapolation Protocol}

We construct three long-context evaluation corpora from held-out FineWeb-Edu validation shards disjoint from both pretraining and cooldown data:
(i) $N_{\mathrm{docs}} = 64$ single contiguous documents truncated to $S = 131{,}072$ target tokens each ($8.39\text{M}$ target tokens, with a single leading BOS token and no internal BOS or EOS resets);
(ii) $N_{\mathrm{docs}} = 64$ packed validation control slices of $S = 131{,}072$ target tokens each, formed by concatenating standard validation documents separated by EOS tokens; and
(iii) $N_{\mathrm{streams}} = 3$ continuous long-document streams of $S = 10{,}485{,}760$ ($10 \times 2^{20}$) target tokens each ($31.46\text{M}$ target tokens, formed from concatenated $\ge\!16\text{K}$ documents with a single leading BOS token and internal EOS tokens stripped so recurrent state accumulation proceeds uninterrupted across $10\text{M}$ tokens).
For each evaluation window $T$ dividing $S,$ each sequence is partitioned into $S / T$ non-overlapping windows with recurrent state or KV cache reset at window boundaries so every $T$ scores the identical target tokens; on the $128\text{K}$ corpora, $\Delta_{\mathrm{tail}}$ pairs per-sequence mean NLL over the final $1{,}024$ target positions $[127\text{K}, 128\text{K})$ between $T = 128\text{K}$ and $T = 4\text{K}.$

%
All $1.34\text{B}$ evaluations run on $8$ TPU7x chips ($16$ TensorCores) without rematerialization or chunked KV prefill: data-parallel RoPE \tf (\Cref{tab:eval_longctx}) fits through $T = 128\text{K},$ whereas unchunked evaluation and prefill in our $16$-way model-sharded harnesses (\Cref{tab:eval_longctx_10m,tab:eval_longctx_niah}) exceed HBM at $T \ge 128\text{K}$ (OOM), while recurrent models carry a fixed $25\text{ MB}$ \texttt{float32} state across fixed-length segments through $T = 10\text{M}.$
\Cref{tab:eval_longctx} reports cross-scale ($166\text{M}\text{--}1.34\text{B}$) zero-shot extrapolation from $4\text{K}$ to $128\text{K}$ on contiguous documents and packed control streams, and \Cref{tab:eval_longctx_10m} reports $1.34\text{B}$ zero-shot and $64\text{K}$-cooldown extrapolation across the $128\text{K}$ documents and $10\text{M}$-token streams.


\input{tables/tab18_abl_suppl}

\xhdr{Long-Context Passkey Retrieval Protocol}

To evaluate passkey retrieval up to $1\text{M}$ tokens (\Cref{tab:eval_longctx_niah}), we construct RULER S-NIAH-1 prompts across $11$ geometric context lengths $T = 2^k \times 1{,}024$ for $k \in \{0, \dotsc, 10\}$ using the S-NIAH-1 template in \S\ref{app:eval:recall}, reserving $128$ tokens for the query suffix and $32$-token greedy continuation.
At each context length, $N = 160$ examples are stratified evenly across $10$ relative needle-depth bins ($16$ examples per $10\%$ depth bin from $0\%$ to $100\%$), which directly populate the depth-by-length heatmaps in \Cref{fig:longctx_niah} and the per-length accuracies in \Cref{tab:eval_longctx_niah}.

\Cref{tab:eval_longctx_niah} also evaluates zero-shot \ours with dynamic query modulation frozen at inference time: setting $\vgamma_t \gets \vgamma^{\mathrm{stat}}$ (using $\vz^q_t \gets \vzero$) breaks co-adaptation with $\vgamma^{\mathrm{stat}}$ by $8\text{K}\text{--}16\text{K},$ while setting $\vgamma_t \gets \bar\vgamma$ ($2\sigmoid(\vz^q_t)$ fixed to its held-out text mean) preserves recall through $16\text{K}$ yet remains below active \ours at $\ge\!32\text{K},$ confirming that the $\ge\!64\text{K}$ passkey gap reflects halved write rank ($r_w = 128$ vs.\ $256$) rather than query-gate drift.


\subsection{Architecture and Gate-Allocation Controls}
\label{app:eval:abl}

\Cref{tab:eval_abl_406m} reports the parameter counts, gate-rank allocations, FineWeb-Edu validation NLL and token perplexity, and full-window ($T = 4{,}096$) WikiText word-level perplexity for the dense dynamic-readout, fixed-FFN gate-compression, and unconstrained wide-FFN controls at the $406\text{M}$ scale.

%% file: tables/tab12_lm_suppl.tex
\begin{table}[!t]
    \centering
    \caption{%
        \textbf{Supplementary language modeling perplexities, zero-shot BoolQ accuracy, and 9-task commonsense average across all four scales.}
        Complementing \Cref{tab:main_lm}, we report FineWeb-Edu validation token perplexity (PPL), full-window ($T = 4{,}096$) WikiText word-level perplexity, zero-shot BoolQ accuracy, and the 9-task zero-shot commonsense average including BoolQ (``Avg.\ (9)'').%
    }
    \label{tab:eval_lm_supp}
    \resizebox{0.65\textwidth}{!}{%
        \setlength{\tabcolsep}{6.5pt}%
        \renewcommand{\arraystretch}{1.08}%
        \begin{tabular}{@{}c@{\hspace{4pt}}|l|cc|cc@{}}
        \toprule
            \multicolumn{1}{@{}c@{\hspace{5pt}}}{} & & \multicolumn{2}{c|}{{Perplexity}} & \multicolumn{2}{c@{}}{{Zero-Shot Accuracy (\%)}} \\
            \cmidrule(lr){3-4} \cmidrule(l){5-6}
            \multicolumn{1}{@{}c@{\hspace{5pt}}}{} & {Model} & {FW-Edu} & {WikiText} & {BoolQ} & {Avg.\ (9)} \\
            \multicolumn{1}{@{}c@{\hspace{5pt}}}{} & & {Token PPL $\downarrow$} & {($4\text{K}$) PPL $\downarrow$} & {acc $\uparrow$} & {w/ BoolQ $\uparrow$} \\
        \midrule
            \multirowrotate{5}{90}{\footnotesize 166M}
            & \tf & 11.67 & 30.15 & \tbf{59.11} & 42.04 \\
            & \gdntwo & 11.55 & 29.32 & 53.46 & 41.94 \\
            & \ours (no $\vgamma$) & 11.38 & 28.87 & \underline{58.99} & \tbf{43.49} \\
            & \ours (static $\vgamma$) & \tbf{11.33} & \tbf{28.31} & 57.13 & \underline{42.74} \\
            & \tbf{\ours (ours)} & \tbf{11.33} & \underline{28.81} & 53.64 & 42.28 \\
        \midrule
            \multirowrotate{5}{90}{\footnotesize 406M}
            & \tf & 9.42 & 21.94 & 58.32 & 47.01 \\
            & \gdntwo & 9.35 & 21.76 & \tbf{60.98} & 47.40 \\
            & \ours (no $\vgamma$) & 9.24 & \tbf{21.16} & 57.34 & \underline{47.50} \\
            & \ours (static $\vgamma$) & \underline{9.20} & 21.24 & \underline{60.64} & \tbf{47.75} \\
            & \tbf{\ours (ours)} & \tbf{9.18} & \underline{21.22} & 53.61 & 46.69 \\
        \midrule
            \multirowrotate{5}{90}{\footnotesize 834M}
            & \tf & 8.32 & 18.44 & \tbf{62.78} & 50.08 \\
            & \gdntwo & 8.22 & 18.26 & \underline{59.60} & 50.47 \\
            & \ours (no $\vgamma$) & 8.07 & \underline{17.57} & 57.43 & \tbf{51.30} \\
            & \ours (static $\vgamma$) & \underline{8.06} & \tbf{17.53} & 54.31 & 50.60 \\
            & \tbf{\ours (ours)} & \tbf{8.05} & 17.67 & 55.14 & \underline{50.90} \\
        \midrule
            \multirowrotate{5}{90}{\footnotesize 1.34B}
            & \tf & 7.64 & 16.28 & \tbf{63.36} & 53.31 \\
            & \gdntwo & 7.52 & 15.96 & 53.27 & 52.55 \\
            & \ours (no $\vgamma$) & 7.39 & 15.69 & 57.86 & \tbf{54.24} \\
            & \ours (static $\vgamma$) & \underline{7.38} & \tbf{15.59} & \underline{59.14} & \underline{54.04} \\
            & \tbf{\ours (ours)} & \tbf{7.37} & \underline{15.63} & 53.79 & 53.84 \\
            \bottomrule
        \end{tabular}%
    }
\end{table}

%% file: tables/tab13_niah_suppl.tex
\begin{table}[!t]
    \centering
    \caption{%
        \textbf{Complete synthetic associative recall accuracy on the six RULER benchmarks at $\boldsymbol{1.34\text{B}}$ parameters.}
        (Top) Single-needle retrieval on S-NIAH-1/2/3, and their 3-task mean.
        (Bottom) Multi-hop VT, multi-value (MV-NIAH) and multi-query (MQ-NIAH) retrieval, and suite-wide averages across multi-needle retrieval (MV+MQ), all six tasks at $1\text{K}\text{--}4\text{K},$ and all evaluated lengths ($1\text{K}\text{--}8\text{K}$).%
    }
    \label{tab:eval_ruler}
    \resizebox{\textwidth}{!}{%
        \setlength{\tabcolsep}{2pt}%
        \renewcommand{\arraystretch}{1.10}%
        \begin{tabular}{@{}l|cccc|cccc|cccc|cccccc@{}}
            \toprule
            & \multicolumn{4}{c|}{{S-NIAH-1}} & \multicolumn{4}{c|}{{S-NIAH-2}} & \multicolumn{4}{c|}{{S-NIAH-3}} & \multicolumn{6}{c}{{S-NIAH-1/2/3 Avg.}} \\
            \cmidrule(lr){2-5} \cmidrule(lr){6-9} \cmidrule(lr){10-13} \cmidrule(lr){14-19}
            {Model ($1.34\text{B}$)} & {1K} & {2K} & {4K} & {8K} & {1K} & {2K} & {4K} & {8K} & {1K} & {2K} & {4K} & {8K} & {1K} & {2K} & {4K} & {8K} & {1--4K} & {1--8K} \\
            \midrule
            \tf & \tbf{100.0} & \tbf{100.0} & \underline{99.6} & 0.0 & \underline{99.8} & \tbf{100.0} & \tbf{100.0} & 0.0 & \tbf{98.0} & \tbf{96.4} & \tbf{80.8} & 0.0 & \tbf{99.3} & \tbf{98.8} & \tbf{93.5} & 0.0 & \tbf{97.2} & 72.9 \\
            \gdntwo & \underline{99.8} & \underline{99.2} & 86.2 & 40.8 & \tbf{100.0} & 99.6 & 83.6 & \tbf{51.6} & 77.4 & 57.6 & 20.6 & 9.8 & 92.4 & 85.5 & 63.5 & 34.1 & 80.4 & 68.9 \\
            \ours (no $\vgamma$) & \tbf{100.0} & \tbf{100.0} & \underline{99.6} & \underline{99.8} & 98.2 & \underline{99.8} & 93.8 & 39.2 & 84.0 & 80.4 & 41.6 & \underline{10.4} & 94.1 & 93.4 & 78.3 & \underline{49.8} & 88.6 & \underline{78.9} \\
            \ours (static $\vgamma$) & \tbf{100.0} & \tbf{100.0} & \tbf{100.0} & 99.6 & \tbf{100.0} & \tbf{100.0} & \underline{99.0} & \underline{41.0} & \underline{96.2} & \underline{80.6} & \underline{43.8} & \tbf{13.6} & \underline{98.7} & \underline{93.5} & \underline{80.9} & \tbf{51.4} & \underline{91.1} & \tbf{81.2} \\
            \tbf{\ours (ours)} & \tbf{100.0} & \tbf{100.0} & \tbf{100.0} & \tbf{100.0} & \tbf{100.0} & \underline{99.8} & 96.2 & 32.0 & 88.2 & 67.4 & 37.8 & 7.0 & 96.1 & 89.1 & 78.0 & 46.3 & 87.7 & 77.4 \\
            \midrule
            \midrule
            & \multicolumn{5}{c|}{{Variable Tracking (VT)}} & \multicolumn{4}{c|}{{MV-NIAH}} & \multicolumn{4}{c|}{{MQ-NIAH}} & \multicolumn{5}{c}{{RULER Suite Avg.}} \\
            \cmidrule(lr){2-6} \cmidrule(lr){7-10} \cmidrule(lr){11-14} \cmidrule(lr){15-19}
            {Model ($1.34\text{B}$)} & {1K} & {2K} & {4K} & {8K} & \multicolumn{1}{c|}{{Avg.}} & {1K} & {2K} & {4K} & \multicolumn{1}{c|}{{Avg.}} & {1K} & {2K} & {4K} & \multicolumn{1}{c|}{{Avg.}} & \multicolumn{2}{c}{{MV+MQ}} & \multicolumn{2}{c}{{1--4K}} & {1--8K} \\
            \midrule
            \tf & 9.9 & 3.6 & \underline{21.5} & 0.0 & \multicolumn{1}{c|}{8.8} & \underline{80.9} & 57.8 & 35.4 & \multicolumn{1}{c|}{\underline{58.0}} & \tbf{88.9} & \underline{72.0} & 31.2 & \multicolumn{1}{c|}{\underline{64.0}} & \multicolumn{2}{c}{\underline{61.0}} & \multicolumn{2}{c}{\underline{70.9}} & 58.0 \\
            \gdntwo & \underline{47.7} & \underline{26.9} & 10.5 & 0.9 & \multicolumn{1}{c|}{21.5} & 65.1 & \underline{58.2} & \underline{38.0} & \multicolumn{1}{c|}{53.8} & 64.2 & 50.5 & 28.0 & \multicolumn{1}{c|}{47.6} & \multicolumn{2}{c}{50.7} & \multicolumn{2}{c}{61.8} & 55.3 \\
            \ours (no $\vgamma$) & 44.4 & 15.5 & 18.1 & 23.8 & \multicolumn{1}{c|}{\underline{25.5}} & 59.5 & 46.0 & 28.6 & \multicolumn{1}{c|}{44.7} & 74.9 & 56.0 & 26.9 & \multicolumn{1}{c|}{52.6} & \multicolumn{2}{c}{48.7} & \multicolumn{2}{c}{64.9} & 60.9 \\
            \ours (static $\vgamma$) & 19.0 & 16.9 & 20.4 & \underline{23.9} & \multicolumn{1}{c|}{20.1} & 57.4 & 44.2 & 33.0 & \multicolumn{1}{c|}{44.8} & 85.2 & 54.7 & \underline{32.8} & \multicolumn{1}{c|}{57.6} & \multicolumn{2}{c}{51.2} & \multicolumn{2}{c}{65.7} & \underline{61.9} \\
            \tbf{\ours (ours)} & \tbf{60.0} & \tbf{27.6} & \tbf{24.0} & \tbf{25.6} & \multicolumn{1}{c|}{\tbf{34.3}} & \tbf{86.2} & \tbf{75.4} & \tbf{45.0} & \multicolumn{1}{c|}{\tbf{68.9}} & \underline{88.1} & \tbf{76.5} & \tbf{42.6} & \multicolumn{1}{c|}{\tbf{69.1}} & \multicolumn{2}{c}{\tbf{69.0}} & \multicolumn{2}{c}{\tbf{73.0}} & \tbf{67.2} \\
            \bottomrule
        \end{tabular}%
    }
\end{table}

%% file: tables/tab14_based_suppl.tex
\begin{table}[!t]
    \centering
    \caption{%
        \textbf{Complete real-world extractive recall accuracy on the six Based/JRT benchmarks across all four model scales.}
        Per-task accuracies are reported under the official $2\text{K}$ context-cut JRT protocol ($12{,}077$ validation examples across structured attribute extraction and passage question answering; the $1.34\text{B}$ $2\text{K}$ context-cut average is reported in \Cref{tab:abl}), alongside the 6-task average under $1\text{K}$ context-cut truncation and un-cut $2\text{K}$ left-truncation.%
    }
    \label{tab:eval_based}
    \resizebox{0.88\textwidth}{!}{%
        \setlength{\tabcolsep}{5.2pt}%
        \renewcommand{\arraystretch}{1.08}%
        \begin{tabular}{@{}c@{\hspace{4pt}}|l|cc|cccc|ccc@{}}
            \toprule
            \multicolumn{1}{@{}c@{\hspace{5pt}}}{} & & \multicolumn{2}{c|}{{Extraction (\%) $\uparrow$}} & \multicolumn{4}{c|}{{Passage Q/A ($2\text{K}$ Cut; \%) $\uparrow$}} & \multicolumn{3}{c@{}}{{6-Task Avg. (\%) $\uparrow$}} \\
            \cmidrule(lr){3-4} \cmidrule(lr){5-8} \cmidrule(l){9-11}
            \multicolumn{1}{@{}c@{\hspace{5pt}}}{} & {Model} & {SWDE} & {FDA} & {SQuAD} & {TQA} & {NQ} & {DROP} & {$2\text{K}$ Cut} & {$1\text{K}$ Cut} & {$2\text{K}$ Trunc.} \\
            \midrule
            \multirowrotate{5}{90}{\footnotesize 166M}
            & \tf & \tbf{29.80} & \tbf{39.51} & \tbf{30.47} & 44.91 & \tbf{17.83} & 16.20 & \tbf{29.79} & \tbf{31.41} & \tbf{23.33} \\
            & \gdntwo & \underline{12.37} & \underline{9.90} & 25.56 & 44.25 & 12.13 & 17.01 & 20.20 & 21.60 & 19.41 \\
            & \ours (no $\vgamma$) & 12.00 & 8.90 & 27.38 & \tbf{46.80} & \underline{13.97} & \tbf{18.97} & \underline{21.34} & \underline{22.29} & \underline{20.41} \\
            & \ours (static $\vgamma$) & 10.50 & 7.63 & 25.26 & 45.85 & 12.48 & 17.87 & 19.93 & 22.01 & 18.96 \\
            & \tbf{\ours (ours)} & 10.97 & 5.27 & \underline{27.68} & \underline{46.27} & 13.40 & \underline{18.78} & 20.39 & 21.78 & 19.51 \\
            \midrule
            \multirowrotate{5}{90}{\footnotesize 406M}
            & \tf & \tbf{41.05} & \tbf{54.31} & \tbf{38.36} & \tbf{54.03} & \tbf{19.77} & 19.41 & \tbf{37.82} & \tbf{40.90} & \tbf{29.81} \\
            & \gdntwo & 18.18 & 11.35 & \underline{32.75} & 52.25 & 17.10 & \tbf{21.42} & 25.51 & 27.89 & 24.05 \\
            & \ours (no $\vgamma$) & 17.34 & 14.35 & 31.81 & \underline{53.20} & \underline{18.12} & \underline{21.32} & \underline{26.02} & 28.49 & 24.52 \\
            & \ours (static $\vgamma$) & 16.40 & \underline{15.71} & 31.41 & 52.49 & 17.52 & 20.03 & 25.59 & \underline{28.80} & 24.08 \\
            & \tbf{\ours (ours)} & \underline{18.37} & 14.99 & 30.94 & 52.13 & 17.71 & 21.13 & 25.88 & 28.79 & \underline{24.65} \\
            \midrule
            \multirowrotate{5}{90}{\footnotesize 834M}
            & \tf & \tbf{42.08} & \tbf{58.67} & \tbf{39.47} & \tbf{60.07} & \tbf{24.58} & \tbf{22.62} & \tbf{41.25} & \tbf{43.40} & \tbf{32.03} \\
            & \gdntwo & 22.59 & 15.80 & 33.32 & 58.06 & 21.41 & 21.80 & 28.83 & 31.55 & 26.81 \\
            & \ours (no $\vgamma$) & 23.43 & 16.44 & 33.66 & \underline{58.65} & 21.35 & 21.95 & 29.25 & 32.56 & 27.15 \\
            & \ours (static $\vgamma$) & \underline{26.15} & \underline{22.62} & \underline{34.97} & 57.70 & \underline{21.57} & 19.50 & \underline{30.42} & \underline{33.20} & \underline{28.43} \\
            & \tbf{\ours (ours)} & 23.62 & 14.71 & 34.93 & 58.29 & 20.78 & \underline{22.42} & 29.13 & 32.57 & 27.09 \\
            \midrule
            \multirowrotate{5}{90}{\footnotesize 1.34B}
            & \tf & \tbf{49.30} & \tbf{70.12} & \tbf{44.68} & \underline{61.43} & \tbf{26.51} & \underline{23.48} & \tbf{45.92} & \tbf{47.00} & \tbf{35.01} \\
            & \gdntwo & 26.71 & 24.80 & 36.08 & 61.32 & 22.36 & 21.23 & 32.08 & 35.44 & 29.93 \\
            & \ours (no $\vgamma$) & 29.90 & 25.16 & 36.58 & \tbf{61.55} & 22.33 & 21.66 & 32.86 & 36.20 & 30.70 \\
            & \ours (static $\vgamma$) & 29.80 & \underline{26.16} & 36.61 & 60.72 & 22.05 & 22.76 & 33.02 & 36.50 & 30.10 \\
            & \tbf{\ours (ours)} & \underline{35.15} & 25.34 & \underline{37.92} & \underline{61.43} & \underline{24.20} & \tbf{23.57} & \underline{34.60} & \underline{37.53} & \underline{31.97} \\
            \bottomrule
        \end{tabular}%
    }
\end{table}

%% file: tables/tab15_longctx_suppl.tex
\begin{table}[!t]
    \centering
    \caption{%
        \textbf{Zero-shot length extrapolation from $\boldsymbol{4\text{K}}$ to $\boldsymbol{128\text{K}}$ on contiguous FineWeb-Edu long documents and packed control streams.}
        For each evaluation window $T \in \{4\text{K}, 8\text{K}, 16\text{K}, 32\text{K}, 64\text{K}, 128\text{K}\},$ we report aggregate token perplexity alongside the paired tail NLL differential $\Delta_{\mathrm{tail}} \defas \mathrm{NLL}_{[127\mathrm{K},128\mathrm{K})}(T = 128\text{K}) - \mathrm{NLL}_{[127\mathrm{K},128\mathrm{K})}(T = 4\text{K})$ (mean $\pm$ paired SE in nats across $N_{\mathrm{docs}}$ sequences).%
    }
    \label{tab:eval_longctx}
    \resizebox{0.71\textwidth}{!}{%
        \setlength{\tabcolsep}{4.5pt}%
        \renewcommand{\arraystretch}{1.08}%
        \begin{tabular}{@{}c@{\hspace{4pt}}|l|cccccc|c@{}}
        \toprule
            \multicolumn{1}{@{}c@{\hspace{5pt}}}{} & & \multicolumn{6}{c|}{{PPL $\downarrow$}} & {$\Delta_{\mathrm{tail}}$ (nats) $\downarrow$} \\
            \cmidrule(lr){3-8}
            \multicolumn{1}{@{}c@{\hspace{5pt}}}{} & {Model} & {$4\text{K}$} & {$8\text{K}$} & {$16\text{K}$} & {$32\text{K}$} & {$64\text{K}$} & {$128\text{K}$} & {} \\
        \midrule
            \multicolumn{9}{c}{Contiguous FineWeb-Edu Long Documents} \\
        \midrule
            \multirowrotate{5}{90}{\footnotesize 166M}
            & \tf & 11.55 & 50.97 & 199.85 & 473.31 & 728.84 & 895.74 & $+$4.5349$\,\pm\,$0.0798 \\
            & \gdntwo & 11.69 & 11.44 & 11.32 & 11.29 & 11.28 & 11.28 & \tbf{$+$0.0051$\,\pm\,$0.0018} \\
            & \ours (no $\vgamma$) & 11.49 & 11.25 & 11.14 & 11.11 & 11.10 & 11.10 & \tbf{$+$0.0051$\,\pm\,$0.0015} \\
            & \ours (static $\vgamma$) & \tbf{11.39} & \tbf{11.14} & \tbf{11.03} & \tbf{11.00} & \tbf{10.99} & \tbf{10.99} & \underline{$+$0.0076$\,\pm\,$0.0020} \\
            & \tbf{\ours (ours)} & \underline{11.40} & \underline{11.15} & \underline{11.04} & \underline{11.02} & \underline{11.03} & \underline{11.05} & $+$0.0113$\,\pm\,$0.0027 \\
        \midrule
            \multirowrotate{5}{90}{\footnotesize 406M}
            & \tf & 9.22 & 40.93 & 163.44 & 393.91 & 591.18 & 663.24 & $+$4.3629$\,\pm\,$0.0771 \\
            & \gdntwo & 9.26 & 9.05 & 8.95 & 8.91 & 8.90 & 8.90 & \tbf{$+$0.0022$\,\pm\,$0.0014} \\
            & \ours (no $\vgamma$) & 9.17 & 8.96 & 8.87 & 8.86 & 8.88 & 8.91 & $+$0.0154$\,\pm\,$0.0036 \\
            & \ours (static $\vgamma$) & \underline{9.11} & \underline{8.90} & \underline{8.81} & \underline{8.79} & \underline{8.79} & \underline{8.80} & $+$0.0075$\,\pm\,$0.0027 \\
            & \tbf{\ours (ours)} & \tbf{9.09} & \tbf{8.88} & \tbf{8.78} & \tbf{8.74} & \tbf{8.73} & \tbf{8.73} & \underline{$+$0.0026$\,\pm\,$0.0015} \\
        \midrule
            \multirowrotate{5}{90}{\footnotesize 834M}
            & \tf & 8.09 & 35.59 & 175.10 & 476.67 & 906.21 & 1253.03 & $+$5.3833$\,\pm\,$0.0666 \\
            & \gdntwo & 8.09 & 7.91 & 7.83 & 7.81 & 7.81 & 7.81 & \underline{$+$0.0064$\,\pm\,$0.0020} \\
            & \ours (no $\vgamma$) & 7.93 & 7.75 & 7.68 & 7.66 & 7.66 & 7.68 & $+$0.0100$\,\pm\,$0.0027 \\
            & \ours (static $\vgamma$) & \underline{7.92} & \underline{7.74} & \underline{7.66} & \underline{7.65} & \underline{7.65} & \underline{7.67} & $+$0.0112$\,\pm\,$0.0021 \\
            & \tbf{\ours (ours)} & \tbf{7.90} & \tbf{7.73} & \tbf{7.64} & \tbf{7.61} & \tbf{7.60} & \tbf{7.60} & \tbf{$+$0.0037$\,\pm\,$0.0014} \\
        \midrule
            \multirowrotate{5}{90}{\footnotesize 1.34B}
            & \tf & 7.40 & 35.26 & 214.64 & 532.59 & 775.44 & 978.62 & $+$5.2102$\,\pm\,$0.0753 \\
            & \gdntwo & 7.36 & 7.20 & 7.12 & 7.10 & 7.09 & 7.10 & $+$0.0069$\,\pm\,$0.0024 \\
            & \ours (no $\vgamma$) & 7.23 & 7.07 & 7.00 & 6.97 & \underline{6.97} & \underline{6.97} & \tbf{$+$0.0043$\,\pm\,$0.0018} \\
            & \ours (static $\vgamma$) & \underline{7.21} & \underline{7.05} & \underline{6.98} & \underline{6.96} & 6.98 & 7.01 & $+$0.0171$\,\pm\,$0.0035 \\
            & \tbf{\ours (ours)} & \tbf{7.19} & \tbf{7.03} & \tbf{6.95} & \tbf{6.92} & \tbf{6.92} & \tbf{6.92} & \underline{$+$0.0047$\,\pm\,$0.0020} \\
        \midrule
            \multicolumn{9}{c}{Packed Validation Control Streams} \\
        \midrule
            \multirowrotate{5}{90}{\footnotesize 166M}
            & \tf & 11.59 & 54.11 & 241.07 & 525.78 & 827.66 & 976.74 & $+$4.5664$\,\pm\,$0.0707 \\
            & \gdntwo & 11.48 & 11.35 & 11.29 & 11.27 & 11.25 & 11.24 & \tbf{$+$0.0002$\,\pm\,$0.0004} \\
            & \ours (no $\vgamma$) & 11.34 & 11.22 & 11.16 & 11.14 & 11.12 & 11.12 & \underline{$+$0.0006$\,\pm\,$0.0005} \\
            & \ours (static $\vgamma$) & \underline{11.29} & \tbf{11.16} & \tbf{11.10} & \underline{11.08} & \underline{11.07} & \underline{11.06} & $+$0.0012$\,\pm\,$0.0006 \\
            & \tbf{\ours (ours)} & \tbf{11.28} & \tbf{11.16} & \tbf{11.10} & \tbf{11.07} & \tbf{11.06} & \tbf{11.05} & $+$0.0010$\,\pm\,$0.0004 \\
        \midrule
            \multirowrotate{5}{90}{\footnotesize 406M}
            & \tf & 9.37 & 47.88 & 232.23 & 547.77 & 785.32 & 847.72 & $+$4.5869$\,\pm\,$0.0637 \\
            & \gdntwo & 9.31 & 9.21 & 9.15 & 9.13 & 9.12 & 9.11 & \tbf{$+$0.0002$\,\pm\,$0.0004} \\
            & \ours (no $\vgamma$) & 9.19 & 9.09 & 9.04 & 9.02 & 9.02 & 9.02 & $+$0.0027$\,\pm\,$0.0009 \\
            & \ours (static $\vgamma$) & \underline{9.16} & \underline{9.05} & \underline{9.00} & \underline{8.98} & \underline{8.97} & \underline{8.97} & \underline{$+$0.0012$\,\pm\,$0.0007} \\
            & \tbf{\ours (ours)} & \tbf{9.14} & \tbf{9.04} & \tbf{8.98} & \tbf{8.96} & \tbf{8.95} & \tbf{8.94} & \tbf{$+$0.0002$\,\pm\,$0.0005} \\
        \midrule
            \multirowrotate{5}{90}{\footnotesize 834M}
            & \tf & 8.27 & 35.88 & 226.16 & 565.41 & 1071.82 & 1465.40 & $+$5.4523$\,\pm\,$0.0569 \\
            & \gdntwo & 8.17 & 8.08 & 8.03 & 8.01 & 8.00 & 8.00 & \underline{$+$0.0014$\,\pm\,$0.0009} \\
            & \ours (no $\vgamma$) & 8.02 & \underline{7.92} & 7.88 & \underline{7.86} & \underline{7.85} & 7.85 & \underline{$+$0.0014$\,\pm\,$0.0009} \\
            & \ours (static $\vgamma$) & \tbf{8.01} & \underline{7.92} & \tbf{7.87} & \underline{7.86} & \underline{7.85} & \underline{7.84} & $+$0.0017$\,\pm\,$0.0008 \\
            & \tbf{\ours (ours)} & \tbf{8.01} & \tbf{7.91} & \tbf{7.87} & \tbf{7.85} & \tbf{7.84} & \tbf{7.83} & \tbf{$+$0.0006$\,\pm\,$0.0006} \\
        \midrule
            \multirowrotate{5}{90}{\footnotesize 1.34B}
            & \tf & 7.59 & 37.67 & 245.28 & 609.58 & 835.47 & 993.22 & $+$5.1502$\,\pm\,$0.0578 \\
            & \gdntwo & 7.47 & 7.38 & 7.34 & 7.32 & 7.31 & 7.31 & \underline{$+$0.0014$\,\pm\,$0.0009} \\
            & \ours (no $\vgamma$) & 7.34 & 7.25 & 7.21 & \underline{7.19} & \underline{7.18} & \underline{7.18} & \tbf{$+$0.0013$\,\pm\,$0.0007} \\
            & \ours (static $\vgamma$) & \underline{7.33} & \underline{7.24} & \underline{7.20} & \underline{7.19} & 7.19 & 7.19 & $+$0.0048$\,\pm\,$0.0011 \\
            & \tbf{\ours (ours)} & \tbf{7.31} & \tbf{7.22} & \tbf{7.18} & \tbf{7.16} & \tbf{7.15} & \tbf{7.15} & $+$0.0018$\,\pm\,$0.0005 \\
        \bottomrule
        \end{tabular}%
    }
\end{table}

%% file: tables/tab16_longctx_10m_suppl.tex
\begin{table}[!t]
    \centering
    \caption{%
        \textbf{Zero-shot and post-cooldown language-modeling extrapolation up to $\boldsymbol{10\text{M}}$-token streams at $\boldsymbol{1.34\text{B}}$ parameters.}
        Complementing \Cref{fig:longctx,tab:eval_longctx}, we evaluate $1.34\text{B}$ checkpoints both zero-shot ($T_0 = 4\text{K}$) and after the $300$-step $64\text{K}$ continual cooldown ($T_0 = 64\text{K},$ $0.315\text{B}$ tokens) on (top) $64$ contiguous $128\text{K}$-token FineWeb-Edu documents and (bottom) three continuous $10\text{M}$-token streams \emph{without} EOS resets; \tf (YaRN) applies Dynamic YaRN for $T > T_0$ ($\theta = 5\times 10^5$ during cooldown), and unchunked evaluation exceeds HBM (OOM) at $T \ge 128\text{K}.$%
    }
    \label{tab:eval_longctx_10m}
    \resizebox{0.72\textwidth}{!}{%
        \setlength{\tabcolsep}{4.5pt}%
        \renewcommand{\arraystretch}{1.08}%
        \vbox{\offinterlineskip\lineskip=\doublerulesep
            \hbox{\begin{tabular*}{390pt}{@{}w{c}{11pt}@{\hspace{0pt}}|w{l}{95pt}|@{\hspace{4.5pt}\extracolsep{\fill}}cccccc|c@{\hspace{4.5pt}}}
            \toprule
                \multicolumn{1}{@{}w{c}{11pt}@{\hspace{0.4pt}}}{} & & \multicolumn{6}{c|}{{Contiguous $128\text{K}$ Documents PPL $\downarrow$}} & {$\Delta_{\mathrm{tail}}$ (nats) $\downarrow$} \\
                \cmidrule(lr){3-8}
                \multicolumn{1}{@{}w{c}{11pt}@{\hspace{0.4pt}}}{} & {Model ($1.34\text{B}$)} & {$4\text{K}$} & {$8\text{K}$} & {$16\text{K}$} & {$32\text{K}$} & {$64\text{K}$} & {$128\text{K}$} & {} \\
            \midrule
                \multirowrotate{5}{90}{\footnotesize Zero-Shot}
                & \tf (YaRN) & 7.40 & 7.26 & 7.36 & 7.77 & 8.74 & OOM & -- \\
                & \gdntwo & 7.36 & 7.20 & 7.12 & 7.10 & 7.09 & 7.10 & $+$0.0069$\,\pm\,$0.0024 \\
                & \ours (no $\vgamma$) & 7.23 & 7.07 & 7.00 & 6.97 & \underline{6.97} & \underline{6.97} & \tbf{$+$0.0043$\,\pm\,$0.0018} \\
                & \ours (static $\vgamma$) & \underline{7.21} & \underline{7.05} & \underline{6.98} & \underline{6.96} & 6.98 & 7.01 & $+$0.0171$\,\pm\,$0.0035 \\
                & \tbf{\ours (ours)} & \tbf{7.19} & \tbf{7.03} & \tbf{6.95} & \tbf{6.92} & \tbf{6.92} & \tbf{6.92} & \underline{$+$0.0047$\,\pm\,$0.0020} \\
                \midrule
                \multirowrotate{5}{90}{\footnotesize 64K Cooldown}
                & \tf (YaRN) & 7.45 & 7.20 & 7.02 & \underline{6.87} & \tbf{6.77} & OOM & -- \\
                & \gdntwo & 7.31 & 7.14 & 7.04 & 6.99 & 6.96 & 6.95 & $-$0.0115$\,\pm\,$0.0025 \\
                & \ours (no $\vgamma$) & 7.21 & 7.04 & 6.95 & 6.90 & 6.87 & 6.86 & $-$0.0119$\,\pm\,$0.0030 \\
                & \ours (static $\vgamma$) & \underline{7.19} & \underline{7.02} & \underline{6.93} & \underline{6.87} & 6.85 & \underline{6.84} & \underline{$-$0.0122$\,\pm\,$0.0027} \\
                & \tbf{\ours (ours)} & \tbf{7.15} & \tbf{6.98} & \tbf{6.88} & \tbf{6.83} & \underline{6.80} & \tbf{6.79} & \tbf{$-$0.0124$\,\pm\,$0.0029} \\
                \specialrule{\lightrulewidth}{\aboverulesep}{0pt}
            \end{tabular*}}
            \hbox{\begin{tabular*}{390pt}{@{}w{c}{11pt}@{\hspace{0pt}}|w{l}{95pt}|@{\hspace{6pt}\extracolsep{\fill}}*{7}{w{c}{28pt}}@{\hspace{6pt}}}
                \specialrule{\lightrulewidth}{0pt}{\belowrulesep}
                \multicolumn{1}{@{}w{c}{11pt}@{\hspace{0.4pt}}}{} & & \multicolumn{7}{c@{\hspace{6pt}}}{{Continuous $10\text{M}$-Token Streams PPL $\downarrow$}} \\
                \cmidrule(l){3-9}
                \multicolumn{1}{@{}w{c}{11pt}@{\hspace{0.4pt}}}{} & {Model ($1.34\text{B}$)} & {$4\text{K}$} & {$16\text{K}$} & {$64\text{K}$} & {$256\text{K}$} & {$1\text{M}$} & {$5\text{M}$} & {$10\text{M}$} \\
                \midrule
                \multirowrotate{5}{90}{\footnotesize Zero-Shot}
                & \tf (YaRN) & 8.69 & 8.73 & 10.51 & OOM & OOM & OOM & OOM \\
                & \gdntwo & 8.60 & 8.35 & 8.31 & 8.31 & 8.31 & 8.31 & 8.31 \\
                & \ours (no $\vgamma$) & 8.45 & 8.19 & \underline{8.16} & \underline{8.16} & \underline{8.16} & \underline{8.16} & \underline{8.16} \\
                & \ours (static $\vgamma$) & \underline{8.43} & \underline{8.17} & 8.17 & 8.21 & 8.23 & 8.23 & 8.23 \\
                & \tbf{\ours (ours)} & \tbf{8.41} & \tbf{8.15} & \tbf{8.11} & \tbf{8.11} & \tbf{8.11} & \tbf{8.11} & \tbf{8.11} \\
                \midrule
                \multirowrotate{5}{90}{\footnotesize 64K Cooldown}
                & \tf (YaRN) & 8.74 & 8.29 & 8.04 & OOM & OOM & OOM & OOM \\
                & \gdntwo & 8.54 & 8.25 & 8.16 & 8.13 & 8.13 & 8.13 & 8.13 \\
                & \ours (no $\vgamma$) & 8.41 & 8.13 & 8.04 & 8.02 & 8.01 & 8.01 & 8.01 \\
                & \ours (static $\vgamma$) & \underline{8.38} & \underline{8.09} & \underline{8.00} & \underline{7.98} & \underline{7.98} & \underline{7.98} & \underline{7.98} \\
                & \tbf{\ours (ours)} & \tbf{8.36} & \tbf{8.07} & \tbf{7.98} & \tbf{7.96} & \tbf{7.95} & \tbf{7.95} & \tbf{7.95} \\
                \bottomrule
            \end{tabular*}}%
        }%
    }
\end{table}

%% file: tables/tab17_longctx_niah_suppl.tex
\begin{table}[!t]
    \centering
    \caption{%
        \textbf{Zero-shot and post-cooldown RULER S-NIAH-1 passkey retrieval from $\boldsymbol{1\text{K}}$ to $\boldsymbol{1\text{M}}$ tokens at $\boldsymbol{1.34\text{B}}$ parameters.}
        Complementing the depth-by-length heatmaps in \Cref{fig:longctx_niah}, each context length $T \in \{1\text{K}, \dotsc, 1\text{M}\}$ evaluates $N = 160$ examples stratified uniformly across $10$ needle-depth bins under greedy decoding, both zero-shot ($4\text{K}$) and after the $64\text{K}$ continual cooldown ($0.315\text{B}$ FineWeb-Edu tokens); unchunked \tf (YaRN) prefill exceeds HBM (OOM) at $T \ge 128\text{K}.$
        Under Zero-Shot, the bottom two rows freeze dynamic readout in \ours at inference time to $\vgamma^{\mathrm{stat}}$ ($\vz^q_t \gets \vzero$) or its held-out text mean $\bar\vgamma = \E[\vgamma_t].$%
    }
    \label{tab:eval_longctx_niah}
    \resizebox{\textwidth}{!}{%
        \setlength{\tabcolsep}{3.8pt}%
        \renewcommand{\arraystretch}{1.08}%
        \begin{tabular}{@{}c@{\hspace{4pt}}|l|ccccccccccc|cc@{}}
        \toprule
            \multicolumn{1}{@{}c@{\hspace{5pt}}}{} & & \multicolumn{11}{c|}{{RULER S-NIAH-1 Accuracy (\%) $\uparrow$}} & \multicolumn{2}{c@{}}{{Avg.\ (\%) $\uparrow$}} \\
            \cmidrule(lr){3-13} \cmidrule(l){14-15}
            \multicolumn{1}{@{}c@{\hspace{5pt}}}{} & {Model ($1.34\text{B}$)} & {$1\text{K}$} & {$2\text{K}$} & {$4\text{K}$} & {$8\text{K}$} & {$16\text{K}$} & {$32\text{K}$} & {$64\text{K}$} & {$128\text{K}$} & {$256\text{K}$} & {$512\text{K}$} & {$1\text{M}$} & {$1\text{--}32\text{K}$} & {$1\text{K--}1\text{M}$} \\
        \midrule
            \multirowrotate{7}{90}{\footnotesize Zero-Shot}
            & \tf (YaRN) & \tbf{100.0} & \tbf{100.0} & \underline{90.0} & 90.6 & 78.8 & 68.1 & 45.6 & OOM & OOM & OOM & OOM & 87.9 & -- \\
            & \gdntwo & \tbf{100.0} & \underline{98.8} & 80.0 & 40.0 & 13.8 & 11.9 & 34.4 & 23.8 & 15.0 & 10.0 & 10.0 & 57.4 & 39.8 \\
            & \ours (no $\vgamma$) & \tbf{100.0} & \tbf{100.0} & \tbf{100.0} & \tbf{100.0} & \underline{98.1} & 73.8 & \underline{46.2} & \underline{46.9} & \underline{26.2} & \underline{15.6} & 11.9 & 95.3 & \underline{65.3} \\
            & \ours (static $\vgamma$) & \tbf{100.0} & \tbf{100.0} & \tbf{100.0} & \underline{99.4} & \tbf{100.0} & \tbf{98.8} & \tbf{98.1} & \tbf{98.8} & \tbf{95.6} & \tbf{91.9} & \tbf{90.0} & \tbf{99.7} & \tbf{97.5} \\
            & \tbf{\ours (ours)} & \tbf{100.0} & \tbf{100.0} & \tbf{100.0} & \tbf{100.0} & \tbf{100.0} & \underline{85.6} & 26.2 & 22.5 & 17.5 & 13.1 & \underline{12.5} & \underline{97.6} & 61.6 \\
        \cmidrule(l){2-15}
            & \ours ($\vgamma_t \gets \vgamma^{\mathrm{stat}}$) & 99.4 & 100.0 & 90.0 & 54.4 & 23.8 & 13.1 & 10.6 & 10.0 & 10.0 & 10.0 & 10.0 & 63.4 & 39.2 \\
            & \ours ($\vgamma_t \gets \bar\vgamma$) & 99.4 & 100.0 & 100.0 & 100.0 & 99.4 & 55.6 & 20.6 & 16.9 & 16.2 & 12.5 & 11.2 & 92.4 & 57.4 \\
        \midrule
            \multirowrotate{5}{90}{\footnotesize 64K Cooldown}
            & \tf (YaRN) & \tbf{100.0} & \tbf{100.0} & \tbf{100.0} & \tbf{100.0} & \tbf{100.0} & \tbf{100.0} & \underline{82.5} & OOM & OOM & OOM & OOM & \tbf{100.0} & -- \\
            & \gdntwo & \tbf{100.0} & \tbf{100.0} & \underline{94.4} & 60.6 & 17.5 & 11.9 & 15.6 & 29.4 & 19.4 & 10.0 & 10.6 & 64.1 & 42.7 \\
            & \ours (no $\vgamma$) & \tbf{100.0} & \tbf{100.0} & \tbf{100.0} & \tbf{100.0} & \tbf{100.0} & \underline{89.4} & 65.6 & \underline{56.2} & \underline{41.9} & \underline{39.4} & \underline{32.5} & 98.2 & \underline{75.0} \\
            & \ours (static $\vgamma$) & \tbf{100.0} & \tbf{100.0} & \tbf{100.0} & \underline{99.4} & \tbf{100.0} & \tbf{100.0} & \tbf{100.0} & \tbf{98.8} & \tbf{98.8} & \tbf{97.5} & \tbf{99.4} & \underline{99.9} & \tbf{99.4} \\
            & \tbf{\ours (ours)} & \tbf{100.0} & \tbf{100.0} & \tbf{100.0} & \tbf{100.0} & \underline{99.4} & 83.8 & 33.1 & 24.4 & 18.1 & 15.0 & 12.5 & 97.2 & 62.4 \\
        \bottomrule
        \end{tabular}%
    }
\end{table}

%% file: tables/tab18_abl_suppl.tex
\begin{table}[!t]
    \centering
    \caption{%
        \textbf{Fixed-FFN gate-compression, dense dynamic-readout, and unconstrained wide-FFN controls at the $\boldsymbol{406\text{M}}$ scale.}
        Complementing the multi-scale rank ablations in \Cref{tab:abl}, we evaluate fixed-FFN gate compression ($d_{\mathrm{ffn}} = 4{,}352$), dense dynamic-readout ($r_q = d,$ $d_{\mathrm{ffn}} = 3{,}584$), and non-iso-parameter wide-FFN ($d_{\mathrm{ffn}} = 5{,}504,$ $448.7\text{M}$) controls at the $406\text{M}$ scale ($L = 11,$ $d = 1{,}536,$ $100\text{B}$ tokens); $\Delta\text{NLL}$ is reported in millinats relative to \gdntwo, with best and second-best iso-parameter models in \textbf{bold} and \underline{underlined}.\looseness=-1%
    }
    \label{tab:eval_abl_406m}
    \resizebox{\textwidth}{!}{%
        \setlength{\tabcolsep}{5.0pt}%
        \renewcommand{\arraystretch}{1.08}%
        \begin{tabular}{@{}c@{\hspace{3pt}}|l|c|cccccc|ccc|c@{}}
            \toprule
            \multicolumn{1}{@{}c@{\hspace{5pt}}}{} & & & \multicolumn{6}{c|}{{Gate Ranks \& FFN ($d = 1{,}536$)}} & \multicolumn{3}{c|}{{FineWeb-Edu Val}} & {WikiText} \\
            \cmidrule(lr){4-9} \cmidrule(lr){10-12} \cmidrule(l){13-13}
            \multicolumn{1}{@{}c@{\hspace{5pt}}}{} & {Model ($406\text{M}$)} & {Params} & {$r_f$} & {$r_b$} & {$r_g$} & {$r_w$} & {$r_q$} & {$d_{\mathrm{ffn}}$} & {Token} & {$\Delta\text{NLL}$} & {Token} & {($4\text{K}$)} \\
            \multicolumn{1}{@{}c@{\hspace{5pt}}}{} & & {(M)} & & & & & & & {NLL $\downarrow$} & {(mnat) $\downarrow$} & {PPL $\downarrow$} & {PPL $\downarrow$} \\
            \midrule
            \multirowrotate{2}{90}{\footnotesize Base}
            & \tf & 409.7 & -- & -- & -- & -- & -- & 6{,}144 & 2.2424 & $+$7.0 & 9.42 & 21.94 \\
            & \gdntwo & 409.7 & 128 & $d$ & 128 & $d$ & -- & 4{,}352 & 2.2354 & 0.0 & 9.35 & 21.76 \\
            \midrule
            \multirowrotate{3}{90}{\footnotesize Ours}
            & \ours (no $\vgamma$) & 409.7 & 64 & 64 & 128 & 256 & 0 & 5{,}632 & 2.2230 & $-$12.4 & 9.24 & \tbf{21.16} \\
            & \ours (static $\vgamma$) & 409.7 & 64 & 64 & 128 & 256 & 0 & 5{,}632 & \underline{2.2191} & \underline{$-$16.3} & \underline{9.20} & 21.24 \\
            & \tbf{\ours (ours)} & 409.7 & 64 & 64 & 128 & 128 & 128 & 5{,}632 & \tbf{2.2174} & \tbf{$-$18.0} & \tbf{9.18} & \underline{21.22} \\
            \midrule
            \multirowrotate{4}{90}{\footnotesize Controls}
            & \gdntwo + dense $\vgamma_t$ (no $\vc_q$) & 409.7 & 128 & $d$ & 128 & $d$ & $d$ & 3{,}584 & 2.2457 & $+$10.3 & 9.45 & 22.13 \\
            & \gdntwo + dense $\vgamma_t$ (with $\vc_q$) & 409.7 & 128 & $d$ & 128 & $d$ & $d$ & 3{,}584 & 2.2437 & $+$8.3 & 9.43 & 22.22 \\
            & \ours (fixed $d_{\mathrm{ffn}}$) & 370.8 & 128 & 128 & 128 & 128 & 128 & 4{,}352 & 2.2428 & $+$7.4 & 9.42 & 22.02 \\
            & \gdntwo (wide $d_{\mathrm{ffn}}$) & 448.7 & 128 & $d$ & 128 & $d$ & -- & 5{,}504 & 2.2149 & $-$20.5 & 9.16 & 21.01 \\
            \bottomrule
        \end{tabular}%
    }
\end{table}

%% file: apps/i_analysis.tex
\clearpage
\section{Mechanistic Probe Protocols and Supplementary Diagnostics}
\label{app:mech}

All mechanistic probes in \S\ref{sec:mech} and this appendix are evaluated without retraining, on the $100\text{B}$-token checkpoints, over held-out FineWeb-Edu validation sequences of length $T = 4{,}096$ ($16$ sequences, $65{,}536$ tokens per evaluation pass).
%
To eliminate floating-point accumulation drift when comparing closed-form stationary moments against empirical state trajectories, we execute the token-level recurrence in \texttt{float64} when recording channel-wise state norms and logit sensitivities, and discard the initial burn-in window $t < 1{,}024$ ($t \in [1{,}024, 4{,}096)$).
All counterfactual validation loss changes $\Delta\text{NLL} \defas \mathrm{NLL}_{\mathrm{intervene}} - \mathrm{NLL}_{\mathrm{base}}$ are paired per sequence across the held-out evaluation batch, and we report the mean difference alongside its paired standard error (SE).


\subsection{Gate-Sensitivity Diagnostics}
\label{app:mech:sens}

\input{tables/tab19_svd_suppl}

\xhdr{Post-Hoc Gate-Rank Truncation}

To evaluate the marginal value of bottleneck rank across all five gates $(f, b, g, w, q)$ (\Figref{fig:mech_sens}{a} and \Cref{tab:eval_mech_svd}), we inspect the trained $406\text{M}$ uniform-rank \ours checkpoint ($L = 11,$ $d = 1{,}536,$ $N_h = 12,$ $d_k = d_v = 128$), in which every gate is parameterized with an identical bottleneck rank of $128.$
For a single gate $m \in \{f, b, g, w, q\}$ at a time, we compute the singular value decomposition (SVD) of its rank-$128$ factor $\mW_{m\downarrow}\mW_{m\uparrow} \in \R{d \times d}$ in every layer, zero out the trailing $128 - r$ singular values for $r \in \{0, 16, 32, 48, 64, 96\}$ while keeping the learned bias $\vc_m$ intact (so $r = 0$ reduces the gate to its static channel bias), and evaluate validation NLL with the remaining four gates fixed at rank $128.$
As an architecture-independent cross-check on dense gates, applying the same SVD rank truncation to the dense $d \times d$ erase and write projections $(\mW_b, \mW_w)$ of the $166\text{M}$ \gdntwo checkpoint yields $\Delta\text{NLL}_w / \Delta\text{NLL}_b = 8.3\text{--}11.3\times$ across $r \in \{32, 64, 128, 256\}$ (e.g., $+145.9$ vs.\ $+13.2\text{ mnats}$ at $r = 64$), which confirms that the write-erase sensitivity asymmetry is intrinsic to the channel-wise delta recurrence and not an artifact of the low-rank parameterization.


\xhdr{Write/Erase Perturbations}

To verify \Cref{prop1} layer by layer on the $166\text{M}$ checkpoints, we evaluate the expected squared Frobenius logit sensitivities $\E[\norm{\D_{\vz^w}\mS_t[\delta\vz^w]}_F^2]$ and $\E[\norm{\D_{\vz^b}\mS_t[\delta\vz^b]}_F^2]$ induced by \eqref{prop1:eq2} under isotropic pre-activation perturbations $\delta\vz^b \sim \gN(\vzero, \I_{d_k})$ and $\delta\vz^w \sim \gN(\vzero, \I_{d_v}).$
Across individual layers, the write-to-erase sensitivity ratio $\E[\norm{\D_{\vz^w}\mS_t[\delta\vz^w]}_F^2] / \E[\norm{\D_{\vz^b}\mS_t[\delta\vz^b]}_F^2]$ spans $2.5\times$ to $29.7\times,$ with a layer-wise geometric mean of $11.9\times$ in \gdntwo and $11.0\times$ in \ours (arithmetic mean of $13.2\times$ and $13.3\times$).
Furthermore, under $\delta\vb \sim \gN(\vzero, \I_{d_k}),$ evaluating the ratio of parallel-to-total key-erase sensitivity $\E[(\D_{\vb}\evmu_t[\delta\vb])^2] / \E[\norm{\D_{\vb}\hat\vk_t[\delta\vb]}_2^2]$ reproduces the empirical fourth-moment key concentration $\E[\norm{\vk_t}_4^4] = \E[\eta_{\mathrm{eff}}(\vk_t)\inv]$ to within $0.3\%$ across all layers, which verifies Proposition~\ref{prop1:item3} on real token activations.


\subsection{Stationary Row-Energy Law Ablation and Readout Scaling}
\label{app:mech:row}

\input{tables/tab20_mech_suppl}

\xhdr{Empirical Channel-Moment Estimators and Row-Law Ablation}

To evaluate the closed-form stationary row-energy law $\evm_j^\ast = \evc_j\evsigma_v^2 / (1 - \evlambda_j + 2\evlambda_j\bar\evb_j\evc_j)$ in \eqref{prop2:eq2} on the $166\text{M}$ \gdntwo and \ours (static $\vgamma$) checkpoints, we estimate for each head and channel $j \in \{1, \dotsc, d_k\}$ over the post-burn-in trajectory $t \in [1{,}024, 4{,}096)$ the empirical moments $\evc_j = \E[\evk_{t,j}^2],$ $\evlambda_j = \E[\evalpha_{t,j}^2],$ $\bar\evb_j = \E[\evb_{t,j}],$ and head-level written value energy $\evsigma_v^2 = \E[\norm{\hat\vv_t}_2^2],$ excluding dead channels with $\evrho_j = \sqrt{\smash{\evm_j}\vphantom{\evm}} < 10^{-10}$ ($<\!1.5\%$ of channels).
\Cref{tab:mech_supp} compares the full key-and-erasure-corrected fixed point $\evm_j^\ast$ \eqref{prop2:eq2} against the naive decay-only approximation $\evm_{j,\mathrm{naive}}^\ast \propto (1 - \evlambda_j)\inv$ that omits heterogeneous key write energy $\evc_j$ and diagonal erasure $2\evlambda_j\bar\evb_j\evc_j.$
%
Across live channels, $\evm_j^\ast$ achieves a median within-head log-Pearson correlation $r(\log_{10}\evm_j, \log_{10}\evm_j^\ast)$ of $0.951$ in both \gdntwo and \ours ($0.932$ and $0.916$ pooled across all heads), whereas dropping $\evc_j$ in $(1 - \evlambda_j)\inv$ lowers median within-head log-Pearson correlation to $0.840$ and $0.803$ ($0.869$ and $0.827$ pooled).
This confirms that both heterogeneous decay $\evlambda_j$ and non-uniform key channel variance $\evc_j$ are required to predict stationary state energy (\Cref{prop2:rem1}).


\xhdr{Cross-Scale Readout Scaling and Within-Head Alignment}

Across all heads of the $166\text{M}$ \ours (static $\vgamma$) checkpoint, the learned readout gain magnitude $\abs{\evgamma_j}$ has a median within-head Spearman rank correlation with inverse stationary row RMS $\evrho_j\inv$ of $r_s(\abs{\evgamma_j}, \evrho_j\inv) = +0.71,$ with positive alignment ($r_s > 0$) in $100\%$ of heads ($+0.706$ and $98\%$ across live channels in uniform \ours static $\vgamma$; \Cref{tab:mech_supp}).
In the $1.34\text{B}$ \ours model with token-dependent $\vgamma_t,$ regressing the log time-averaged readout gain $\log_{10}\bar\evgamma_j = \log_{10}\E[\evgamma_{t,j}]$ against $\log_{10}\evrho_j$ within each of the $384$ heads yields a median power-law slope $-\mathrm{d}\log\bar\evgamma_j / \mathrm{d}\log\evrho_j = 0.44$ ($\bar\evgamma_j \propto \evrho_j^{-0.44}$; \Cref{tab:mech_supp}), which matches the $166\text{M}$ static exponent $p = 0.35$ and lies inside the fractional interval $p \in (0, 1)$ of \Cref{prop3:cor1}.
On the amplitude (RMS) scale at $1.34\text{B},$ weighting state rows by $\bar\evgamma_j$ reduces the median within-head $\mathrm{p90}/\mathrm{p10}$ readout row-RMS ratio from $3.83\times$ in \gdntwo (unweighted $\evrho_j = \sqrt{\smash{\evm_j}\vphantom{\evm}},$ i.e., $\bar\evgamma_j = 1$) to $3.10\times$ for $\bar\evgamma_j\evrho_j$ in \ours ($-19.1\%$).


\subsection{Counterfactual Interventions and Memory-Horizon Diagnostics}
\label{app:mech:intervene}

\input{tables/tab21-22_counterfactual-logit}

\xhdr{Position-Stratified Gate Freezing, Isotropic Erase, and Erase-Logit Shifts}

\Cref{tab:eval_mech_m8} provides the complete position-stratified breakdown ($[0, 16),$ $[16, 256),$ and $[256, 4{,}096)$) and induced parallel erase rates $\bar\evmu$ for the causal gate interventions on the $834\text{M}$ checkpoints (\Figref{fig:mech_sens}{b} and \Figref{fig:mech_cf}{b}).
In addition to temporal gate freezing ($\vw_t \gets \bar\vw$ and $\vb_t \gets \bar\vb$), we evaluate an isotropic channel-erase intervention $\vb_t \gets \evmu_t\vone$ that preserves the exact token-by-token parallel erase scalar $\evmu_t = \inner{\vk_t}{\vb_t \odot \vk_t}$ while zeroing out the oblique erase residual $\vs_t = (\vb_t - \evmu_t\vone) \odot \vk_t$ (Proposition~\ref{prop1:item3} and \Cref{lem1}).
%
Zeroing out the oblique residual $\vs_t$ degrades validation NLL by only $+0.084\text{ nats}$ in \gdntwo and $+0.049\text{ nats}$ in \ours ($8\text{--}10\times$ smaller than freezing $\vw_t$; slightly above $\vb_t \gets \bar\vb$ because $\bar\vb$ retains static per-channel baseline erase rates across rows whereas $\evmu_t\vone$ enforces a uniform channel scalar, \Cref{prop1:rem2}), which confirms that the dynamic action of the erase gate is captured mainly by $\evmu_t$ along the active key subspace.


\xhdr{Counterfactual Readout Gain}

\Cref{tab:eval_mech_m13} reports the full counterfactual readout sweep (\Figref{fig:mech_cf}{a}) on both $166\text{M}$ static-$\vgamma$ checkpoints (\ours static $\vgamma$ with gate ranks $(64, 64, 128, 256, 0)$ and uniform \ours static $\vgamma$ with gate ranks $(128, 128, 128, 128, 0)$).
In each head, the surrogate profile is rescaled to match the $\ell_2$ norm of the learned $\vgamma^{\mathrm{stat}}$ (setting dead channels with $\evrho_j < 10^{-10}$ to the within-head median live $\evrho_j$ prior to norm-matching) so that $\Delta\text{NLL}$ isolates the effect of the relative channel-weighting shape.

Beyond the power-law family $\evgamma_j(p) \propto \evrho_j^{-p}$ ($p \in [0, 1]$), we evaluate three additional controls:
(i) randomly permuting the learned $\vgamma^{\mathrm{stat}}$ entries across channels within each head (which preserves the exact per-head gain distribution while breaking alignment with $\evrho_j\inv$) degrades validation NLL by $+0.130 \pm 0.010\text{ nats}$ in \ours and $+0.141 \pm 0.011\text{ nats}$ in uniform \ours (worse than unit gain $p = 0$), which shows that the benefit of $\vgamma$ comes specifically from ordering gains inversely with channel row energy;
(ii) inverting the power law to amplify slow rows ($\evgamma_j \propto \evrho_j^{+1}$) causes a loss spike of $+2.291 \pm 0.037\text{ nats}$ in \ours ($+2.541 \pm 0.038\text{ nats}$ in uniform \ours); and
(iii) replacing $\vgamma^{\mathrm{stat}}$ with the unregularized single-horizon ratio $g_H(\evlambda_j)/\evm_j^\ast$ (\Cref{tab:eval_mech_m13}, right) achieves its lowest loss at intermediate horizons $H \in [64, 256]$ tokens ($+0.768\text{ nats}$ at $H = 64$), which confirms that language modeling queries integrate a broad mixture of target lags (\Cref{prop3:cor1:rem1}).

Across all horizons $H,$ however, the raw ratio $g_H(\evlambda_j)/\evm_j^\ast$ incurs a larger $\Delta\text{NLL}$ than the fractional power law for two reasons.
%
First, because post-$\silu$ keys concentrate energy onto a few active channels ($\eta_{\mathrm{eff}}(\vk_t) \approx 6\text{--}7$; \Cref{tab:mech}), dividing directly by $\evm_j^\ast \propto \evc_j$ assigns large weights $\evc_j\inv$ to weakly active channels ($\evc_j \ll 1/d_k$), dominating the head norm budget and suppressing active channels.
%
Second, on slow channels with $2\tau_{\mathrm{eff},j} > H - 1,$ the local exponent $p(\tau_{\mathrm{eff},j})$ in \Cref{prop3:cor1} exceeds $1$ ($p \to 2$), over-damping the slow rows that store long-range context.


\xhdr{State Spectral Effective Rank and Memory Horizon}

Finally, to test the prediction of \Cref{prop4} and \Cref{prop4:rem1} that relaxing write-time erasure $\bar\evmu$ (\Cref{tab:mech}) preserves more historical associations in $\mS_t,$ we measure two complementary dimensions of stored memory across all $384$ heads at $1.34\text{B}$: \emph{spatial subspace capacity} (how many orthogonal key-value directions coexist in $\mS_t$ at timestep $t$) and \emph{temporal lookback range} (how far back in token lag the readout $\vo_t$ attributes its output).

First, we quantify the spatial subspace capacity via spectral effective rank $\mathrm{erank}(\mS_t) \defas \exp(-\sum_{j=1}^{d_k} \evp_j \log \evp_j),$ where $\evp_j = \sigma_j^2(\mS_t) / \norm{\mS_t}_F^2$ is the fraction of state Frobenius energy in the $j$-th singular direction of the single-head state $\mS_t$ ($\mathrm{erank}(\mS_t) = K$ when $K$ orthogonal key-value pairs are stored with equal energy).
Second, to quantify temporal lookback range, unrolling the linear recurrence $\mS_t = \mA_t\mS_{t-1} + \vk_t\tr\hat\vv_t$ with single-step transition $\mA_s \defas (\I - \vk_s\tr\hat\vk_s)\Diag(\valpha_s)$ expresses the readout $\vo_t = \hat\vq_t\mS_t$ ($\hat\vq_t = \vgamma_t \odot \vq_t$) as a linear-attention sum over all past written values $\hat\vv_{t-\ell}$ ($\ell \in \{0, \dotsc, t-1\}$):
\[
    \vo_t = \sum_{\ell=0}^{t-1} \eva_{t,t-\ell}\,\hat\vv_{t-\ell},
    \qquad
    \eva_{t,t-\ell} \defas \hat\vq_t\,\mPhi_{t,t-\ell}\,\vk_{t-\ell}\tr,
    \qquad
    \mPhi_{t,s} \defas \mA_t\mA_{t-1}\cdots\mA_{s+1},
\]
with $\mPhi_{t,t} = \I,$ where $\eva_{t,t-\ell} \in \R{}$ is the effective attention weight from query timestep $t$ to the token written $\ell$ steps earlier.
For late query positions $t \ge 2{,}048$ (where at least $2{,}048$ steps of context are available), we define the $50\%$-mass memory horizon $L_{50}$ as the smallest lag $L$ at which the cumulative attribution $\sum_{\ell=0}^{L} \abs{\eva_{t,t-\ell}}$ reaches $50\%$ of the total historical attribution $\sum_{\ell=0}^{t-1} \abs{\eva_{t,t-\ell}}.$
Across held-out sequences, mean $\mathrm{erank}(\mS_t)$ increases by $16.1\%$ ($6.16 \to 7.15$) and layer-mean median $L_{50}$ by $59.6\%$ ($49.3 \to 78.7$ tokens).


\input{tables/tab23_mech_ruler}

\subsection{Multi-Value Recall and Multi-Hop Variable Tracking}
\label{app:mech:vt}

\Cref{tab:eval_mech_vt_mv} reports the complete write-order MV-NIAH ($V_1\text{--}V_4$) and hop-resolved VT ($H_1\text{--}H_5$) recall accuracies across the $1.34\text{B}$ models at $1\text{K}$ and $4\text{K}$ context lengths ($N = 500$ prompts per task).

Using the $1\text{K}$ RULER VT protocol and per-target scoring from \Cref{app:eval:recall}, we track how each written VT hop statement $H_m$ at $s_m$ ($m \in \{1, \dotsc, 5\}$) is preserved in $\mS_t$ and read out at generation time across $N = 16$ held-out prompts.
At each sequence milestone $t \in \{s_m, \dotsc, s_5, t_{\mathrm{q}}\},$ we measure the remaining initial-query-projected signal of needle $H_m$ in $\mS_t$ (summed over the token span $\gT_m$ of statement $H_m$ ending at $s_m$) via
\[
    R_m(t) \defas \sum_{s \in \gT_m} \abs{\hat\vq_{t_{\mathrm{q}}}\,\mPhi_{t,s}\,\vk_s\tr}\,\norm{\hat\vv_s}_2,
\]
averaged across the top-quartile active retrieval heads (ranked by total needle signal $\sum_{m=1}^5 R_m(t_{\mathrm{q}})$ at $t_{\mathrm{q}}$) in the middle associative-memory layers (layers $8\text{--}19$ of our $1.34\text{B}$ model).
Because per-head $\rms(\vo_t)$ cancels uniform multiplicative scaling of $\hat\vq_{t_{\mathrm{q}}}$ and $\hat\vv_s,$ within each prompt and model we normalize the head-mean needle trace to its initial value at write milestone $s_m,$ $\tilde{R}_m(t) \defas R_m(t) / R_m(s_m)$ ($\tilde{R}_m(s_m) = 100\%$), and average $\tilde{R}_m(t)$ across the $16$ prompts to isolate retention across $\mPhi_{t,s}$ from pre-$\rms$ gate scale and random needle placement.
\Figref{fig:mech_vt_interp}{c} plots prompt-mean $\tilde{R}_1(t)$ across milestones $s_1\text{--}s_5$ to $t_{\mathrm{q}}$ (with interquartile bands across the $16$ prompts), and \Figref{fig:mech_vt_interp}{d} reports prompt-mean $\tilde{R}_m(t)$ across all five needles $H_1\text{--}H_5.$
Across the four teacher-forced answer-step transitions $A_1\text{--}A_4$ ($t_1 < t_2 < t_3 < t_4,$ as the model emits $\texttt{VAR\_1}\to\dotsb\to\texttt{VAR\_4}$ to chain each hop to the next), \Figref{fig:mech_vt_interp}{e} reports the share of readout weight over the haystack $\gH$ placed on each needle $H_m$ (excluding the trailing question and answer suffix, averaged over the top-$10$ active heads per answer step across the $16$ prompts):
\[
    \tilde{a}_m(A_k) \defas \frac{\sum_{s \in \gT_m} \abs{\eva_{t_k,s}}\,\norm{\hat\vv_s}_2}{\sum_{s \in \gH} \abs{\eva_{t_k,s}}\,\norm{\hat\vv_s}_2},
    \qquad k \in \{1, 2, 3, 4\}.
\]

%% file: tables/tab19_svd_suppl.tex
\begin{table}[!t]
    \centering
    \caption{%
        \textbf{Post-hoc SVD gate-rank truncation sensitivity on the trained $\boldsymbol{406\text{M}}$ uniform-rank checkpoint.}
        For each gate $m \in \{f, b, g, w, q\},$ we replace its rank-$128$ factor $\mW_{m\downarrow}\mW_{m\uparrow}$ across all $11$ layers with its best rank-$r$ SVD approximation (retaining the additive bias $\vc_m$) and measure $\Delta\text{NLL}(r) \defas \mathrm{NLL}(r) - \mathrm{NLL}(128)$ (in nats $\pm$ paired SE) on held-out FineWeb-Edu sequences.%
    }
    \label{tab:eval_mech_svd}
    \resizebox{\textwidth}{!}{%
        \setlength{\tabcolsep}{3pt}%
        \renewcommand{\arraystretch}{1.08}%
        \begin{tabular}{@{}l|cccccc@{}}
        \toprule
        & \multicolumn{6}{c@{}}{{Kept SVD Rank $r$ (Validation $\Delta\text{NLL}$ vs.\ $r = 128$ in nats $\pm$ SE $\downarrow$)}} \\
        \cmidrule(l){2-7}
            {Gate} & {$r = 0$ (bias)} & {$r = 16$} & {$r = 32$} & {$r = 48$} & {$r = 64$} & {$r = 96$} \\
        \midrule
            Decay $f$ ($\valpha_t$) & $+$0.5801$\,\pm\,$0.2147 & $+$0.0259$\,\pm\,$0.0084 & $+$0.0043$\,\pm\,$0.0018 & $+$0.0020$\,\pm\,$0.0015 & $-$0.0007$\,\pm\,$0.0015 & $-$0.0001$\,\pm\,$0.0004 \\
            Erase $b$ ($\vb_t$) & $+$0.1193$\,\pm\,$0.0154 & $+$0.0039$\,\pm\,$0.0004 & $+$0.0016$\,\pm\,$0.0009 & $+$0.0006$\,\pm\,$0.0008 & $+$0.0006$\,\pm\,$0.0004 & $+$0.0000$\,\pm\,$0.0001 \\
            \midrule
            Write $w$ ($\vw_t$) & $+$0.4454$\,\pm\,$0.2248 & $+$0.0486$\,\pm\,$0.0048 & $+$0.0185$\,\pm\,$0.0022 & $+$0.0108$\,\pm\,$0.0015 & $+$0.0064$\,\pm\,$0.0010 & $+$0.0013$\,\pm\,$0.0007 \\
            Query $q$ ($\vgamma_t$) & $+$0.5639$\,\pm\,$0.2352 & $+$0.0296$\,\pm\,$0.0057 & $+$0.0118$\,\pm\,$0.0023 & $+$0.0077$\,\pm\,$0.0013 & $+$0.0045$\,\pm\,$0.0013 & $+$0.0015$\,\pm\,$0.0008 \\
            Output $g$ ($\vg_t$) & $+$2.7725$\,\pm\,$0.2189 & $+$0.7489$\,\pm\,$0.2328 & $+$0.4824$\,\pm\,$0.2483 & $+$0.3492$\,\pm\,$0.2225 & $+$0.2472$\,\pm\,$0.1783 & $+$0.0376$\,\pm\,$0.0209 \\
        \bottomrule
        \end{tabular}%
    }
\end{table}

%% file: tables/tab20_mech_suppl.tex
\begin{table}[!ht]
    \centering
    \caption{%
        \textbf{Supplementary stationary row-energy and readout-scaling diagnostics across trained \gdntwo and \ours checkpoints.}
        Complementing \Cref{tab:mech}, row-law ablations and static-gain alignment are evaluated across live channels ($\evrho_j \ge 10^{-10}$) on the $166\text{M}$ \gdntwo and \ours (static $\vgamma$) checkpoints, and dynamic-gain scaling and readout row-RMS spread are evaluated across all $24$ layers ($384$ heads) at $1.34\text{B}$ (with $\bar\evgamma_j = 1$ in \gdntwo); values report the median across heads.%
    }
    \label{tab:mech_supp}
    \resizebox{\textwidth}{!}{%
        \setlength{\tabcolsep}{3pt}%
        \renewcommand{\arraystretch}{1.12}%
        \begin{tabular}{@{}l|l|c|cc|c|l@{}}
            \toprule
            {Diagnostic} & {Within-Head Statistic} & {Scale} & {\gdntwo} & {\ours} & {Effect} & {Theory} \\
            \midrule
            Full row-law log-Pearson $r$ & $r(\log_{10}\evm_j, \log_{10}\evm_j^\ast)$ & 166M & 0.951 & 0.951 & $0.932/0.916$ pooled & Prop.~\ref{prop2} \\
            Decay-only log-Pearson $r$ & $r(\log_{10}\evm_j, -\log_{10}(1 - \evlambda_j))$ & 166M & 0.840 & 0.803 & $-0.13$ ($0.869/0.827$ pooled) & Rem.~\ref{prop2:rem1} \\
        \midrule
            Static gain alignment & $r_s(\abs{\evgamma_j}, \evrho_j\inv)$ & 166M & -- & $+$0.710 & $100\%$ heads $\!>\!0$ & Cor.~\ref{prop3:cor1} \\
            Readout power-law slope & $-\mathrm{d}\log\bar\evgamma_j / \mathrm{d}\log\evrho_j$ & 1.34B & 0.00 & 0.44 & $p \in (0, 1)$ & Cor.~\ref{prop3:cor1} \\
            Readout row-RMS spread & $\mathrm{p90}/\mathrm{p10}$ of $\bar\evgamma_j\evrho_j$ & 1.34B & 3.83$\times$ & \tbf{3.10$\times$} & $-$19.1\% & Cor.~\ref{prop3:cor1} \\
            \bottomrule
        \end{tabular}%
    }
\end{table}

%% file: tables/tab21-22_counterfactual-logit.tex
\begin{table}[!t]
    \centering
        \caption{%
        \textbf{Causal gate-freezing and erase-logit shift interventions on trained $\boldsymbol{834\text{M}}$ checkpoints.}
        (Top) Validation $\Delta\text{NLL}$ and mean erase rate $\bar\evmu$ when replacing the dynamic write gate $\vw_t$ or erase gate $\vb_t$ with its stream mean ($\bar\vw$ or $\bar\vb$) or isotropic key-space projection ($\evmu_t\vone$) across all layers.
        (Bottom) Additive erase-gate logit shifts $\vz^b_t \gets \vz^b_t + \delta_b$ for $\delta_b \in \{-1.0, -0.5, +0.5\},$ reported overall ($[0, 4{,}096)$) and stratified by token position bucket ($[0, 16),$ $[16, 256),$ and $[256, 4{,}096)$).%
    }
    \label{tab:eval_mech_m8}
    \resizebox{\textwidth}{!}{%
        \setlength{\tabcolsep}{2.8pt}%
        \renewcommand{\arraystretch}{1.08}%
        \begin{tabular}{@{}c@{\hspace{4pt}}|l|l|c|cccc@{}}
            \toprule
            \multicolumn{1}{@{}c@{\hspace{5pt}}}{} & \multicolumn{3}{c|}{} & \multicolumn{4}{c@{}}{{Validation $\Delta\text{NLL}$ ($\text{nats} \pm \text{SE}$) Across Token Positions}} \\
            \cmidrule(l){5-8}
            \multicolumn{1}{@{}c@{\hspace{5pt}}}{} & {Intervention} & {Model} & {Mean Erase} & {Overall} & {Early} & {Mid} & {Late} \\
            \multicolumn{1}{@{}c@{\hspace{5pt}}}{} & & {($834\text{M}$)} & {$\bar\evmu$} & {$[0, 4{,}096)$} & {$[0, 16)$} & {$[16, 256)$} & {$[256, 4{,}096)$} \\
            \midrule
            \multirowrotate{6}{90}{\footnotesize Freeze}
            & Freeze write gate & \gdntwo & 0.733 & $+$0.8325$\,\pm\,$0.0581 & $+$0.5799$\,\pm\,$0.1467 & $+$0.9517$\,\pm\,$0.1275 & $+$0.8261$\,\pm\,$0.0566 \\
            & Freeze erase gate & \gdntwo & 0.672 & $+$0.0574$\,\pm\,$0.0066 & $+$0.0153$\,\pm\,$0.0143 & $+$0.0523$\,\pm\,$0.0152 & $+$0.0578$\,\pm\,$0.0062 \\
            & Isotropic erase $\vb_t \gets \evmu_t\vone$ & \gdntwo & 0.732 & $+$0.0844$\,\pm\,$0.0044 & $+$0.0189$\,\pm\,$0.0116 & $+$0.0658$\,\pm\,$0.0157 & $+$0.0859$\,\pm\,$0.0044 \\
            \cmidrule(lr){2-8}
            & Freeze write gate & \ours & 0.610 & $+$0.3845$\,\pm\,$0.0368 & $+$0.2534$\,\pm\,$0.0767 & $+$0.4877$\,\pm\,$0.1309 & $+$0.3786$\,\pm\,$0.0328 \\
            & Freeze erase gate & \ours & 0.559 & $+$0.0266$\,\pm\,$0.0039 & $+$0.0068$\,\pm\,$0.0051 & $+$0.0316$\,\pm\,$0.0101 & $+$0.0264$\,\pm\,$0.0038 \\
            & Isotropic erase $\vb_t \gets \evmu_t\vone$ & \ours & 0.611 & $+$0.0485$\,\pm\,$0.0040 & $-$0.0039$\,\pm\,$0.0135 & $+$0.0374$\,\pm\,$0.0059 & $+$0.0494$\,\pm\,$0.0043 \\
            \midrule
            \multirowrotate{6}{90}{\footnotesize Shift}
            & Erase shift $\delta_b = -1.0$ & \gdntwo & $0.737 \to 0.440$ & $+$0.0437$\,\pm\,$0.0038 & $+$0.0146$\,\pm\,$0.0060 & $+$0.0367$\,\pm\,$0.0115 & $+$0.0442$\,\pm\,$0.0037 \\
            & Erase shift $\delta_b = -0.5$ & \gdntwo & $0.737 \to 0.579$ & $+$0.0082$\,\pm\,$0.0013 & $+$0.0032$\,\pm\,$0.0030 & $+$0.0019$\,\pm\,$0.0032 & $+$0.0087$\,\pm\,$0.0014 \\
            & Erase shift $\delta_b = +0.5$ & \gdntwo & $0.737 \to 0.907$ & $+$0.0158$\,\pm\,$0.0017 & $+$0.0039$\,\pm\,$0.0027 & $+$0.0222$\,\pm\,$0.0042 & $+$0.0154$\,\pm\,$0.0018 \\
            \cmidrule(lr){2-8}
            & Erase shift $\delta_b = -1.0$ & \ours & $0.611 \to 0.369$ & $+$0.0313$\,\pm\,$0.0032 & $+$0.0109$\,\pm\,$0.0060 & $+$0.0309$\,\pm\,$0.0115 & $+$0.0314$\,\pm\,$0.0034 \\
            & Erase shift $\delta_b = -0.5$ & \ours & $0.611 \to 0.481$ & $+$0.0079$\,\pm\,$0.0013 & $+$0.0047$\,\pm\,$0.0041 & $+$0.0091$\,\pm\,$0.0054 & $+$0.0078$\,\pm\,$0.0014 \\
            & Erase shift $\delta_b = +0.5$ & \ours & $0.611 \to 0.756$ & $+$0.0078$\,\pm\,$0.0018 & $-$0.0026$\,\pm\,$0.0045 & $+$0.0066$\,\pm\,$0.0043 & $+$0.0080$\,\pm\,$0.0019 \\
            \bottomrule
        \end{tabular}%
    }

\vspace{1em}

    \centering
    \caption{%
        \textbf{Counterfactual readout gain replacement on trained $\boldsymbol{166\text{M}}$ static-$\boldsymbol{\vgamma}$ checkpoints.}
        In every head and layer, we replace the learned static readout gain $\vgamma^{\mathrm{stat}}$ (which has empirical log-log slope $-\mathrm{d}\log\abs{\evgamma_j}/\mathrm{d}\log\evrho_j = 0.354$ in \ours static $\vgamma$ and $0.353$ in uniform \ours static $\vgamma$) with a norm-matched power-law profile $\evgamma_j \propto \evrho_j^{-p}$ for $p \in \{0.00, 0.25, 0.50, 0.75, 1.00\}$ (left) or a single-horizon \Cref{prop3:cor1} profile $g_H(\evlambda_j)/\evm_j^\ast$ for $H \in \{16, 64, 256, 1{,}024, 4{,}096\}$ (right).%
    }
    \label{tab:eval_mech_m13}
    \resizebox{\textwidth}{!}{%
        \setlength{\tabcolsep}{3.5pt}%
        \renewcommand{\arraystretch}{1.08}%
        \begin{tabular}{@{}l|cc|l|cc@{}}
            \toprule
            \multicolumn{3}{@{}c|}{{Power-Law Surrogate $\evgamma_j \propto \evrho_j^{-p}$ ($\Delta\text{NLL} \downarrow$)}} & \multicolumn{3}{c@{}}{{Single-Horizon Filter $g_H(\evlambda_j)/\evm_j^\ast$ ($\Delta\text{NLL} \downarrow$)}} \\
            \cmidrule(r){1-3} \cmidrule(l){4-6}
            {Exponent $p$} & {\ours} & {Uniform \ours} & {Horizon $H$} & {\ours} & {Uniform \ours} \\
            {(Norm-Matched)} & {(static $\vgamma$)} & {(static $\vgamma$)} & {(Norm-Matched)} & {(static $\vgamma$)} & {(static $\vgamma$)} \\
            \midrule
            Unit gain $p = 0.00$ ($\evgamma_j = 1$) & $+$0.1007$\,\pm\,$0.0066 & $+$0.1142$\,\pm\,$0.0094 & $H = 16$ toks & $+$0.9417$\,\pm\,$0.0346 & $+$1.1014$\,\pm\,$0.0400 \\
            Fractional power $p = 0.25$ & \tbf{$+$0.0369$\,\pm\,$0.0030} & \tbf{$+$0.0406$\,\pm\,$0.0039} & $H = 64$ toks & \tbf{$+$0.7681$\,\pm\,$0.0293} & \tbf{$+$1.0367$\,\pm\,$0.0398} \\
            Half-power $p = 0.50$ ($1/\sqrt{\smash{\evrho_j}\vphantom{\evrho}}$) & \underline{$+$0.0633$\,\pm\,$0.0053} & \underline{$+$0.0630$\,\pm\,$0.0056} & $H = 256$ toks & \underline{$+$0.7911$\,\pm\,$0.0262} & \underline{$+$1.0964$\,\pm\,$0.0388} \\
            Fractional power $p = 0.75$ & $+$0.1568$\,\pm\,$0.0109 & $+$0.1549$\,\pm\,$0.0109 & $H = 1{,}024$ toks & $+$0.9537$\,\pm\,$0.0229 & $+$1.3139$\,\pm\,$0.0315 \\
            Full whitening $p = 1.00$ ($1/\evrho_j$) & $+$0.3110$\,\pm\,$0.0179 & $+$0.3128$\,\pm\,$0.0171 & $H = 4{,}096$ toks & $+$1.1032$\,\pm\,$0.0246 & $+$1.4444$\,\pm\,$0.0324 \\
            \bottomrule
        \end{tabular}%
    }
\end{table}

%% file: tables/tab23_mech_ruler.tex
\begin{table}[!t]
    \centering
    \caption{%
        \textbf{Write-order MV-NIAH and hop-resolved variable tracking recall at $\boldsymbol{1.34\text{B}}$ parameters.}
        All results are evaluated over $N = 500$ examples per task at $1\text{K}$ (top) and $4\text{K}$ context lengths (bottom).
        MV-NIAH columns $V_1\text{--}V_4$ order the four target values bound to the shared key from oldest write ($V_1$) to most recent write ($V_4$), and VT columns $H_1\text{--}H_5$ order the five target variables from earliest root binding ($H_1,$ \texttt{VAR\_1 = V}) to latest alias ($H_5,$ \texttt{VAR\_5 = VAR\_4}).%
    }
    \label{tab:eval_mech_vt_mv}
    \resizebox{0.69\textwidth}{!}{%
        \setlength{\tabcolsep}{3.4pt}%
        \renewcommand{\arraystretch}{1.08}%
        \begin{tabular}{@{}c@{\hspace{4pt}}|l|cccc|c|ccccc|c@{}}
            \toprule
            \multicolumn{1}{@{}c@{\hspace{5pt}}}{} & & \multicolumn{5}{c|}{{MV-NIAH Recall (\%) $\uparrow$}} & \multicolumn{6}{c@{}}{{5-Hop VT Recall (\%) $\uparrow$}} \\
            \cmidrule(lr){3-7} \cmidrule(l){8-13}
            \multicolumn{1}{@{}c@{\hspace{5pt}}}{} & {Model ($1.34\text{B}$)} & {$V_1$} & {$V_2$} & {$V_3$} & {$V_4$} & {Avg.} & {$H_1$} & {$H_2$} & {$H_3$} & {$H_4$} & {$H_5$} & {Avg.} \\
            \midrule
            \multirowrotate{5}{90}{\footnotesize 1K}
            & \tf & \underline{84.8} & \tbf{87.2} & \underline{85.6} & 66.0 & \underline{80.9} & 25.8 & 5.8 & 6.0 & 6.6 & 5.2 & 9.9 \\
            & \gdntwo & 67.0 & 57.0 & 62.0 & \underline{74.4} & 65.1 & 28.4 & 34.4 & \underline{49.4} & \underline{61.6} & \underline{64.6} & \underline{47.7} \\
            & \ours (no $\vgamma$) & \tbf{85.6} & 46.0 & 49.0 & 57.4 & 59.5 & \underline{29.2} & \underline{36.0} & 49.0 & 57.2 & 50.8 & 44.4 \\
            & \ours (static $\vgamma$) & 57.4 & 55.2 & 57.2 & 59.6 & 57.4 & 13.2 & 17.2 & 20.8 & 23.8 & 20.2 & 19.0 \\
            & \tbf{\ours (ours)} & 82.4 & \underline{84.4} & \tbf{86.6} & \tbf{91.6} & \tbf{86.2} & \tbf{48.0} & \tbf{46.0} & \tbf{61.2} & \tbf{69.2} & \tbf{75.4} & \tbf{60.0} \\
            \midrule
            \multirowrotate{5}{90}{\footnotesize 4K}
            & \tf & \underline{15.4} & \tbf{43.2} & \underline{45.4} & 37.6 & 35.4 & \tbf{34.8} & 13.8 & 20.0 & 25.6 & 13.2 & \underline{21.5} \\
            & \gdntwo & 10.6 & 20.6 & 40.2 & \tbf{80.6} & \underline{38.0} & 3.6 & 5.6 & 11.8 & 22.4 & 9.2 & 10.5 \\
            & \ours (no $\vgamma$) & 15.0 & 14.0 & 32.4 & 53.0 & 28.6 & 14.4 & 13.6 & 22.6 & 31.8 & 8.2 & 18.1 \\
            & \ours (static $\vgamma$) & 14.6 & 21.0 & 36.8 & 59.4 & 33.0 & 8.6 & \underline{15.4} & \underline{22.8} & \tbf{34.8} & \tbf{20.2} & 20.4 \\
            & \tbf{\ours (ours)} & \tbf{18.2} & \underline{34.6} & \tbf{53.4} & \underline{73.6} & \tbf{45.0} & \underline{16.2} & \tbf{22.0} & \tbf{29.6} & \underline{32.8} & \underline{19.4} & \tbf{24.0} \\
            \bottomrule
        \end{tabular}%
    }
\end{table}

%% file: apps/j_code.tex
\clearpage
\section{Reference Implementation}
\label{app:code}

In this appendix, we provide a JAX implementation of the \ours token mixer and the boundary-pivoted chunkwise forward and backward passes for wide-matrix-unit accelerators.
On GPUs, because replacing $\vq_t$ with $\hat\vq_t = \vgamma_t \odot \vq_t$ reduces the \ours recurrence to the chunkwise linear system of \gdntwo, the mixer in \Cref{lst:mixer} also runs directly on existing Triton kernels for \gdntwo \citep{Yang2024-vn,Hatamizadeh2026-cw} without kernel changes by $\ell_2$-normalizing queries and keys to unit norm ($\vq_t, \vk_t$) \emph{before} forming $\hat\vq_t = \vgamma_t \odot \vq_t,$ and passing $(\hat\vq_t, \vk_t, \vv_t, \log\valpha_t, \vb_t, \vw_t)$ with in-kernel query-key $\ell_2$ normalization disabled so $\hat\vq_t$ is not re-normalized.\looseness=-1

\vspace{12pt}
\listingcaption{\textbf{\ours token mixer.} Fused low-rank gate projection $(r_f, r_b, r_g, r_w, r_q) = (64, 64, 128, 128, 128),$ dynamic matched-filter readout $\vgamma_t,$ pre-permutation fusion, and post-$\rms$ output gating.}{lst:mixer}
\vspace{4pt}
\begin{minted}{python}
import functools, jax
from jax import numpy as jnp
# Assumes depthwise causal_conv1d(u, w), l2_norm(x), and rms_norm(x, scale).

def muir_mixer(x, params, num_heads=16, chunk_size=256):
  """Muir token mixer (x: (B, T, d) bf16/fp32; c_f, a_log, gamma_stat, w_rms fp32)."""
  batch, seq_len, d_model = x.shape
  d_head, num_chunks = d_model // num_heads, seq_len // chunk_size

  u = causal_conv1d(jnp.dot(x, params['W_qkv']), params['w_conv'])
  q, k, v = jnp.split(jax.nn.silu(u), 3, axis=-1)

  h = jnp.dot(x, params['W_down'])
  h_fb = h[..., :128].reshape(batch, seq_len, 2, 64)
  h_gwq = h[..., 128:].reshape(batch, seq_len, 3, 128)
  z_f, z_b = jnp.einsum('btgr,grd->gbtd', h_fb, params['W_up_fb'])
  z_b = z_b + params['c_b']
  z_g, z_w, z_q = (jnp.einsum('btgr,grd->gbtd', h_gwq, params['W_up_gwq'])
                   + params['c_gwq'][:, None, None, :])

  q, k, v, z_f, z_b, z_w, z_q = [
      t.reshape(batch, seq_len, num_heads, d_head)
      for t in (q, k, v, z_f, z_b, z_w, z_q)
  ]
  q_norm, k_norm = l2_norm(q), l2_norm(k)
  q_hat = params['gamma_stat'] * (2.0 * jax.nn.sigmoid(z_q)) * q_norm
  k_hat = 2.0 * jax.nn.sigmoid(z_b) * k_norm
  v_hat = v * jax.nn.sigmoid(z_w)

  chunk_shape = (batch, num_chunks, chunk_size, num_heads, d_head)
  q_hat, k_norm, k_hat, v_hat, z_f = [
      t.astype(x.dtype).reshape(chunk_shape).transpose(1, 0, 3, 2, 4)
      for t in (q_hat, k_norm, k_hat, v_hat, z_f)
  ]
  log_alpha = -jnp.exp(params['a_log'])[:, None, None] * jax.nn.softplus(
      z_f.astype(jnp.float32) + params['c_f'].reshape(num_heads, 1, d_head))

  out = muir_chunkwise(q_hat, k_norm, k_hat, v_hat, log_alpha)
  out = out.transpose(1, 0, 3, 2, 4).reshape(batch, seq_len, num_heads, d_head)
  out = rms_norm(out, params['w_rms']) * jax.nn.silu(
      z_g.reshape(batch, seq_len, num_heads, d_head))
  return jnp.dot(out.astype(x.dtype).reshape(batch, seq_len, d_model), params['W_o'])
\end{minted}

\vspace{10pt}
\listingcaption{\textbf{Top-level chunkwise recurrence and VMEM cross-chunk state scan.} Computes cumulative log-decay $\mLambda$ on the MXU, invokes the boundary-pivoted solver, forms $(\mP_{\mathrm{read}}, \mP_{\mathrm{bias}}),$ and scans $\mS \in \R{d_k \times d_v}$ in \pycode{float32} (delegating the Stage~(iii) scan backward pass to JAX autodiff for brevity).}{lst:kernel_scan}
\begin{minted}{python}
def _mxu(lhs, rhs, high_prec=False, out_dtype=jnp.float32, fp32=None):
  """Batched MXU matmul (1-pass bf16; 3-pass bf16 if high_prec; full fp32 if both fp32)."""
  if fp32 is None:
    fp32 = (lhs.dtype == jnp.float32) and (rhs.dtype == jnp.float32)
  in_dtype = jnp.float32 if (fp32 or high_prec) else jnp.bfloat16
  prec = (jax.lax.Precision.HIGHEST if fp32 else
          jax.lax.Precision.HIGH if high_prec else jax.lax.Precision.DEFAULT)
  return jnp.matmul(lhs.astype(in_dtype), rhs.astype(in_dtype), precision=prec,
                    preferred_element_type=jnp.float32).astype(out_dtype)

def _clamp_neg(x):
  return jnp.where(x <= 0.0, x, 0.0)

def _leaf_blocks(q_b, k_b, kh_b, lam_b, orig_dtype):
  """Evaluates causal readout M_leaf and strict-tril L_leaf on diagonal leaves."""
  tril = jnp.tri(q_b.shape[-2], q_b.shape[-2], dtype=bool)
  d_raw = _clamp_neg(lam_b[..., :, None, :] - lam_b[..., None, :, :])
  k_dec = k_b[..., None, :, :] * jnp.exp(jnp.where(tril[..., None], d_raw, 0.0))
  L_blk = jnp.tril(jnp.sum(kh_b[..., :, None, :] * k_dec, axis=-1), k=-1)
  M_blk = jnp.tril(jnp.sum(q_b[..., :, None, :] * k_dec, axis=-1))
  return M_blk.astype(orig_dtype), L_blk

def muir_chunkwise(q_hat, k, k_hat, v_hat, log_alpha, leaf_size=8, init_state=None):
  """Top-level chunkwise recurrence and VMEM cross-chunk state scan."""
  _, batch, num_heads, chunk_size, d_k = q_hat.shape
  d_v, orig_dtype = v_hat.shape[-1], v_hat.dtype

  lam = _mxu(jnp.tril(jnp.ones((chunk_size, chunk_size), jnp.float32)),
             log_alpha, high_prec=True, fp32=(orig_dtype == jnp.float32))
  pi = jnp.exp(_clamp_neg(lam))
  rhs = jnp.concatenate([v_hat, (k_hat * pi).astype(orig_dtype)], axis=-1)
  M, Z = _build_and_solve(q_hat, k, k_hat, lam, rhs, leaf_size, orig_dtype)

  q_hat_pi, pi_end = q_hat * pi, pi[..., -1, :]
  k_tilde = (k * jnp.exp(_clamp_neg(lam[..., -1:, :] - lam))).astype(orig_dtype)
  prop = _mxu(jnp.concatenate([M, k_tilde.mT], axis=-2), Z)
  MU, K_tilde_U = prop[..., :chunk_size, :d_v], prop[..., chunk_size:, :d_v]
  MY, K_tilde_Y = prop[..., :chunk_size, d_v:], prop[..., chunk_size:, d_v:]

  eye_k = jnp.eye(d_k, dtype=jnp.float32)
  A_state = pi_end[..., None] * eye_k - K_tilde_Y
  P_read = jnp.concatenate([q_hat_pi - MY, A_state], axis=-2)
  P_bias = jnp.concatenate([MU, K_tilde_U], axis=-2)
  if init_state is None:
    init_state = jnp.zeros((batch, num_heads, d_k, d_v), jnp.float32)

  def scan_step(state, xs):
    packed = _mxu(xs[0], state, high_prec=True, fp32=(orig_dtype == jnp.float32)) + xs[1]
    return packed[..., chunk_size:, :], packed[..., :chunk_size, :].astype(orig_dtype)

  _, out = jax.lax.scan(scan_step, init_state, (P_read, P_bias))
  return out
\end{minted}

\vspace{10pt}
\listingcaption{\textbf{Boundary-pivoted matrix construction and block-triangular inversion.} Exact nilpotent Neumann leaf inversion ($n_0 = 8$) and $r$-level division-free boundary-pivoted merge tree ($C = 256$).}{lst:kernel_fwd}
\begin{minted}{python}
def _neumann_inv(L, mxu_dtype):
  """Exact nilpotent Neumann inverse (I + L)^{-1} on leaf blocks."""
  n0, eye = L.shape[-1], jnp.eye(L.shape[-1], dtype=jnp.float32)
  T_inv, L_pow, step = eye - L, L, 2
  while step < n0:
    L_pow = jnp.tril(_mxu(L_pow, L_pow.astype(mxu_dtype)), k=-step)
    T_inv = jnp.tril(T_inv + _mxu(T_inv, L_pow.astype(mxu_dtype)), k=-1) + eye
    step *= 2
  return T_inv

def _build_and_solve_impl(q_hat, k, k_hat, lam, rhs, leaf_size, orig_dtype):
  """Constructs M and T^{-1} via boundary-pivoted merges and solves T Z = rhs."""
  prefix, (chunk_size, d_k) = q_hat.shape[:-2], q_hat.shape[-2:]
  leaf_shape = (*prefix, chunk_size // leaf_size, leaf_size, d_k)
  q_b, k_b, kh_b, lam_b = [t.reshape(leaf_shape) for t in (q_hat, k, k_hat, lam)]

  M_blk, L_blk = _leaf_blocks(q_b, k_b, kh_b, lam_b, orig_dtype)
  T_inv_blk = _neumann_inv(L_blk, orig_dtype).astype(orig_dtype)

  n = leaf_size
  while n < chunk_size:
    p_shape = (*prefix, chunk_size // (2 * n), 2 * n, d_k)
    q_p, k_p, kh_p, lam_p = [t.reshape(p_shape) for t in (q_hat, k, k_hat, lam)]
    k_L, lam_L = k_p[..., :n, :], lam_p[..., :n, :]
    q_R, kh_R, lam_R = q_p[..., n:, :], kh_p[..., n:, :], lam_p[..., n:, :]
    
    lam_piv = lam_L[..., -1:, :]
    pi_R = jnp.exp(_clamp_neg(lam_R - lam_piv))
    pi_L = jnp.exp(_clamp_neg(lam_piv - lam_L))

    lhs_R = jnp.concatenate([kh_R * pi_R, q_R * pi_R], axis=-2).astype(orig_dtype)
    TM_cross = _mxu(lhs_R, (k_L * pi_L).astype(orig_dtype).mT, out_dtype=orig_dtype)
    T_cross, M_cross = TM_cross[..., :n, :], TM_cross[..., n:, :]

    T_inv_L, T_inv_R = T_inv_blk[..., 0::2, :, :], T_inv_blk[..., 1::2, :, :]
    T_cross_L = _mxu(T_cross, T_inv_L, out_dtype=orig_dtype)
    X_21 = -_mxu(T_inv_R, T_cross_L, out_dtype=orig_dtype)
    T_inv_blk = jnp.block([[T_inv_L, jnp.zeros_like(X_21)], [X_21, T_inv_R]])
    M_L, M_R = M_blk[..., 0::2, :, :], M_blk[..., 1::2, :, :]
    M_blk = jnp.block([[M_L, jnp.zeros_like(M_cross)], [M_cross, M_R]])
    n *= 2

  M, T_inv = M_blk[..., 0, :, :], T_inv_blk[..., 0, :, :]
  return M, _mxu(T_inv, rhs, out_dtype=orig_dtype), T_inv
\end{minted}

\vspace{10pt}
\listingcaption{\textbf{Custom chunkwise backward pass.} Differentiates $\mZ = \mT\inv[\hat\mV\;\,\hat\mK_{\odot\pi}]$ directly from the saved $(\mT\inv, \mZ)$ and recomputes $(\vpi_\gR, \vpi_\gL)$ in VMEM across the $r$ merge levels.}{lst:kernel_bwd}
\begin{minted}{python}
@functools.partial(jax.custom_vjp, nondiff_argnums=(5, 6))
def _build_and_solve(q_hat, k, k_hat, lam, rhs, leaf_size, orig_dtype):
  return _build_and_solve_impl(q_hat, k, k_hat, lam, rhs, leaf_size, orig_dtype)[:2]

def _build_and_solve_fwd(q_hat, k, k_hat, lam, rhs, leaf_size, orig_dtype):
  M, Z, T_inv = _build_and_solve_impl(q_hat, k, k_hat, lam, rhs, leaf_size, orig_dtype)
  return (M, Z), (q_hat, k, k_hat, lam, T_inv, Z)

def _build_and_solve_bwd(leaf_size, orig_dtype, res, cotangents):
  """Memory-efficient backward pass recomputing boundary-pivot decays on the fly."""
  q_hat, k, k_hat, lam, T_inv, Z = res
  dM, dZ = cotangents[0].astype(orig_dtype), cotangents[1]
  prefix, (chunk_size, d_k) = q_hat.shape[:-2], q_hat.shape[-2:]

  def diag_blocks(mat, blk):
    nb = chunk_size // blk
    d = jnp.diagonal(mat.reshape(*prefix, nb, blk, nb, blk), axis1=-4, axis2=-2)
    return jnp.moveaxis(d, -1, -3)

  drhs = _mxu(T_inv.mT, dZ)
  dL = -jnp.tril(_mxu(drhs, Z.mT, high_prec=True), k=-1)
  leaf_in = [t.reshape(*prefix, chunk_size // leaf_size, leaf_size, d_k)
             for t in (q_hat, k, k_hat, lam)]
  _, leaf_vjp = jax.vjp(lambda *xs: _leaf_blocks(*xs, orig_dtype), *leaf_in)
  d_leaf = (diag_blocks(dM, leaf_size), diag_blocks(dL, leaf_size))
  dq, dk, dkh, dlam = [g.reshape(q_hat.shape) for g in leaf_vjp(d_leaf)]

  n = leaf_size
  while n < chunk_size:
    p_shape = (*prefix, chunk_size // (2 * n), 2 * n, d_k)
    q_p, k_p, kh_p, lam_p = [t.reshape(p_shape) for t in (q_hat, k, k_hat, lam)]
    k_L, lam_L = k_p[..., :n, :], lam_p[..., :n, :]
    q_R, kh_R, lam_R = q_p[..., n:, :], kh_p[..., n:, :], lam_p[..., n:, :]
    lam_piv = lam_L[..., -1:, :]
    pi_R = jnp.exp(_clamp_neg(lam_R - lam_piv))
    pi_L = jnp.exp(_clamp_neg(lam_piv - lam_L))
    
    dT_c = diag_blocks(dL, 2 * n)[..., n:, :n].astype(orig_dtype)
    dTM = jnp.concatenate([dT_c, diag_blocks(dM, 2 * n)[..., n:, :n]], axis=-2)
    lhs_R = jnp.concatenate([kh_R * pi_R, q_R * pi_R], axis=-2).astype(orig_dtype)
    dkh_q_R = _mxu(dTM, (k_L * pi_L).astype(orig_dtype))
    dkh_R_pi, dq_R_pi = dkh_q_R[..., :n, :], dkh_q_R[..., n:, :]
    dk_L_pi = _mxu(dTM.mT, lhs_R)

    flux_R = jnp.where(lam_R <= lam_piv, (dkh_R_pi * kh_R + dq_R_pi * q_R) * pi_R, 0.0)
    flux_L = jnp.where(lam_piv <= lam_L, (dk_L_pi * k_L) * pi_L, 0.0)
    dlam_L = (-flux_L).at[..., -1:, :].add(jnp.sum(flux_L - flux_R, axis=-2, keepdims=True))
    zeros = jnp.zeros_like(k_L)
    dq += jnp.concatenate([zeros, dq_R_pi * pi_R], axis=-2).reshape(q_hat.shape)
    dkh += jnp.concatenate([zeros, dkh_R_pi * pi_R], axis=-2).reshape(k_hat.shape)
    dk += jnp.concatenate([dk_L_pi * pi_L, zeros], axis=-2).reshape(k.shape)
    dlam += jnp.concatenate([dlam_L, flux_R], axis=-2).reshape(lam.shape)
    n *= 2

  return (dq.astype(q_hat.dtype), dk.astype(k.dtype), dkh.astype(k_hat.dtype),
          dlam.astype(lam.dtype), drhs.astype(orig_dtype))

_build_and_solve.defvjp(_build_and_solve_fwd, _build_and_solve_bwd)
\end{minted}